\documentclass[10pt,twocolumn,letterpaper]{article}
\usepackage{cvpr}
\usepackage{graphicx}
\usepackage{amsmath}
\usepackage{amssymb}
\usepackage{booktabs}
\usepackage{enumitem}

\usepackage[T1]{fontenc}    
\usepackage{url}            
\usepackage{amsfonts}       
\usepackage{nicefrac}       
\usepackage{xcolor}         
\usepackage{comment}
\usepackage{caption}
\usepackage{tabularx}
\usepackage[export]{adjustbox}
\usepackage{amsmath}
\usepackage{array}
\usepackage{soul,color}
\usepackage{multirow}
\usepackage{multicol}
\usepackage{bm}
\usepackage[pagebackref=true,breaklinks=true,letterpaper=true,colorlinks,bookmarks=false]{hyperref}

\usepackage[capitalize]{cleveref}
\crefname{section}{Sec.}{Secs.}
\Crefname{section}{Section}{Sections}
\Crefname{table}{Table}{Tables}
\crefname{table}{Tab.}{Tabs.}
\crefname{figure}{Figure}{Figures}
\crefname{equation}{Equation}{Equations}
\newcommand\func[1]{\mathsf{#1}} 
\newcommand\vect[1]{\mathbf{#1}}
\def \real {\mathbb{R}}
\def \wspace {\mathcal{W}}
\def \zspace {\mathcal{Z}}
\def \normal {\mathcal{N}}
\def \sphere {\mathcal{S}}
\def \w {\vect{w}}

\def \noise {\bm{\eta}}
\def \z {\vect{z}}
\def \x {\vect{x}}
\def \thetaa {\bm{\theta}}
\def \y {\vect{y}}

\def \F {\func{F}}
\def \G {\func{G}}
\def \D {\func{D}}
\def \dim {d}

\def \layer{L}
\def \n {n}

\def \zero {\vect{0}}
\def \eye {\vect{I}}
\def \Pg {p_{\G}}
\def \Pf {p_{\F}}
\def \deltaa {\delta}
\def \Pdelta {p_{\deltaa}}
\def \Pw {p_{\w}}

\def \prior {\mathcal{P}}
\newcommand\MyIm[1]{%
	\includegraphics[width=1.7cm,height=1.7cm]{#1} 
}
\newcommand\MyImm[1]{%
	\includegraphics[width=1.5cm,height=1.5cm]{#1}
}
\newcommand\MyImbig[1]{%
	\includegraphics[width=3cm,height=3cm]{#1}
}
\newcommand{\blue}[1]{\textcolor{blue}{#1}}
\newcommand{\red}[1]{\textcolor{red}{#1}}

\newcolumntype{C}{ >{\centering\arraybackslash} m{1.7cm}} 
\newcolumntype{D}{ >{\centering\arraybackslash} m{1.6cm}}
\newcolumntype{E}{ >{\centering\arraybackslash} m{1.5cm}}
\newcolumntype{F}{ >{\centering\arraybackslash} m{3cm}}

\DeclareRobustCommand\onedot{\futurelet\@let@token\@onedot}
\def\onedot{. }
 
\def\ie{\emph{i.e}\onedot}

\begin{document}

\title{Super-Resolution through StyleGAN Regularized Latent Search: A Realism-Fidelity Trade-off}

\author{Marzieh Gheisari\\
	École Normale Supérieure\\
	Paris, France\\
	{\tt\small gheisari@bio.ens.psl.eu}
	\and
	Auguste Genovesio\\
	École Normale Supérieure\\
	Paris, France\\
	{\tt\small auguste.genovesio@ens.psl.eu}
}

\maketitle
\begin{abstract}
This paper addresses the problem of super-resolution: constructing a highly resolved (HR) image from a low resolved (LR) one. Recent unsupervised approaches search the latent space of a StyleGAN pre-trained on HR images, for the image that best downscales to the input LR image. However, they tend to produce out-of-domain images and fail to accurately reconstruct HR images that are far from the original domain. Our contribution is twofold. Firstly, we introduce a new regularizer to constrain the search in the latent space, ensuring that the inverted code lies in the original image manifold. Secondly, we further enhanced the reconstruction through expanding the image prior around the optimal latent code. Our results show that the proposed approach recovers realistic high-quality images for large magnification factors. Furthermore, for low magnification factors, it can still reconstruct details that the generator could not have produced otherwise. Altogether, our approach achieves a good trade-off between fidelity and realism for the super-resolution task.
\end{abstract}
\begin{figure}[t!]
	\centering
	\includegraphics[width=\linewidth]{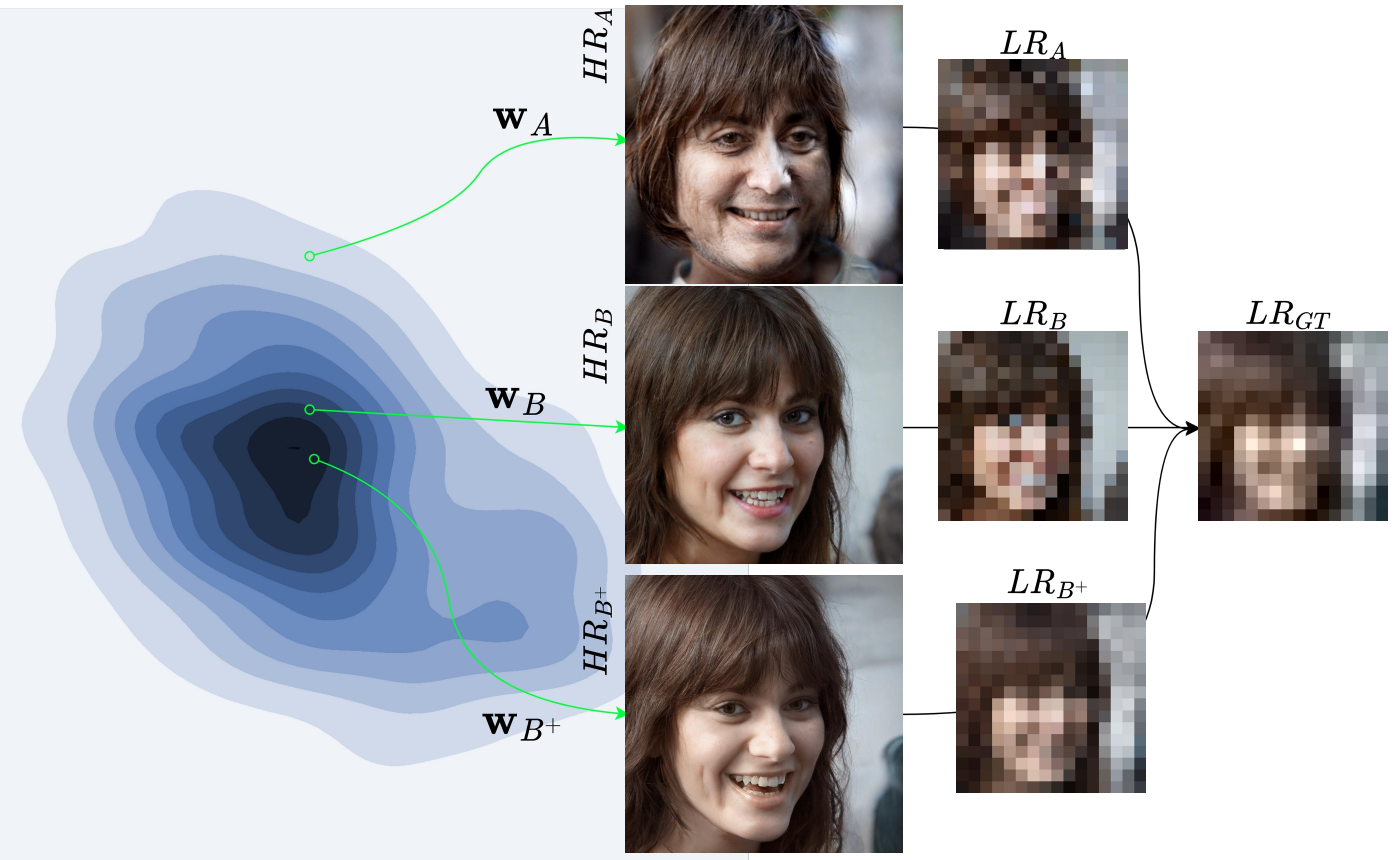}
	\caption{Searching the latent space of a StyleGAN without proper constraints leads to an unrealistic HR image, represented by $\w_A$. However, RLS finds an optimal latent code $\w_B$ located in the dense regions of the image distribution. RLS$^+$ further improves the results by modulating the generator's weights within a small $\ell_1$-norm ball centered on $\w_B$, resulting in $\w_{B^+}$, an image that accurately matches the input LR image when degraded.}
	\label{fig:intro}
\end{figure}

\section{Introduction}
\label{sec:intro}
Super-resolution aims to reconstruct an unknown High Resolution (HR) image $\x\in\real^{\n\times \n}$ from a Low Resolution (LR) image $\y\in\real^{m\times m}$, related to one another by a down-sampling process
described by $\y=\D(\x)+\deltaa$ with  $\D:\real^{n\times n}\rightarrow\real^{m\times m}$ a down-sampling non-invertible forward operator and $\deltaa$ an independent noise with distribution $\Pdelta$.
\begin{figure*}[!h]
	\centering
	\includegraphics[scale=0.5]{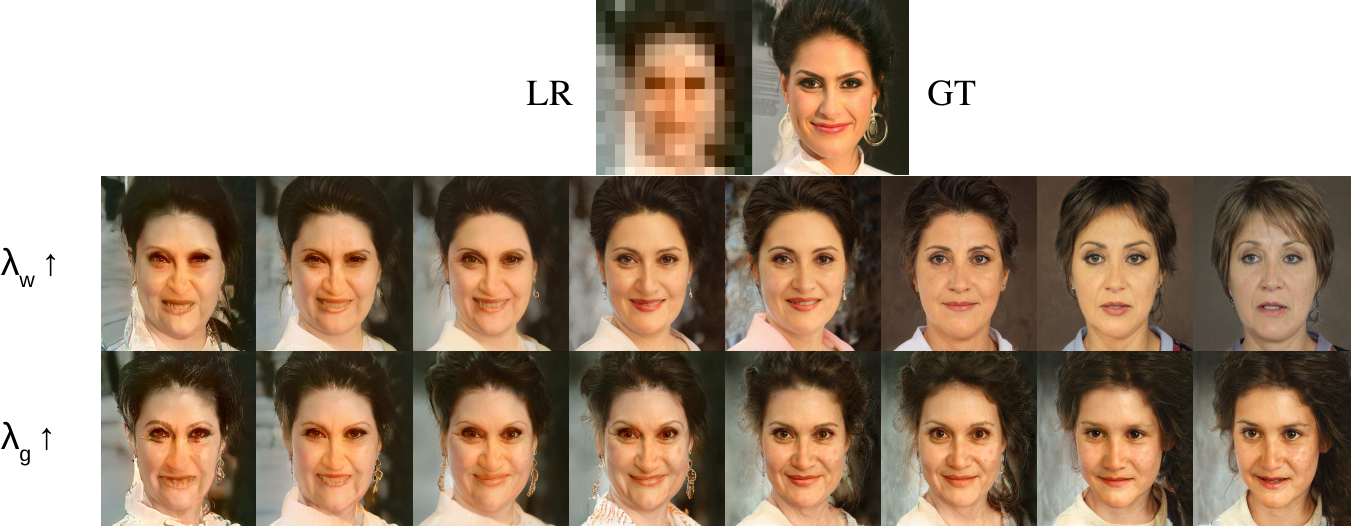}
	\caption{Impact of regularizer parameters $\lambda_w$ and $\lambda_g$ on fidelity and realism. The x-axis represents increasing values of $\lambda_w$ (respectively $\lambda_g$) from left to right, with $\lambda_g=0$ (respectively $\lambda_w=0$) and $\lambda_c=0$.}
	\label{figure tradeoff}
	\vspace{-5pt}
\end{figure*}
As with many data generation tasks, super-resolution has largely benefited in recent years from the advent of Generative models.
Two main research trends have emerged for this task: GAN-based methods and prior-guided methods.

GAN-based methods~\cite{isola2017image, wang2018high,wang2018esrgan}, learn a direct coupling of HR and LR images based on coupled image translation.
Recently, GCFSR~\cite{he2022gcfsr} introduced a generative and controllable face super-resolution model that does not rely on additional priors and has been shown to reconstruct faithful images.
However, GAN-based methods have limitations in that they train the SR generator from scratch using a combined objective function consisting of a fidelity term and an adversarial loss.
This approach requires the generator to capture both the natural image characteristics and the fidelity to the ground-truth, which can result in limitations when approximating the natural image manifold.
As a result, GAN-based methods often produce artifacts and unnatural textures.

Moreover, the super-resolution problem is ill-posed as, for a non-invertible forward operator $\D$ with $m<n$, there are infinitely many HR images that match a given LR image.
Thus reconstruction procedure must be further constrained by prior information to better define the objective and lead to a stable solution.
In addition, GAN-based methods often rely on a specific degradation model during training, which can restrict their ability to handle the true degradation that can be encountered in real-world applications.

Prior-guided methods~\cite{kawar2021snips,bora2017compressed,wang2021towards,yang2021gan} can be considered as blind restoration techniques, as they can adapt to the given problem without requiring re-training. These methods fall into two main categories: posterior sampling-based and optimization-based methods.
DDRM~\cite{kawardenoising} is a posterior sampling method that uses a pre-trained denoising diffusion generative model to gradually denoise a sample to the desired output, conditioned on the LR input image.

Optimization-based methods, learn the distribution of HR images in an unsupervised fashion using a GAN, and then search the latent space of this trained GAN to find
the HR image that, once down-sampled, is the closest to the LR image.
This idea was first introduced by Bora et \textit{al}~\cite{bora2017compressed}, and subsequent work by PULSE~\cite{menon2020pulse} and BRGM~\cite{marinescu2021bayesian} improved upon it using StyleGAN.

To keep the search within the image manifold, BRGM assumed that the intermediate latent space $\wspace^+$ followed a standard Gaussian distribution, $\normal(\zero, \eye_\dim)$, where $\dim$ is the dimensionality of the latent space. On the other hand, PULSE and a few other studies utilized an invertible transformation of $\wspace^+$ that included a leaky rectified linear unit (ReLU)~\cite{goodfellow2016deep} followed by an affine whitening transformation to ensure that transformed latent vectors approximately followed the standard Gaussian distribution. Sampled vectors are then constrained to lie around a hypersphere with radius $\sqrt{\dim}$ hypothesizing that most of the mass of a high-dimensional Gaussian distribution is located at or near $\sqrt{\dim} \sphere^{\dim-1}$, where $\sphere^{\dim-1}$ is the $\dim$-dimensional unit hypersphere. Constraining samples to lie in dense area of the StyleGAN style distribution resulted in increased realism of the generated images.

Although the above approach showed major improvements over previous work, it also presents three important caveats that in practice led to image artifacts.
First, as we will show later, transforming the intermediate latent space this way does not lead to an accurate standard Gaussian distribution, and prevents proper regularization based on this hypothesis. Second, we argue that a search strictly limited to the spherical surface $\sphere^{\dim-1}$ restricts access to the whole variety of images a StyleGAN can generate, thus preventing a close reconstruction of the HR image to be reached. Third, there is an inherent trade-off between realism and fidelity: the generator encoding the image prior can produce images from the learnt domain, however, there is no guarantee a pre-trained generator can produce a specific image we aim to reconstruct.

In this work, firstly, we operate a Regularized Latent Search (RLS) for a latent code located in ``healthy'' regions of the latent space. In this way, the system is constrained to produce images that belong to the original image domain StyleGAN was trained on. 
To do so, we take advantage of normalizing flow to Gaussianize the
latent style sample distribution and show that it leads to a
much closer standard Gaussian distribution.
We then use this revertible transformation to regularize the search in $\wspace^+$
such that it remains in a high-density area of the style vector
distribution.

Secondly, we mitigate the realism-fidelity trade-off issue by slightly modulating the generator's weights
within a small $\ell_1$-norm ball centered to the previously identified latent code.
To this end, we perform a small number of additional iterations to fine-tune the generator's training.
In this way, it becomes possible to faithfully reconstruct a slightly out-of-domain HR image by simply increasing the generator's domain on demand.
We then show experimentally, that the latter produces reconstructed images that are not only realistic but also
more faithful to the original HR image.
We also show that the approach is robust to noise and
other image corruptions.
\section{Prior work}
\subsection{Style-based Generative Adversarial Networks}
\label{sec:stylebased}
StyleGAN models are well known for generating highly realistic images. The StyleGAN architecture consists of two sub-networks: a mapping network $\G_m:\real^\dim \rightarrow \real^\dim$, and an $\layer$-layer synthesis network $\G_s :\real^{\layer\times\dim} \rightarrow \real^{\n}$. The mapping network maps a sample $\z \in \real^\dim$ from a standard normal distribution to a vector $\w \in \wspace$ . 
The $\dim\times \layer$ dimensional vector $\w \in \wspace^+$ containing $\layer$ copies of $\w$ is fed to the $\layer$-layer synthesis network $\G_s$. The $i$-th copy of $\w$ represents the input to the $i$-th layer of $\G_s$, which controls the $i$-th level of detail in the generated image. In addition to these, $\G_s$ also takes as input a collection of latent noise vectors $\noise$ that control minor stochastic variations of the generated image at each scale.

The ability of a StyleGAN to control features of the generated image at different scales is partially due to this architecture, and partially due to the style-mixing regularization occurring during training~\cite{karras2019style, karras2020analyzing}. 
In addition to these basic characteristics, StyleGAN2 introduces path length regularization, which helps in  reducing the representation error. Prior works demonstrated that performing latent search in the extended latent space $\wspace^+$ led to more accurate reconstructions but at the cost of a reduced editability~\cite{wulff2020improving, abdal2019image2stylegan}. 
In PULSE, GEOCROSS is introduced as a penalty term to force the embedding in the extended latent space $\wspace^+$ to remain close to the latent space $\wspace$, which in turn promotes the embedded object to be close to the range of the generator network $\G_s$ with the latent space $\wspace$.
\subsection{Learning the prior with Normalizing Flow}
\label{sec:normalizing_flow}


Using a sequence of invertible mappings, a Normalizing Flow $\F:\real^\dim\rightarrow\real^{\dim}$ is a transformation of an unknown complex distribution into a simple probability distribution that is easy to sample from and whose density is easy to evaluate such as standard Gaussian~\cite{kobyzev2020normalizing}.

Let $\z=\F(\w)$ with probability density function $p(\z)$. Using the change-of-variable formula, we can express the log-density of $\w$ by\cite{papamakarios2021normalizing}:
\begin{equation}
\log \Pf(\w) = \log p(\z) + \log |\det J_{\F} (\w)|, \quad \z=\F(\w)
\label{equation nf}
\end{equation}
where $J_\F(\w)$ is the Jacobian of $\F$ evaluated at $\w$.

In practice the Jacobian determinant in \cref{{equation nf}} should be easy to compute, so that the density $\Pf(\w)$ can be evaluated. Furthermore, as a generative model, the invertibility of $\F$ allows new samples $\w = \F^{-1}(z)$ to be drawn through sampling from the base distribution.

In the literature, several flow models were proposed, such as Real Non-volume Preserving Flow (RealNVP)~\cite{dinh2016density} and Masked Auto-regressive Flow (MAF)~\cite{papamakarios2017masked}.

\setlength{\tabcolsep}{2pt}

\begin{figure}
	\centering
	\includegraphics[width=0.7\linewidth,valign=m]{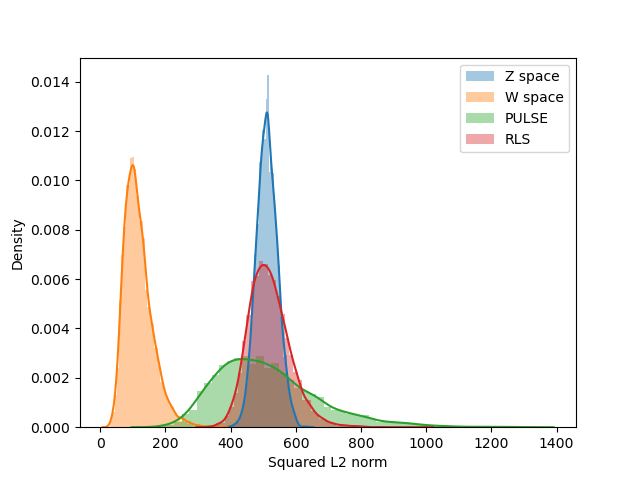} 
	\caption{Evaluation of the gaussianization prior.}
	\label{fig:gaussian}
\end{figure}
\section{Proposed Method}
In this work, we aim to find the latent code of a pre-trained StyleGAN that produces a realistic HR image that, when degraded, matches the input LR image accurately. The proposed method includes two steps to ensure that the results reside within the image domain and are close to the HR image.
 First, in order to leverage the StyleGAN prior, we restrict the optimization solution to remain on the manifold.
This restriction can be efficiently implemented by incorporating the image prior introduced in \cref{sec:RLS}. However, especially in the case of slightly out-of-domain images, this approach yields less faithful reconstructions. Therefore, in a second step (\cref{sec:RLS+}), the generator is refined around this anchor point to recover the missing details without affecting the image prior.

\begin{figure*}[!h]
	\centering
	\setlength{\tabcolsep}{2pt}
	\begin{scriptsize}
		\begin{tabular}{cCCCCCCCC}
			& & PULSE\cite{menon2020pulse} & BRGM\cite{marinescu2021bayesian} & GPEN\cite{yang2021gan} & GFPGAN\cite{wang2021towards} & DDRM\cite{kawardenoising} & RLS & RLS+ \\

   			64x & \MyIm{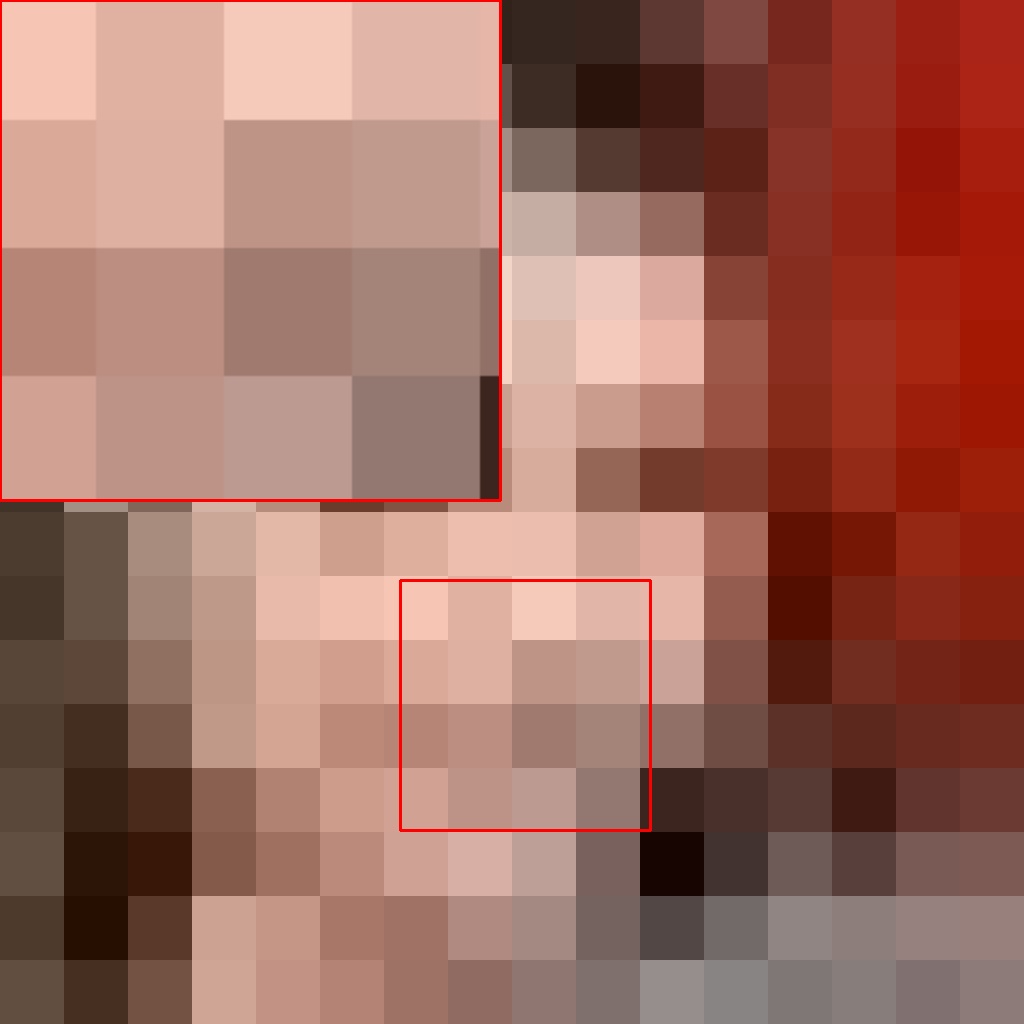}
			& \MyIm{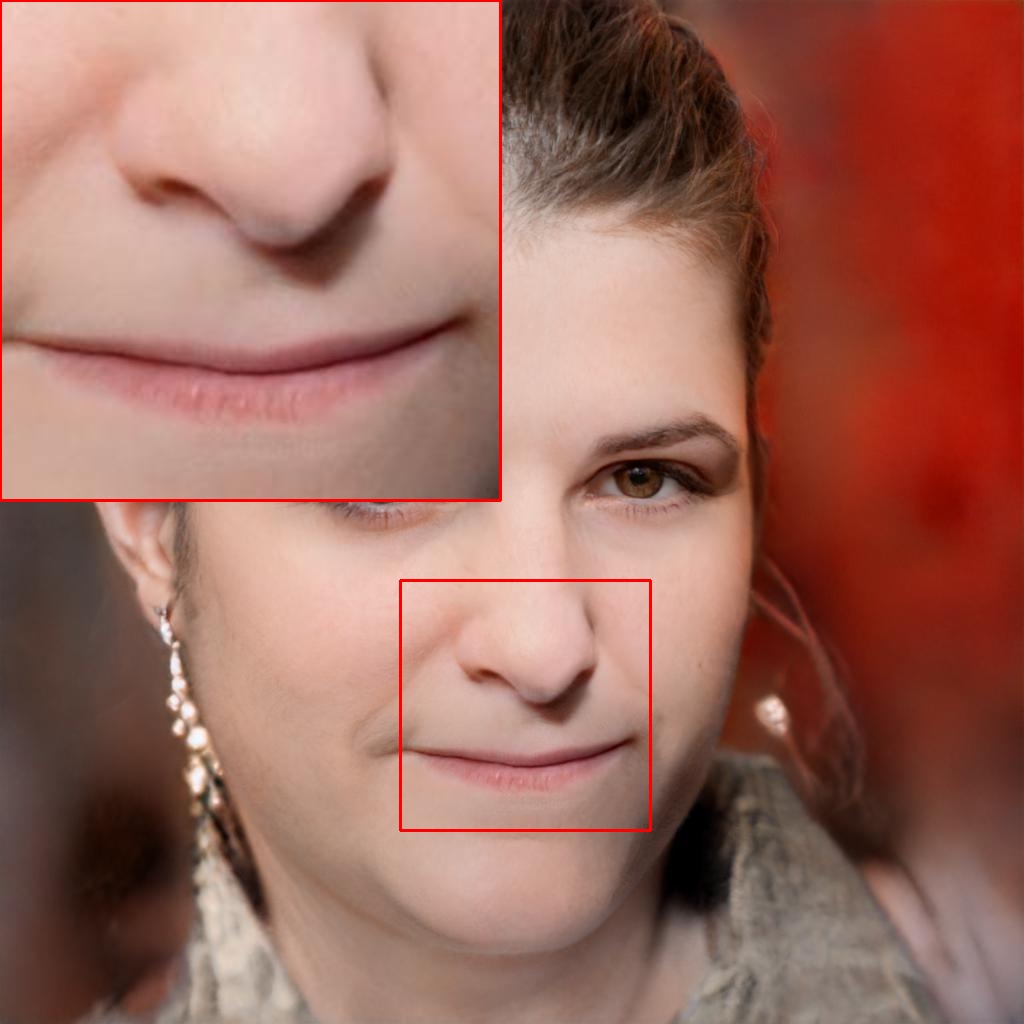}
			& \MyIm{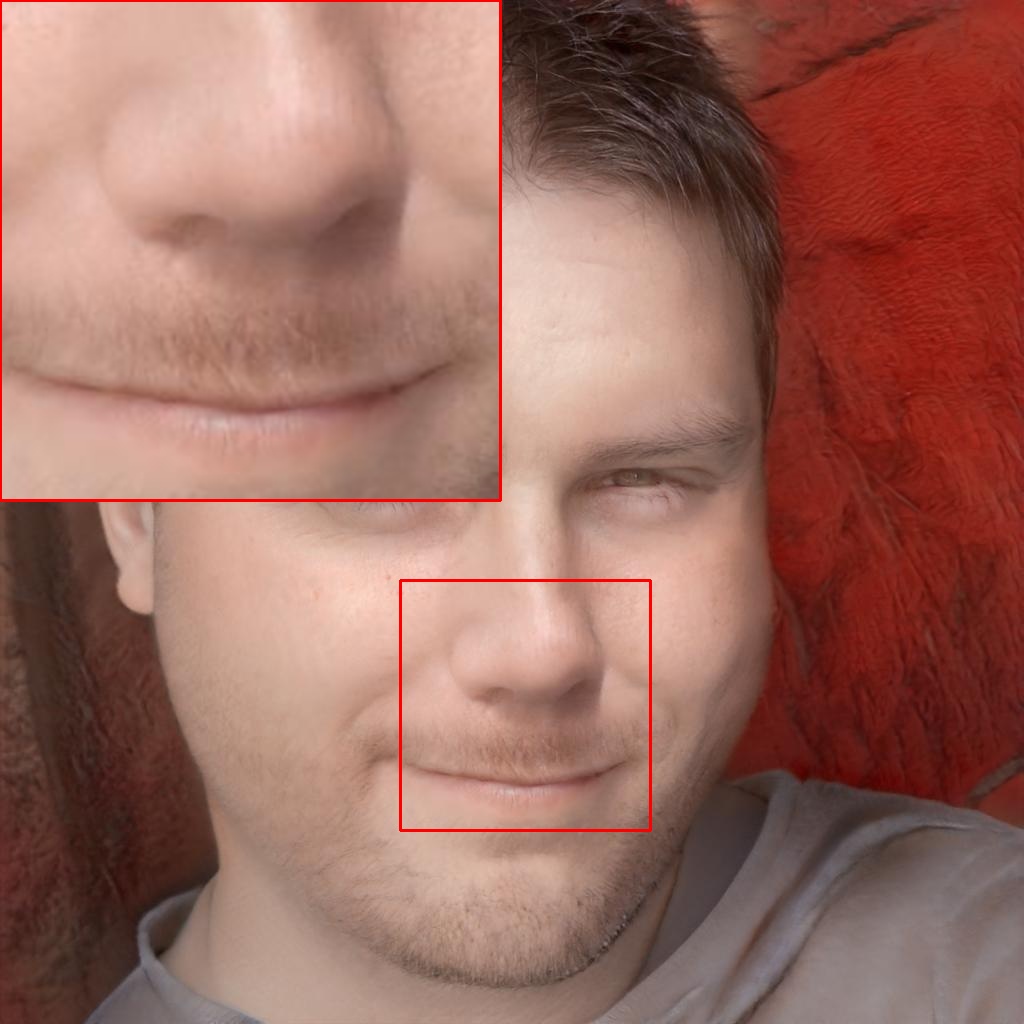}
			& \MyIm{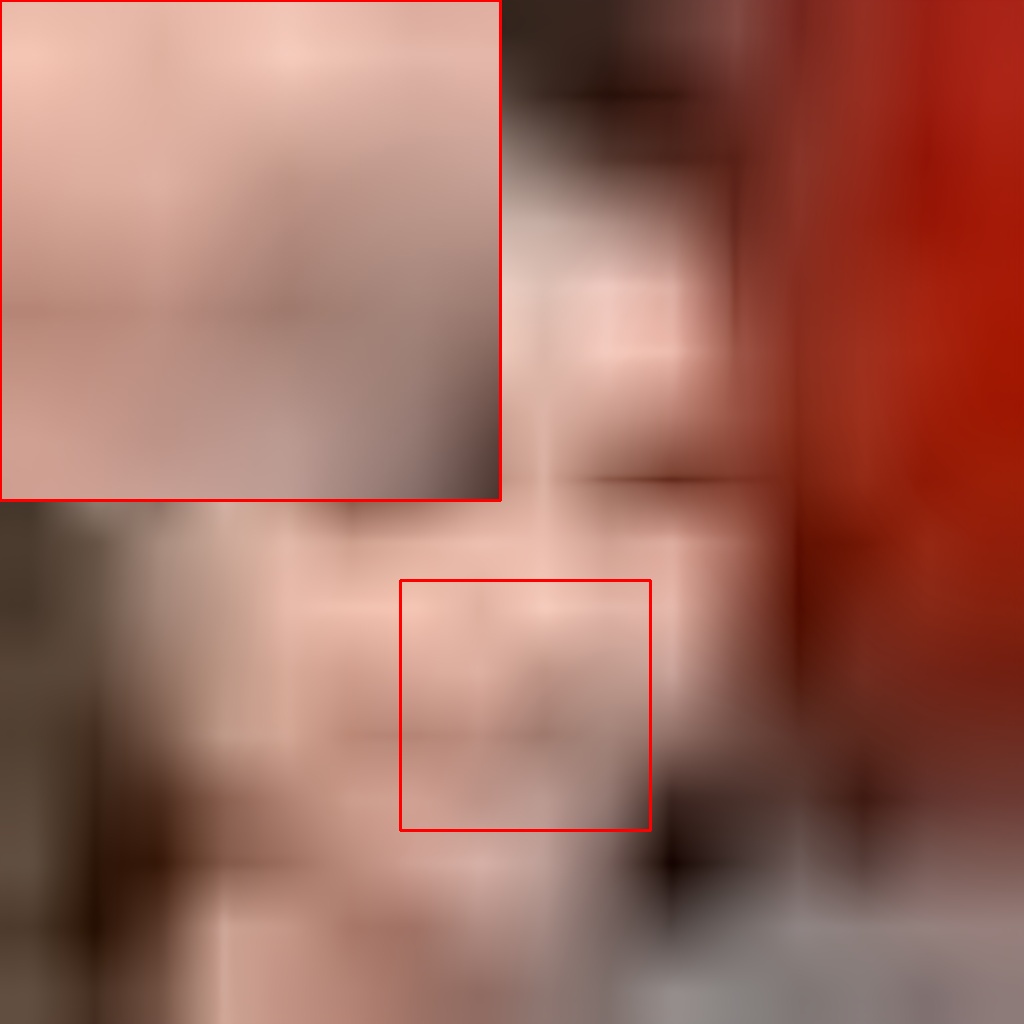}
			& \MyIm{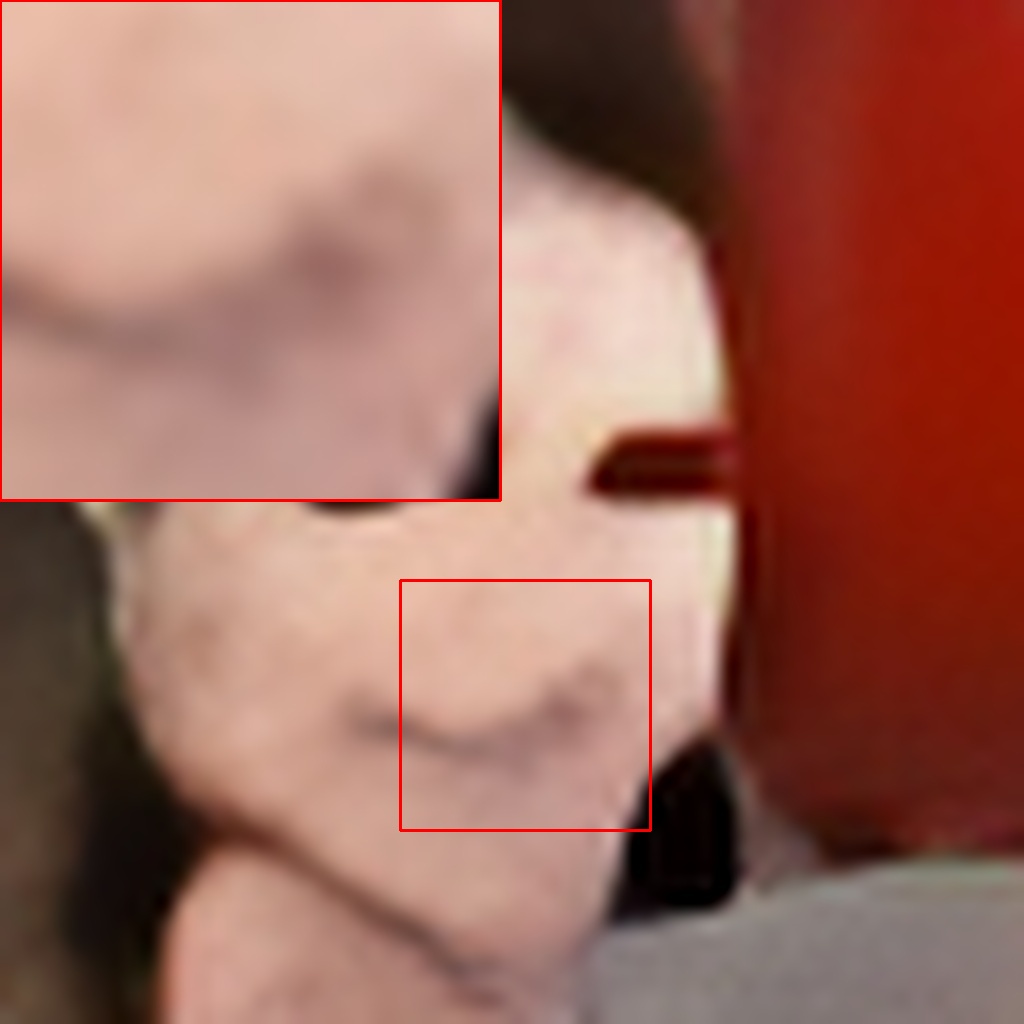}
			& \MyIm{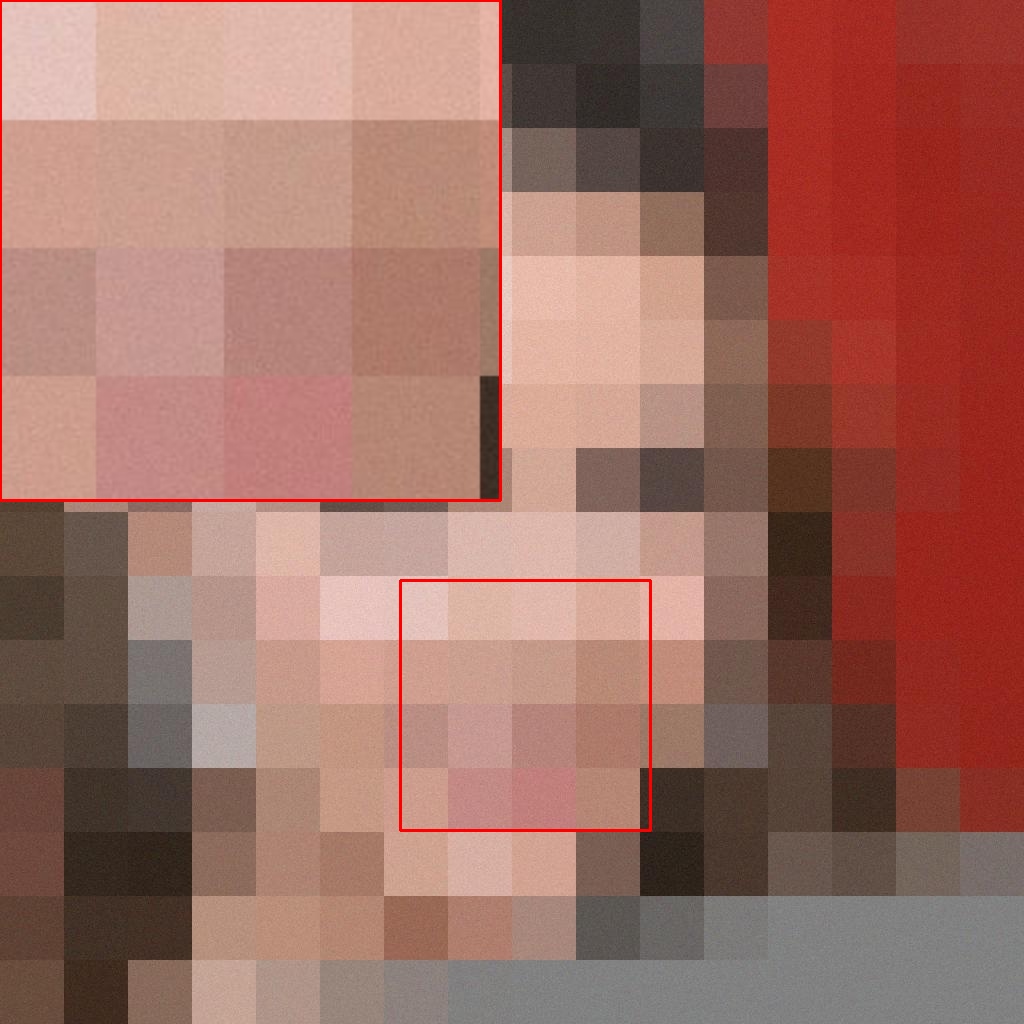}
			& \MyIm{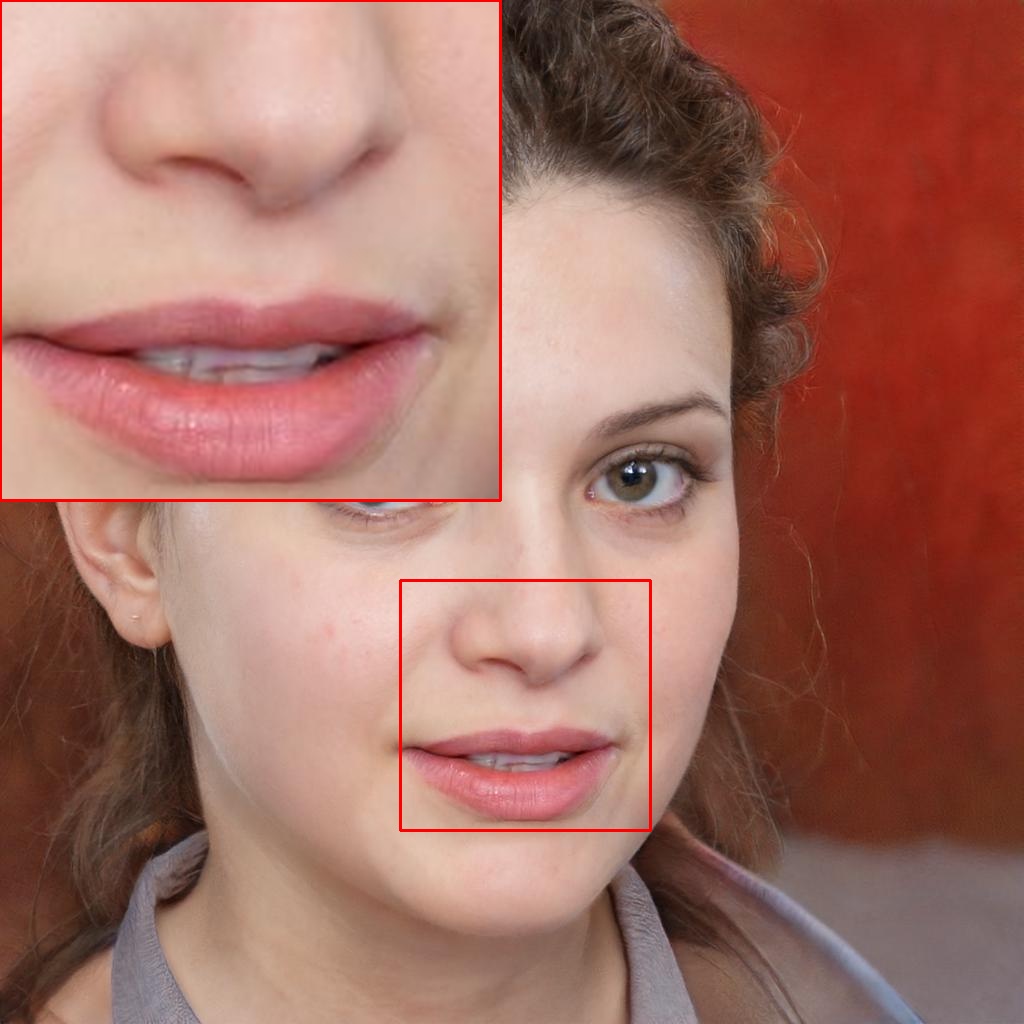}
			& \MyIm{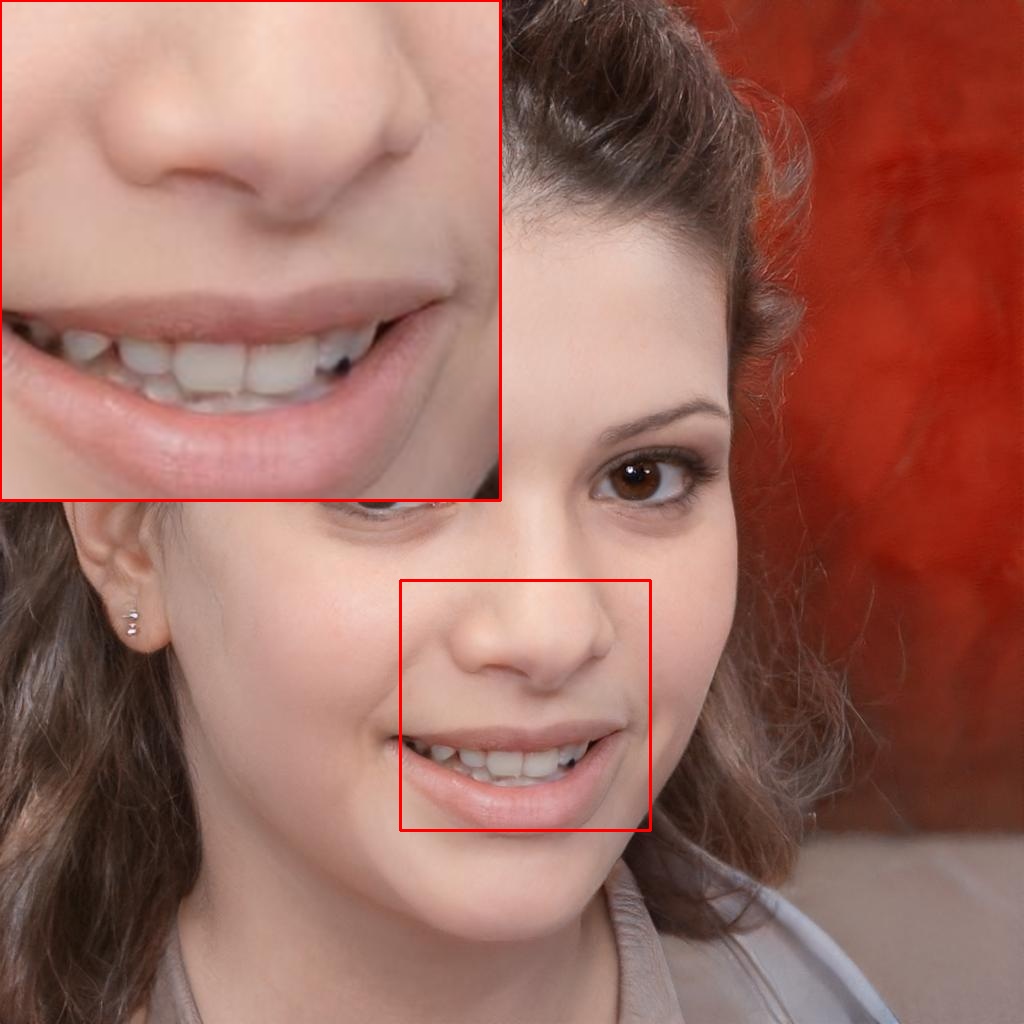} \\
			8x & \MyIm{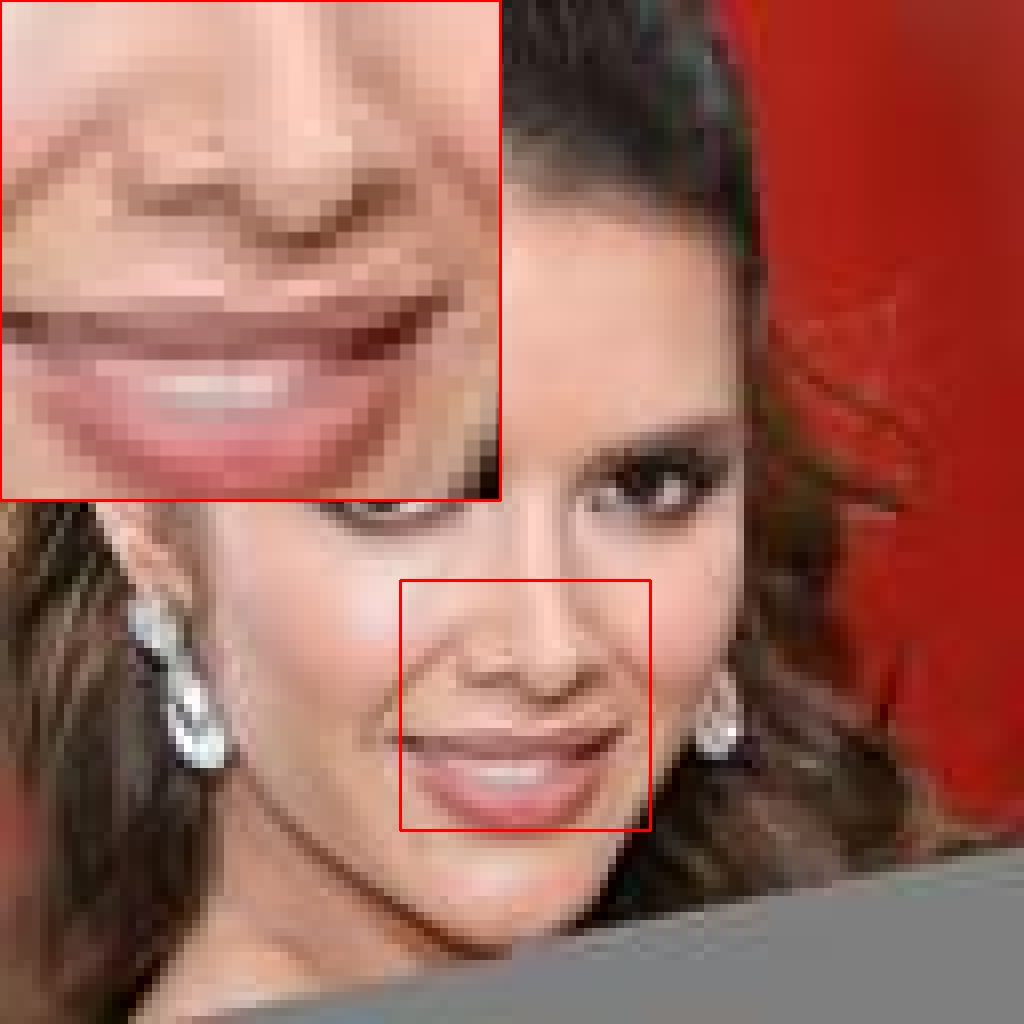}
			& \MyIm{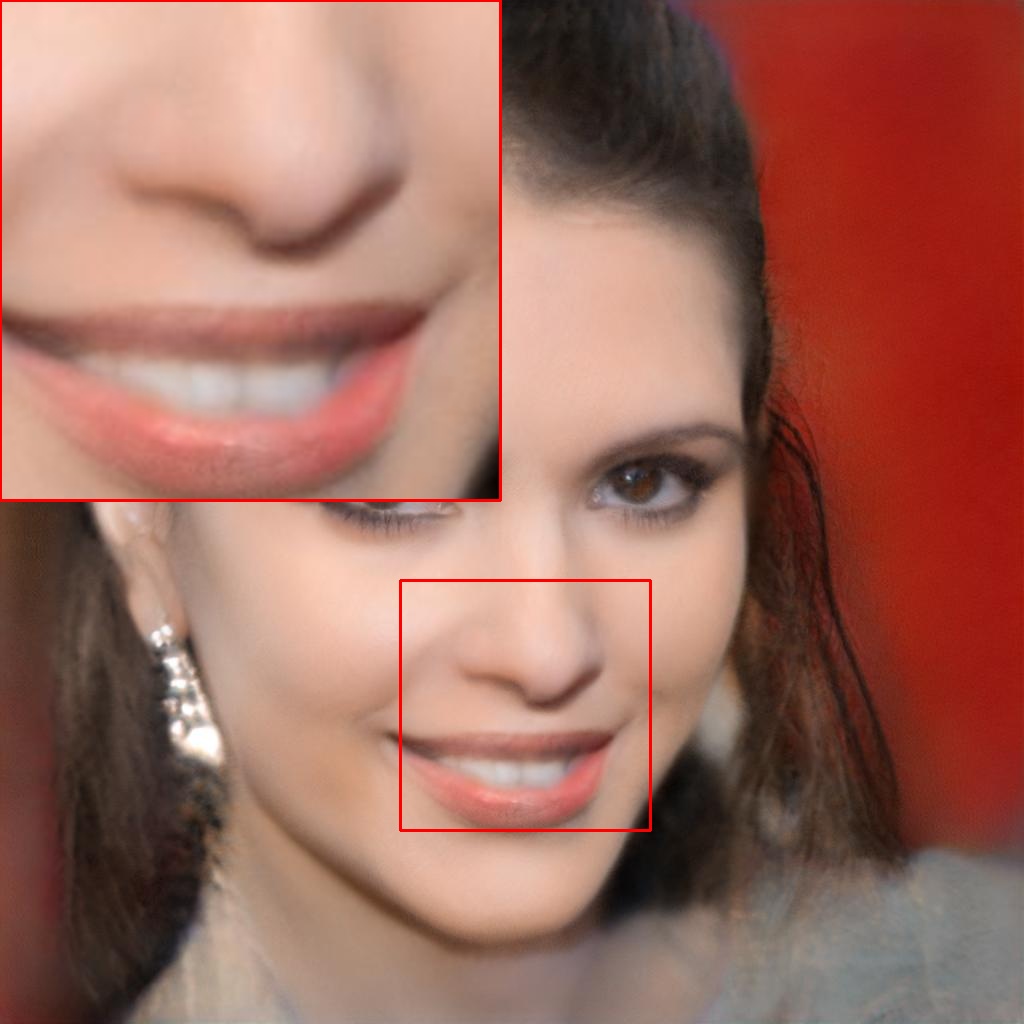}
			& \MyIm{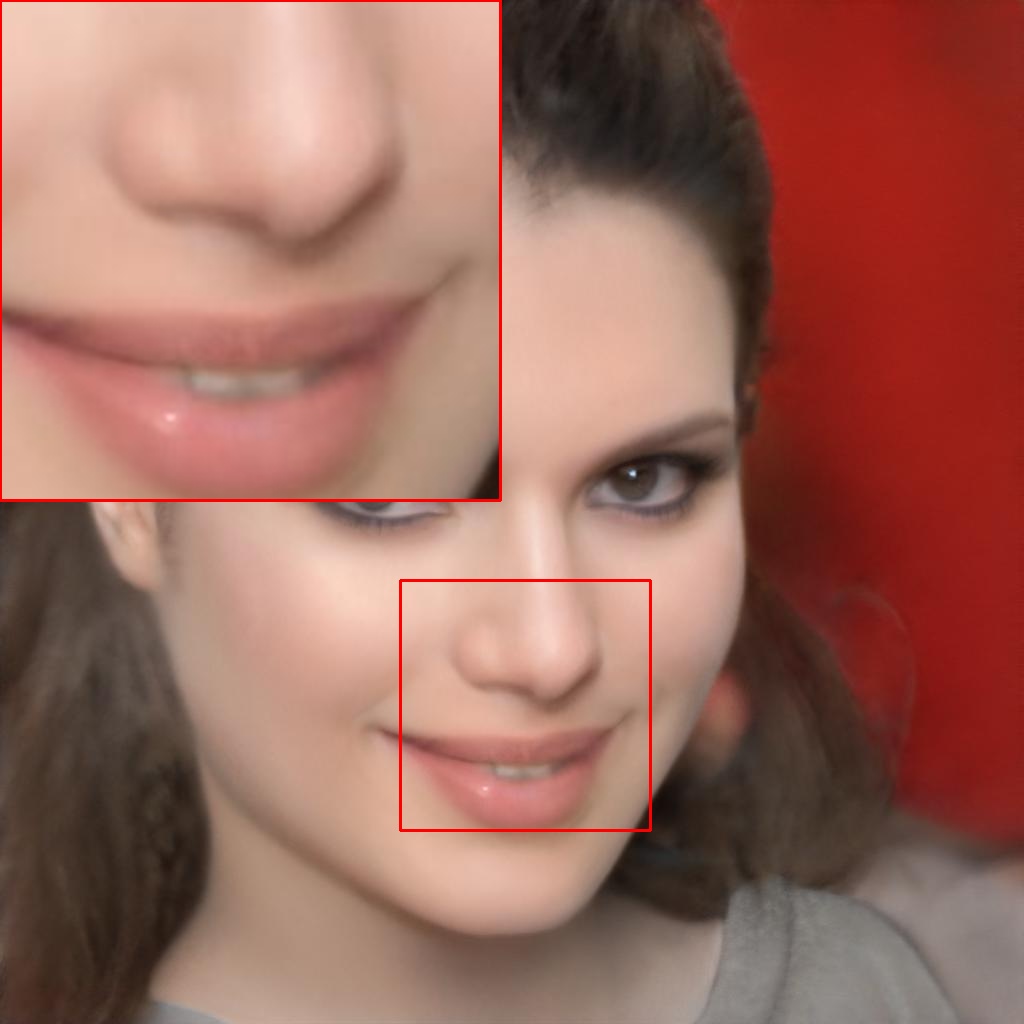}
			& \MyIm{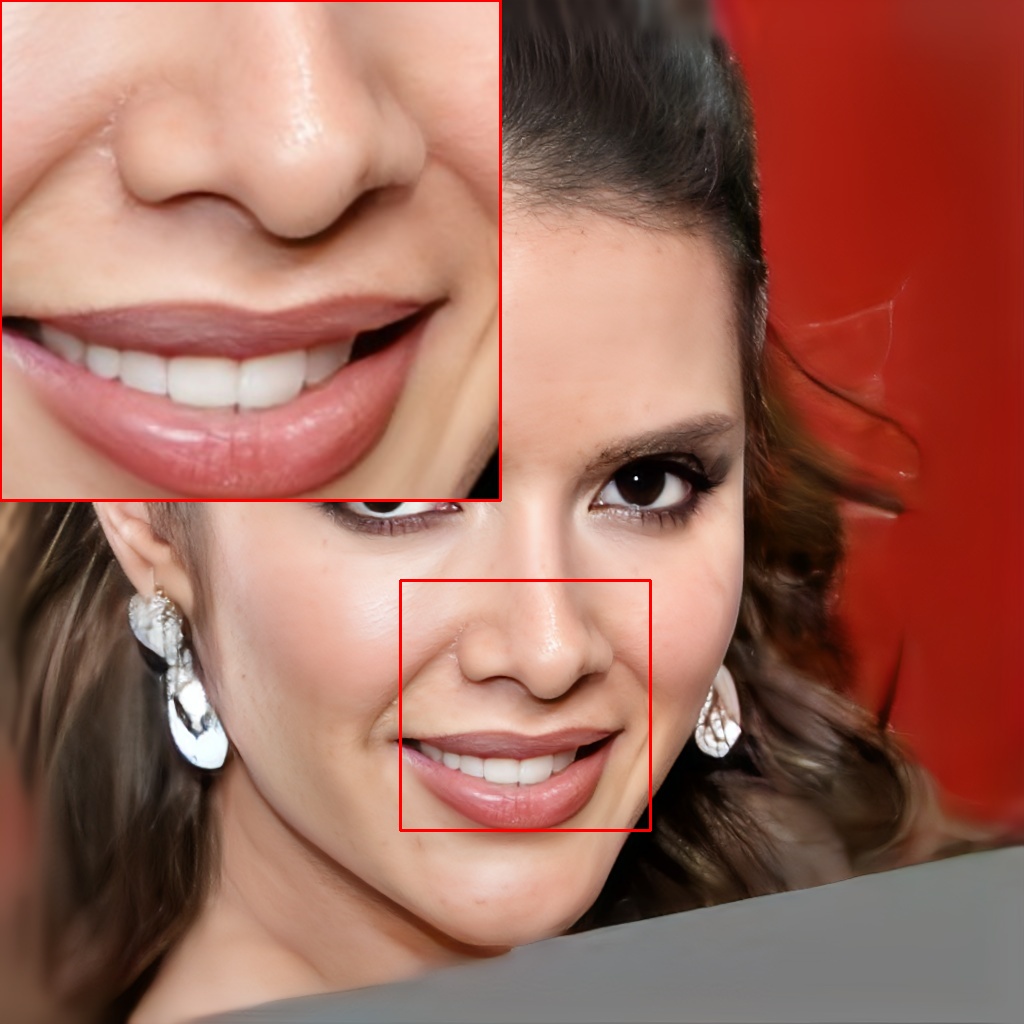}
			& \MyIm{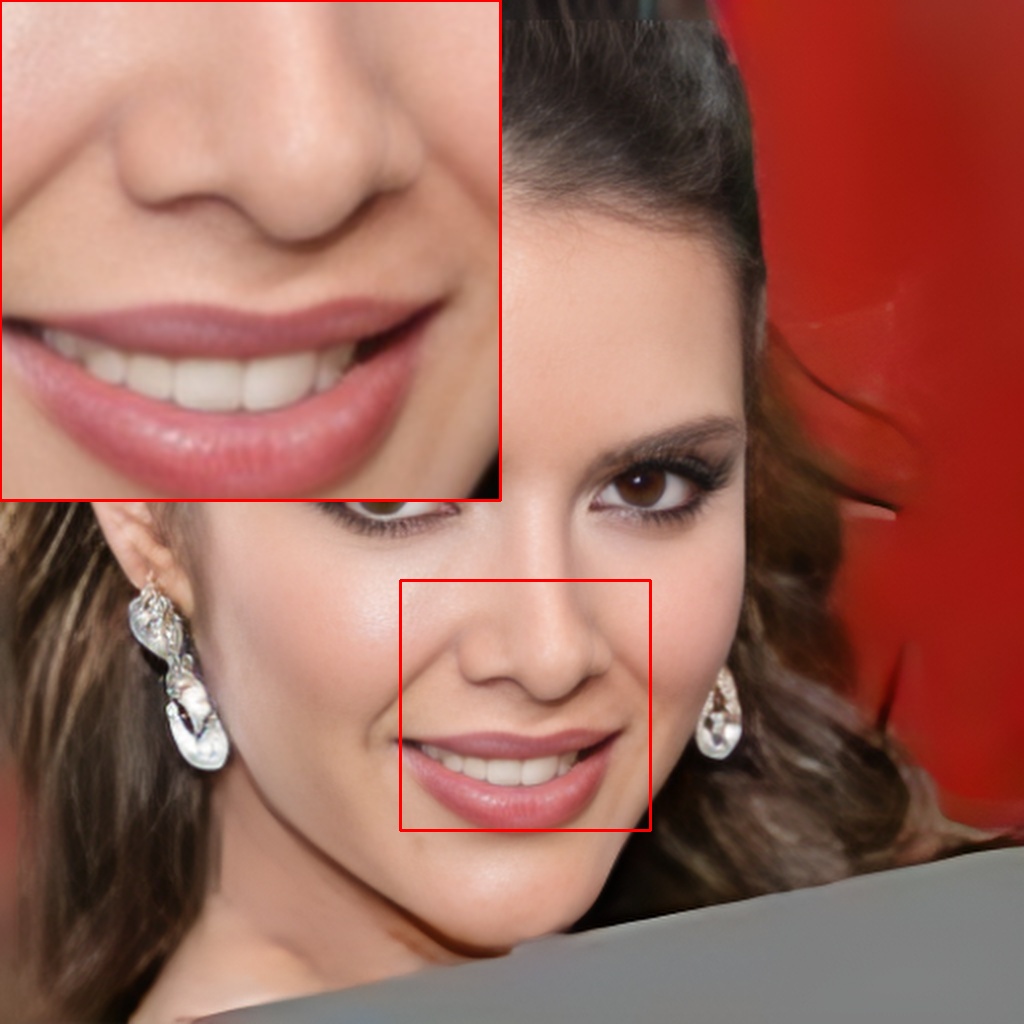}
			& \MyIm{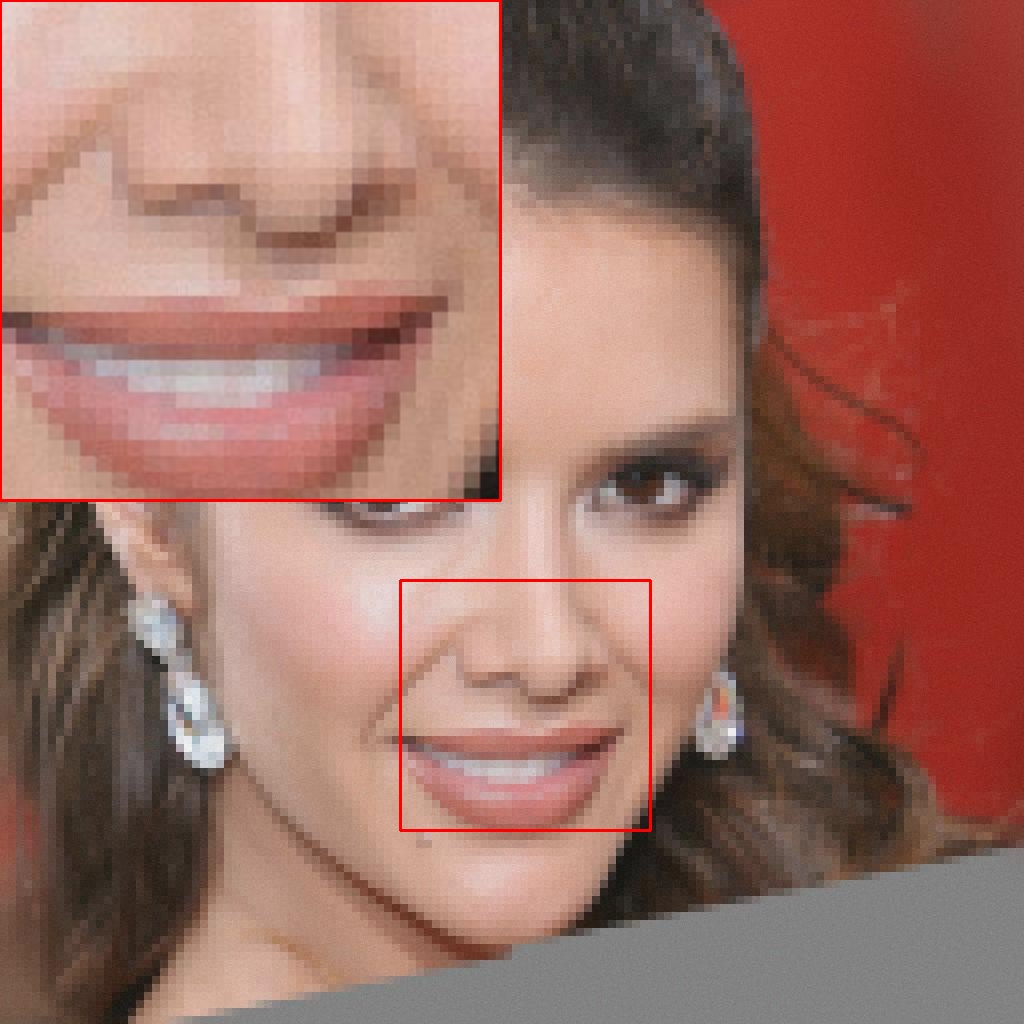}
			& \MyIm{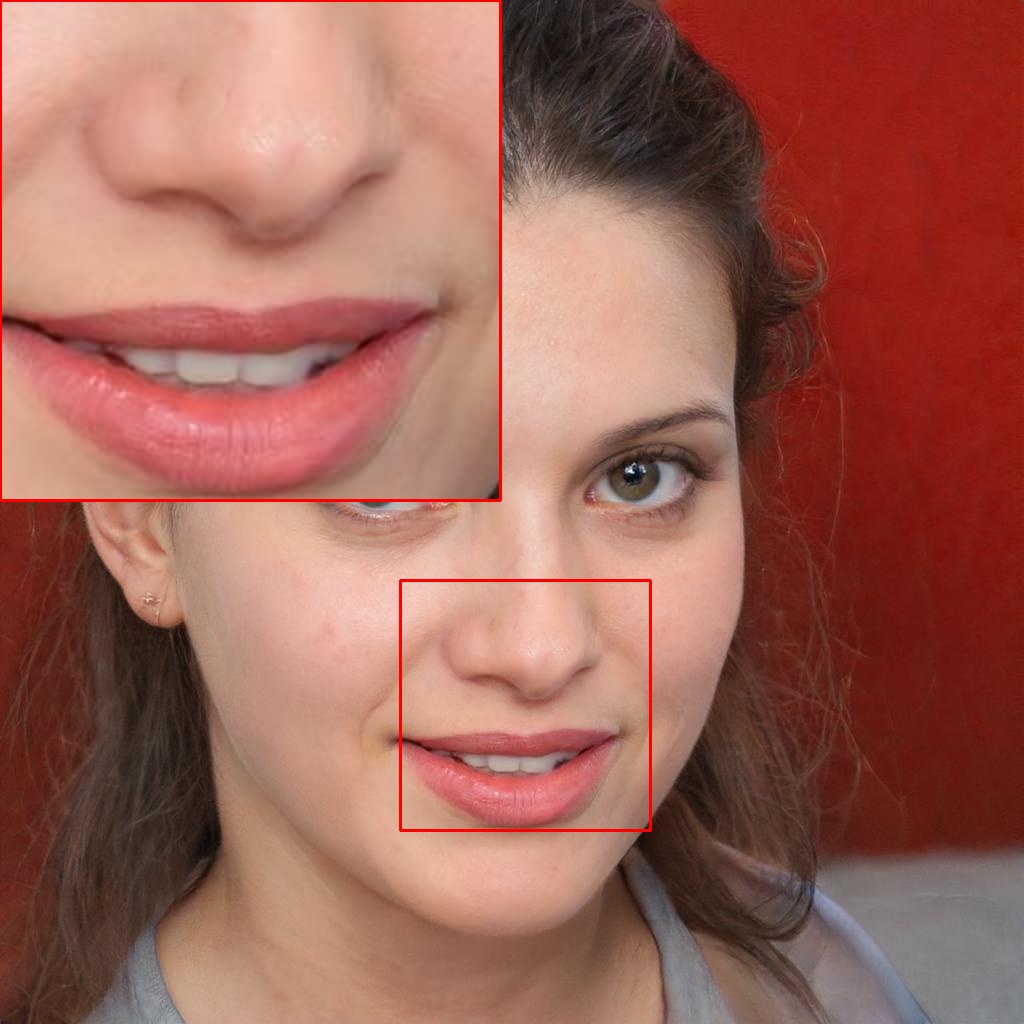}
			& \MyIm{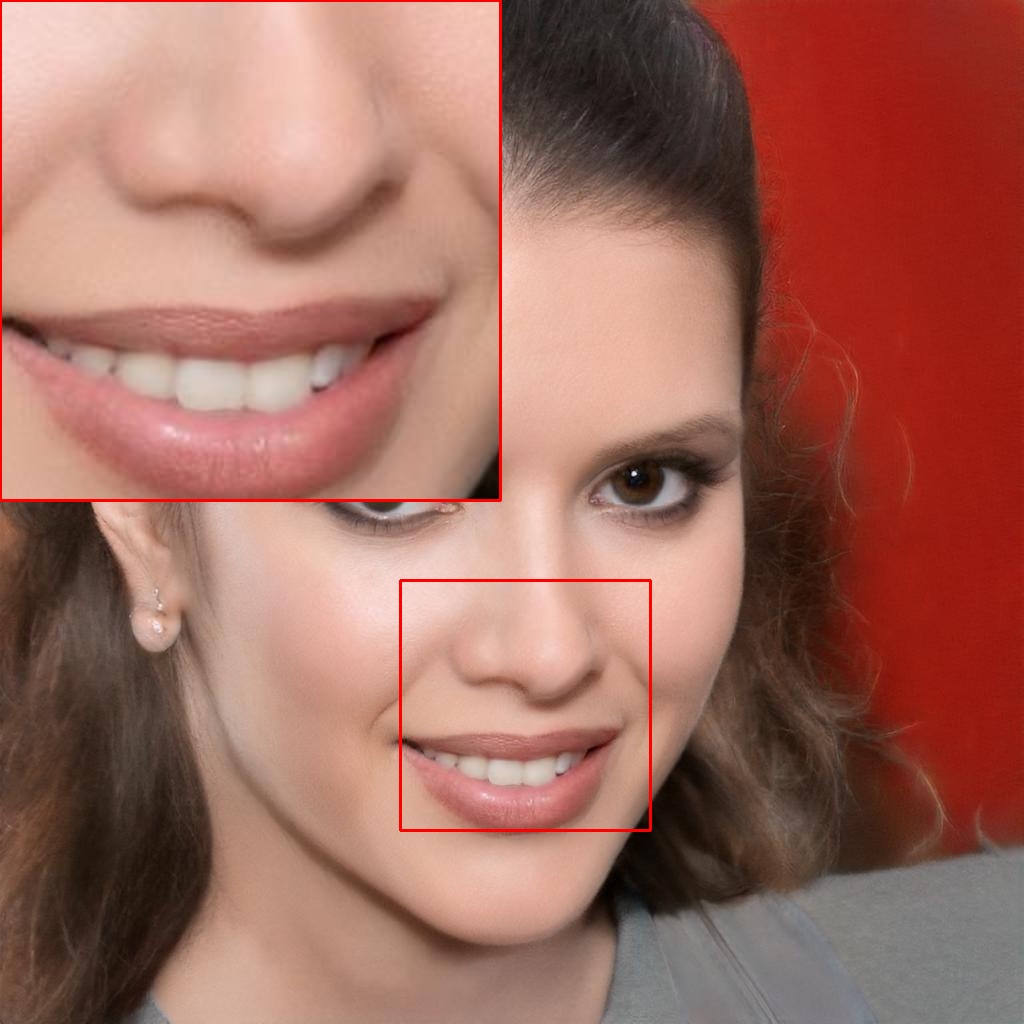} \\

			64x & \MyIm{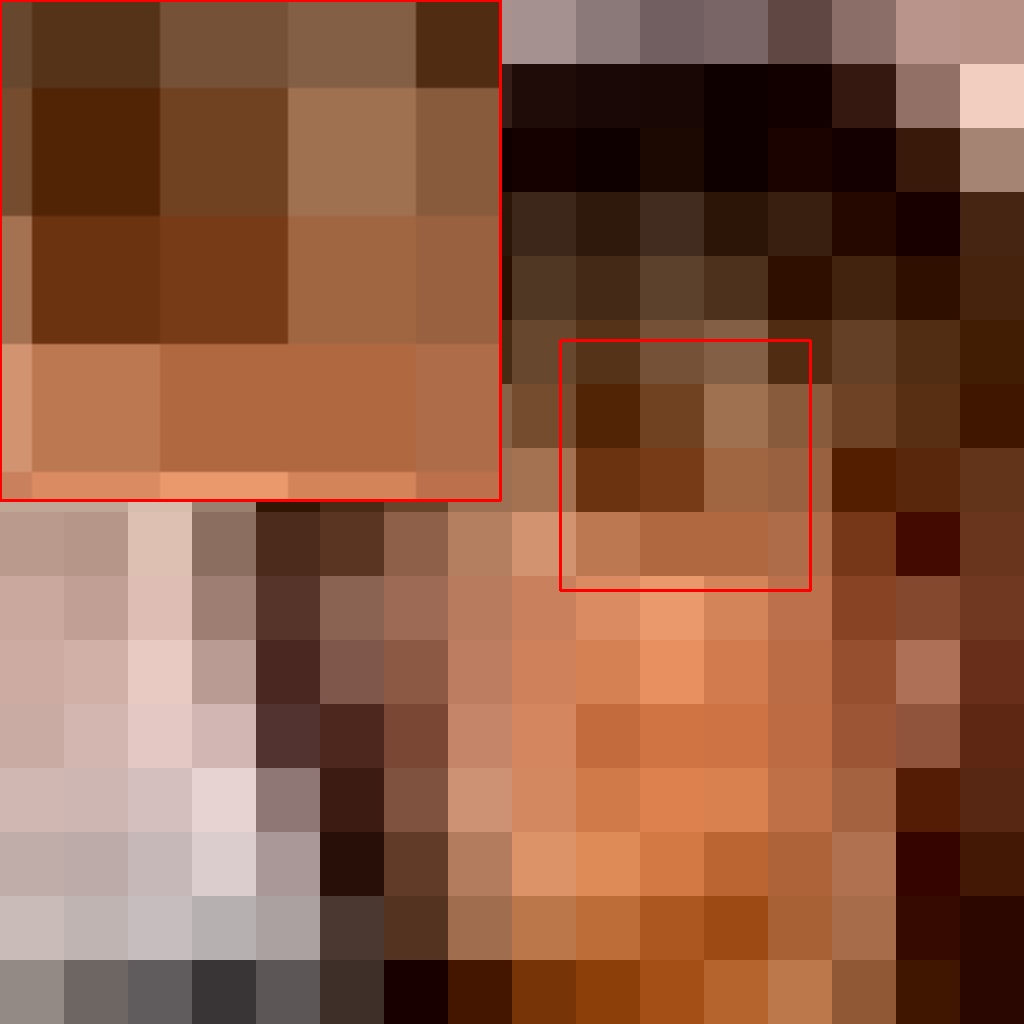}
			& \MyIm{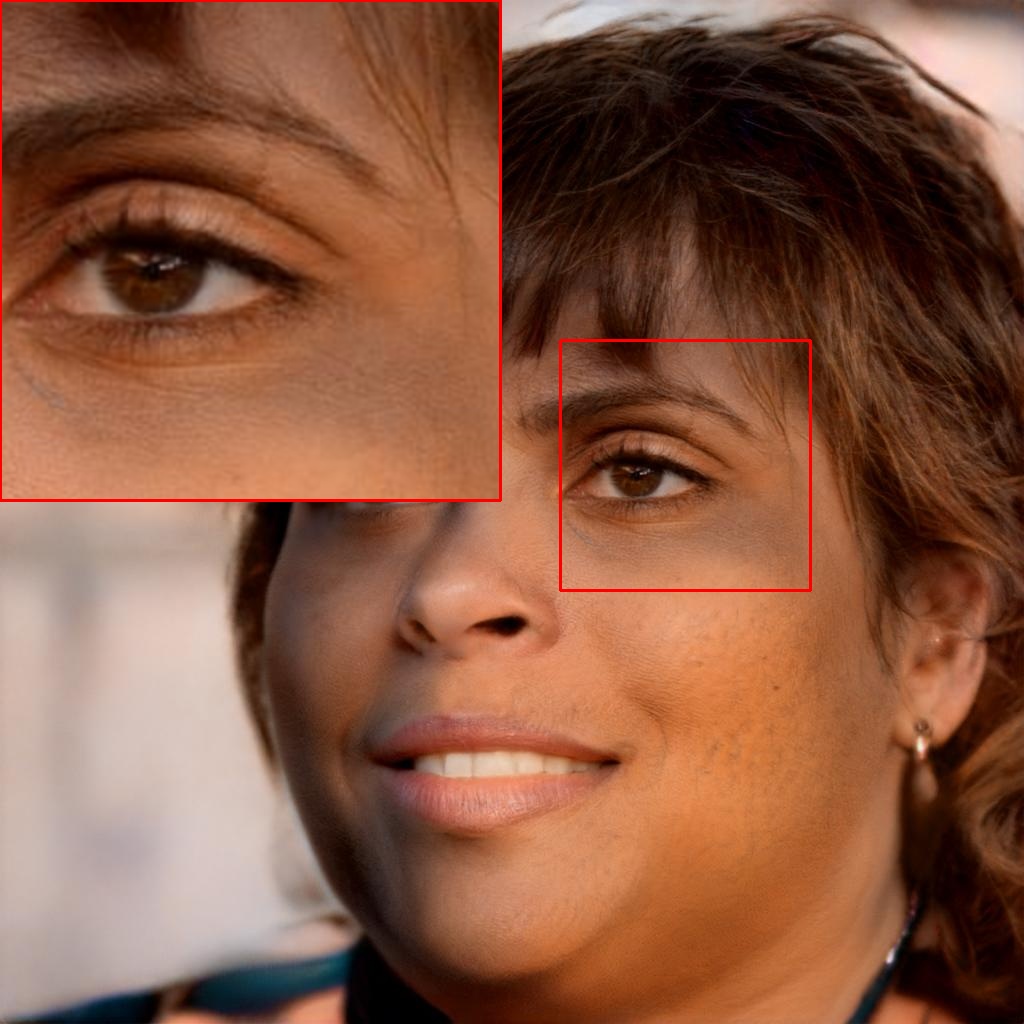}
			& \MyIm{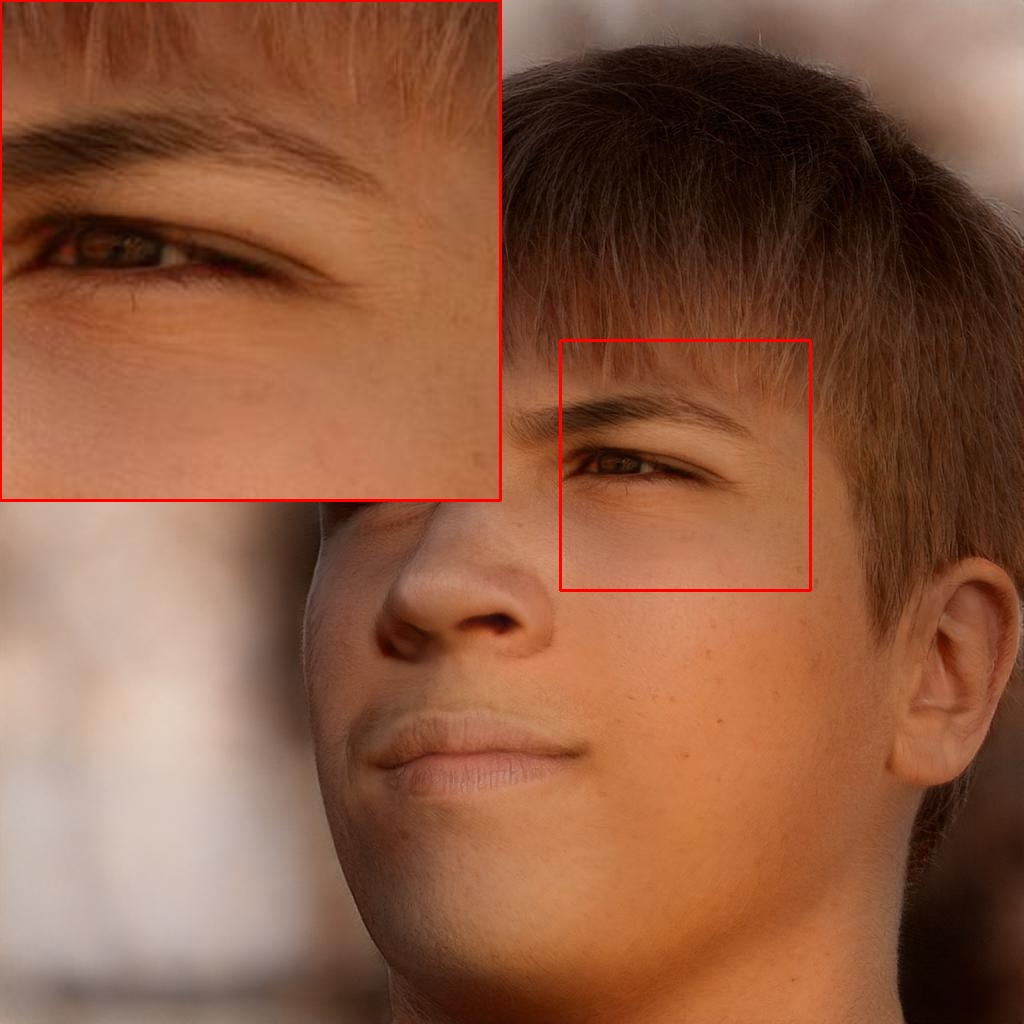}
			& \MyIm{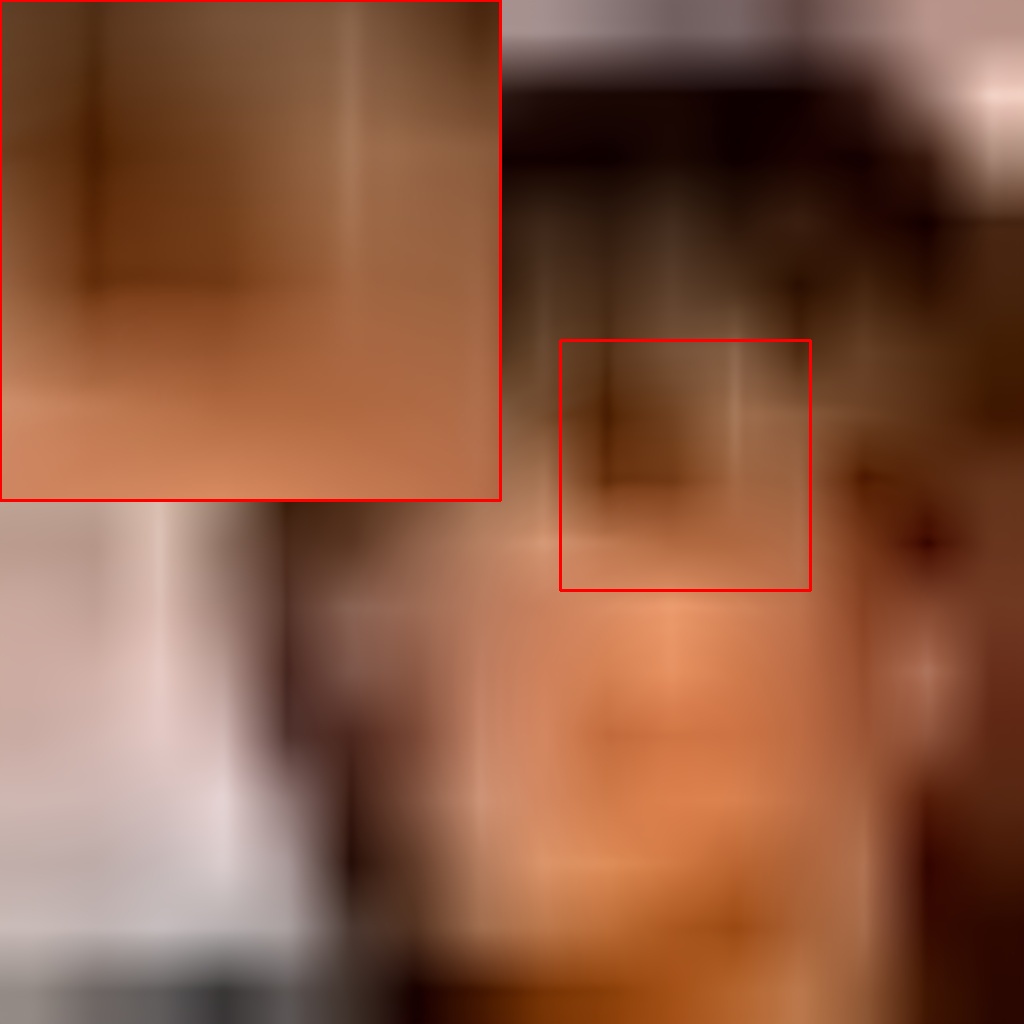}
			& \MyIm{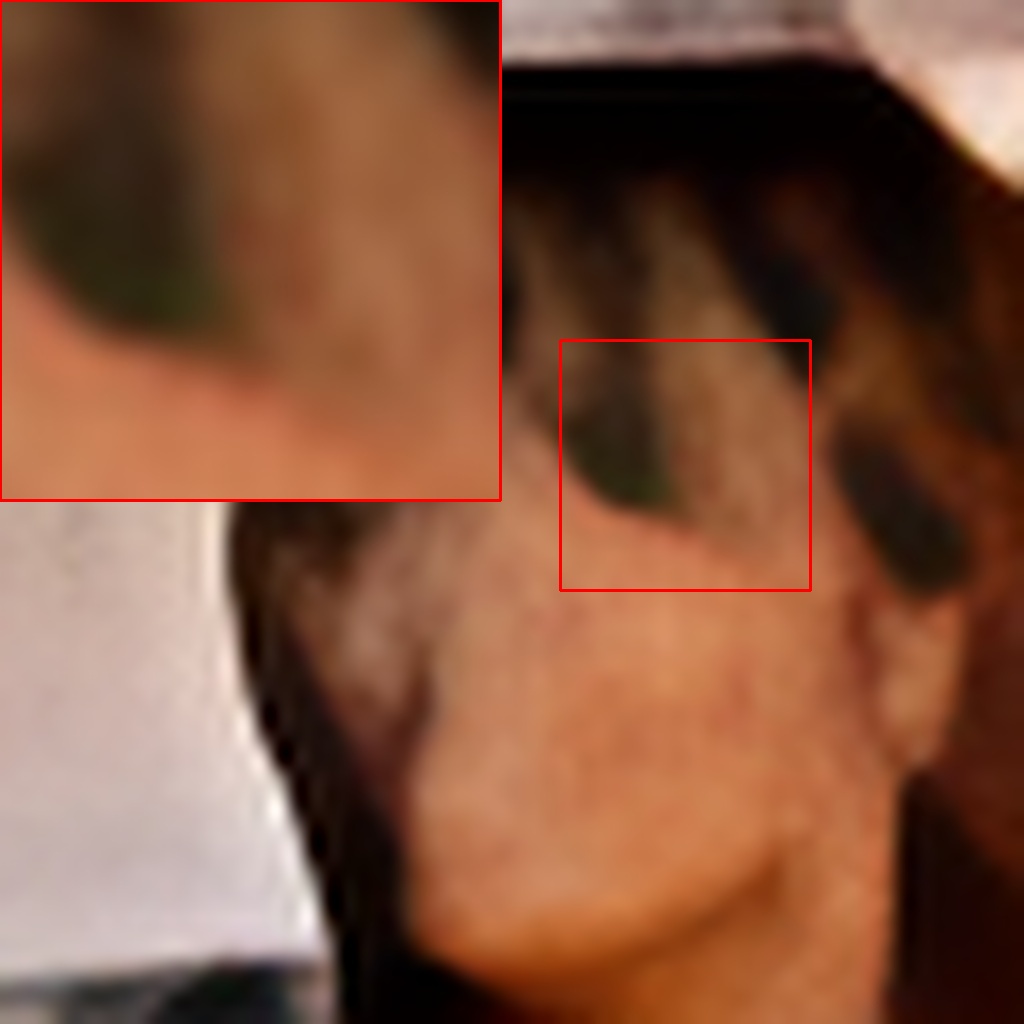}
			& \MyIm{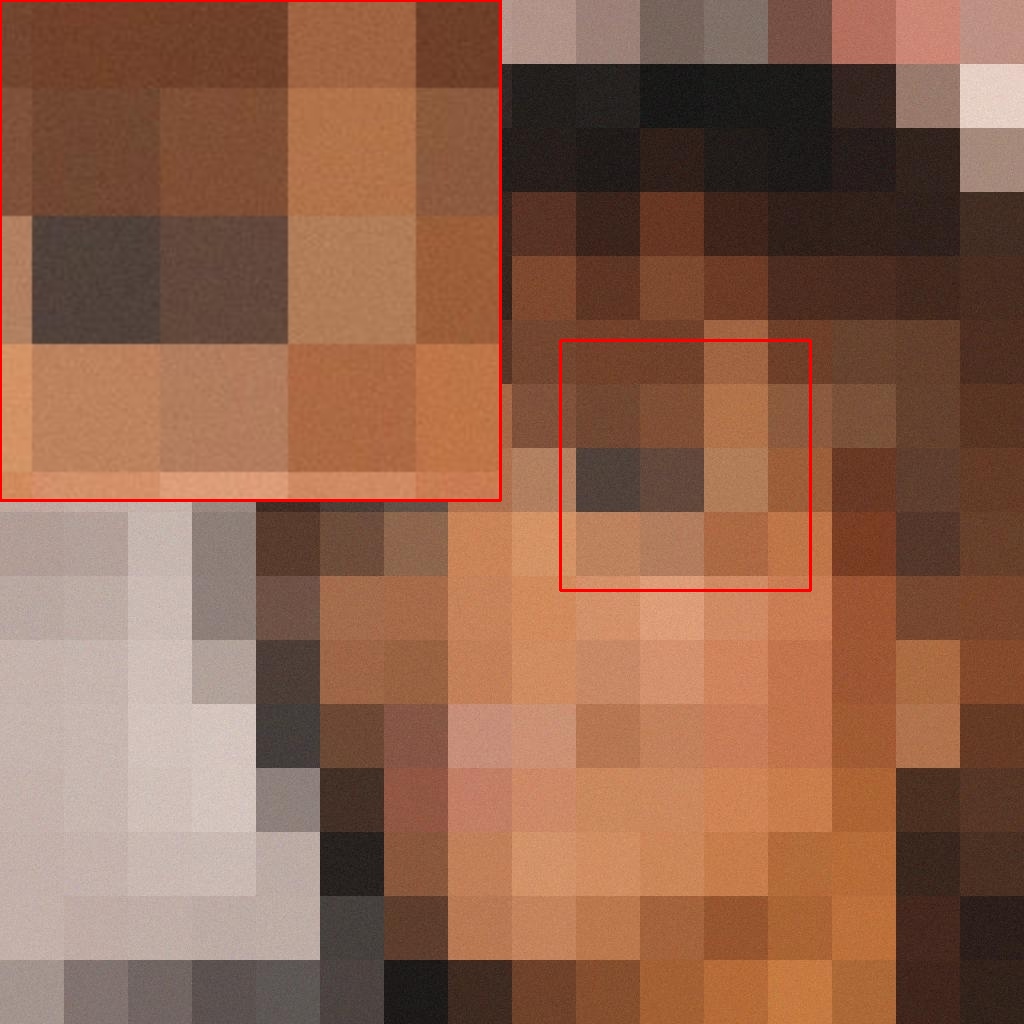}
			& \MyIm{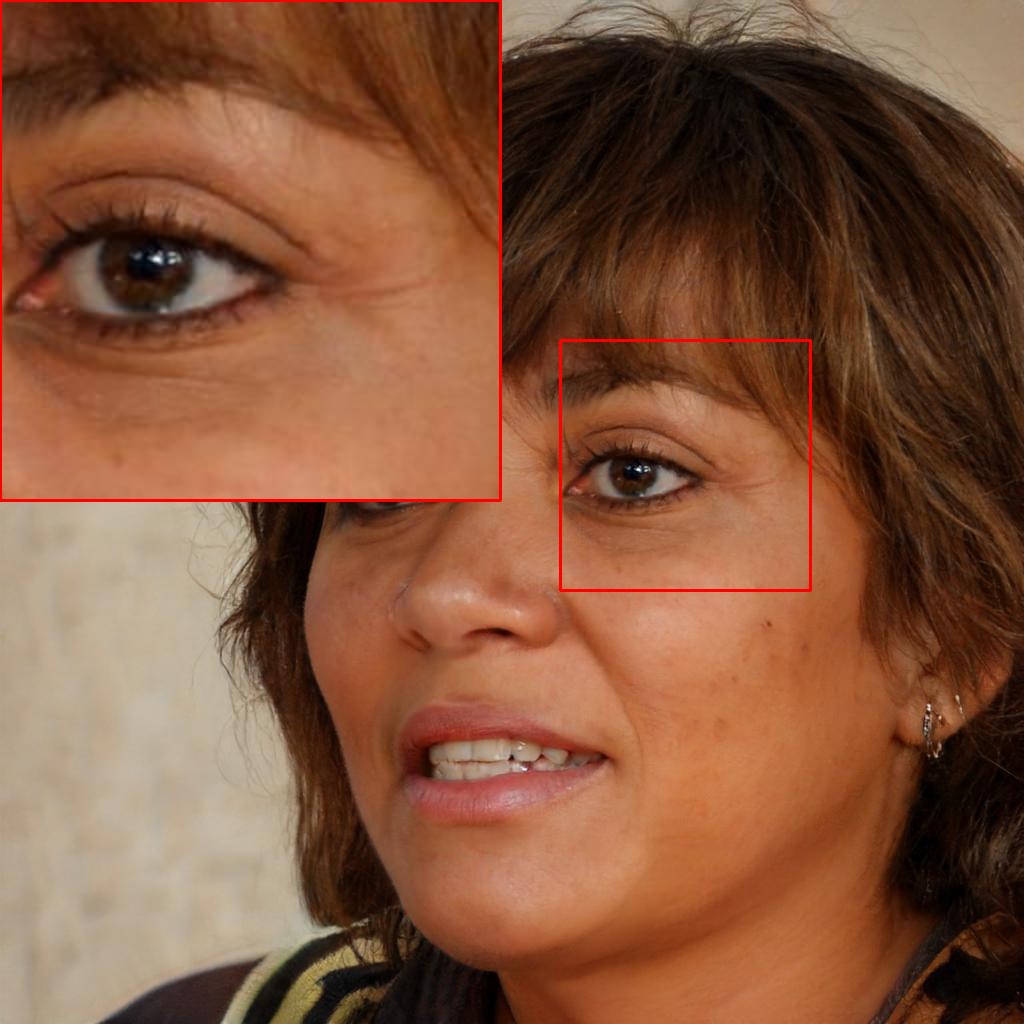}
			& \MyIm{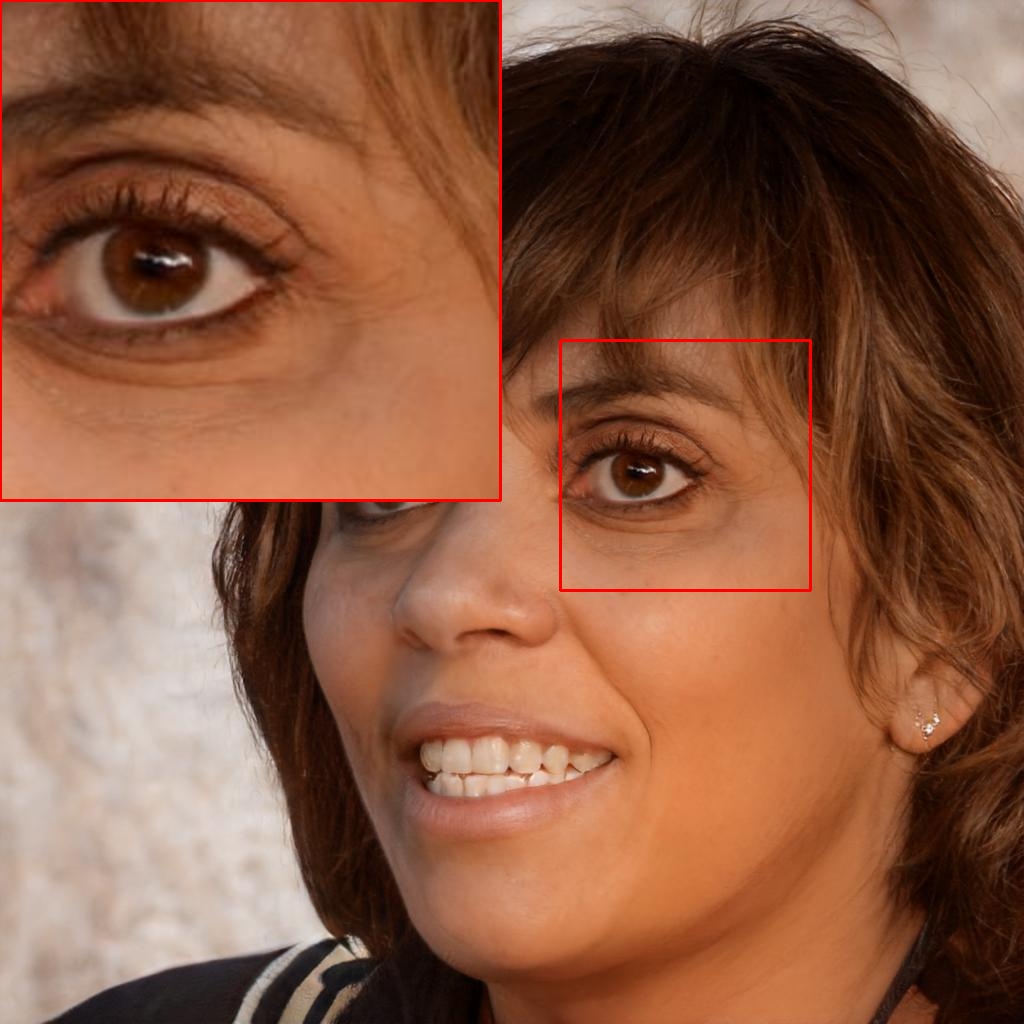} \\
			8x & \MyIm{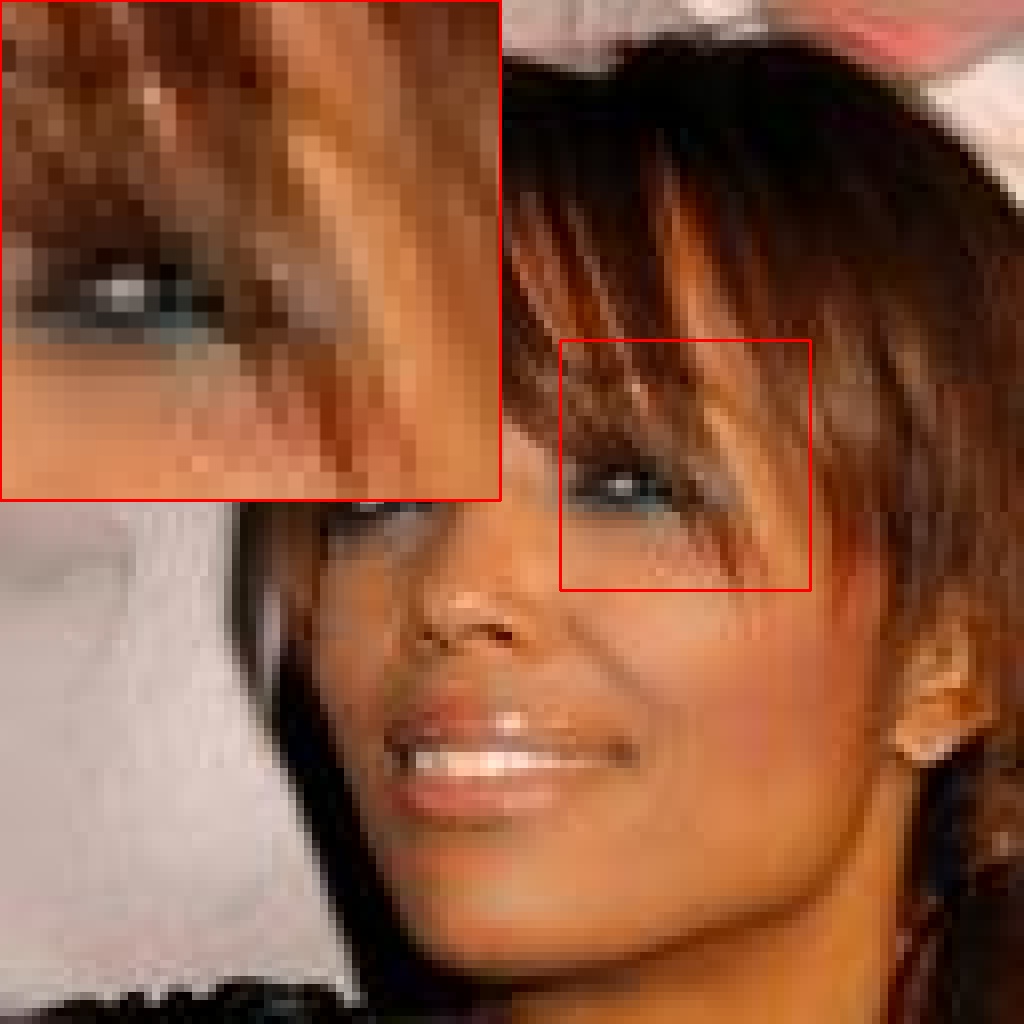}
			& \MyIm{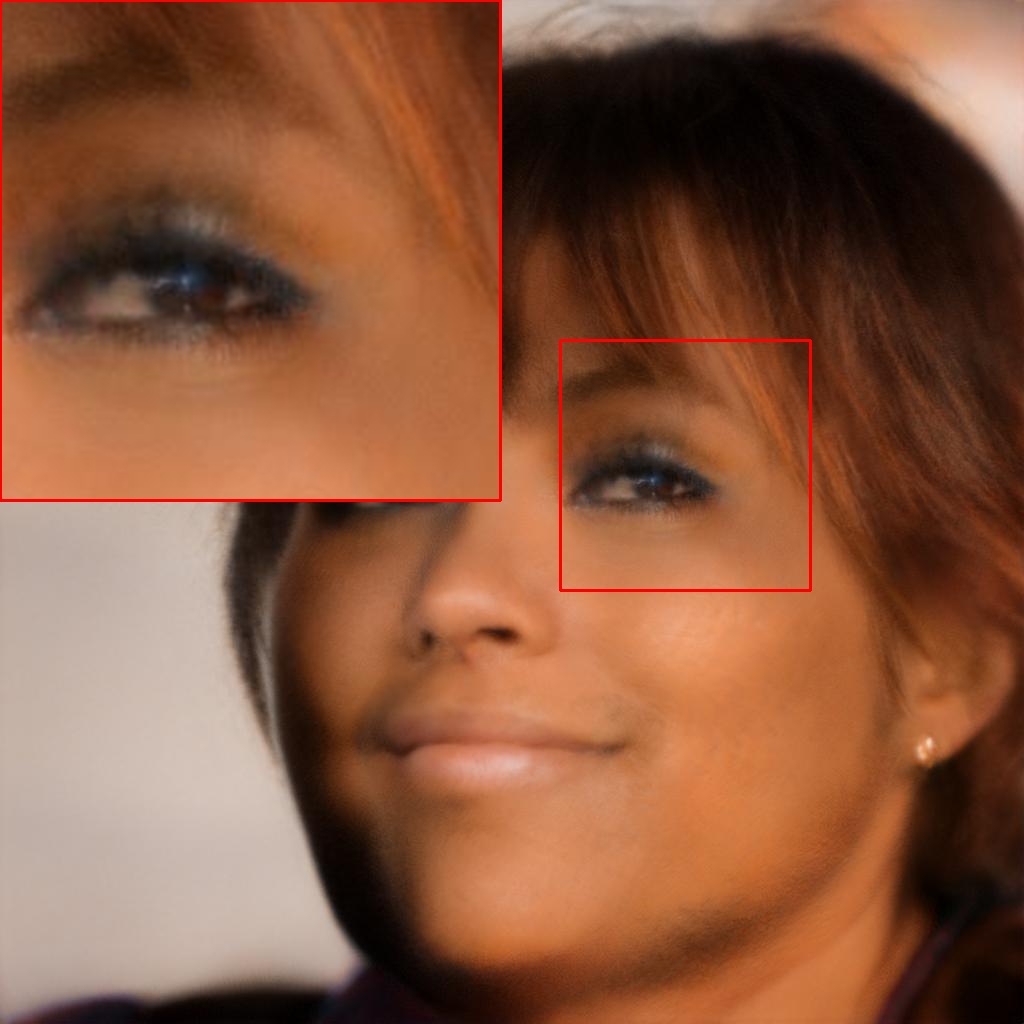}
			& \MyIm{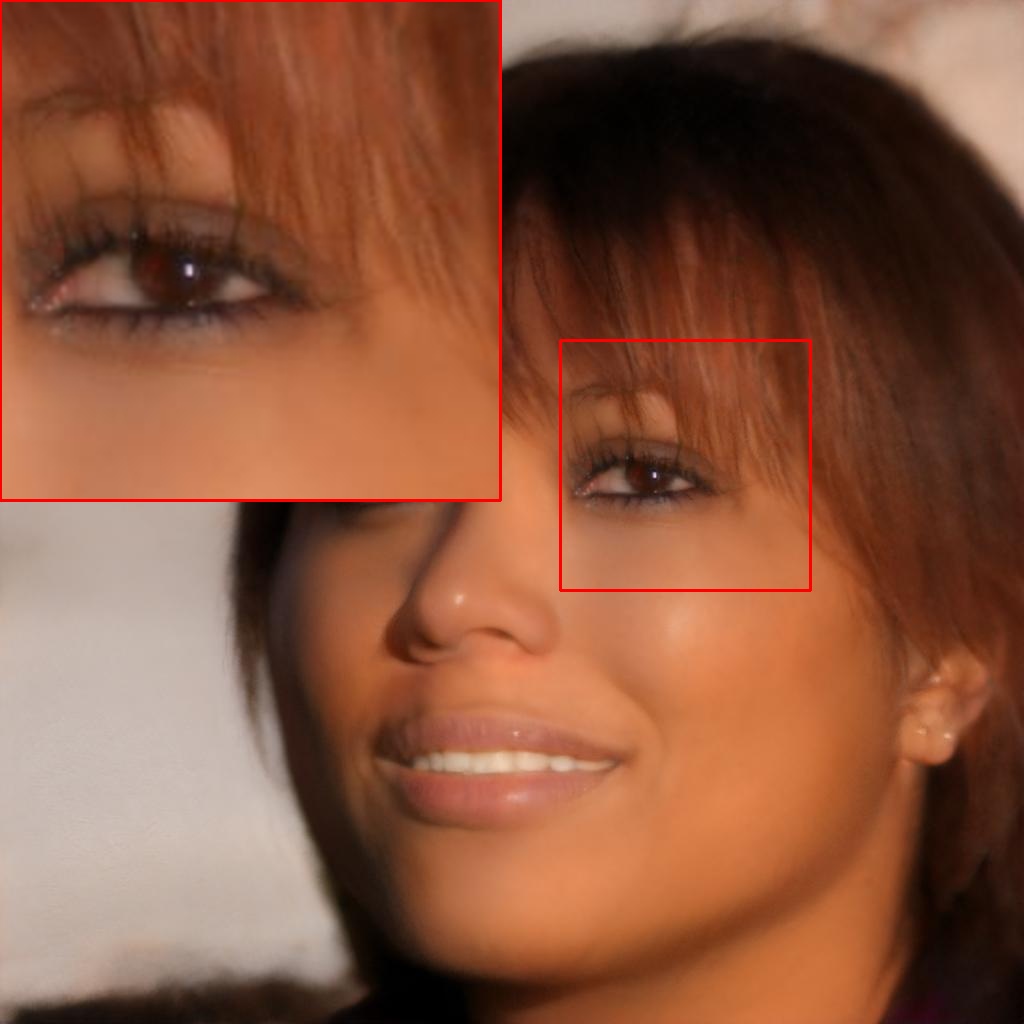}
			& \MyIm{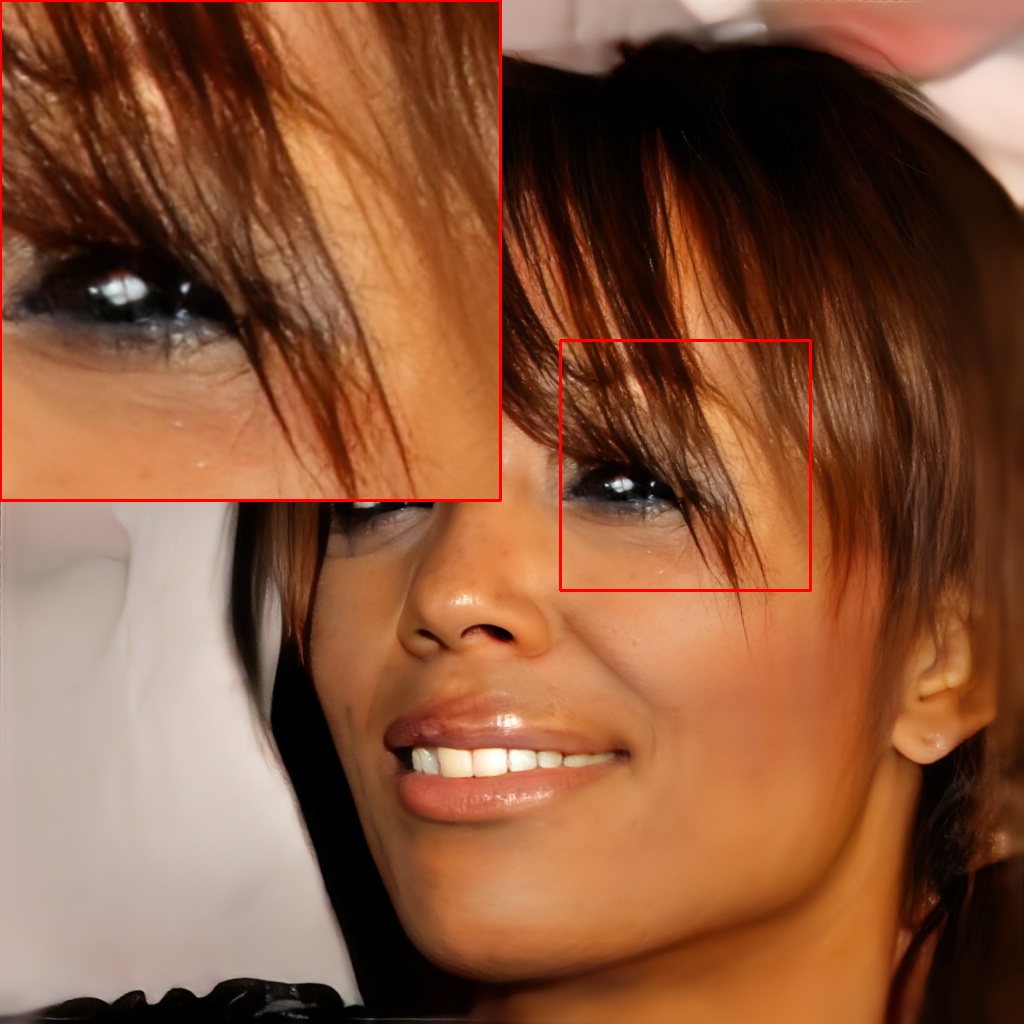}
			& \MyIm{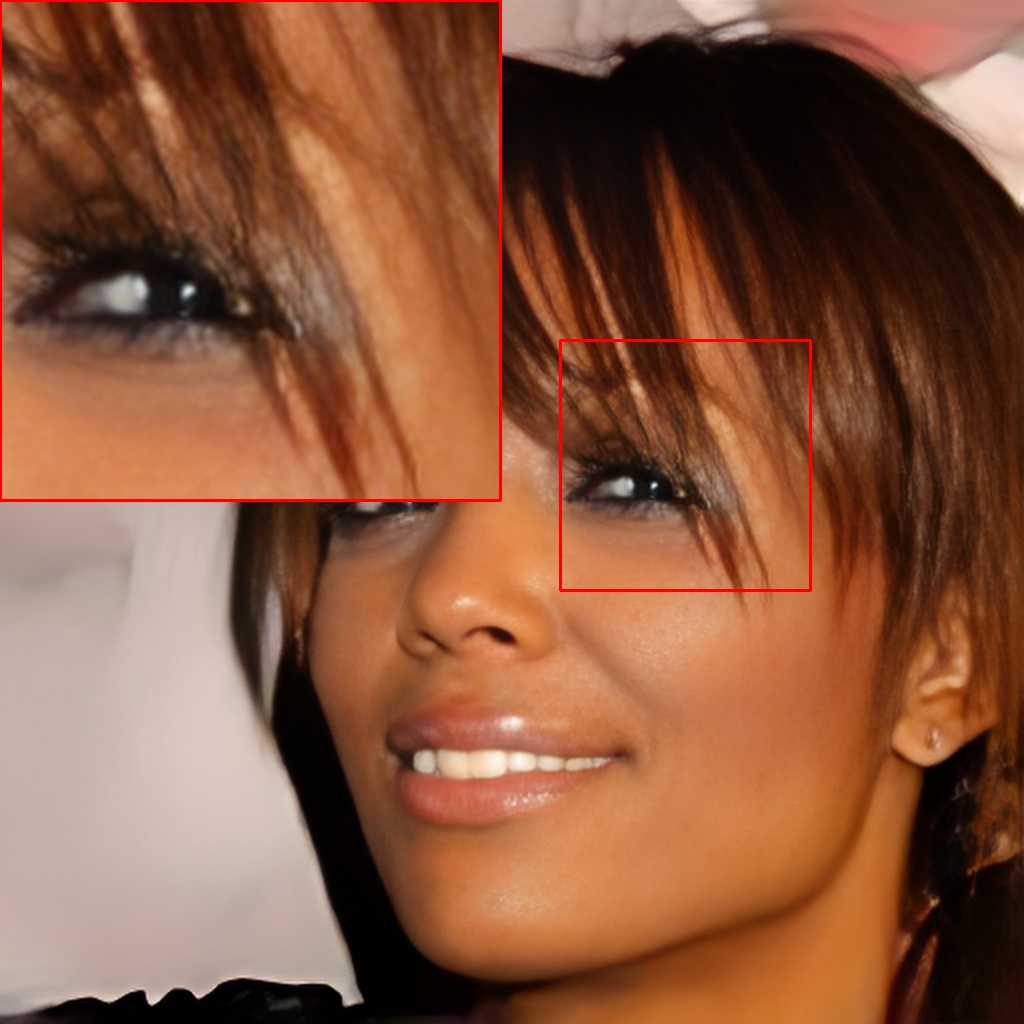}
			& \MyIm{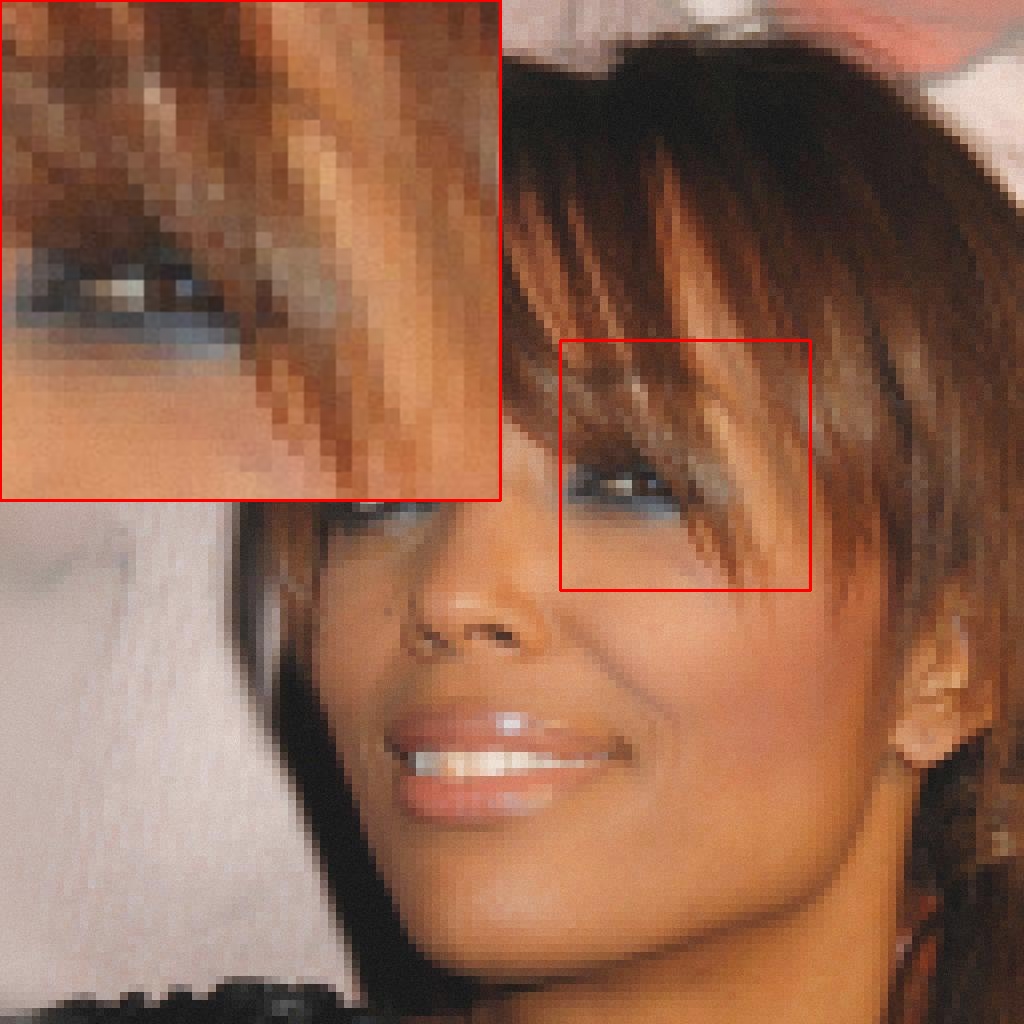}
			& \MyIm{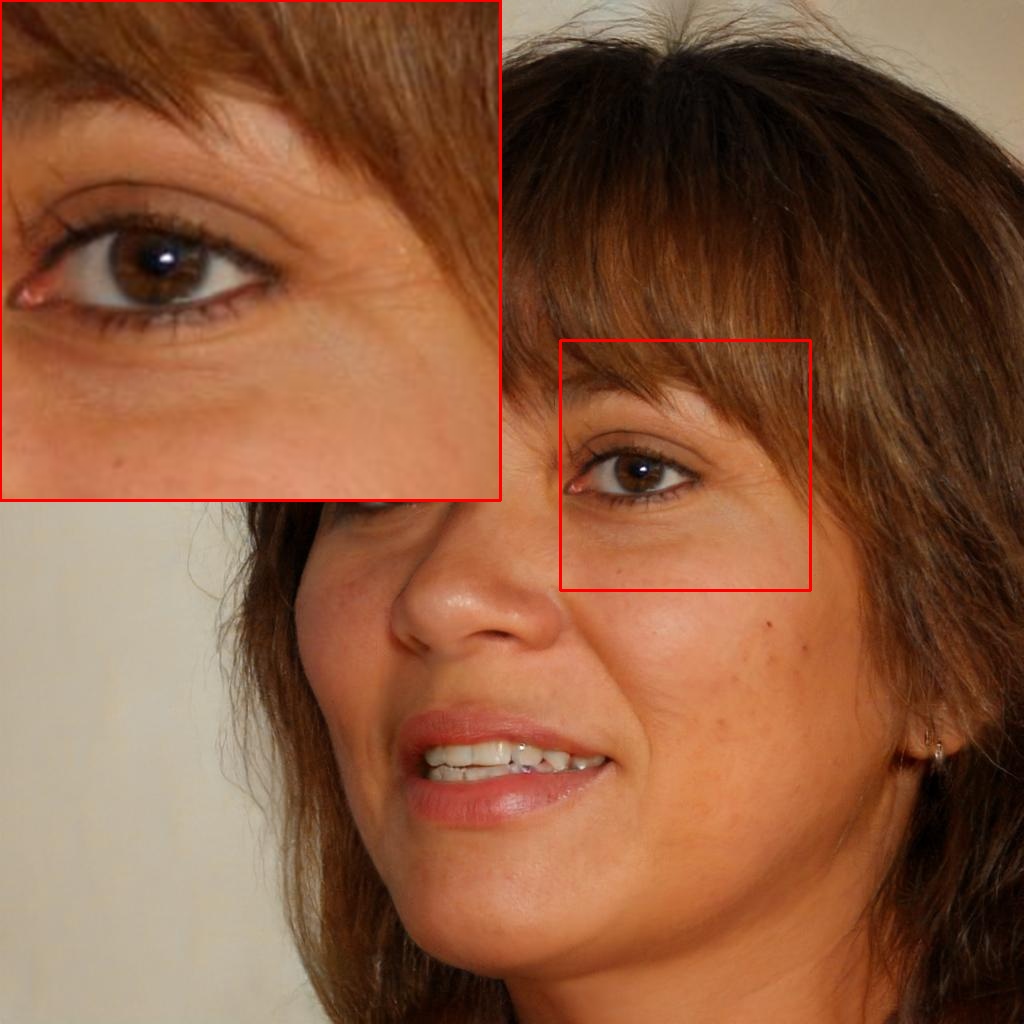}
			& \MyIm{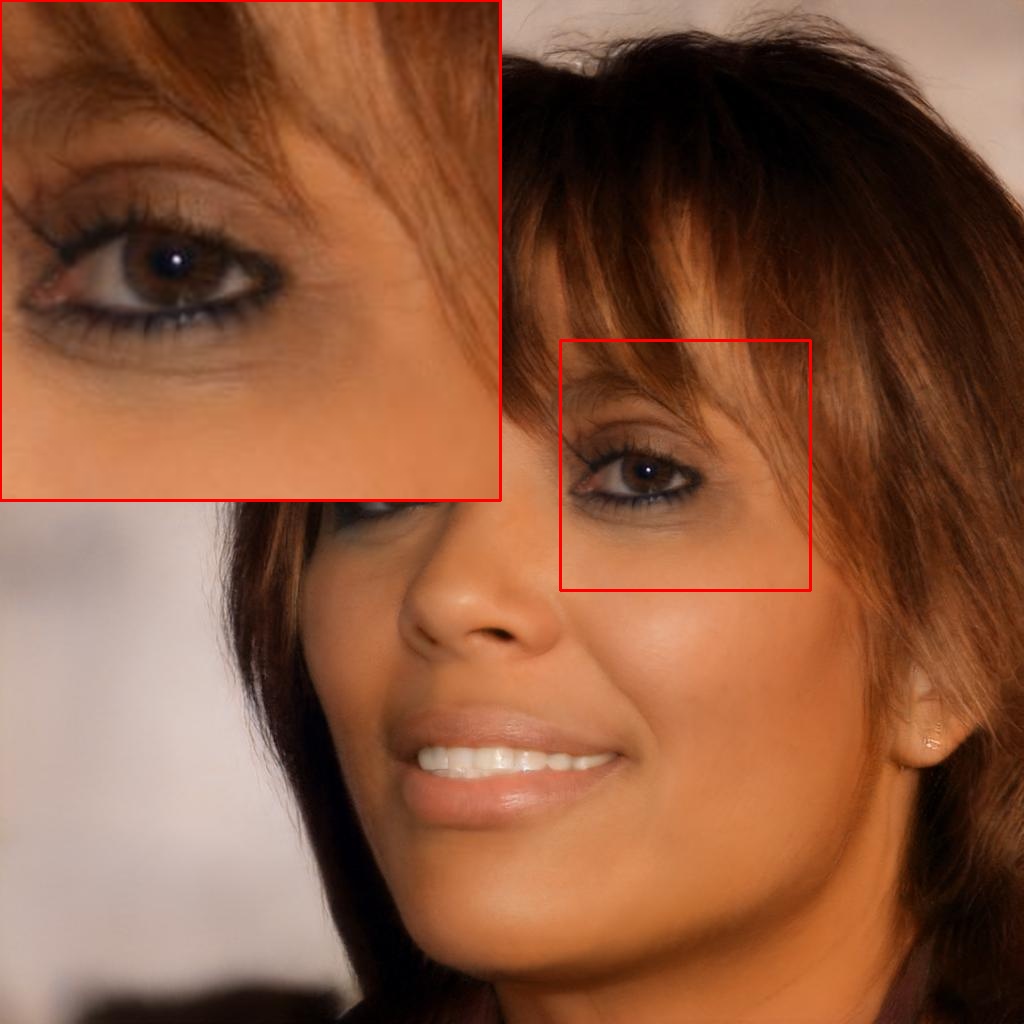} \\
		\end{tabular}
	\end{scriptsize}
	\caption{Comparison of the reconstructions of a high resolution face from CelebA. (Zoom-in for best view)}
	\label{fig:qualitative}
\end{figure*}

\subsection{Regularized Latent Search}
\label{sec:RLS}
MAP estimation is a common approach in Bayesian statistics for estimating an unknown parameter $\x$ based on observed data $\y$.
We can formulate the super-resolution problem in terms of MAP estimation~\cite{marinescu2021bayesian}. The unknown parameter we are trying to estimate is the HR image $\x$, and the observed data is the LR image $\y$. For a given LR image $\y$,
we wish to recover the HR image $\x$ as the MAP estimate of the conditional distribution $\Pg(\x|\y)$:

\begin{equation}
\label{equation mapestimation1}
\begin{aligned}
& \arg\max_{\x}\log \Pg(\x|\y)\\
& \qquad = \arg\max_{\x} [\log p(\y|\x) + \log \Pg(\x) + \log p(\y)]
\end{aligned}
\end{equation}

Since the marginal density $\log p(\y)$ is constant we drop it. We also model the likelihood $p(\y|\x)$ as a delta function $\Pdelta(\y - \D(\x))$, where $\D$ is a degradation operator that maps HR images to LR images. We can then rewrite the MAP objective as:

\begin{equation}
\label{equation mapestimation}
\arg\max_{\x} [\log \Pdelta(\y-\D(\x)) + \log \Pg(\x)]
\end{equation}
where the first term is the likelihood term and the second term is the prior which describes the manifold of real HR images.

\paragraph{Image prior} Let $\G_s$ be the synthesis network of a StyleGAN~\cite{karras2020analyzing} pre-trained on the considered image domain. $\G_s$ takes as input $\w$, produced by the mapping network, and outputs an image \ie $\x = \G_s(\w)$ (\cref{sec:stylebased}),  which is a deterministic transformation of $\w$ through a differentiable function $\G_s$.
A change of variables can be used for a non-invertible mapping~\cite{cvitkovic2019minimal}, then the probability density function of $\x$ can be obtained from the probability density function of $\w$:
$$p_G(\x) = \Pw(\w) \left| \frac{\partial \w}{\partial \x} \right|$$
where $\w = \G_s^{-1}(\x)$ is the inverse transformation of $\G_s$, and $\frac{\partial \w}{\partial \x}$ is the derivative of $\w$ with respect to $\x$.\\

Now, we can express the image prior with respect to the latent variables $\w$, which allows us to work with the more tractable latent space of the StyleGAN network, rather than the high-dimensional space of the high-resolution images:

\begin{equation}
\label{equation changeofvarivale}
\log \Pg(\G_s(\w)) = \log \Pw(\w) + \log |\det J_{\G_s} (\w)|. 
\end{equation}

$J_{\G_s} (\w)$ is the Jacobian matrix of the mapping $\G_s$ evaluated at $\w$ that describes how small changes in the input $\w$ result in changes in the output $\x$.

In StyleGAN2, the authors introduced a new regularization term to encourage smoothness and disentanglement of the latent space. This regularization term is called the "path length regularization," and it is based on the notion of a "path" in the latent space.
In particular, given two latent vectors $\w_1$ and $\w_2$, we can define a "path" in the latent space as a function $\func{w}(s)$ that smoothly interpolates between $\w_1$ and $\w_2$ as $s$ varies from $0$ to $1$. The authors of StyleGAN2 then introduce a penalty term on the length of this path
and claim that this regularization term implies that the Jacobian determinant of the network $\G_s$ is approximately constant for all $\w$.
Based on this property of StyleGAN2 the Jacobian determinant term in the above equation can be dropped and the image prior can be expressed directly by the image prior $\Pw(\w)$, which is defined on $\w \in \wspace^+$ by:
\begin{equation}
\label{equation imageprior}
\log \Pw(\w) =  \lambda_w \prior_{w} + \lambda_g \prior_{gaussian} + \lambda_c \prior_{cross} 
\end{equation}
where:
\begin{itemize}[leftmargin=6mm, label={--}] 
	\item $\prior_{w}$ is a prior that keeps $\w$ in the area of high density in $\wspace^+$: $\prior_{w} = \frac{1}{\layer}\sum_{i=1}^{\layer} {\log \Pf(\w_i)}$
	, where $\Pf(\w)$ is estimated by a normalizing flow model $\F$ explained in \cref{sec:normalizing_flow}.
	\item $\prior_{gaussian}$ is a gaussianization prior. Using the normalizing flow model $\F$, a gaussianized latent vector $\w_n$ can be obtained and a $L2$ regularization applied on it to keep it near the surface of the hypersphere: $\prior_{gaussian} = -\frac{1}{\layer}\sum_{i=1}^{\layer}(||\F(\w)||_2-\sqrt{\dim})^2$
	\item $\prior_{cross}$ is a pairwise euclidean distance prior on $\w  = [\w_1,\dots, \w_\layer]\in \wspace^+$ that ensures $\w \in \wspace^+$ remains close to the trained manifold in $\wspace$: \quad $\prior_{cross} = -\sum_{i=1}^{\layer-1}\sum_{j=i+1}^{\layer}{||\w_i-\w_j||^2_2}$
	
\end{itemize}
\paragraph{Optimization}
In the likelihood term in \cref{equation mapestimation}, we assume that the noise follows a Laplace distribution \ie $\deltaa\sim Laplace(0, \lambda_l I)$, then the log-density of $\deltaa$ becomes: $\log \Pdelta (\deltaa)=-{||\deltaa||_1}-C$
for a constant $C$. With the parameters of $\G_s$ denoted by $\thetaa$, the problem in \cref{equation mapestimation} can be recast as an optimization over $\w$, leading to the final objective function:
\begin{equation}
\label{equation loss}
\begin{aligned}
\hat{\w} = \arg\min_{\w} ||\y - \D(\G_s(\w, \thetaa))||_1 - \log \Pw(\w)
\end{aligned}
\end{equation}

\setlength{\tabcolsep}{6pt}
\begin{table*}
	\centering	
	\begin{tabular}{@{}l||l|llll|lll@{}}
		\toprule
		Scale & Method & FID$\downarrow$ & KID$^{({\times10^3})}$$\downarrow$ & NIQE$\downarrow$ & ID$\uparrow$ & LPIPS$\downarrow$ &PSNR$\uparrow$ & MSSIM$\uparrow$\\
  
		\hline
		\multirow{7}{*}{64x} & PULSE~\cite{menon2020pulse} & \red{42.9331} & \red{30.2643} & 5.0957  & 0.6709 & 0.5197 & 19.5775 & 0.5430 \\
            & BRGM~\cite{marinescu2021bayesian} & 58.3559 & 47.5288 & 4.3817  & 0.6426 & 0.5412 & 18.8989 & 0.5300 \\
		& GPEN~\cite{yang2021gan} & 474.275 & 693.3078 & 14.6695 & 0.6623 & 0.6704 & \blue{20.2558} & \red{0.6014} \\
		& GFPGAN~\cite{wang2021towards} & 197.3977 & 181.3333 & 12.9577 & 0.7151 & 0.6431 & 19.6302 & \blue{0.6047}\\
		& DDRM~\cite{kawardenoising} & 391.2105 &538.9393 & 8.0546 & 0.6938 & 0.6924 & 18.6866 & 0.5296\\
		& RLS & 47.8888 & 30.8534 & 4.1032 & \red{0.7210} & \red{0.5037} & 17.9680 & 0.5183 \\
		& RLS$^{+}$ & \blue{36.3110} & \blue{20.9327} & \red{4.1348} & \blue{0.7335} & \blue{0.4749} & \red{19.7064} & 0.5671\\[0.2cm]
		\hline
		
		\multirow{8}{*}{8x} & Bicubic & 86.9839 & 104.7387 & 9.9253 & 0.8100 & 0.5346 & \blue{28.0067} & \blue{0.8568} \\
		& PULSE~\cite{menon2020pulse} & 34.5038 & 21.6472 & 5.9383 & 0.7511 & 0.4618 & 23.4985 &  0.7090\\
		& BRGM~\cite{marinescu2021bayesian} & 38.0316  & 27.2339 & 7.6593  & 0.7634 & 0.4998 & 21.9977 & 0.6817 \\
		& GPEN~\cite{yang2021gan} & \red{27.9026} & 19.4761 & 4.9814 & \red{0.8746} & \blue{0.3217} & 26.3693 & 0.8472 \\
		& GFPGAN~\cite{wang2021towards} & 28.2971 & \red{18.5216} & 6.0168 & \blue{0.8775} & \red{0.3323} & \red{27.1016} & 0.8512 \\
		& DDRM~\cite{kawardenoising} & 30.1999 & 25.1746 & 7.1486 & 0.8358 & 0.5386 & 23.4707 & \red{0.8566} \\
		& RLS & 45.8778 & 29.6858 & \blue{4.2293}  & 0.7539 & 0.4738 & 18.4833 & 0.5677 \\
		& RLS$^{+}$ & \blue{27.6691} & \blue{13.0044} & \red{4.7241} & 0.8152 & 0.3925 & 24.2802 & 0.7577 \\
		
		\bottomrule
	\end{tabular}
	\caption{Quantitative comparison on CelebA for 64x and 8x super-resolution. (The \blue{best} and the \red{second-best} are emphasized by blue and red respectively.)}
	\label{tab:quantitative}
\end{table*}
\subsection{Boosting Reconstruction Fidelity}
\label{sec:RLS+}

\paragraph{Realism-Fidelity Trade-off}
The impact of regularizer parameters $\lambda_w$ and $\lambda_g$ on the reconstruction quality is depicted in \cref{figure tradeoff}. The experiment considers two variants, where only the corresponding prior is retained, and the others are disconnected. For instance, when examining the effect of $\lambda_w$ (i.e., as $\lambda_w$ increases), $\lambda_g$ and $\lambda_c$ are set to zero. The variant in which $\lambda_g$ is increased showcases the rationale behind $P_{gaussian}$, which maintains $w$ around the surface of a sphere, rather than precisely on it. This variant demonstrates that as $\lambda_g$ approaches infinity, the LR consistency (reconstruction fidelity) decreases, indicating that the exact surface of the sphere does not encompass the entire image distribution learned by StyleGAN.

It is also worth noting that \cref{figure tradeoff} shows that retaining only the prior $\prior_w$ (for some parameters) can result in a reconstructed image that is closer to the ground truth, indicating the consequences of searching beyond regions near the hypersphere. Conversely, when $\lambda_w$ is large, the reconstruction quality decreases. Thus, we can employ $P_{gaussian}$ to strengthen the prior and ensure that the latent code resides in healthy regions.

As depicted in \cref{figure tradeoff}, both variants of RLS yield excellent reconstruction outcomes for specific parameter values. This allows us to strike a balance between realism (with respect to the dataset StyleGAN was trained on) and reconstruction fidelity. However, this realism-fidelity trade-off can be further enhanced by RLS$^+$, which is explained in the following.

\begin{figure*}[!h]
 \setlength{\tabcolsep}{0pt}
	\centering
	\begin{scriptsize}
		\begin{tabular}{CCCCCCCC}
			LR & 
			w/o anchor & w/o noise & w/o g & w/o w & w/o $\ell_1$-ball & RLS$^+$ & GT\\
			\MyIm{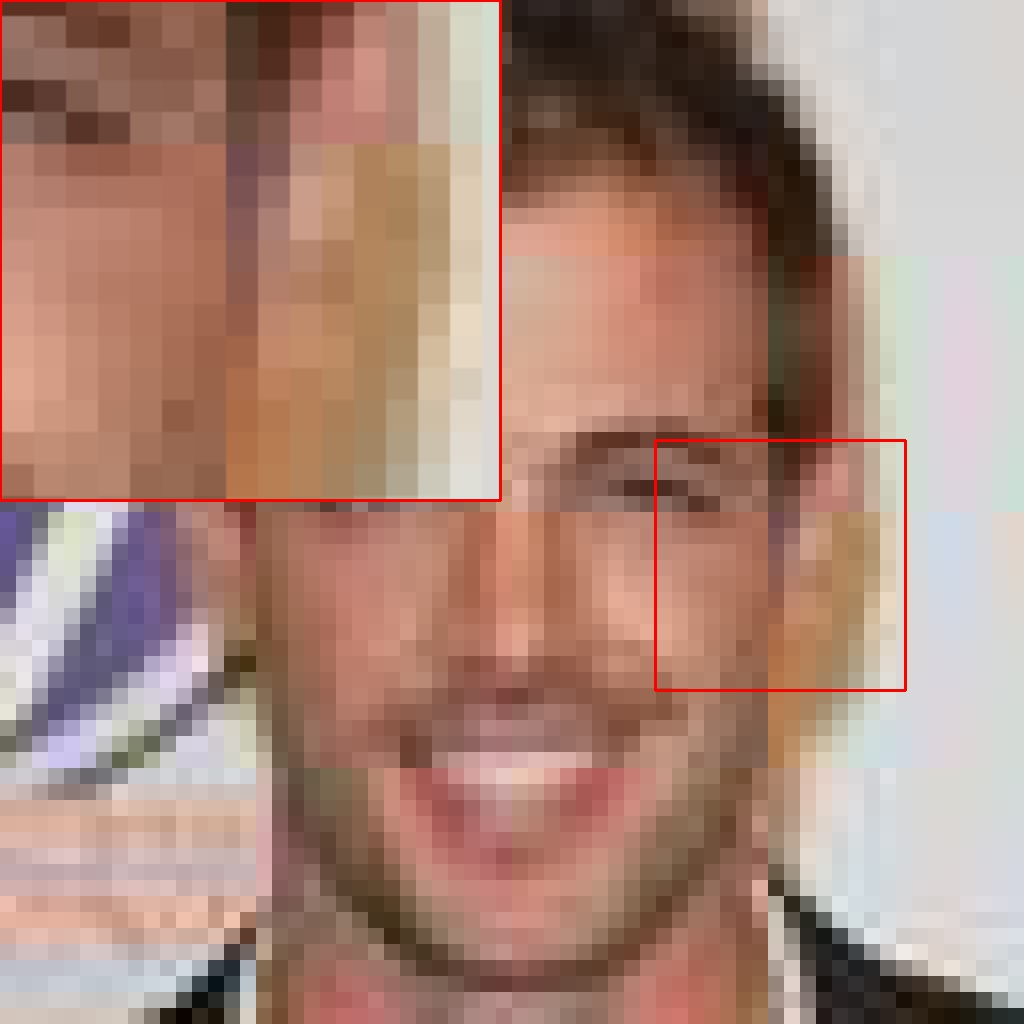} &
			\MyIm{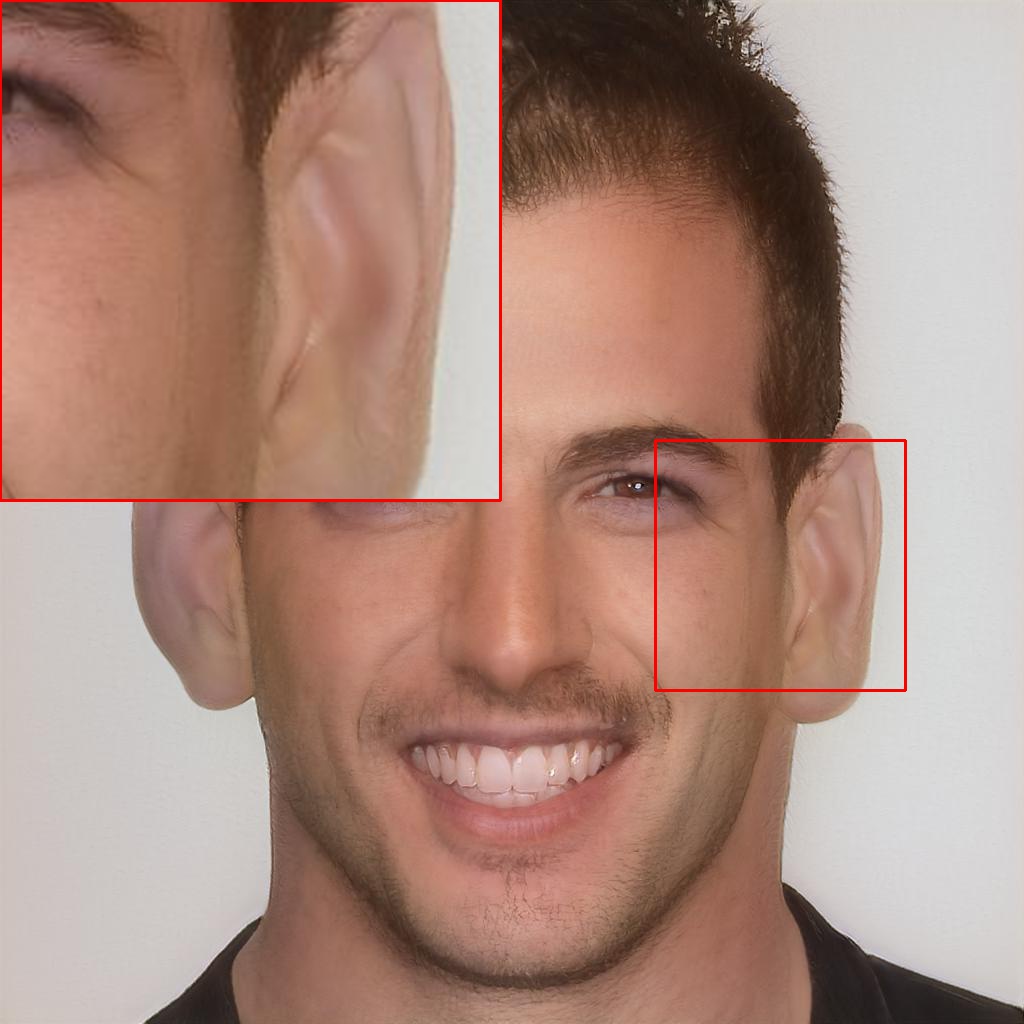} 
			& \MyIm{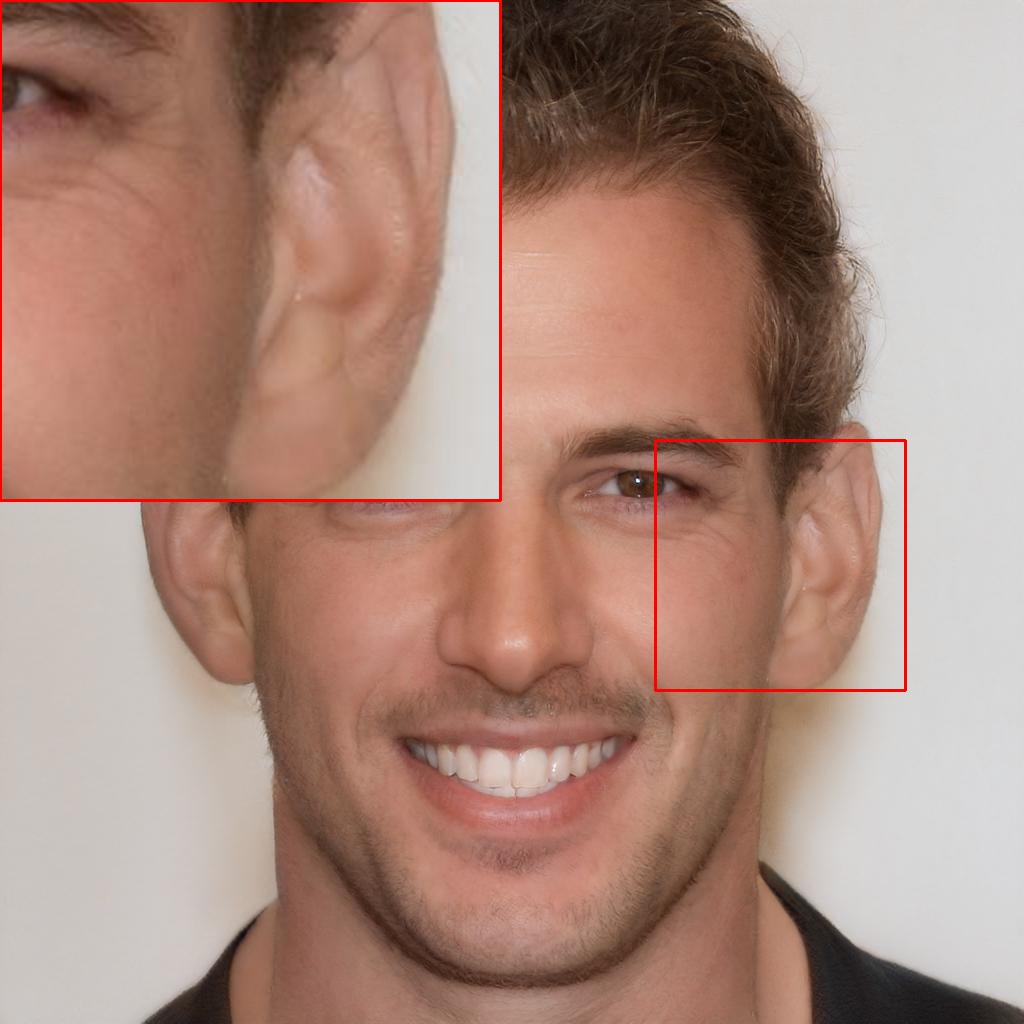} 
			& \MyIm{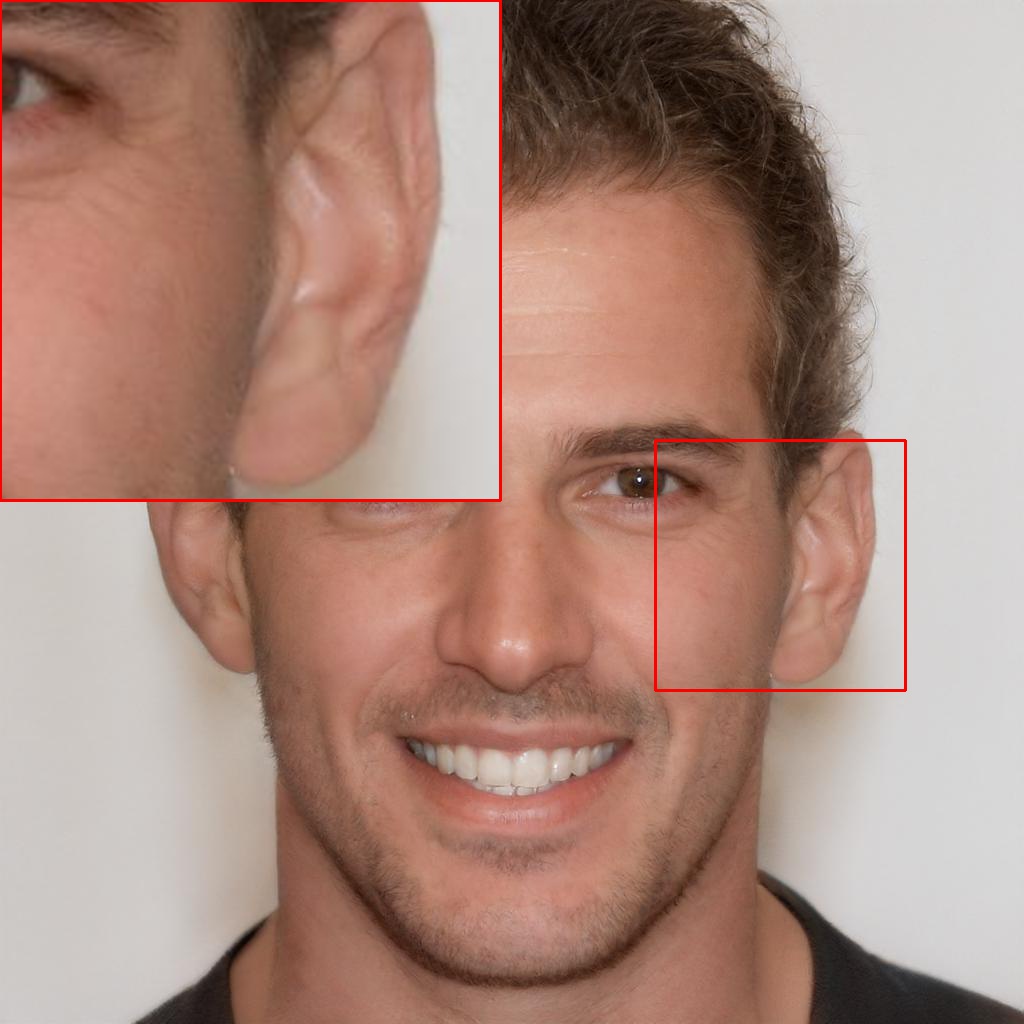}
			& \MyIm{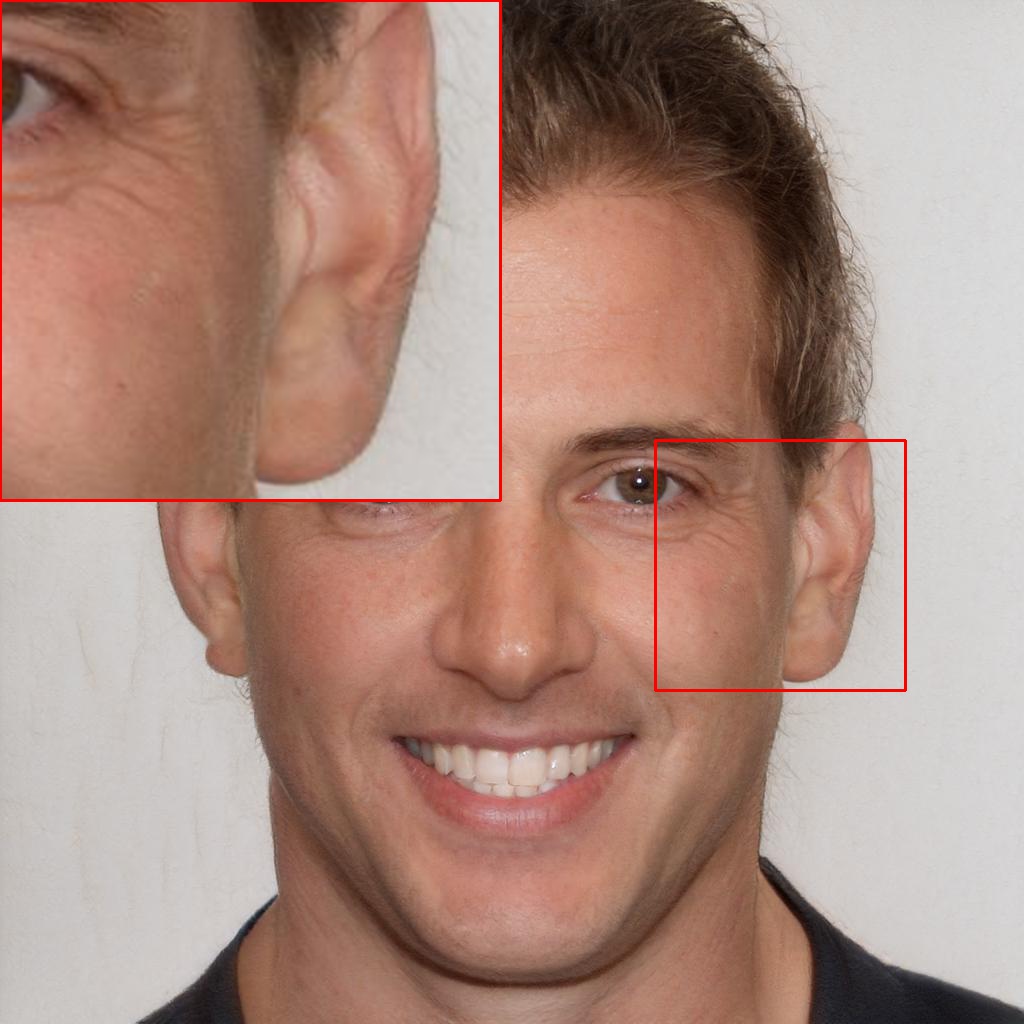}
			& \MyIm{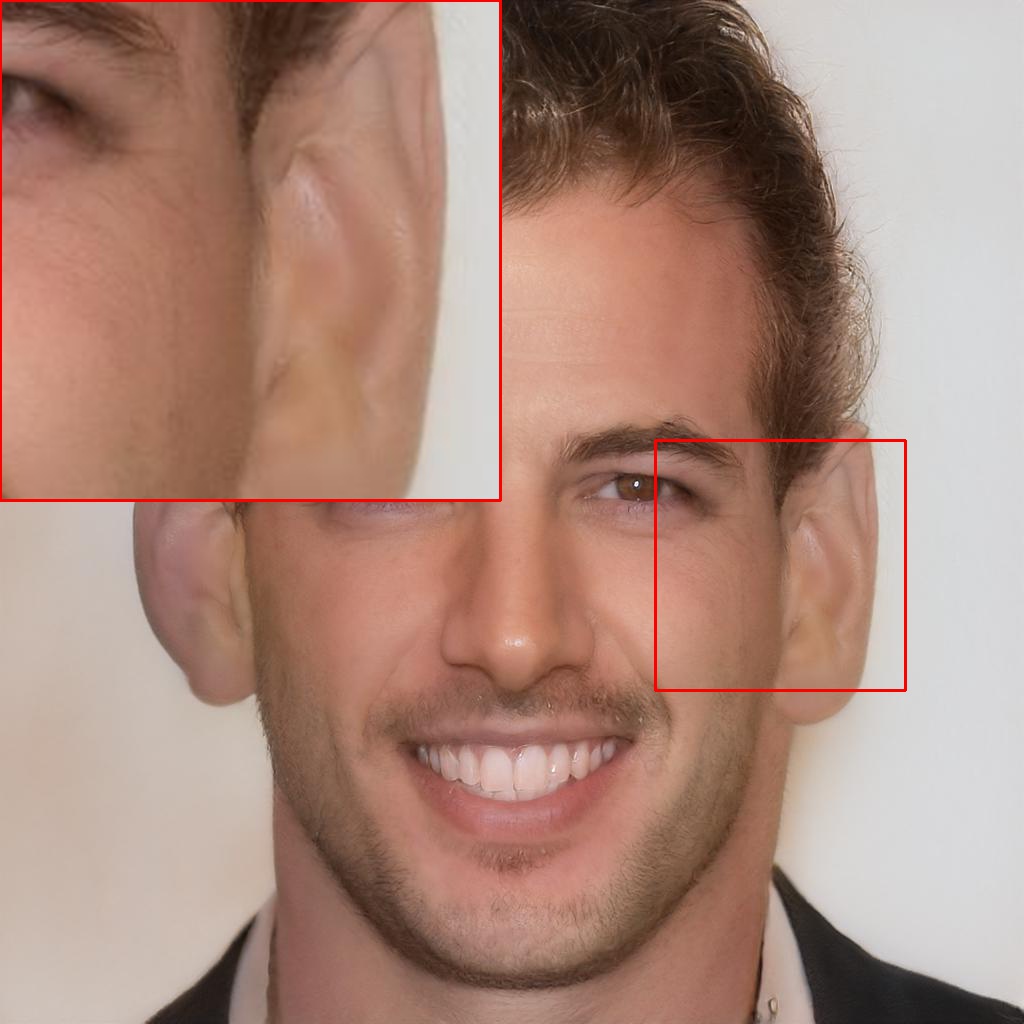}
			& \MyIm{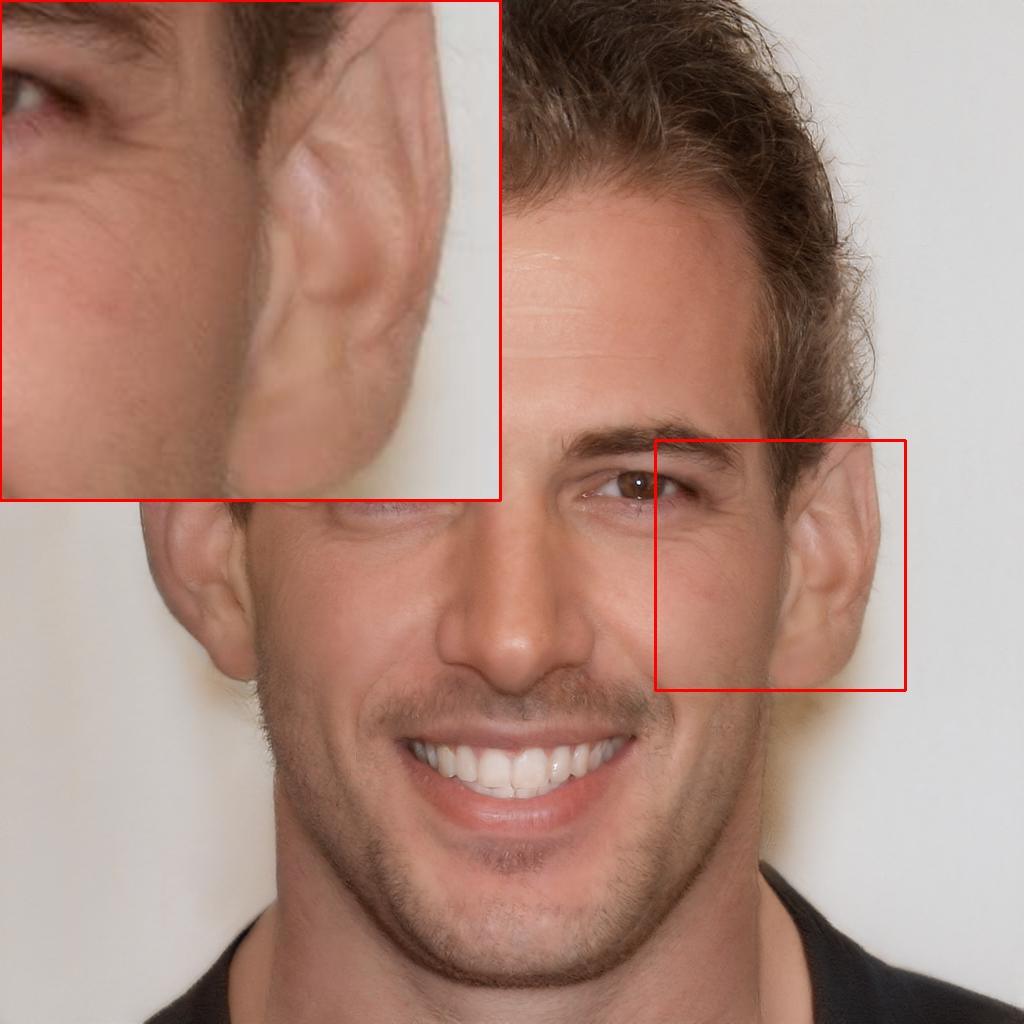}
			& \MyIm{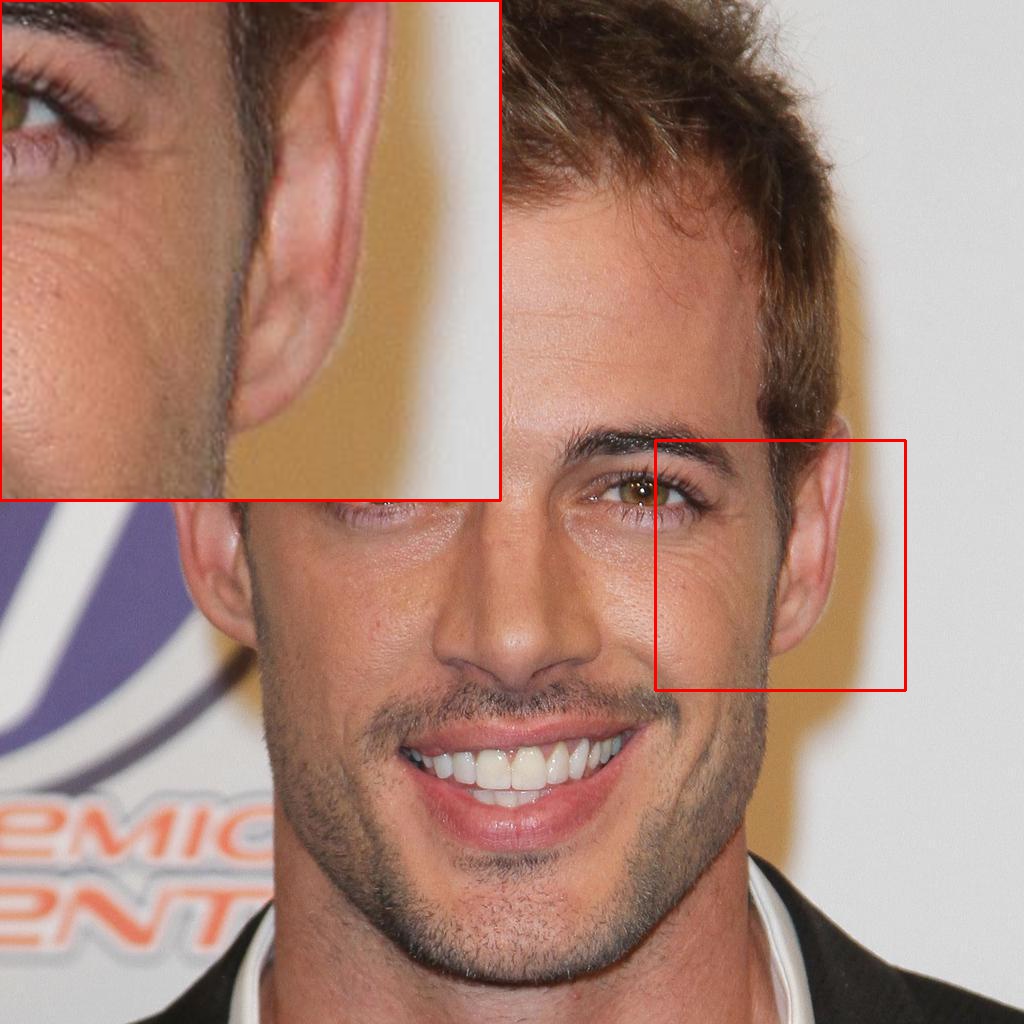}\\
			
			\MyIm{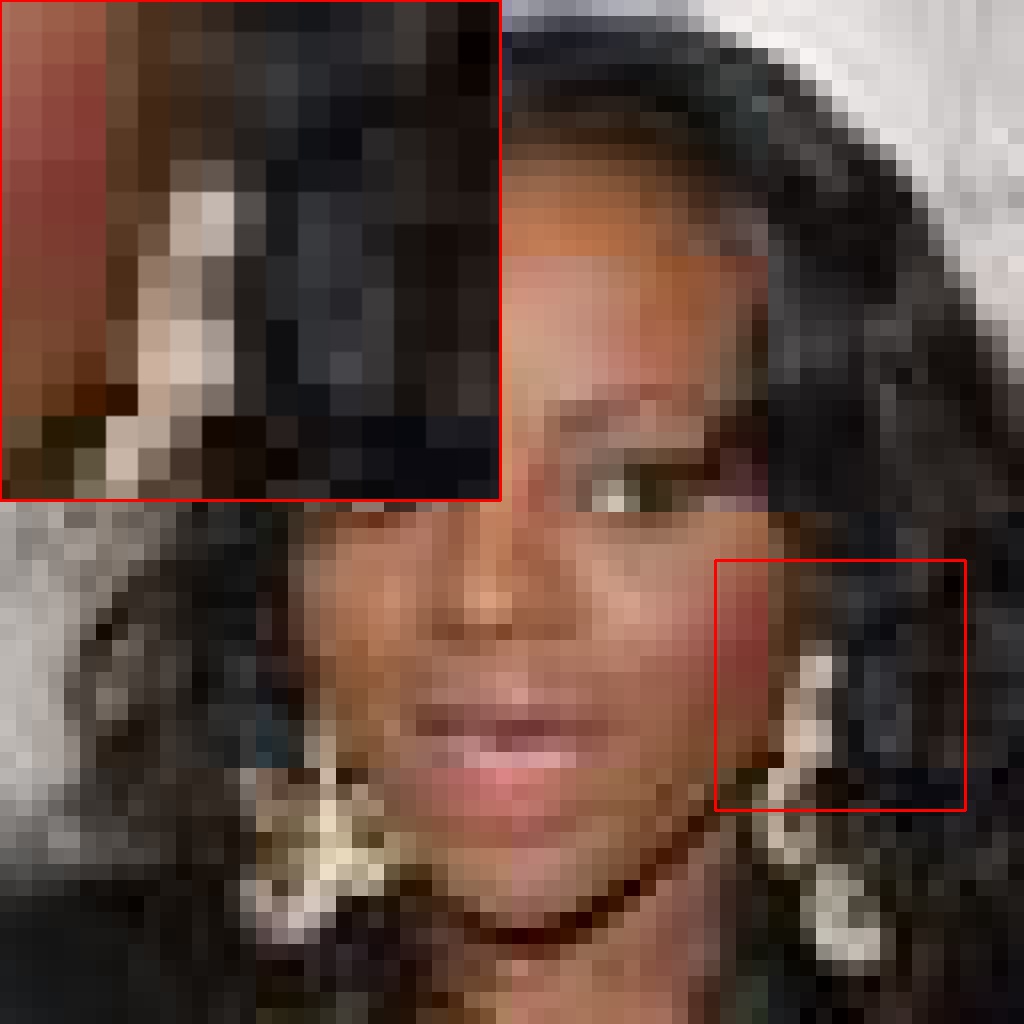} &
			\MyIm{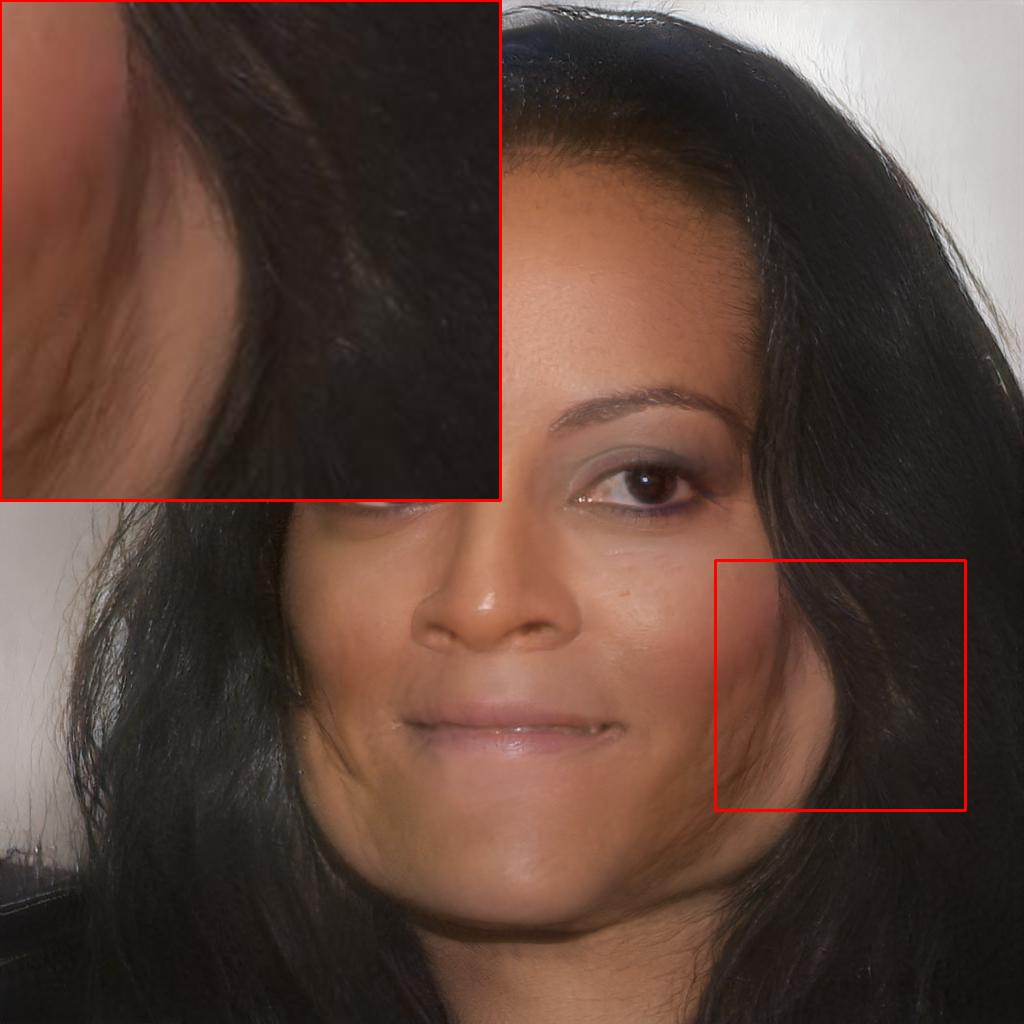} 
			& \MyIm{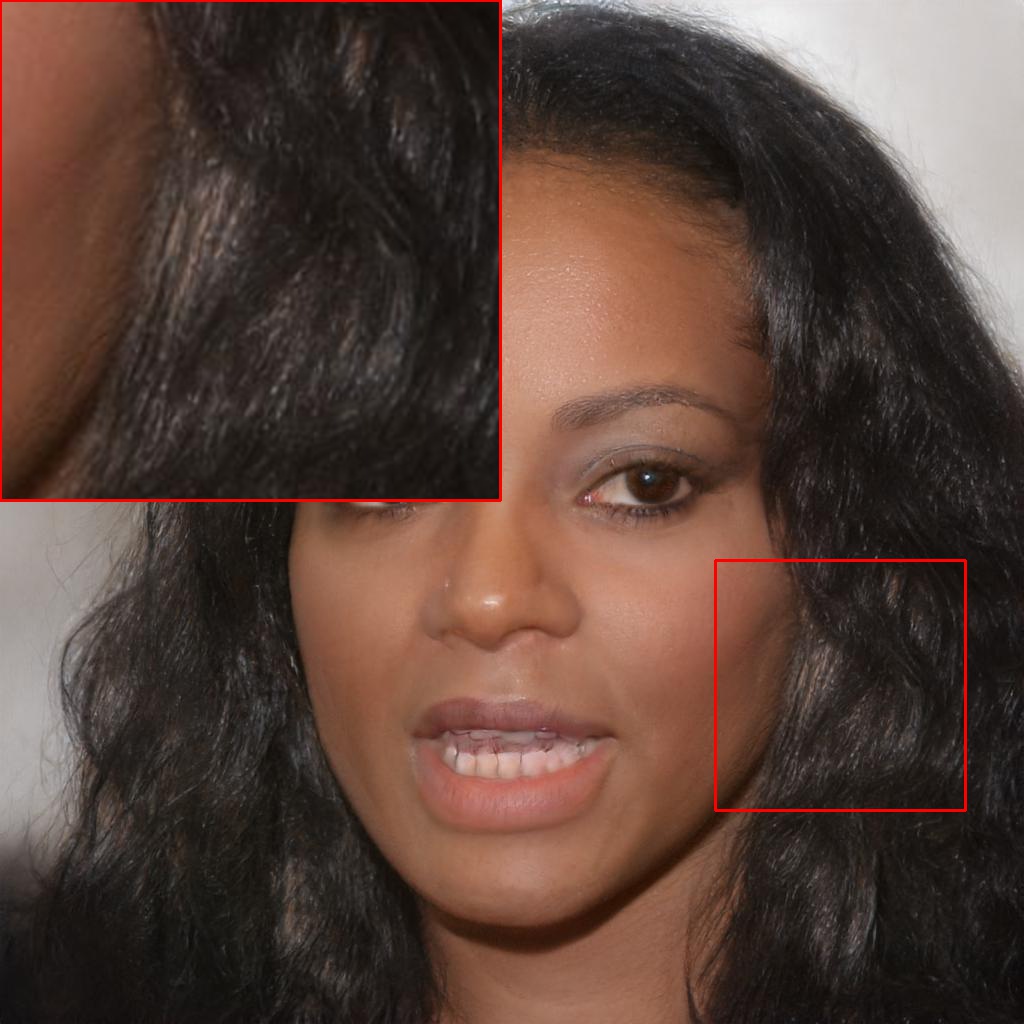} 
			& \MyIm{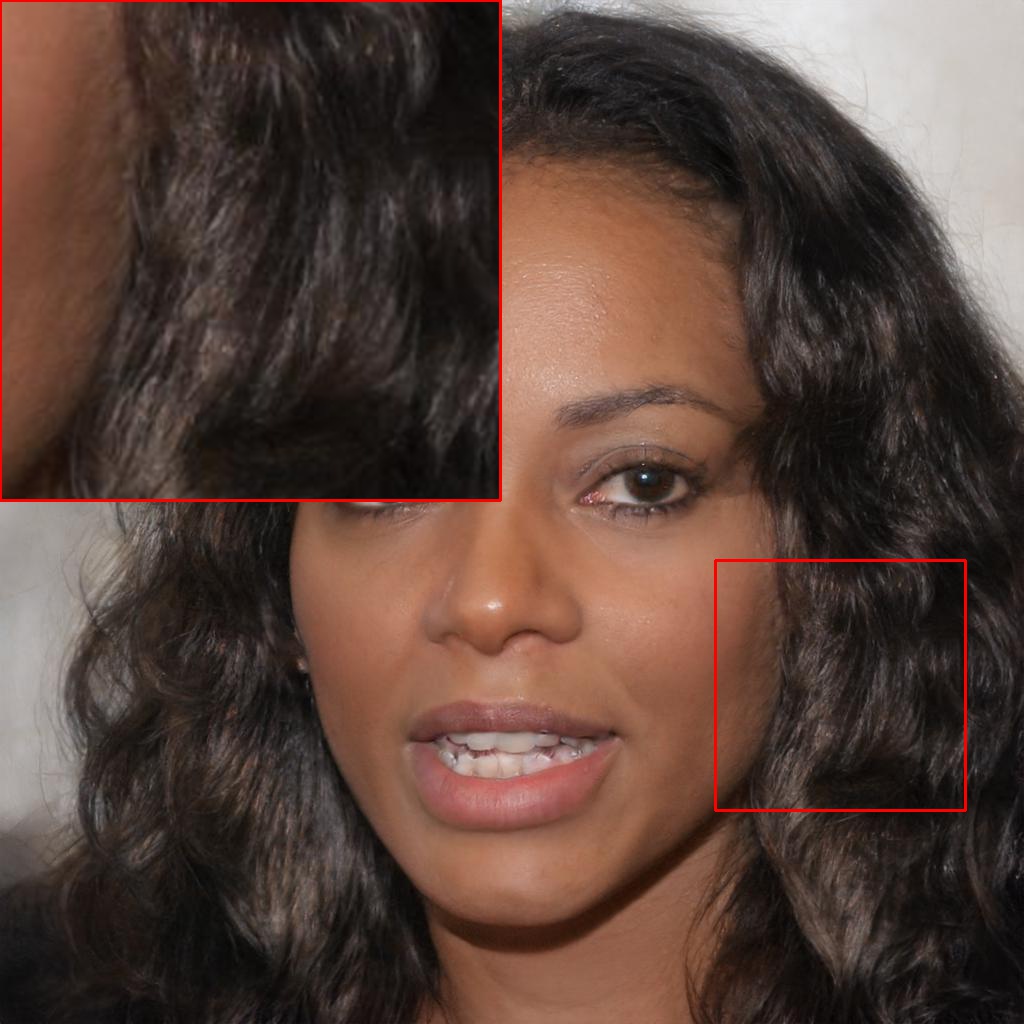}
			& \MyIm{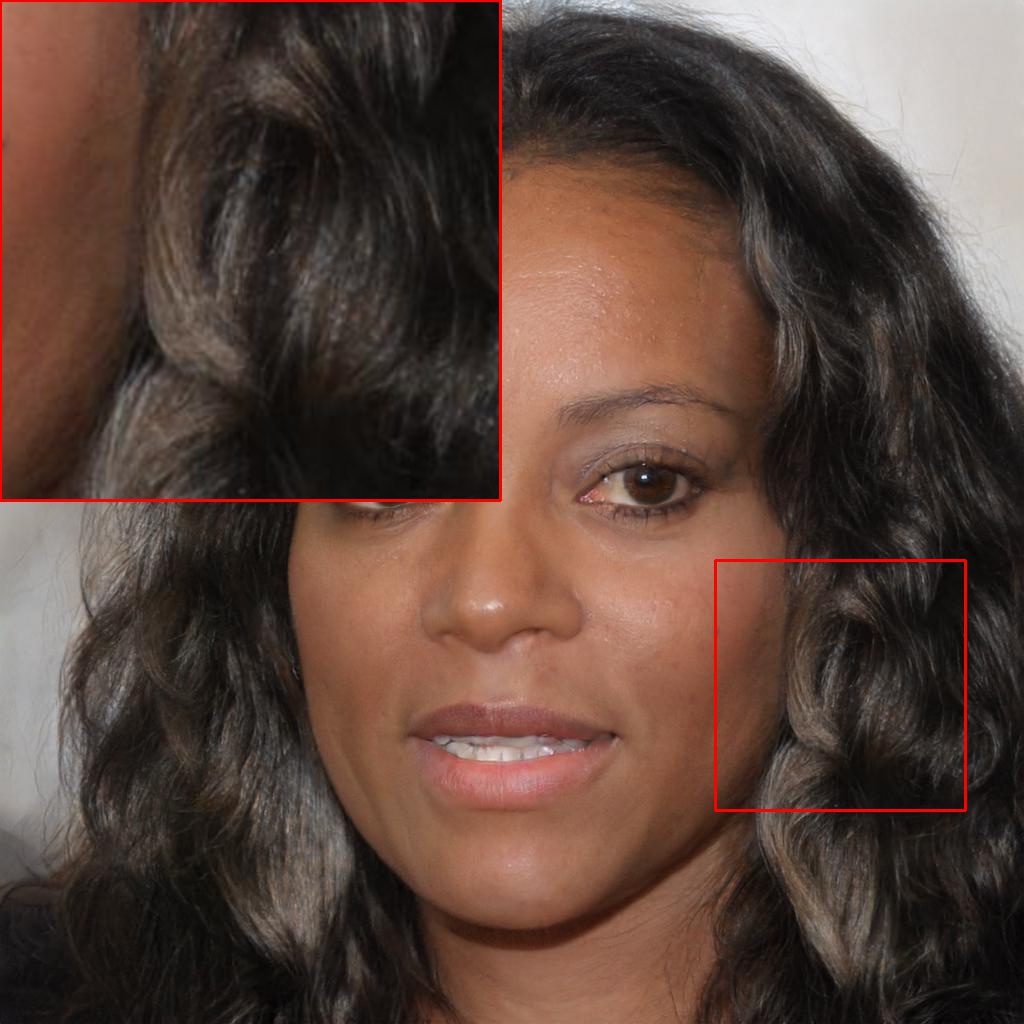}
			& \MyIm{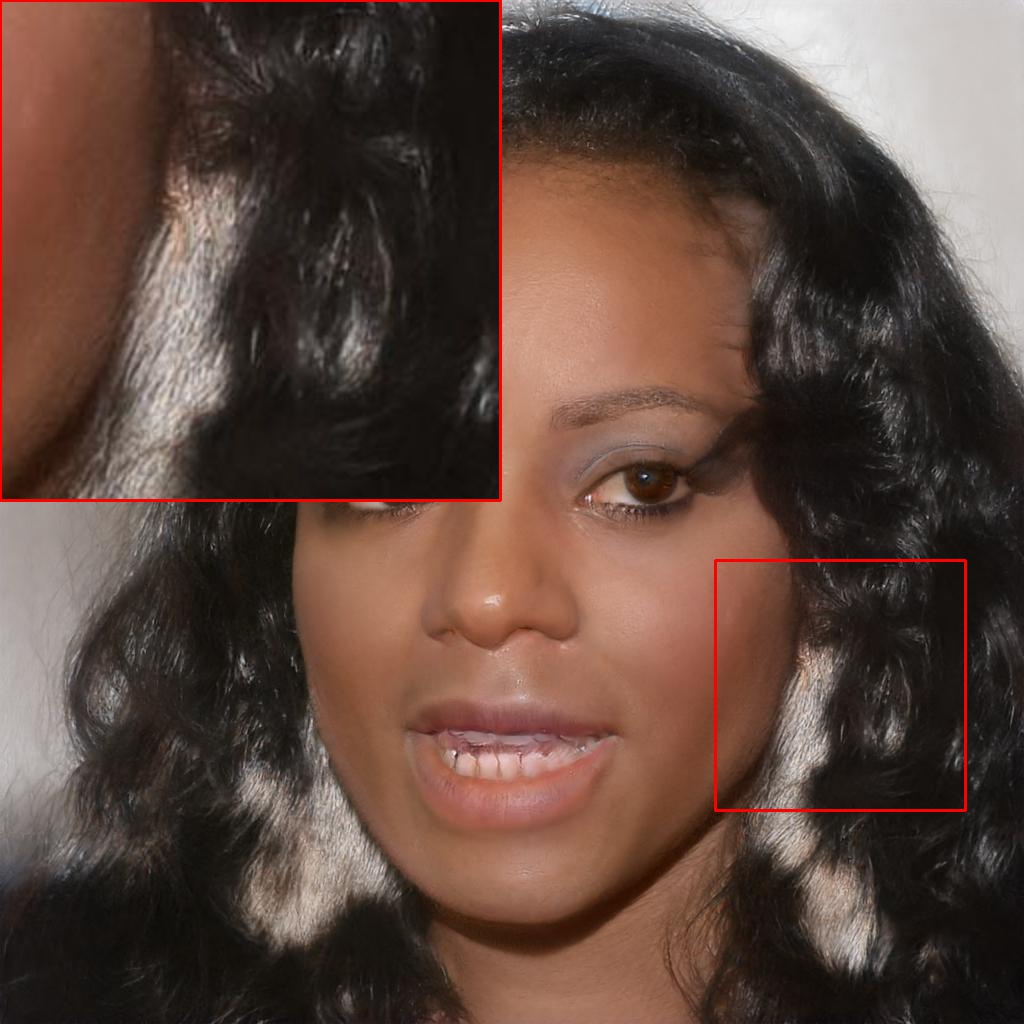}
			& \MyIm{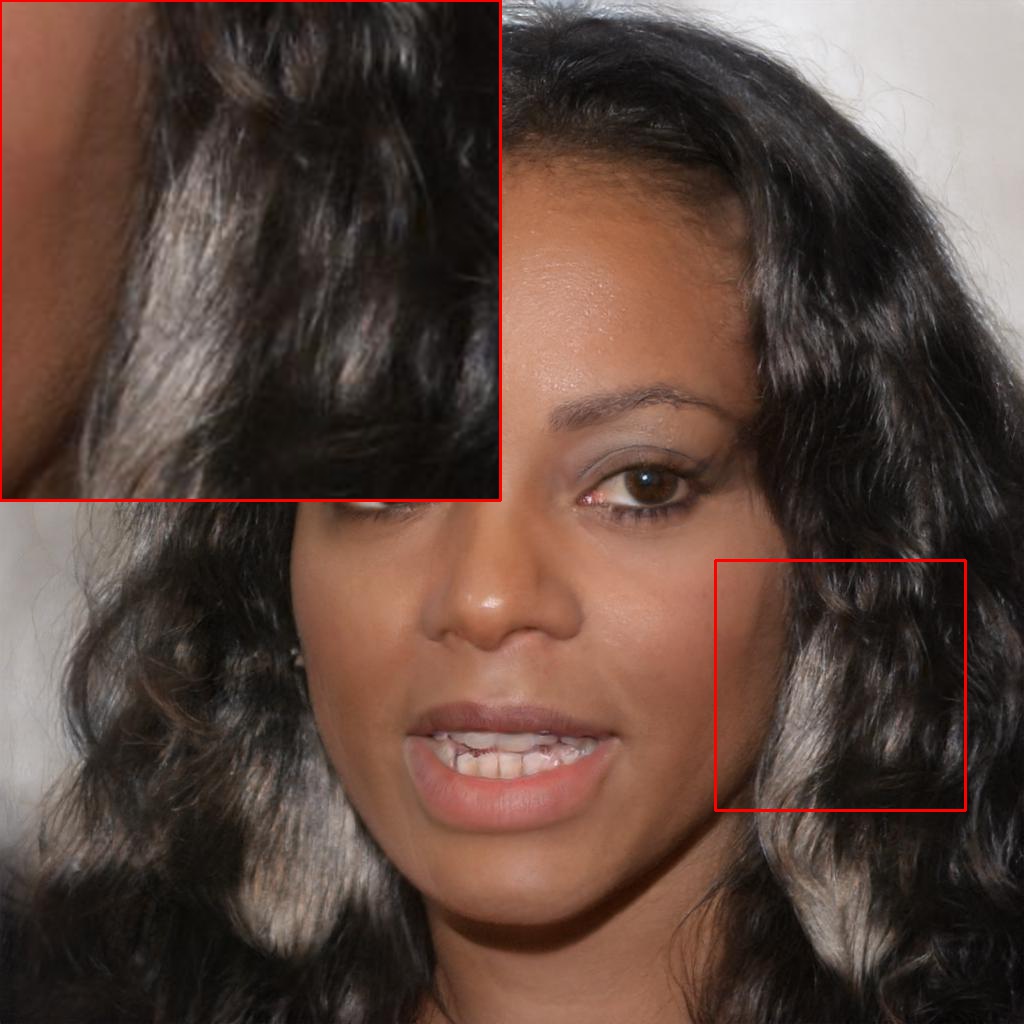}
			& \MyIm{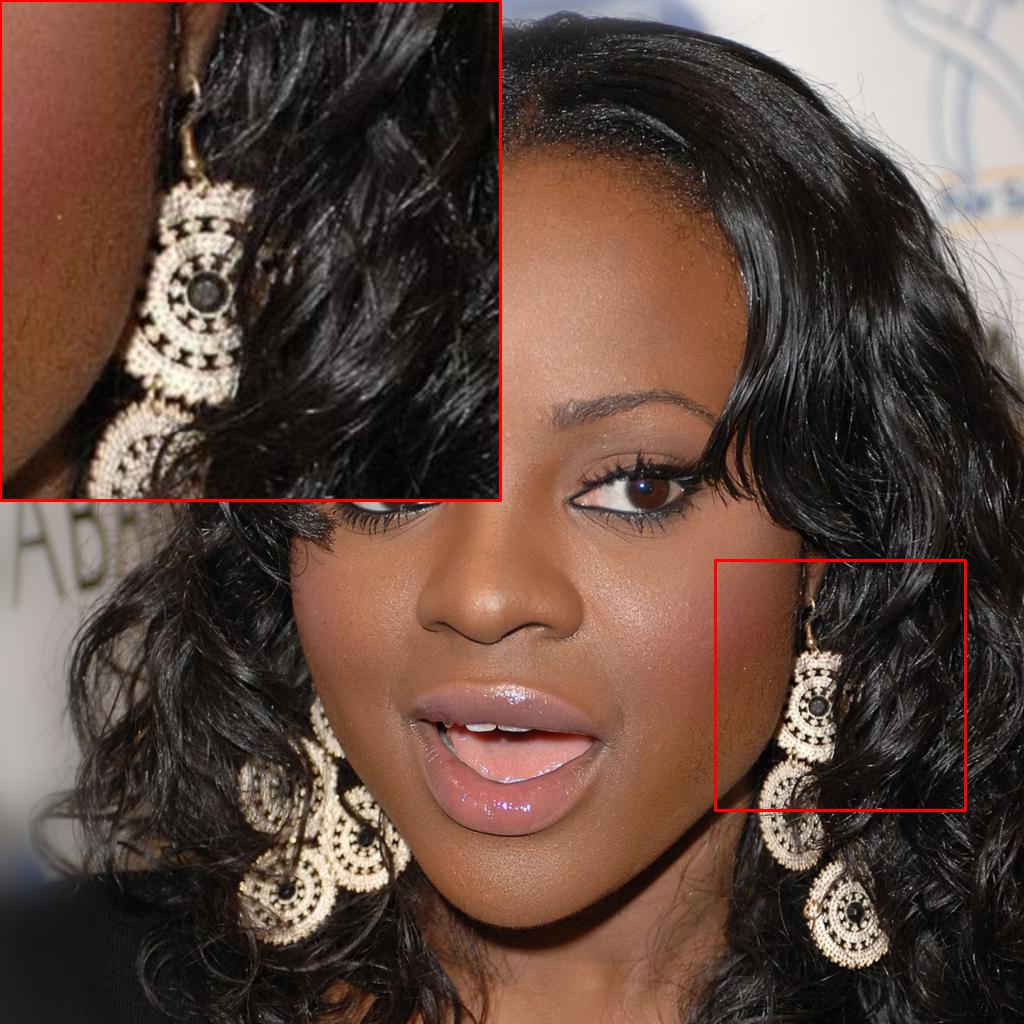}\\
			
			\MyIm{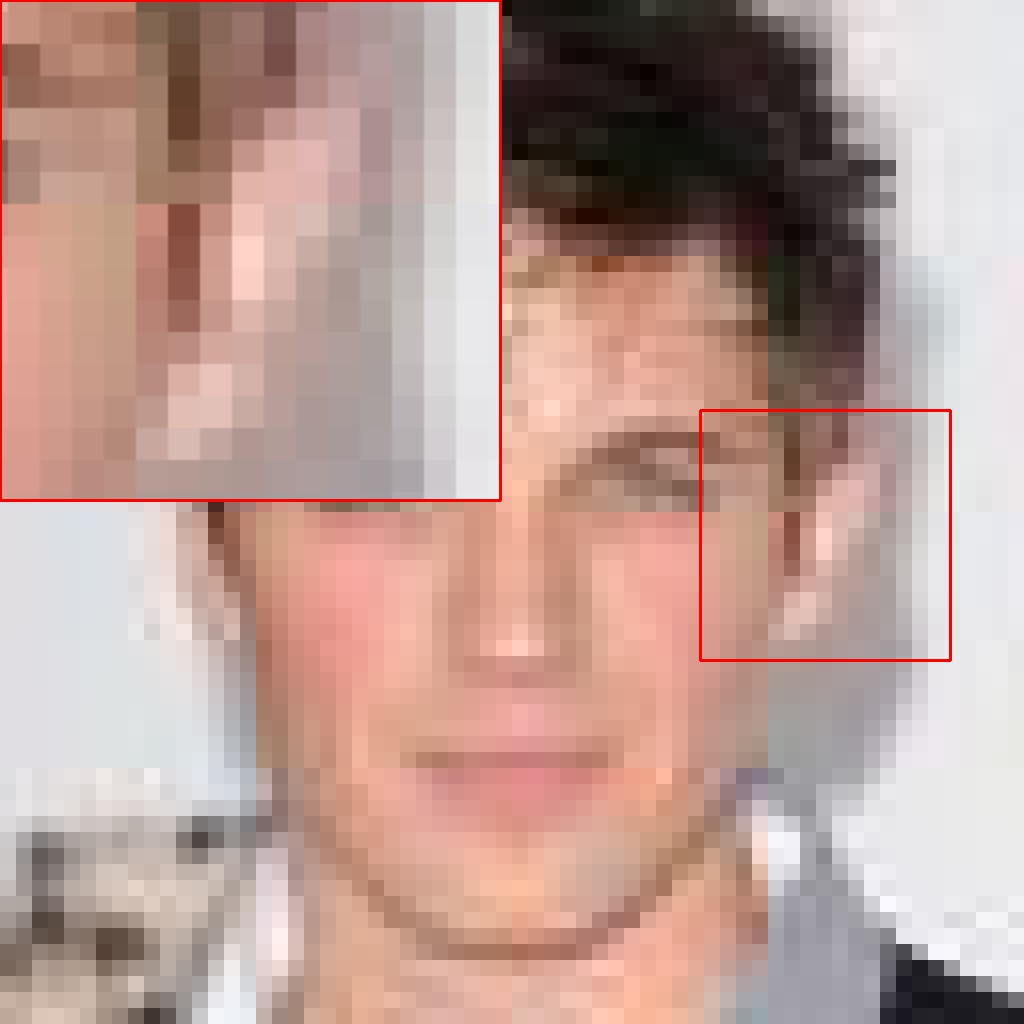} &
			\MyIm{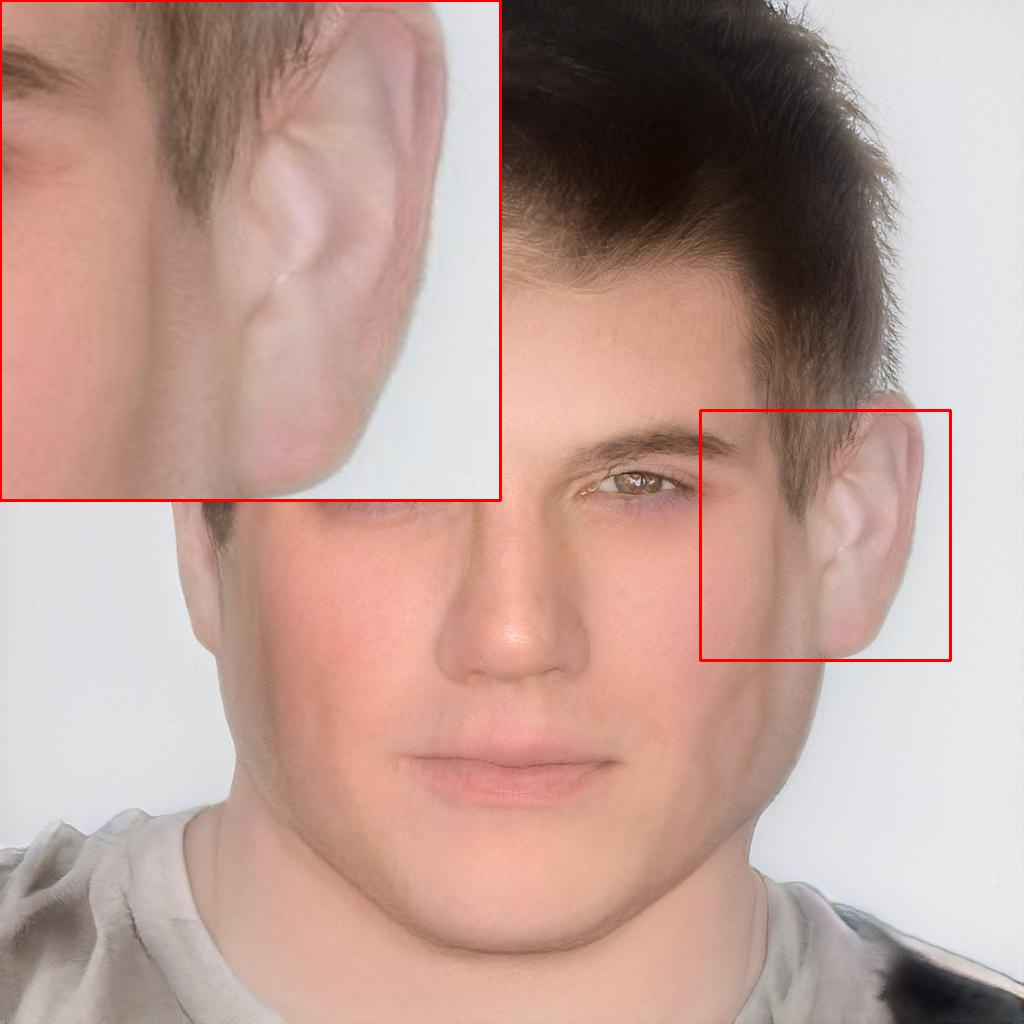} 
			& \MyIm{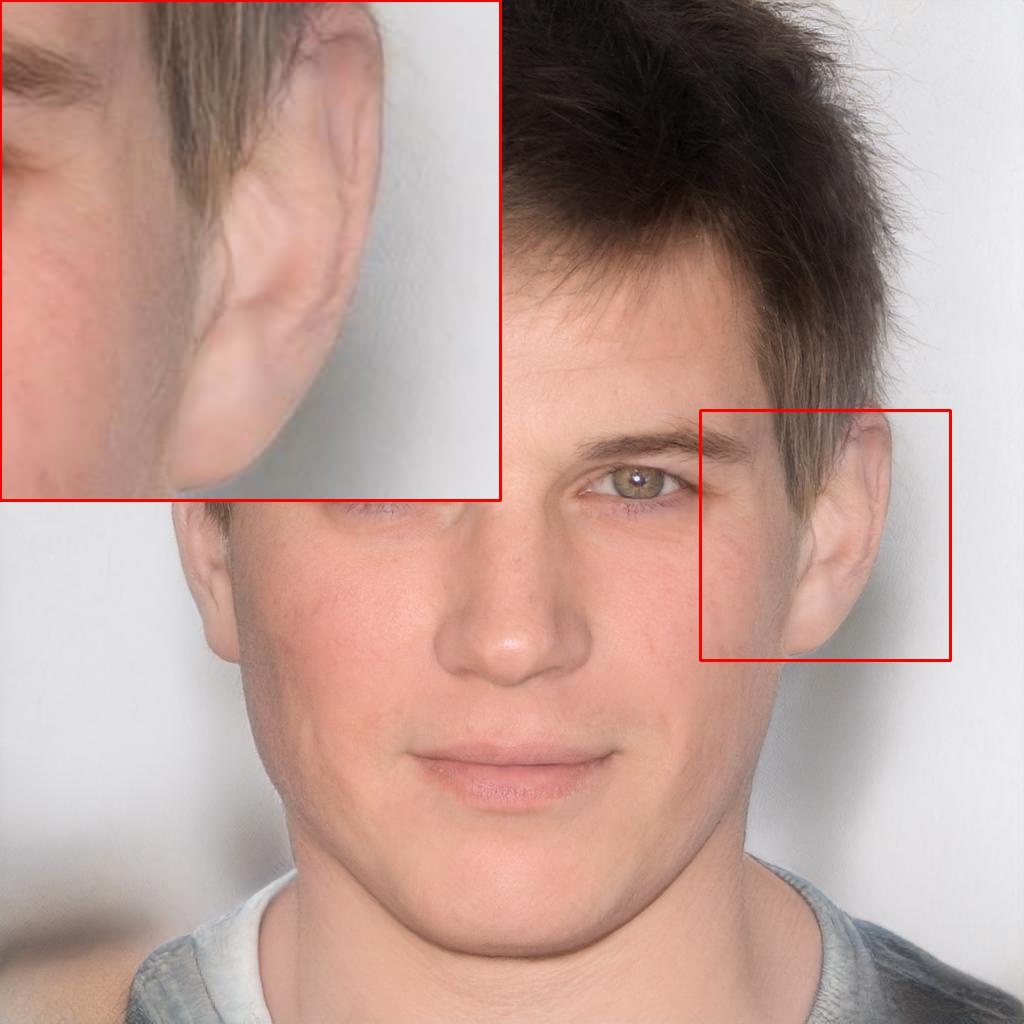}
			& \MyIm{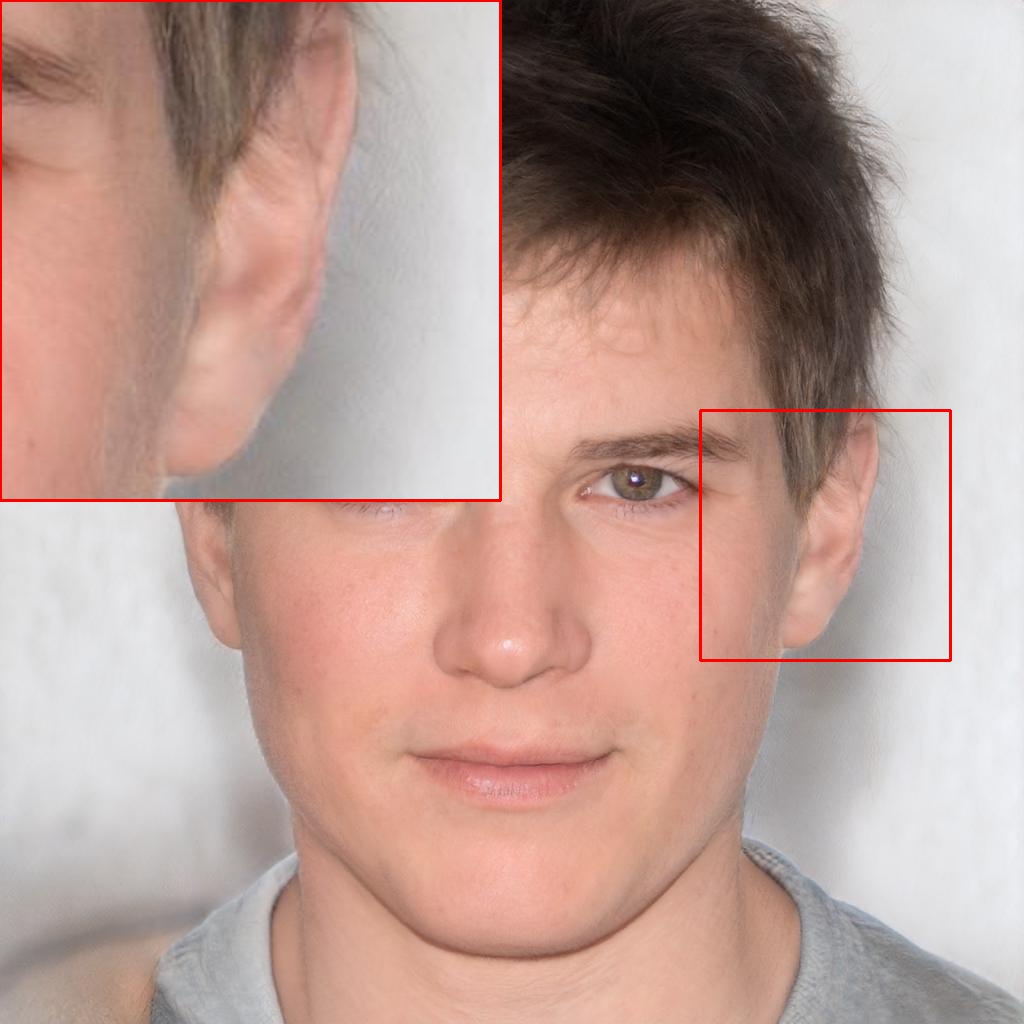}
			& \MyIm{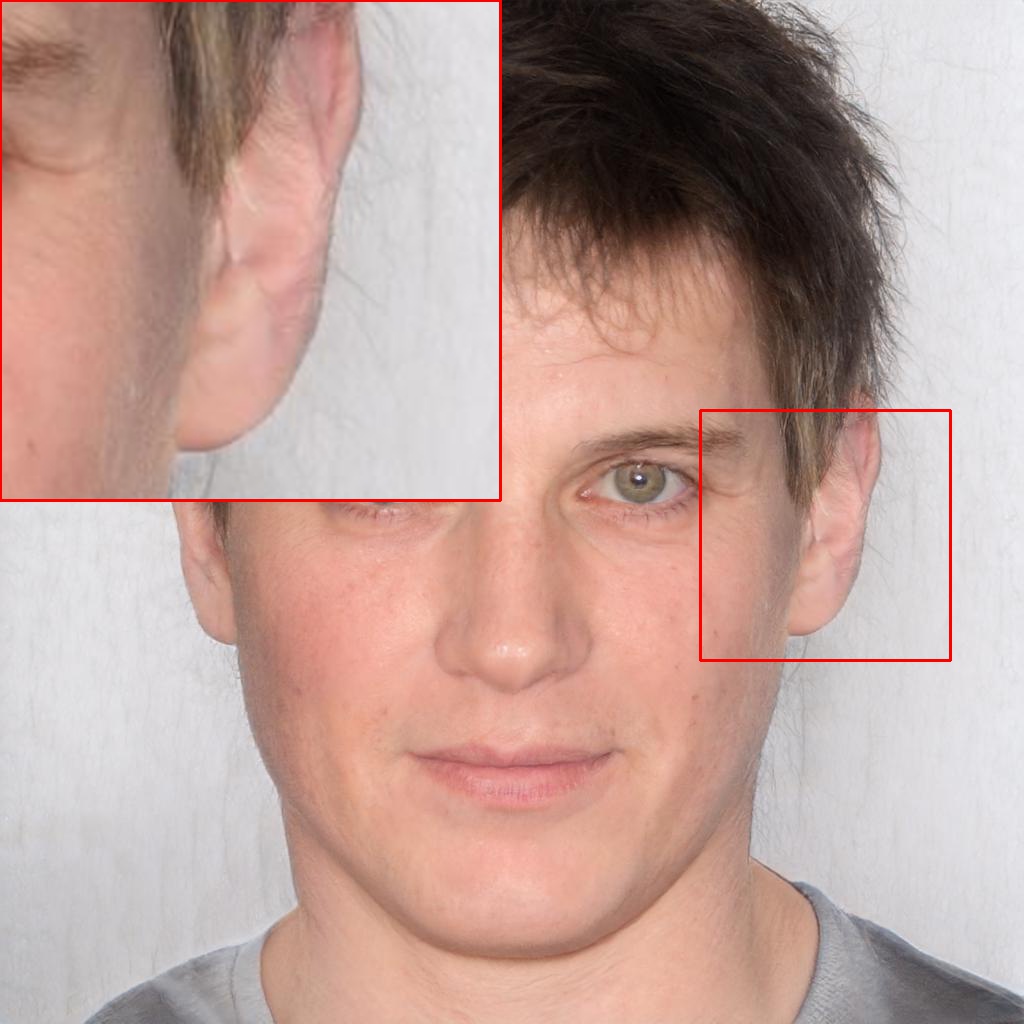}
			& \MyIm{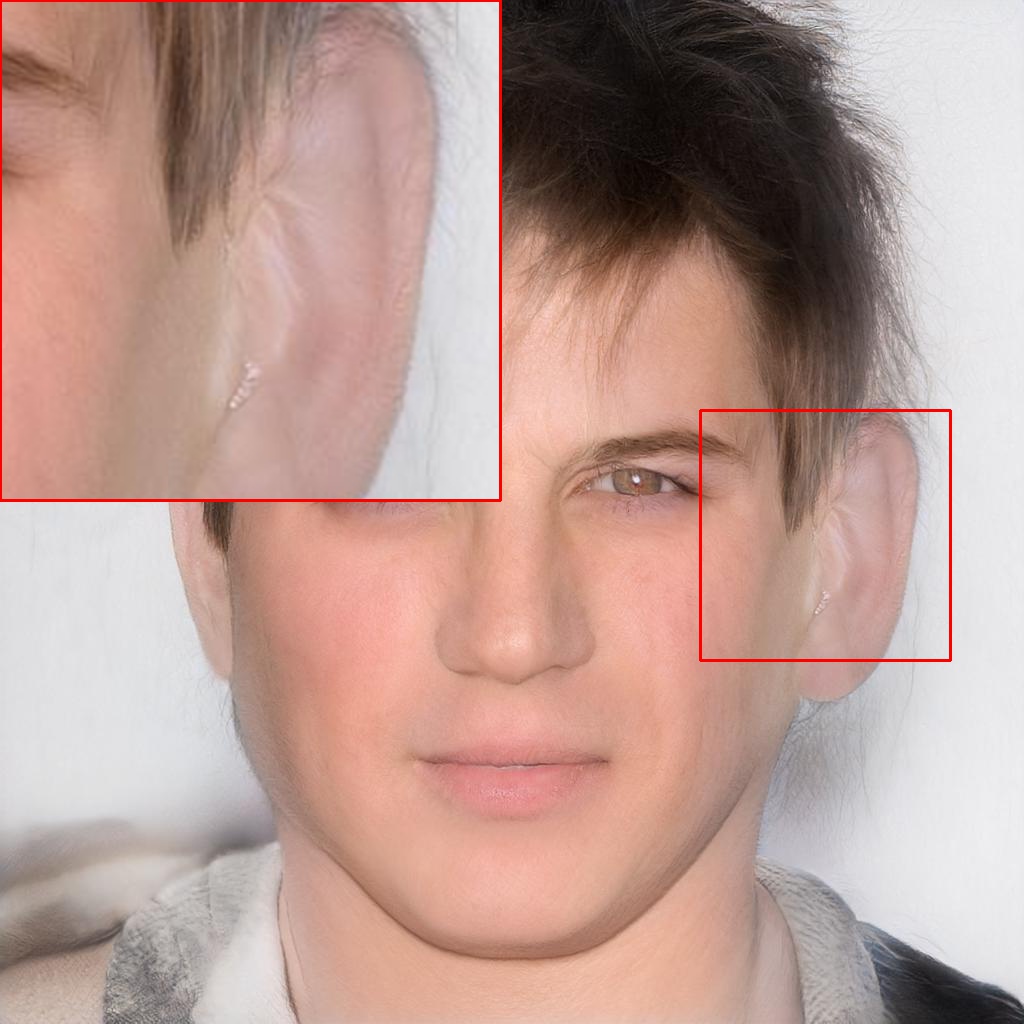}
			& \MyIm{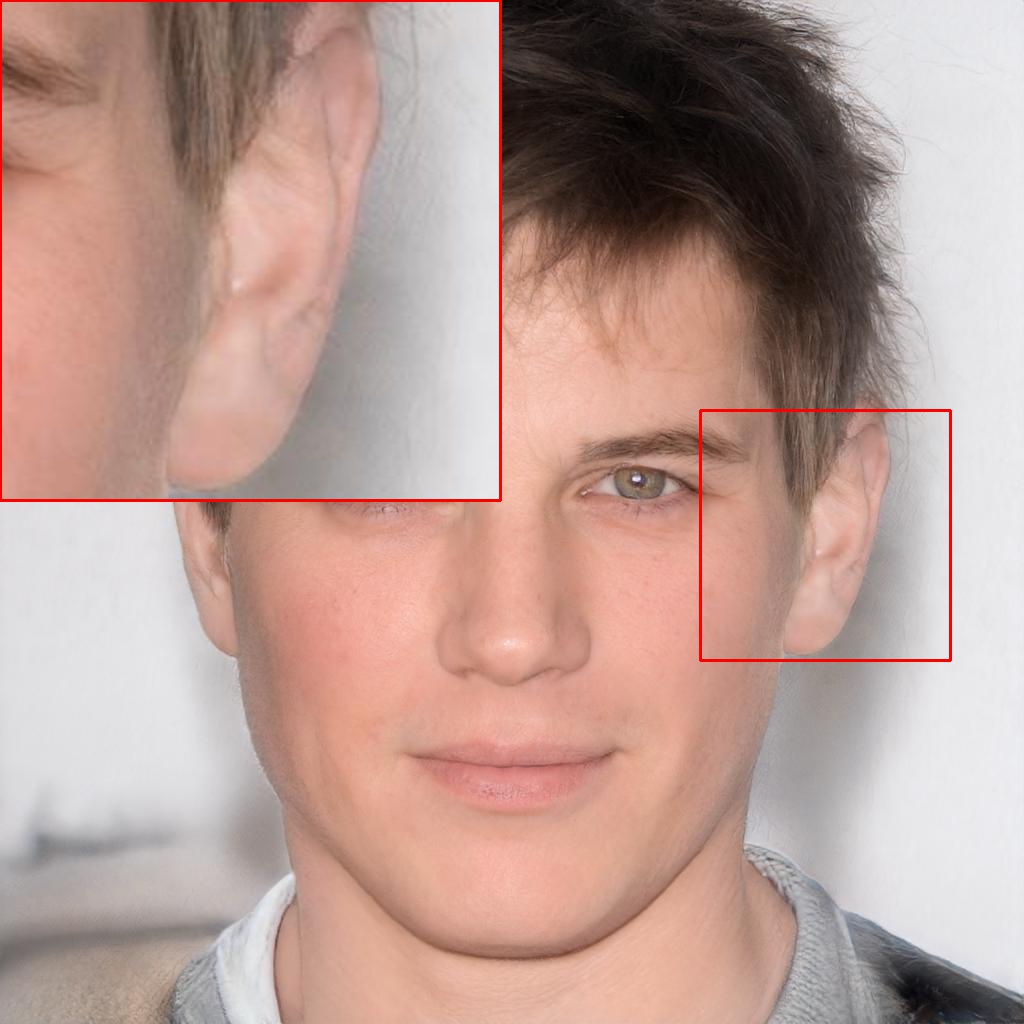}
			& \MyIm{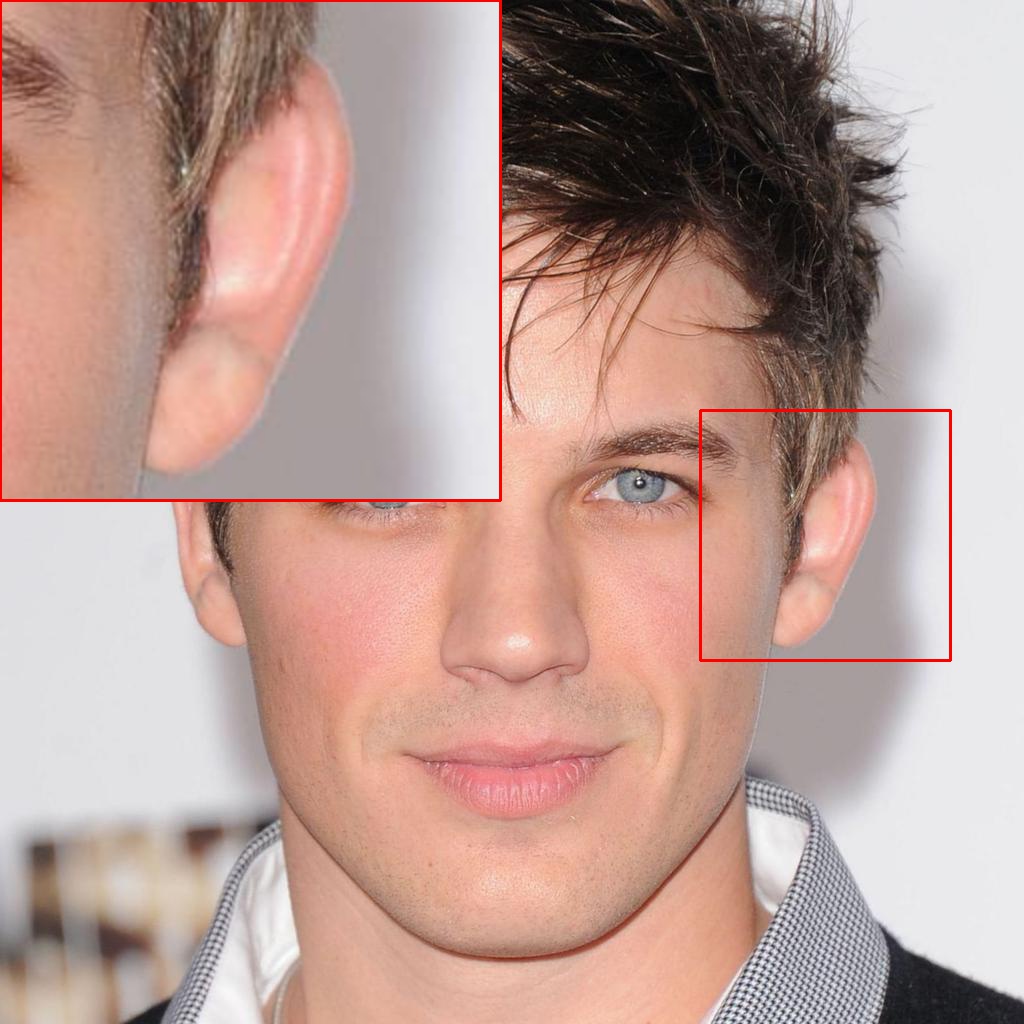}\\
			
			\MyIm{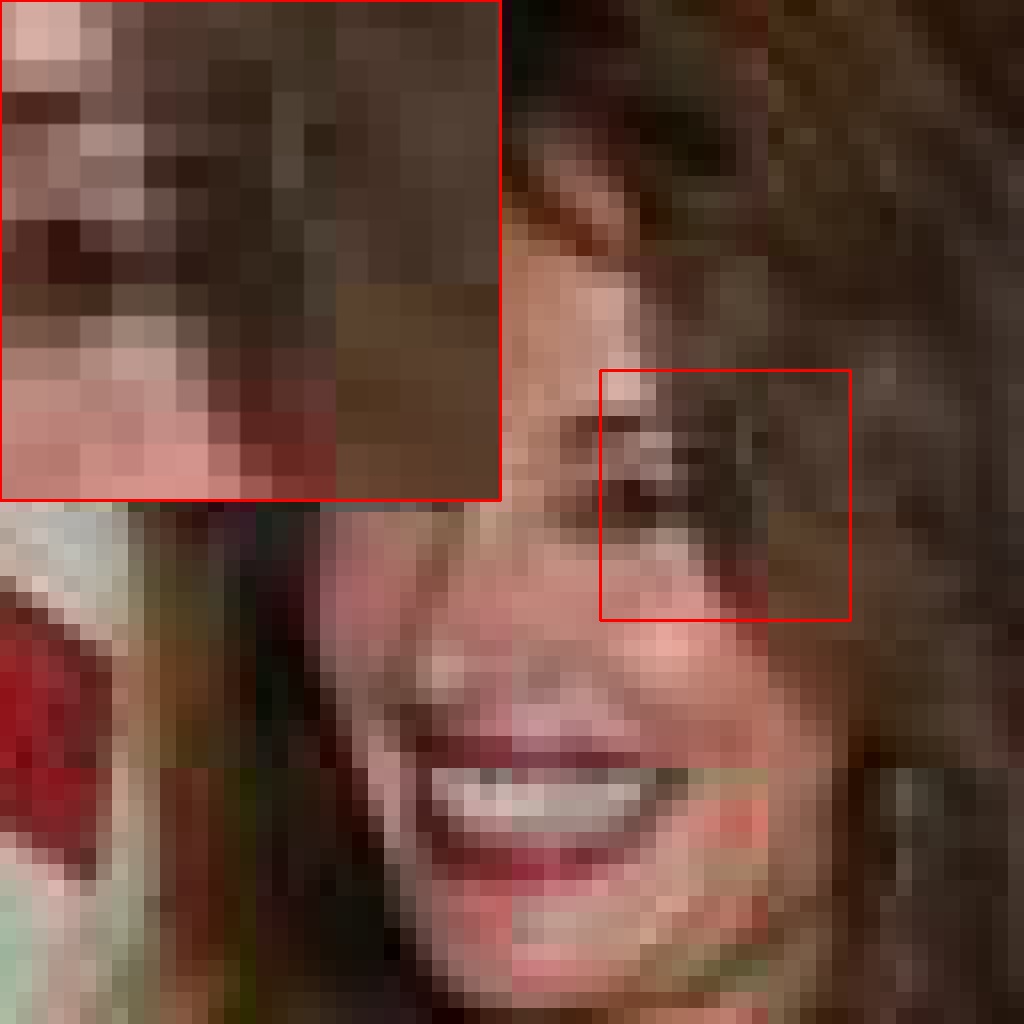} &
			\MyIm{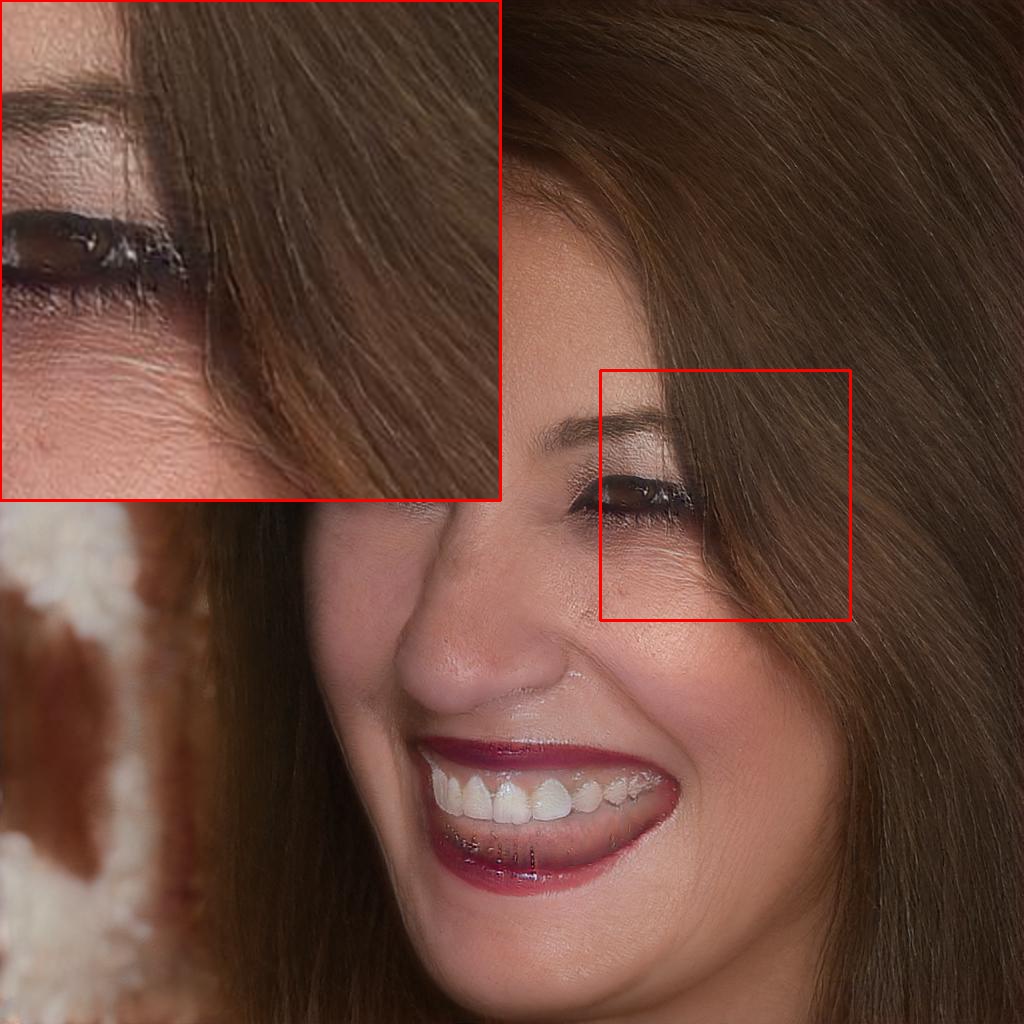} 
			& \MyIm{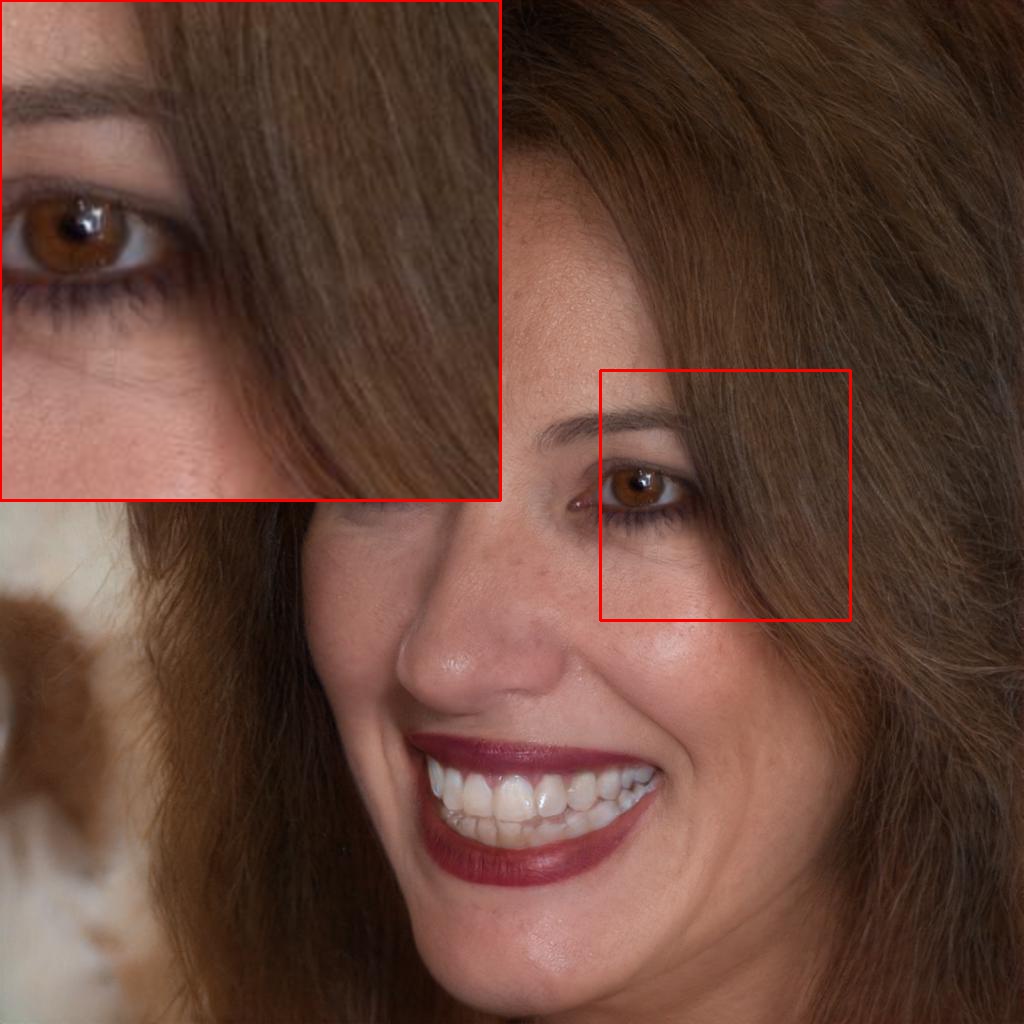} 
			& \MyIm{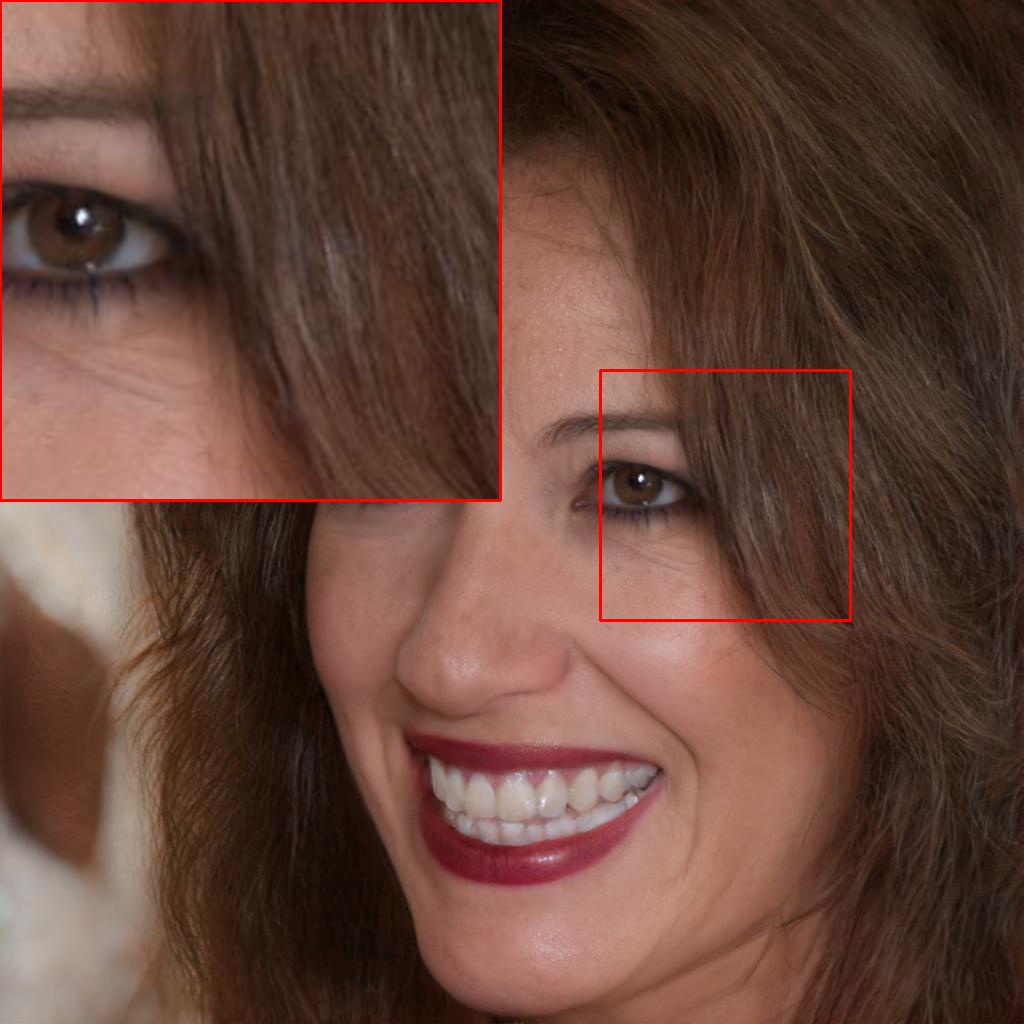}
			& \MyIm{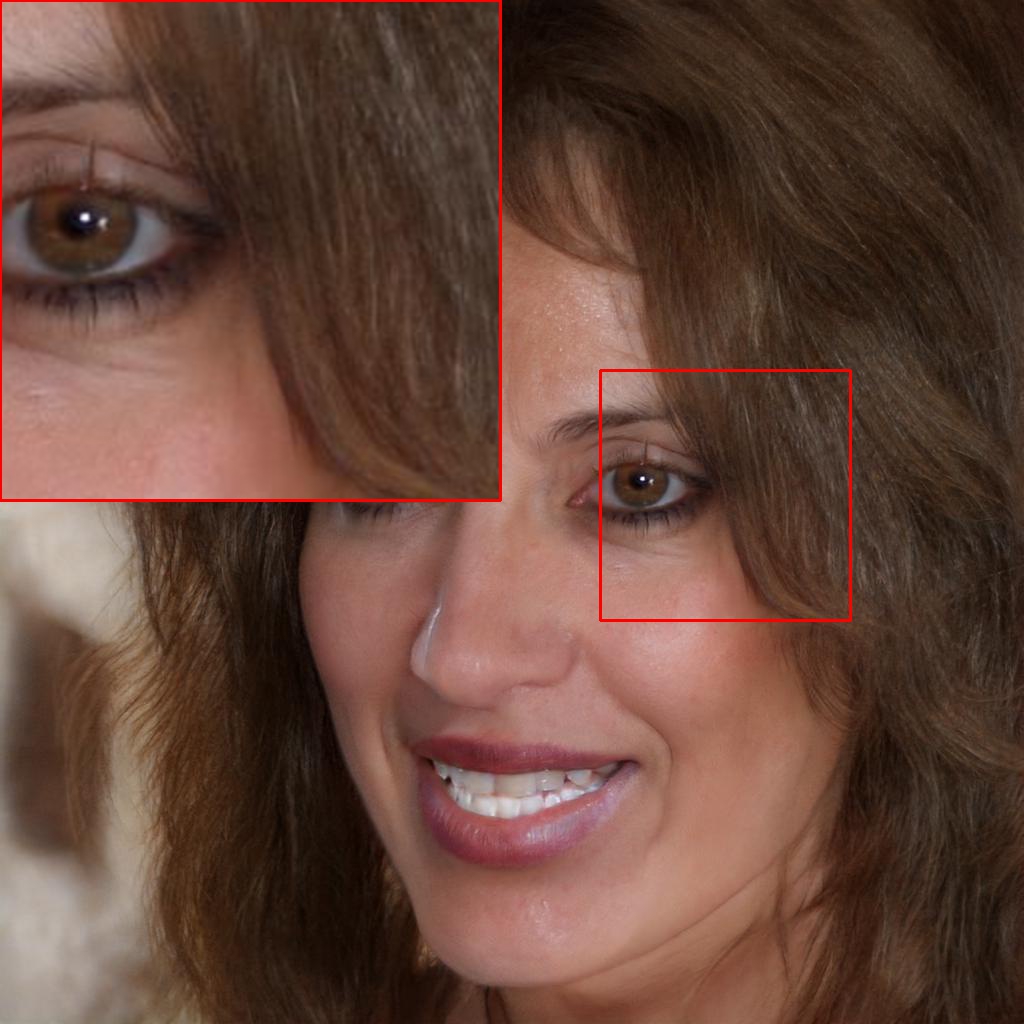}
			& \MyIm{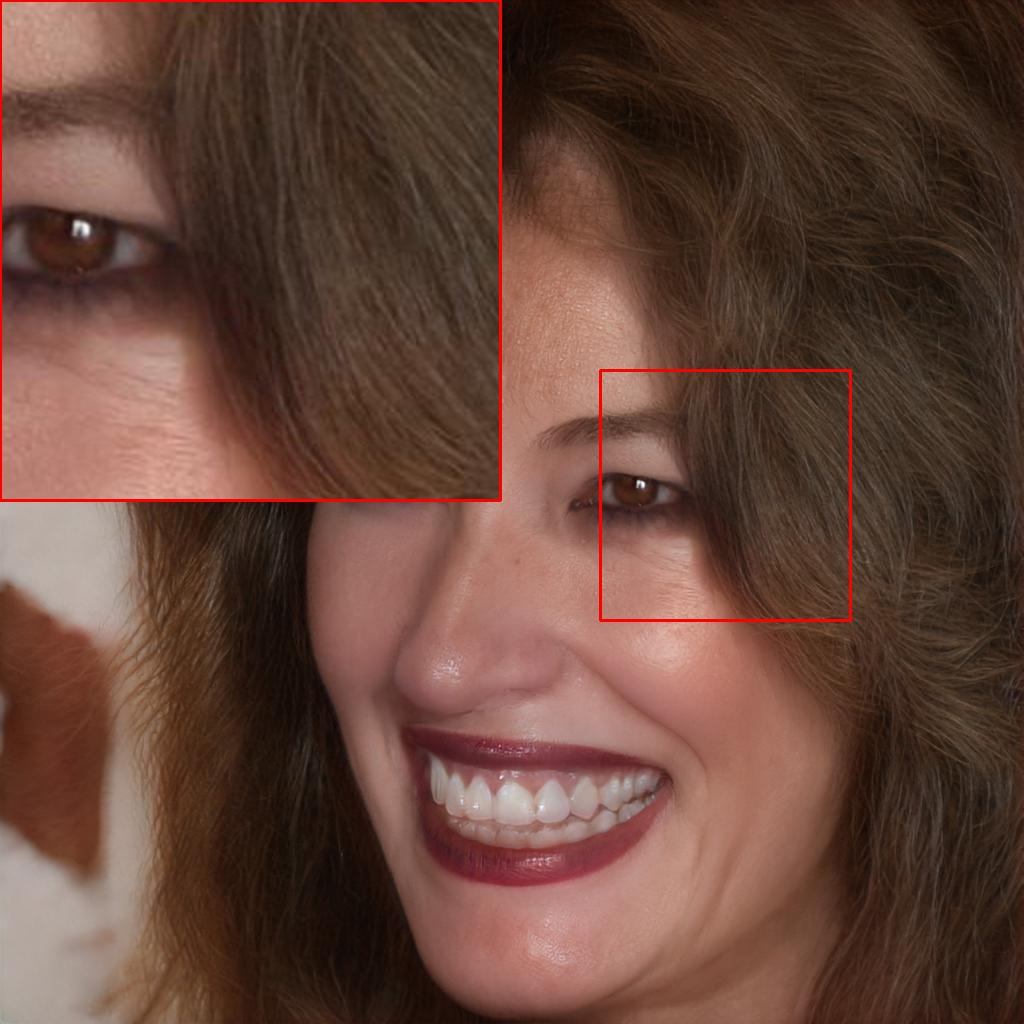}
			& \MyIm{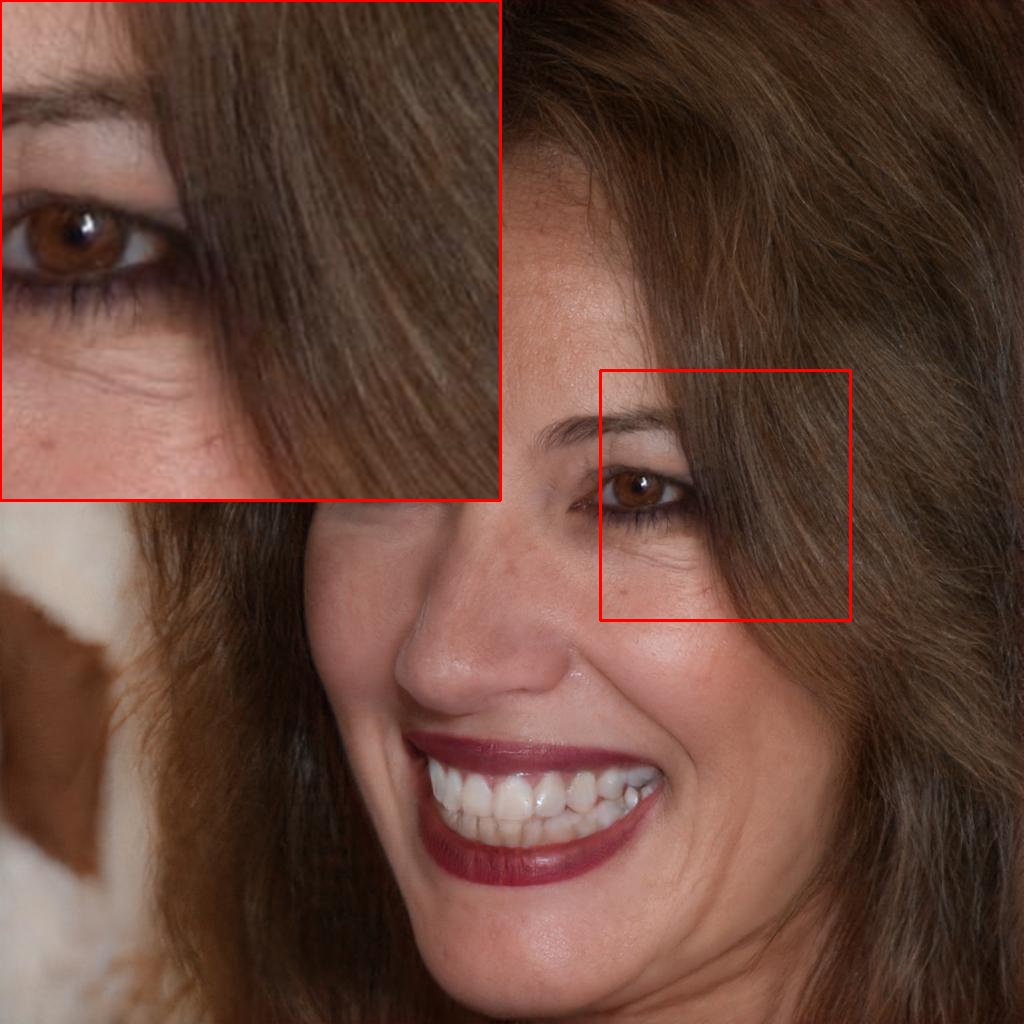}
			& \MyIm{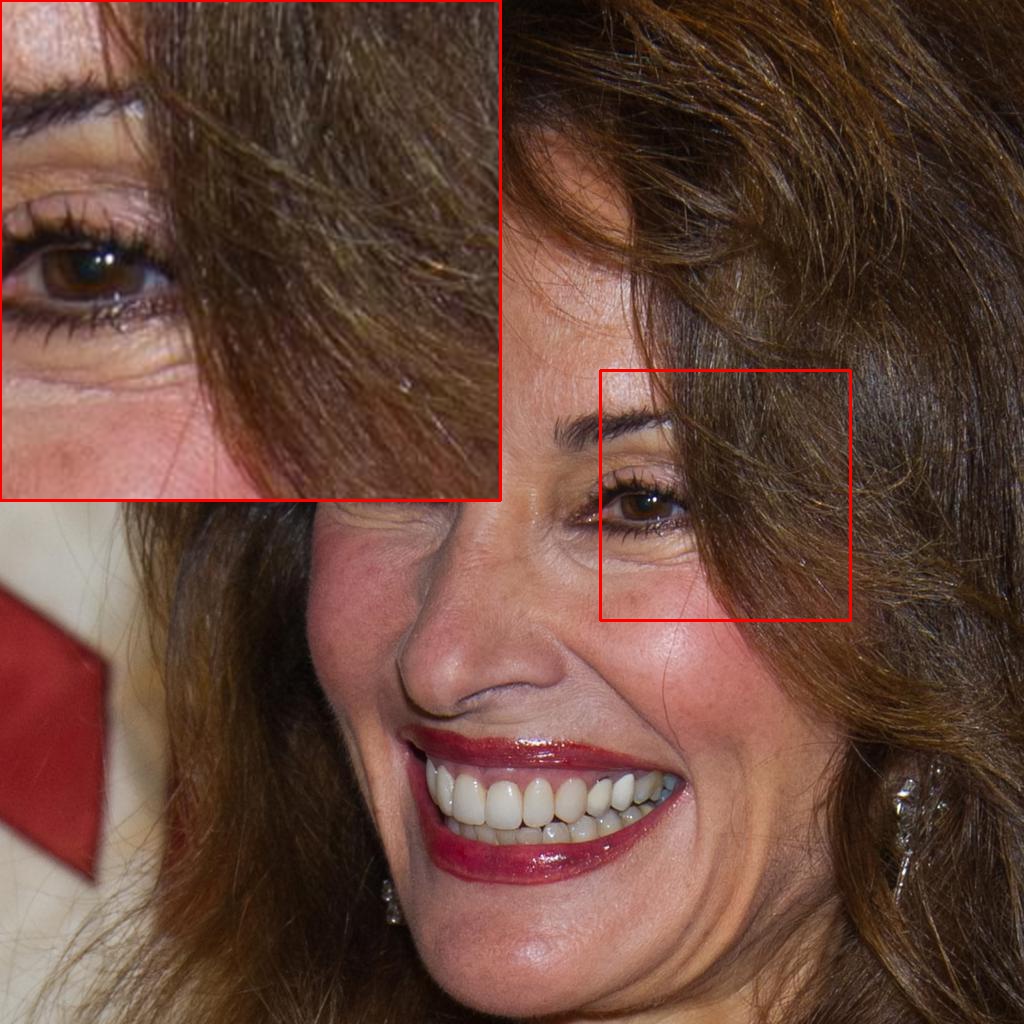}\\
		\end{tabular}
	\end{scriptsize}
	\caption{Ablation study. w/o anchor indicates optimization over $\w$, $\thetaa$ and $\noise$ using a mean value of random samples as initialization. The variants w/o noise, w/o g, and w/o w remove noise inputs, generator's weights, and latent vectors from the optimization, respectively. w/o $\ell_1$-ball removes the locality constraint during the fine-tuning. (16x)}
	\label{fig:ablation_RLS+}
\end{figure*}

\paragraph{RLS$^+$}
Using a pre-trained StyleGAN as the image prior introduces a limitation in maintaining low reolution consistency. As with any GAN model, the representational capacity of StyleGAN is directly linked to the size and diversity of the dataset on which it was trained. In the case of faces, it is unrealistic to expect a given trained model to reconstruct all possible faces. It will typically excel at reconstructing common faces but may struggle to accurately reproduce rare details that were not present during training.

To mitigate the aforementioned issue, we propose RLS$^+$, a refinement step that aims to enhance the reconstruction quality by further training the generator, the latent code, and the noise inputs simultaneously. In this way, the generator range can be expanded to recover an image that better matches  the original HR image.
In this step, we aim at modifying $\G$ as little as possible in order to faithfully reconstruct the HR image still without affecting the semantic prior learned previously that ensures the HR image will remain a face. Moreover, the reconstruction quality can be further improved by optimizing the noise inputs $\noise$, which comprises high-frequency details. 
Finally, after obtaining the anchor point with \cref{equation loss}, the following optimization problem can be solved by initializing the generator with the pre-trained weights $\thetaa$:

\begin{equation}
\label{equation loss+}
\begin{aligned}
\min_{\w, \thetaa, \noise} ||\y - \D(\G_s(\w, \thetaa, \noise))||_1, \quad \w \in \mathcal{B}(\w_{ancor},r)
\end{aligned}
\end{equation}

where $\mathcal{B}(\w_{anchor},r)=\{\w \in \real^\dim|\;||\w-\w_{anchor}||_1 \leq r\}$ denote the set of points with $\ell_1$-norm bounded by radius $r$ from the anchor point. This set comprises latent codes that can generate realistic images with almost the same facial attributes. 
Our algorithm attempts to explore the extended range of $\G_s$, to find the latent vector that best explains the input LR image, but we only allow solutions that lie within an $\ell_1$-norm ball centered at $\w_{anchor}$. Our empirical analysis shows that by keeping the radius of the ball relatively small, we can enhance both realism and identity-similarity.

Intuitively, allowing deviations from the anchor point increases the capability of the generator to produce the closest reconstruction of the target image. However, limiting its deviation, prevent over-fitting to unrealistic details. Obviously, as $r$ increases, $\mathcal{B}$ contains latent vectors that are further away from the anchor point. The last allows finding latent codes that are more expressive and diverse, but also further away from faces and thus producing unrealistic images. Altogether, $r$ offers control over a trade-off between realism and reconstruction error.

We increase the expressivity of the StyleGAN by further optimizing  $\thetaa$. However, to avoid over-expanding the range of the generator to non-realistic images, we employ early stopping to maintain the generative prior. Once the generator is tuned, the final HR image is obtained by $\hat{\x}= \G_s(\w^*,\thetaa^*,\noise^*)$.

\section{Results}
\label{sec:results}
\paragraph{Experimental setup.}
We used a StyleGAN2 generator~\cite{karras2020analyzing} pre-trained on the FFHQ dataset~\cite{karras2019style} that includes 70,000 high-quality face images of resolution $1024\times1024$. For evaluation, we used the first 2000 samples from the CelebA-HQ test set~\cite{karras2017progressive} and simulated degraded faces from the HR images using bicubic downsampling.

\setlength{\tabcolsep}{9pt}
\begin{table*}
	\centering	
	\begin{tabular}{@{}l|llll|lll@{}}
		\toprule
		Method & FID$\downarrow$ & KID$^{({\times10^3})}$$\downarrow$ & NIQE$\downarrow$ & ID$\uparrow$ & LPIPS$\downarrow$ &PSNR$\uparrow$ & MSSIM$\uparrow$ \\
		\hline
		w/o anchor & 32.4085 & 18.9718 & 4.4316 & 0.7817 & 0.4309 & 22.7645 & 0.7009 \\
		w/o noise & 31.1461 & 16.7528 & 4.3895 & 0.7887 & 0.4132 & 22.6009 & 0.6934 \\
		w/o w & 36.0475 & 20.5331 & \blue{4.2209} & 0.7790 & 0.4150 & 22.4965 & 0.6932 \\
		w/o g & 30.2329 & 15.5456 & 4.2967 & 0.7896 & 0.4121 & 22.4442 & 0.6867 \\
		w/o $\ell_1$-ball & \blue{26.8366} & \blue{13.1830} & 4.4294 & \blue{0.8040} & \red{0.3998} & \blue{23.7940} & \blue{0.7288} \\
		RLS$^{+}$ & \red{28.3786} & \red{13.7663} & \red{4.2878} & \red{0.7981} & \blue{0.3972} & \red{23.5242} & \red{0.7195} \\
		
		\bottomrule
	\end{tabular}
	\caption{Ablation study on 16x super-resolution. (The \blue{best} and the \red{second-best} are emphasized by blue and red respectively.)}
	\label{tab:quantitative_ablation}
\end{table*}

The regularization parameters $\lambda_w$, $\lambda_c$ and $\lambda_g$ were set to 0.0002, 0.05, and 0.0004, respectively. The normalizing flow model we use is the MAF, as it tends to work better than RealNVP for density estimation tasks. Five flow blocks are used in our model and all hidden dimensions are set to 1024.
For RLS, we used an Adam optimizer over 200 iterations with a learning rate of 0.5 and initialized the search by the mean of 100,000 randomly generated latent vectors. Then, for RLS$^+$, we further run for only 50 iterations with a learning rate of $0.0001$. To enforce the $\ell_1$-norm ball constraint, we use Projected Gradient Descent and set the radius of the ball as $\sqrt{\dim}$. Also, only the first nine input layers are optimized, and the rest are fixed.

We compared our algorithm with state-of-the-art face restoration methods, including PULSE~\cite{menon2020pulse}, BRGM~\cite{marinescu2021bayesian}, GPEN~\cite{yang2021gan}, GFPGAN~\cite{wang2021towards} and DDRM~\cite{kawardenoising}.
We used the original codes and weights from the official paper repositories for all experiments, and replicated the same parameter settings reported in the original papers.

\begin{figure*}[!h]
	\centering
	\begin{tiny}
 \setlength{\tabcolsep}{2pt}
		\begin{tabular}{CCCCC}
			DS
			& \shortstack{Gaussian Noise \\ ($\sigma=0.1$)} 
			& \shortstack{Salt \& Pepper \\ ($\sigma=0.05$)}
			& \shortstack{Gaussian Blur \\ ($\sigma=1$)}
			& \shortstack{Motion Blur \\ ($length=100$)}\\
		
			\MyIm{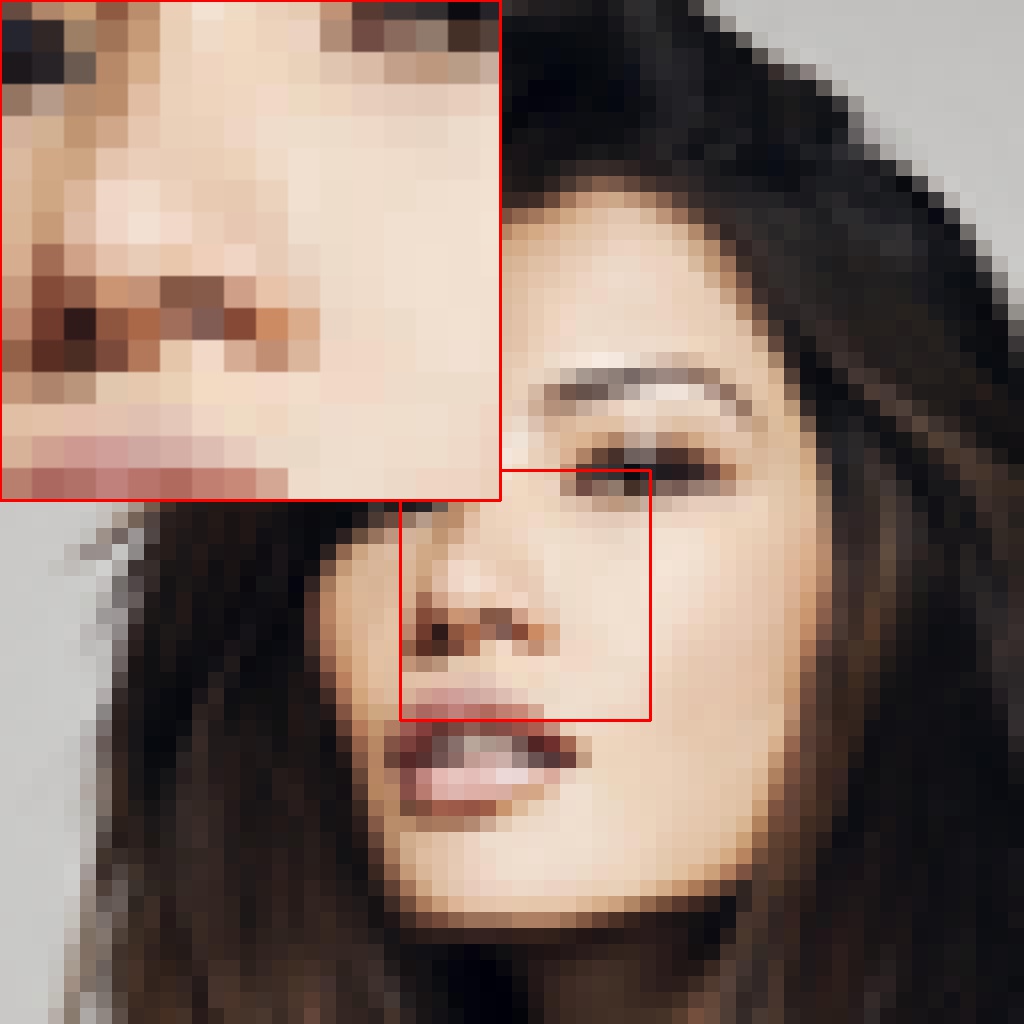}
			&\MyIm{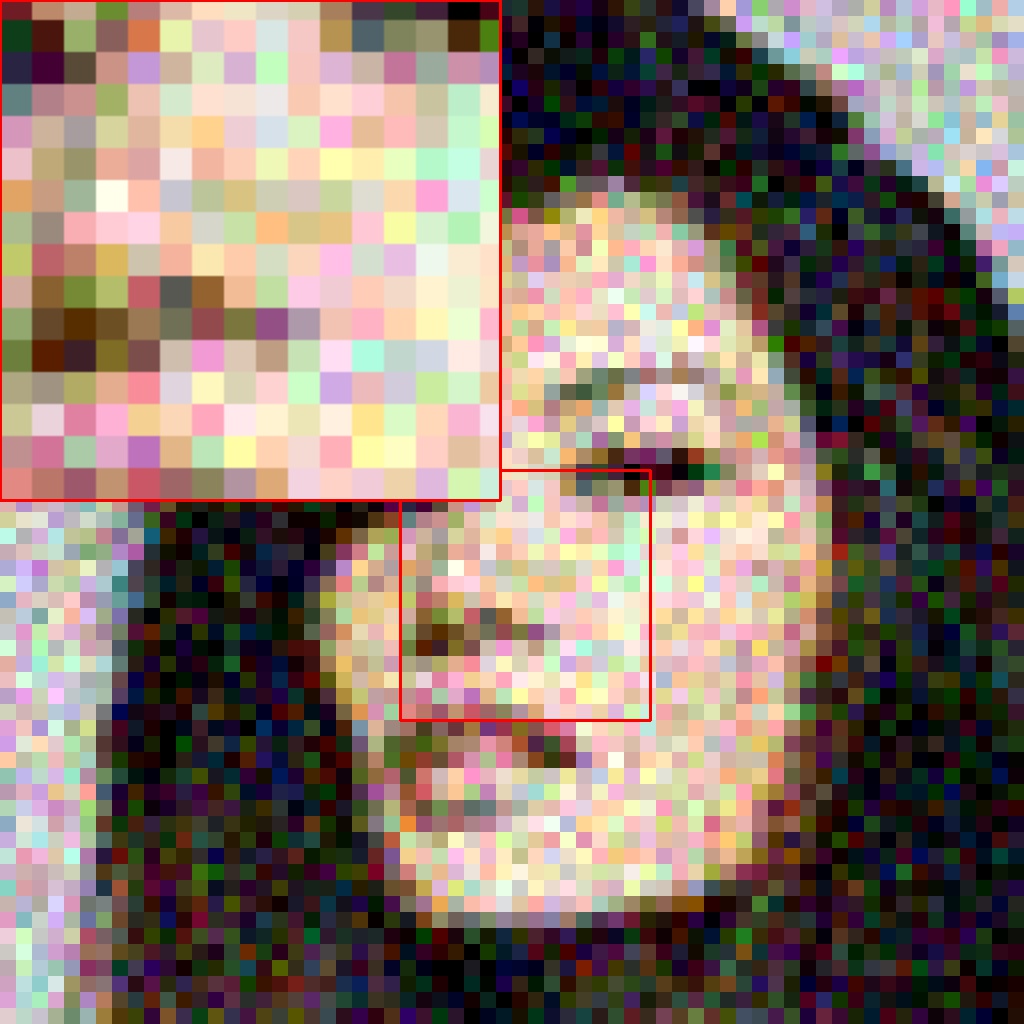}
			& \MyIm{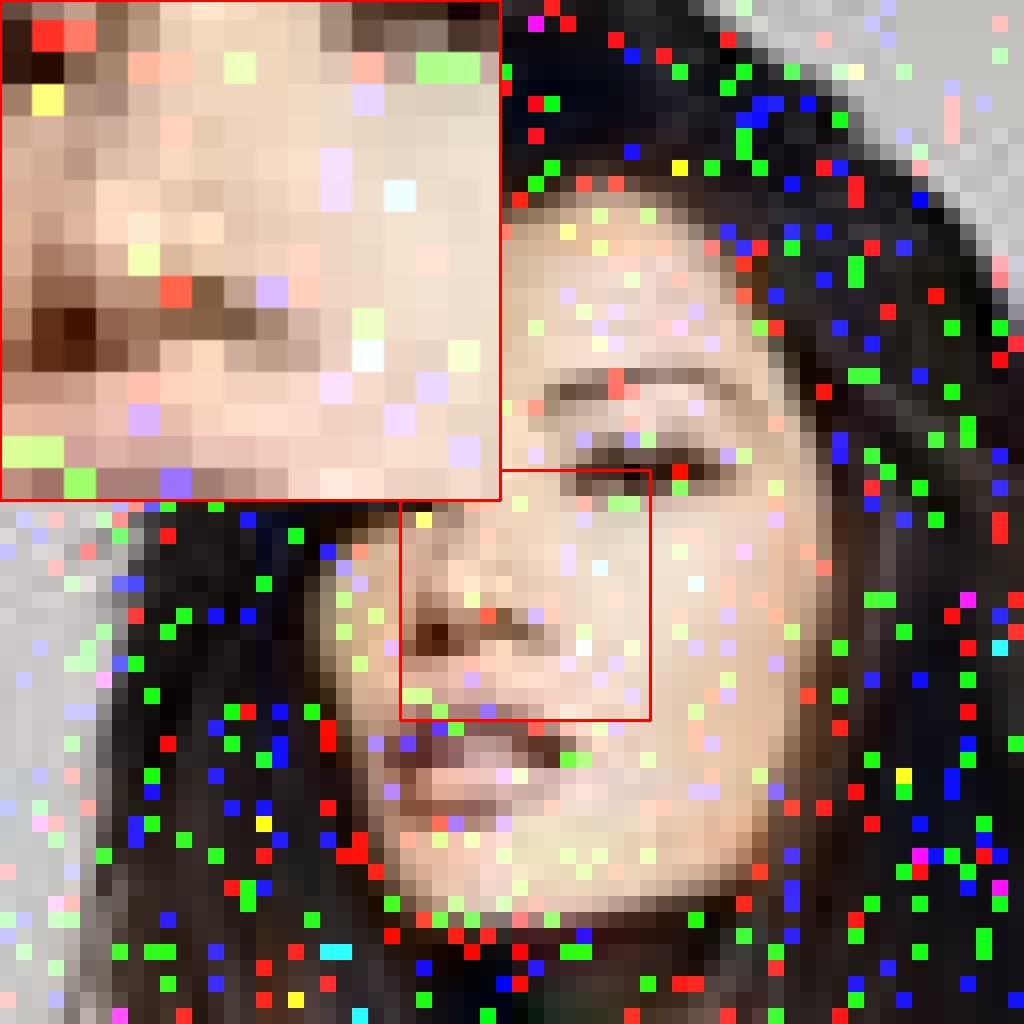} 
			& \MyIm{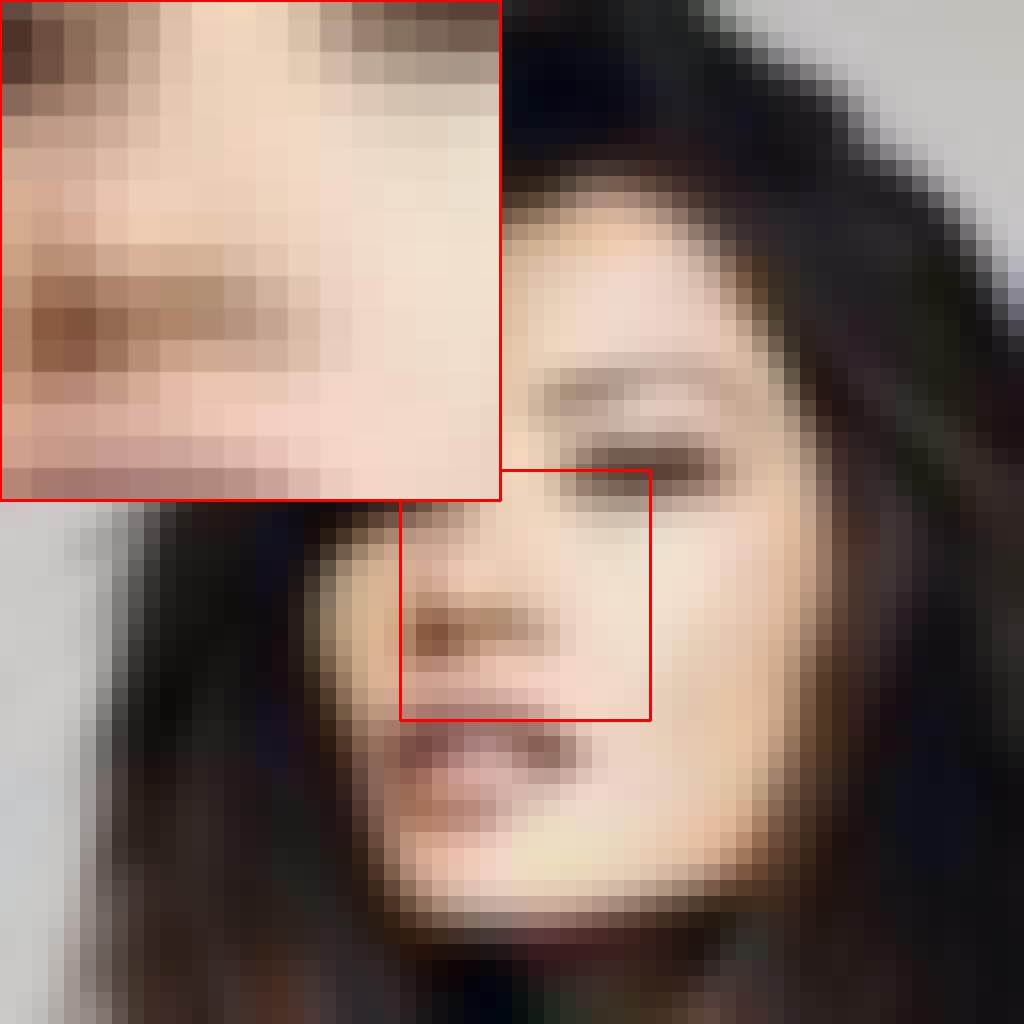} 
			& \MyIm{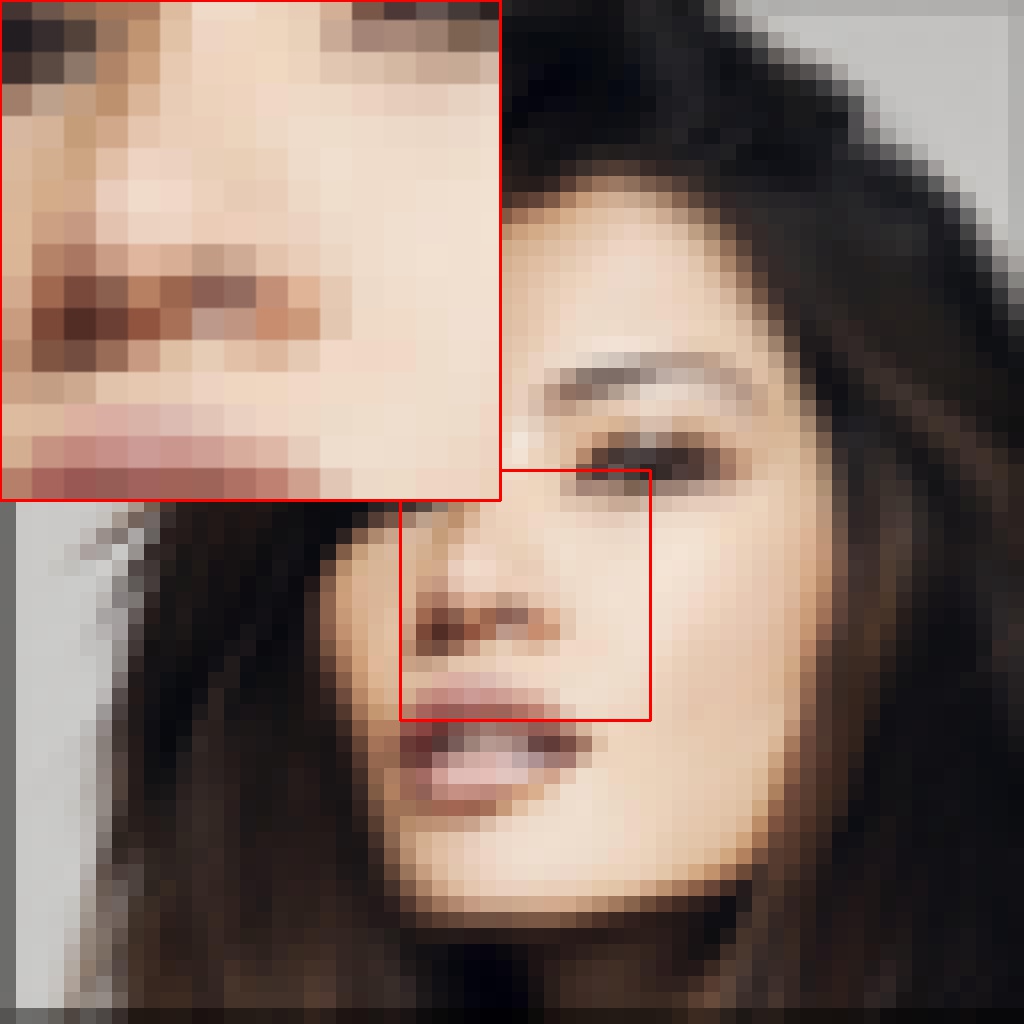}\\
			
			\MyIm{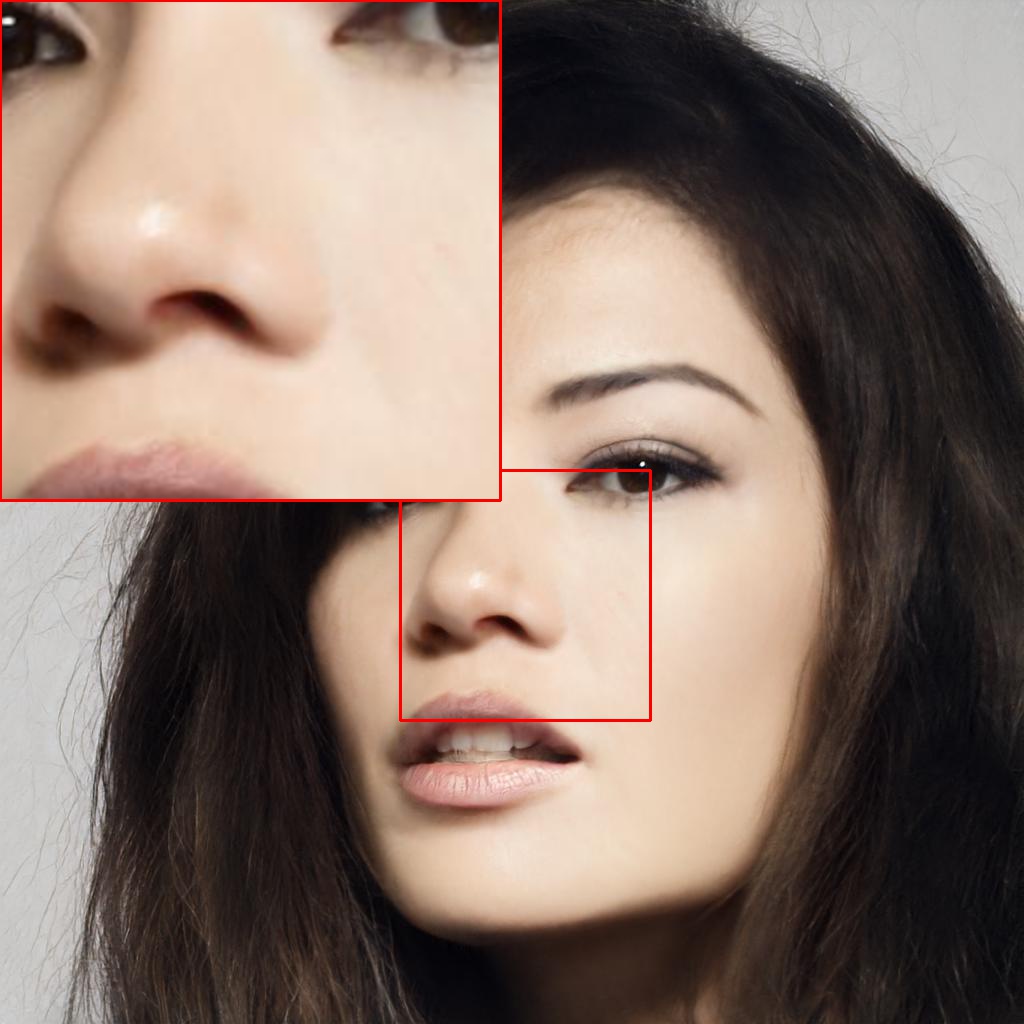}
			& \MyIm{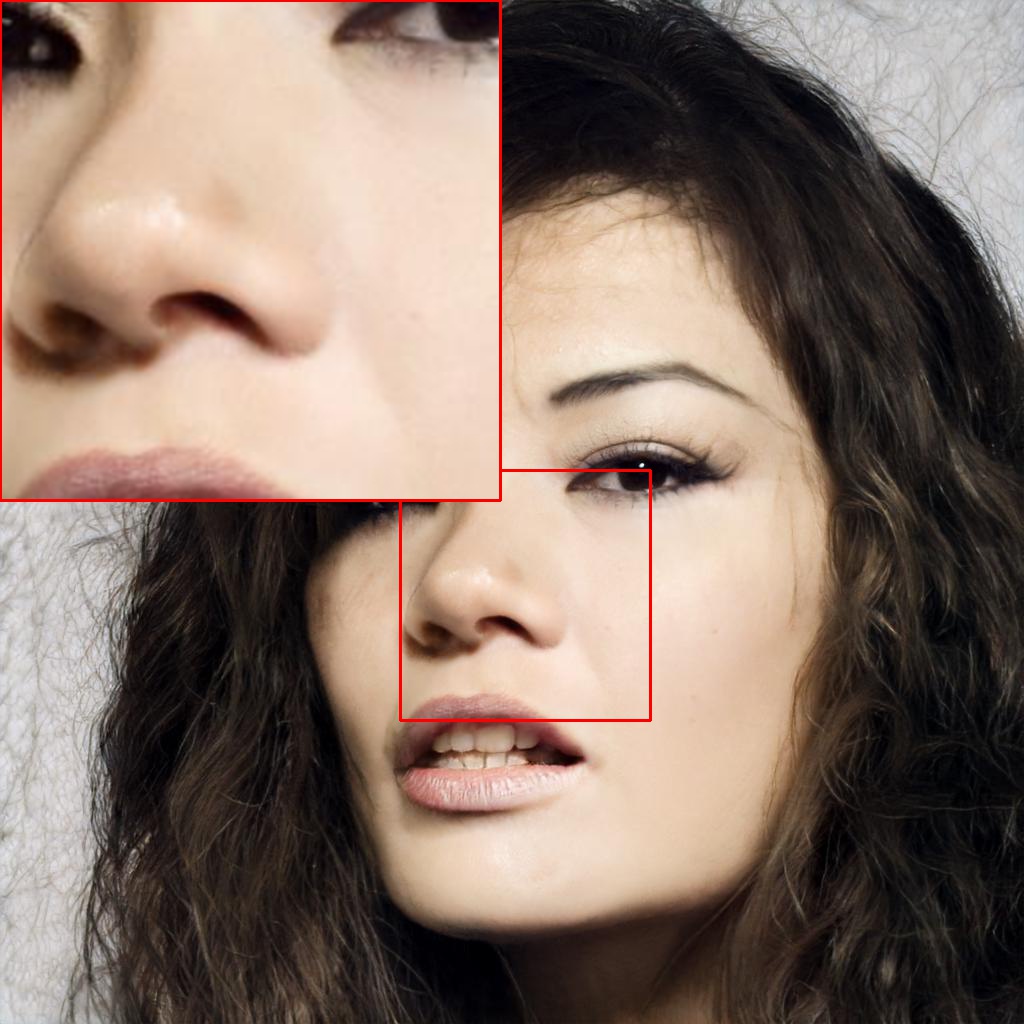} 
			& \MyIm{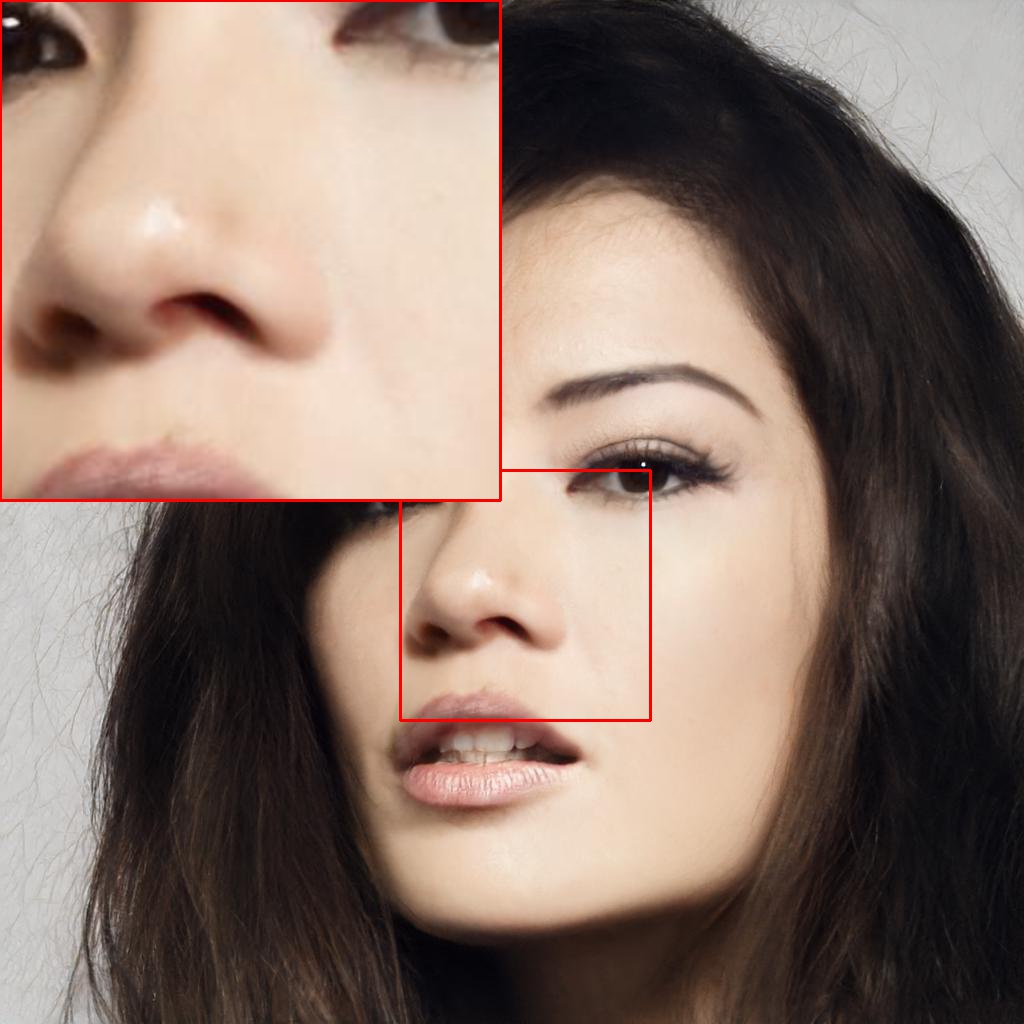} 
			& \MyIm{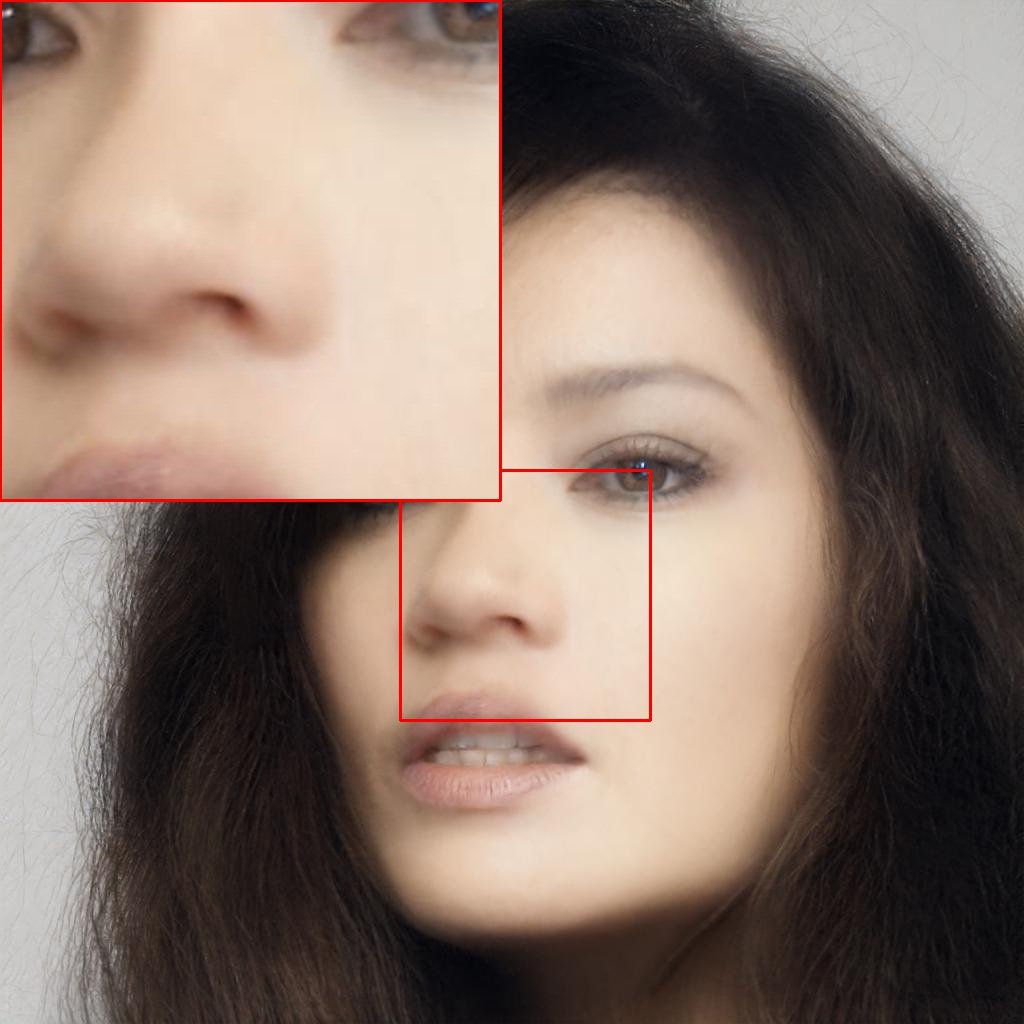}
			& \MyIm{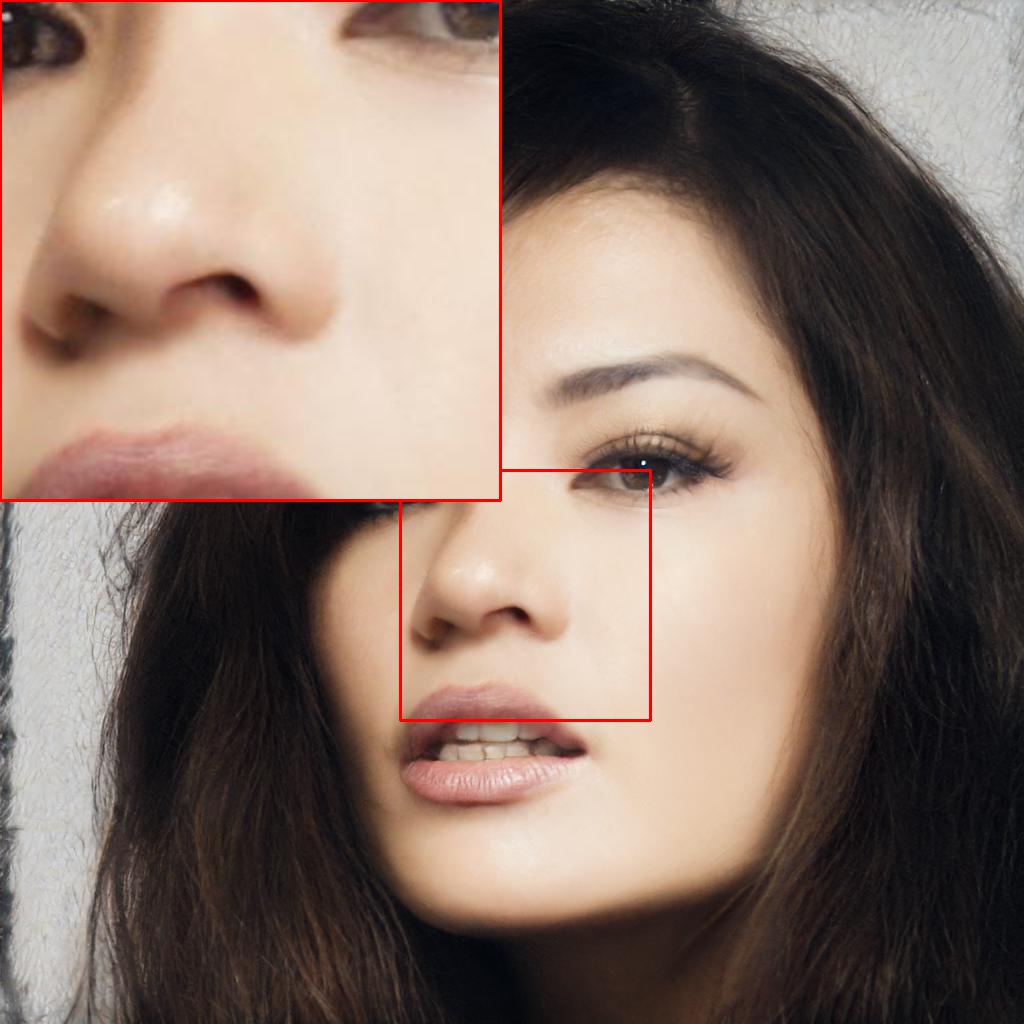}\\
		\end{tabular}
	\end{tiny}
	\caption{Robustness evaluation: degradation includes downscaling followed by corresponding operations, except for the last column, which consists of motion blur followed by downscaling (16x).}
	\label{fig:robustness}
\end{figure*}

\paragraph{Quality of the Gaussianization process.}
We generated 5000 random samples  $\z \sim \normal (\zero, \eye_\dim)$ and their associated style vector $\w=\G_m(\z)$, then gaussianized distribution of $\wspace$ by PULSE and our method. We then computed the squared norm for all of these samples (see \cref{fig:gaussian}). As expected, the squared norm of the standard gaussian distribution $\zspace$ approximately follows $||\z_n||^2_2 \sim \chi^2_\dim$ and thus forms a narrow distribution around $\dim=512$ while the squared norm of the untransformed distribution $\wspace$ does not, which is inconsistent with the prior assumption held by BRGM~\cite{marinescu2021bayesian}. Furthermore, PULSE\cite{menon2020pulse} appears to produce a wider squared norm distribution while ours approaches the squared norm distribution of $\zspace$ more closely.

\paragraph{Preservation of domain integrity.}
At higher magnification factors (e.g. 64x), it is essential to increase the weight of the image domain prior (e.g., faces) over maintaining LR consistency to avoid generating images that do not resemble a face. Whereas, for lower magnification factors (e.g. 8x), where low-resolution images already contain fine grain details, the focus should be on preserving those details and ensure the LR consistency.

The qualitative comparison in \cref{fig:qualitative} shows that at 64x magnification, competing methods fail to generate reasonable facial details. Optimization-based methods such as PULSE and BRGM tend to generate images that are accurately downscaled to the LR image but at the cost of generating distorted faces, thus moving away from the face image manifold. This could be attributed to a lack of proper regularization, leading to over-fitting to the input LR image.

Moreover, GPEN and GFPGAN extract multi-resolution features from the LR input image and use them to modulate the intermediate features of the pre-trained StyleGAN model. However, at higher magnification factors, the input image contains limited spatial information, leading to insufficient features to generate fine facial details. In comparison, both RLS and RLS$^+$ succeed to produce plausible and realistic faces by maintaining the overall structure of the face and generating visually accurate details such as eyes, eyebrows, teeth, mouth, and hair, among others.
It is worth noting that while GPEN and GFPGAN yield similar outcomes at low magnifications (8x), they may generate unrealistic facial features when dealing with higher magnification factors. Additional examples can be found in Appendix D.

To quantitatively assess this gain in performance, we use Frechet Inception Distance (FID)~\cite{heusel2017gans} and Kernel Inception Distance (KID)~\cite{binkowski2018demystifying} to measure the discrepancy between the real HR face images and the reconstructed one. We also employ Natural Image Quality Evaluator (NIQE) ~\cite{mittal2012making} to evaluate the naturalness of reconstructed images. As expected, the scores of FID, KID, and NIQE listed in \cref{tab:quantitative} show that RLS$^+$ improves realism at both large and small magnifications, with improvements especially noticeable at higher magnification factors.

While even on the 8x magnification factor, the baselines still suffer from the realism-fidelity trade-off, RLS$^+$ performs significantly better both at reproducing the distribution of real HR face images and at producing images that have better perceptual quality. Thus, RLS$^+$ does not compromise realism despite modulating the generator. This is in accordance with our strategy that imposes the image to belong to a given domain.

Note that, with 8x magnification, even though RLS achieves the best NIQE scores (showing that the outputs look natural), which is also consistent with the visual results, it is almost the worst in terms of FID and KID.
This is explained by the fact that FID and KID measure the distance between the distributions of real HR images and the reconstructed images, whereas RLS aims to produce a plausible image from the original image domain the StyleGAN was trained on.

Although the super-resolution task is ill-posed and has many plausible solutions, we evaluated the reconstruction quality using perceptual LPIPS \cite{zhang2018unreasonable}, PSNR, and MS$-$SSIM \cite{wang2003multiscale}. The scores in \cref{tab:quantitative} indicate that RLS$^+$ may not perform the best in terms of these metrics at low magnification factors, but it shows significant improvements in reconstruction quality at high magnification factors.

It is worth noting that the naive bicubic interpolator achieves high PSNR and MSSIM scores, but it fails to restore facial details, demonstrating that PSNR and SSIM are inadequate metrics for measuring super-resolution tasks. Moreover, RLS$^+$ achieves low LPIPS scores on both scales, indicating that the generated images are perceptually close to the ground-truth. 
\begin{figure*}[!h]
	\centering
	\includegraphics[width=0.9\linewidth]{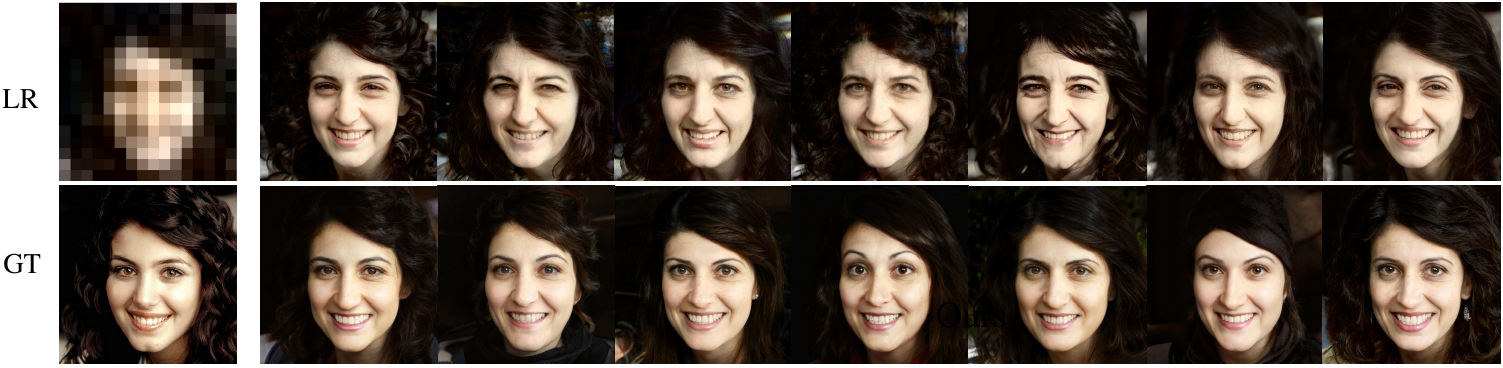}
	\caption{Generating multiple solutions for a given LR image. Apart from the first column that displays Low Resolution and Ground Truth images, top row is PULSE and bottom row is RLS (64x).}
	\label{fig:diversity}
\end{figure*}
Using a pre-trained face recognition model, CurricularFace~\cite{huang2020curricularface}, we measure the identity-similarity between the ground-truth and reconstructed images.
With the 64x magnification factor, RLS reconstructs realistic images resembling the ground-truth, which is improved slightly by RLS$^+$. Note that, for high magnification factors we do not expect the output to perfectly match the ground-truth image as there are many plausible outputs. 

\cref{tab:quantitative appendix} also presents the results of our approach on a 16$x$ super-resolution task and compares it with two baselines that achieved similar results on an 8x task, as shown in \cref{tab:quantitative}. The table demonstrates that while our approach produces fidelity values (such as LPIPS and PSNR) that are comparable to the baselines, it significantly improves realism (as indicated by FID and KID).

\begin{figure*}[!h]
	\centering
	\setlength{\tabcolsep}{0pt}
	\begin{footnotesize}
		\begin{tabular}{FFFFF}
			LR & 
			GT & GPEN~\cite{yang2021gan} & GFPGAN~\cite{wang2021towards} & RLS$^+$\\
			
			\MyImbig{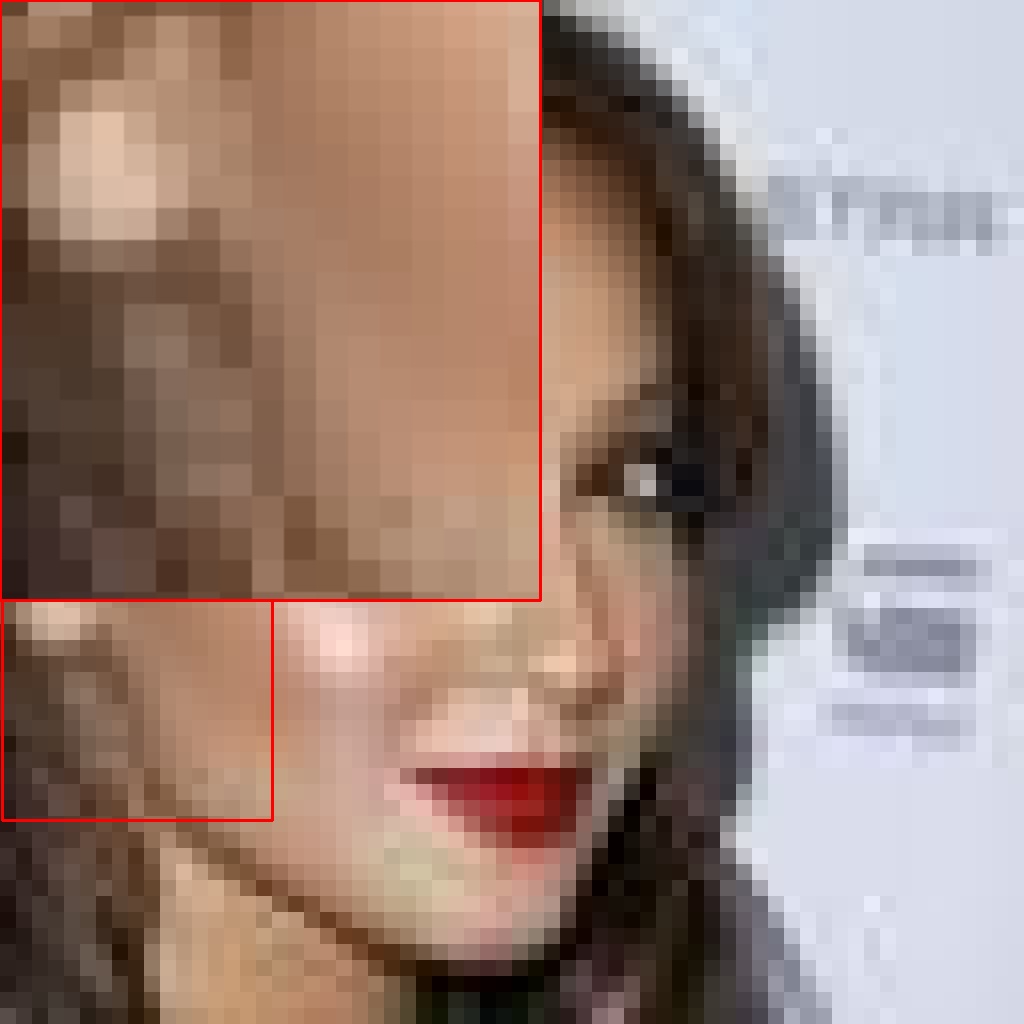} & \MyImbig{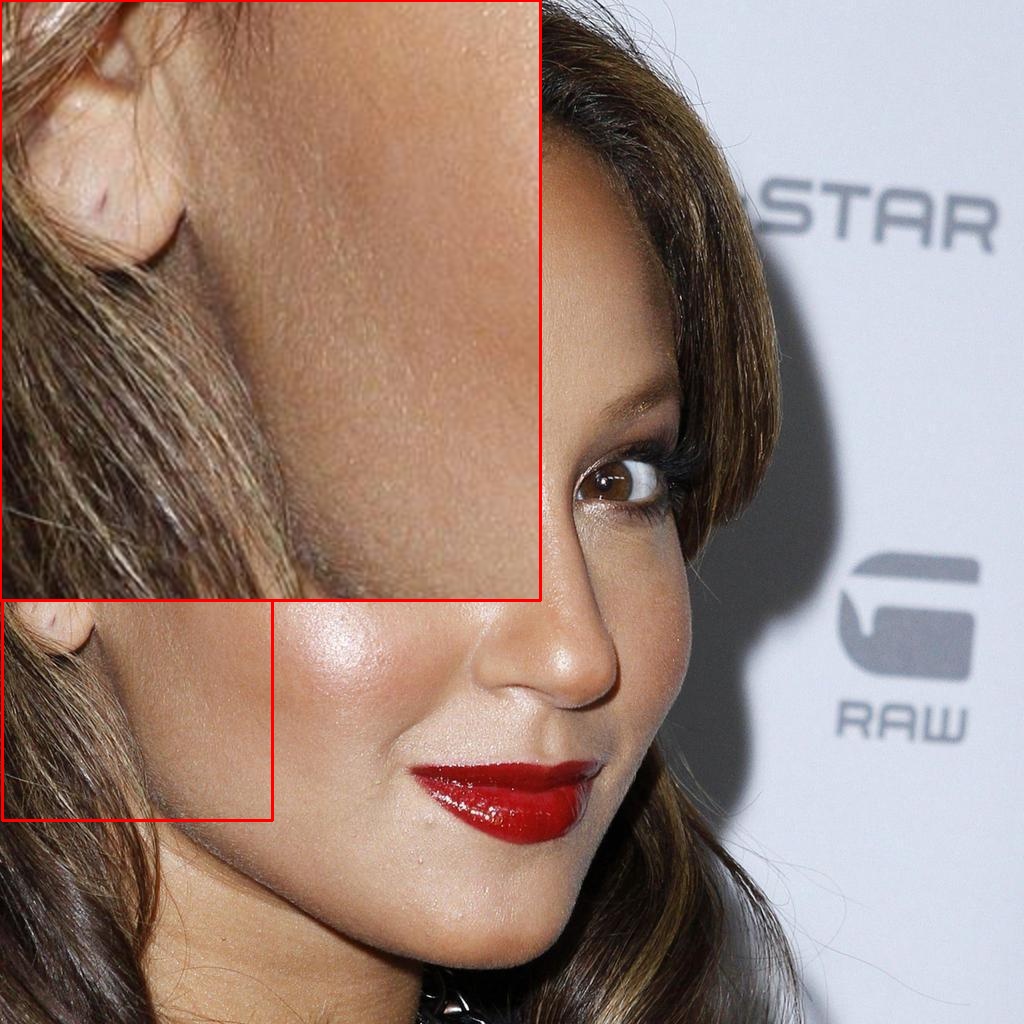}
			& \MyImbig{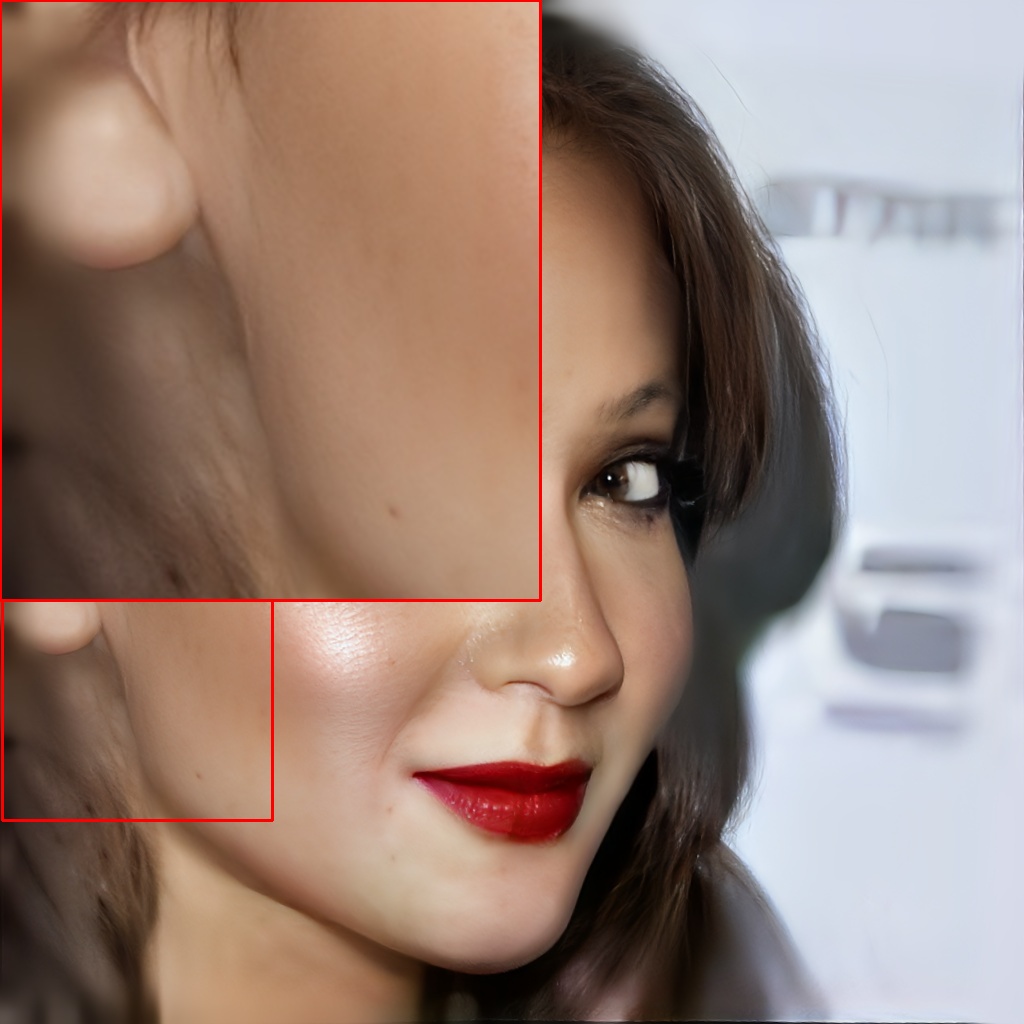} & \MyImbig{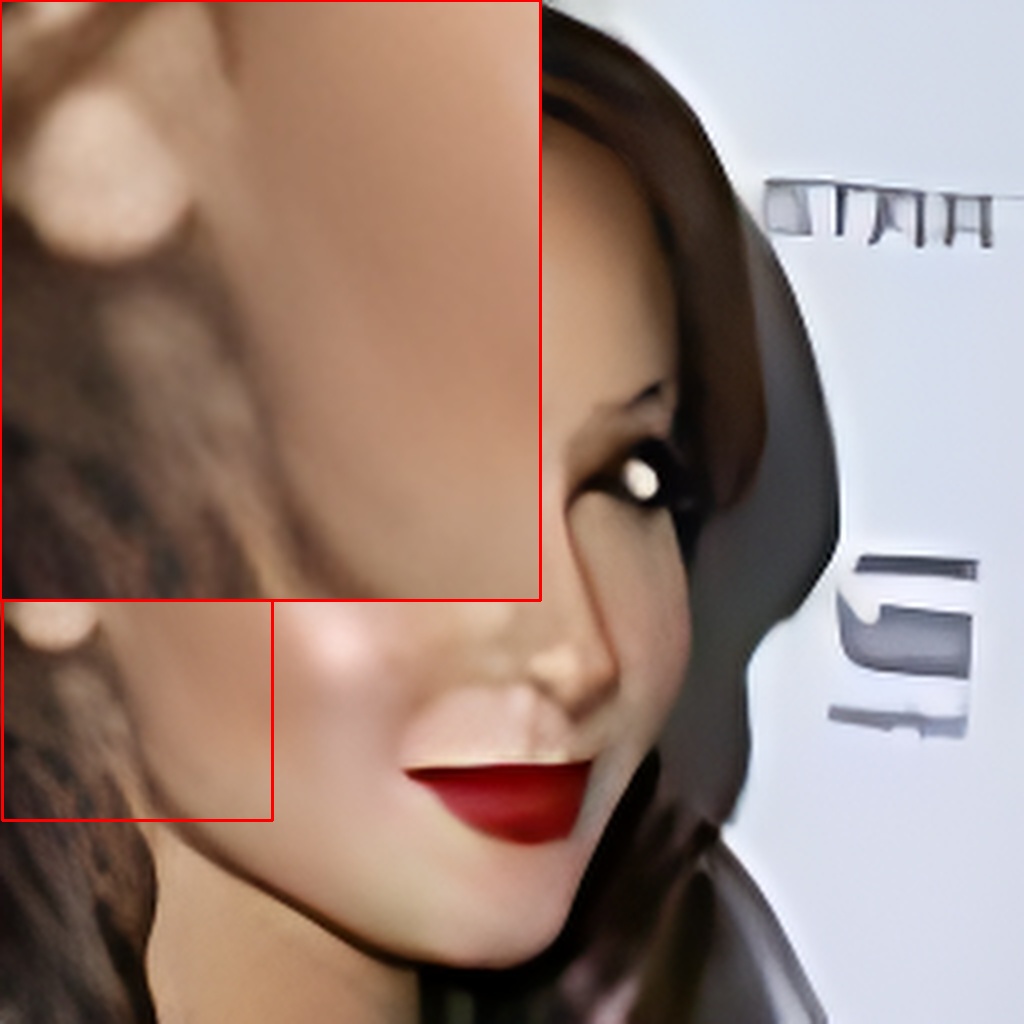}
			& \MyImbig{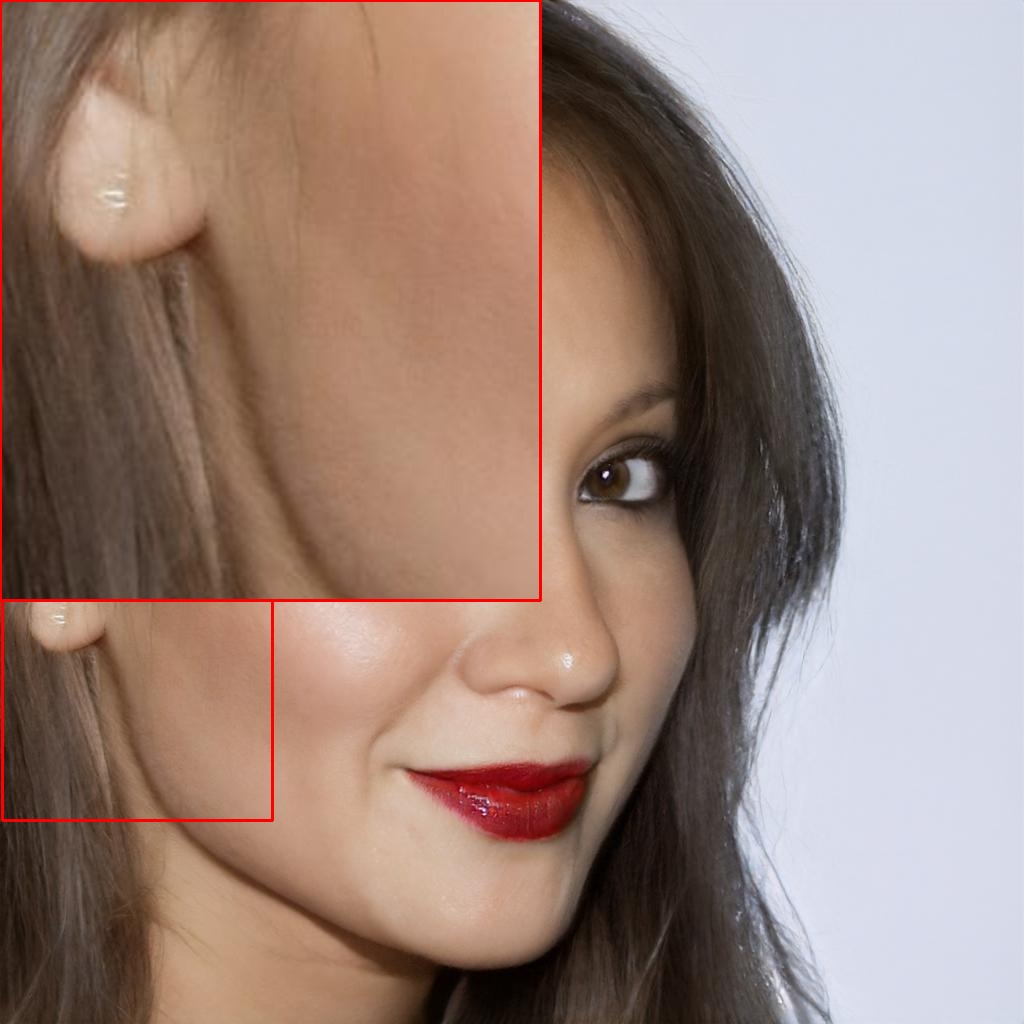}\\
			
			\MyImbig{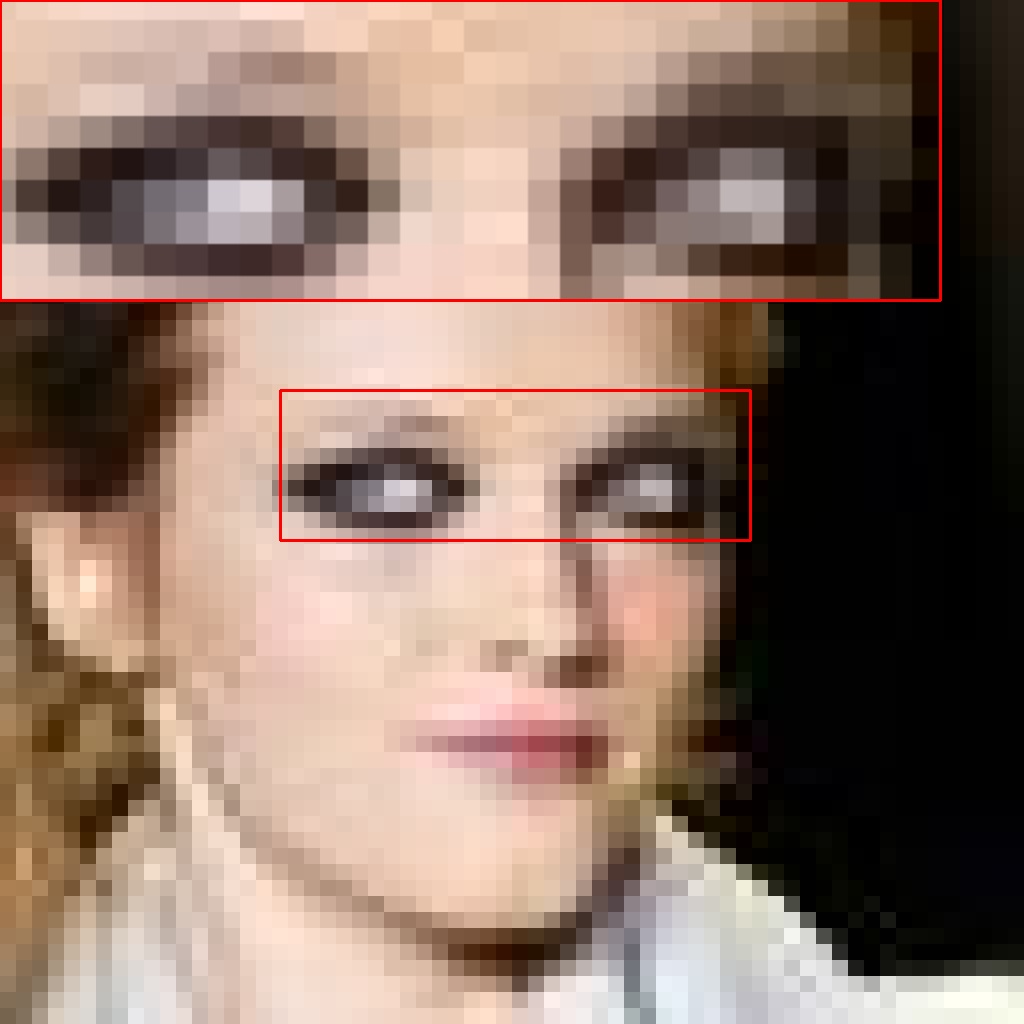} & \MyImbig{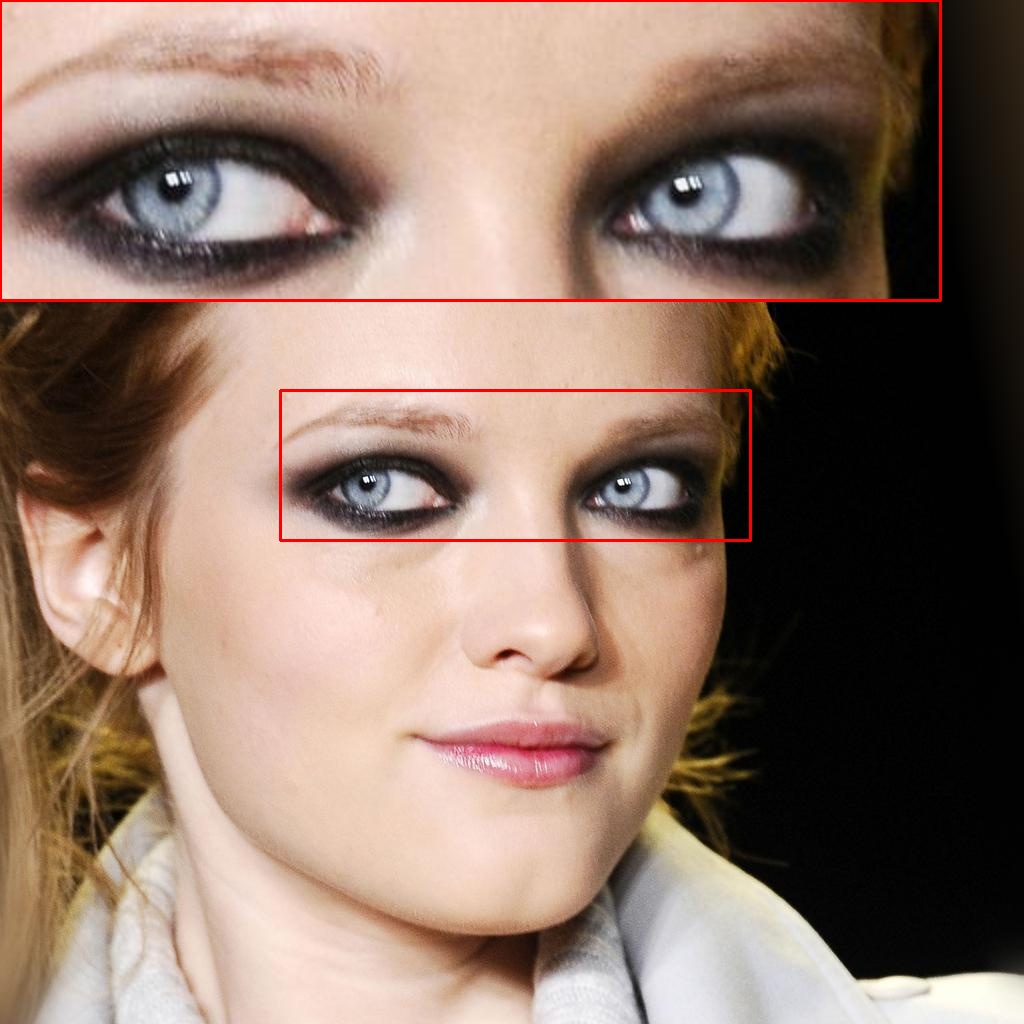}
			& \MyImbig{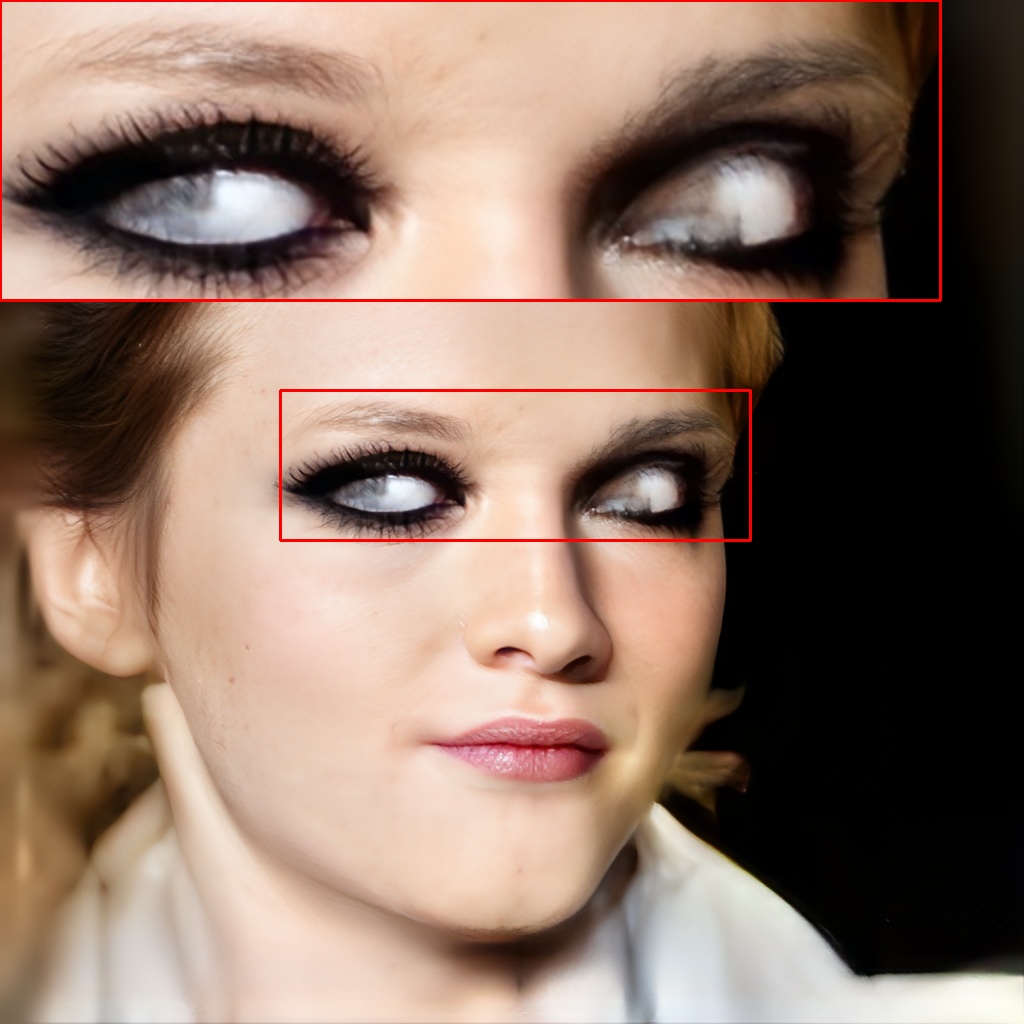} & \MyImbig{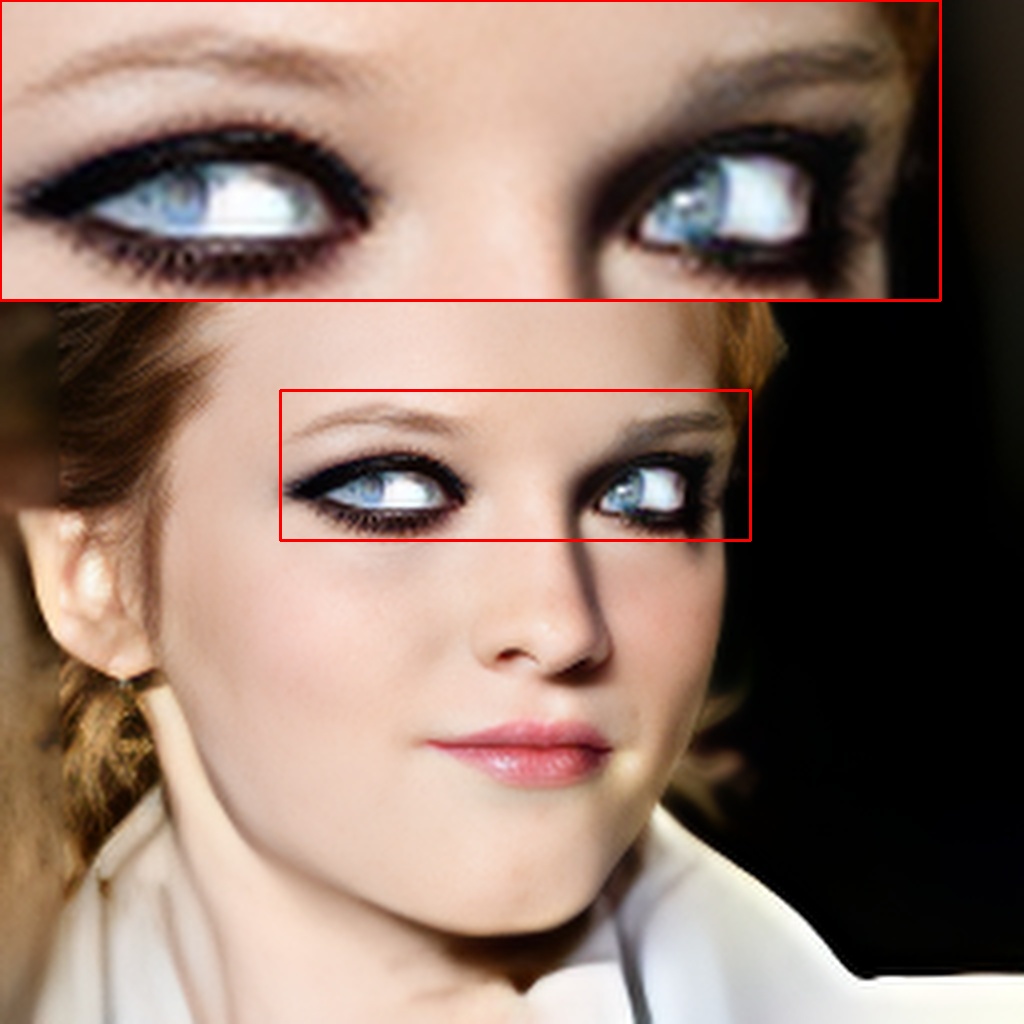}
			& \MyImbig{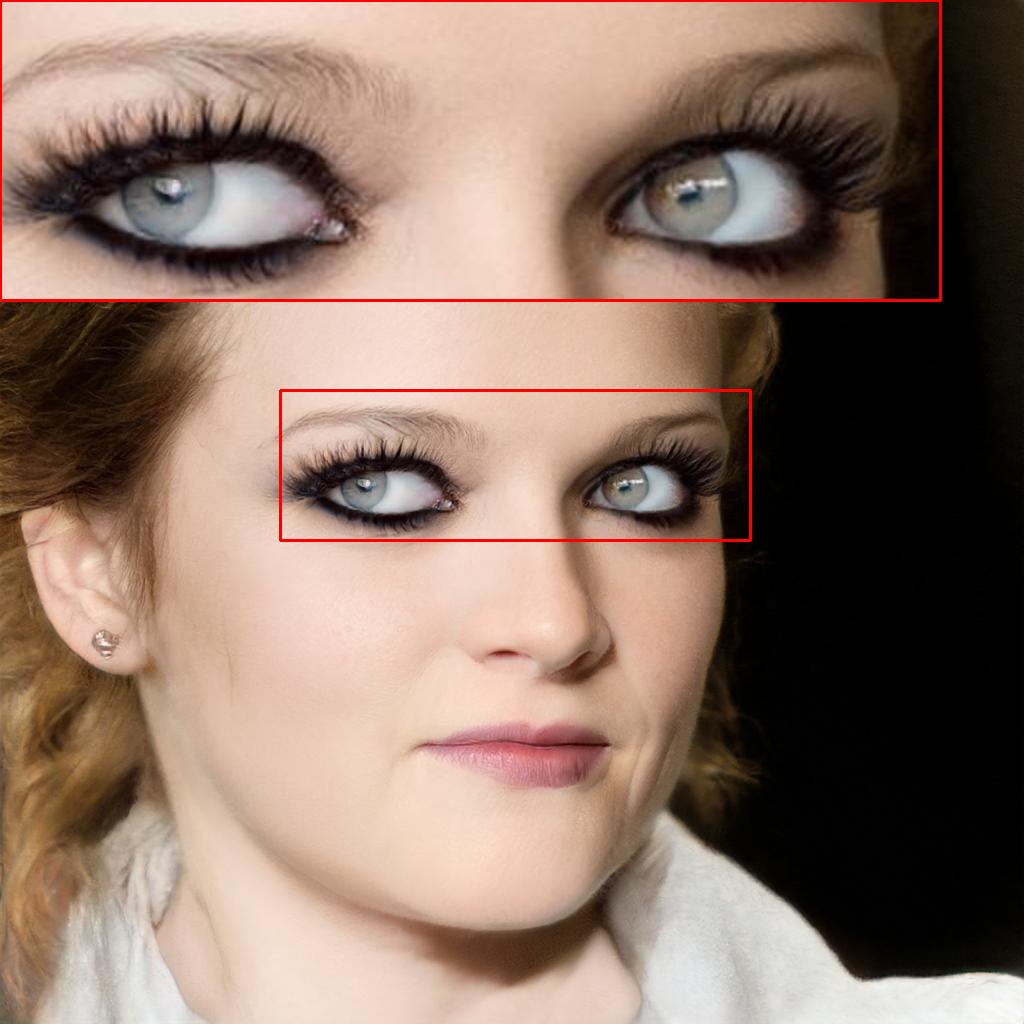}\\
			
			\MyImbig{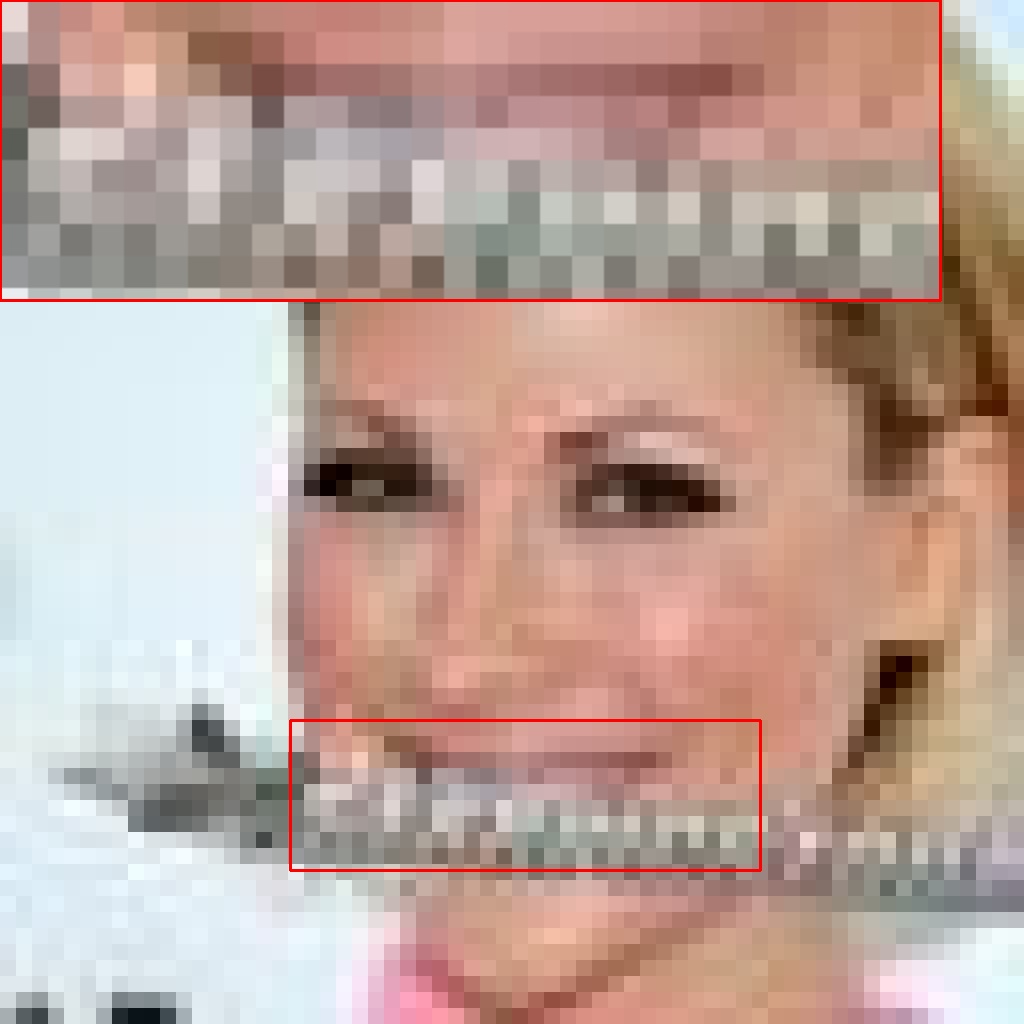} & \MyImbig{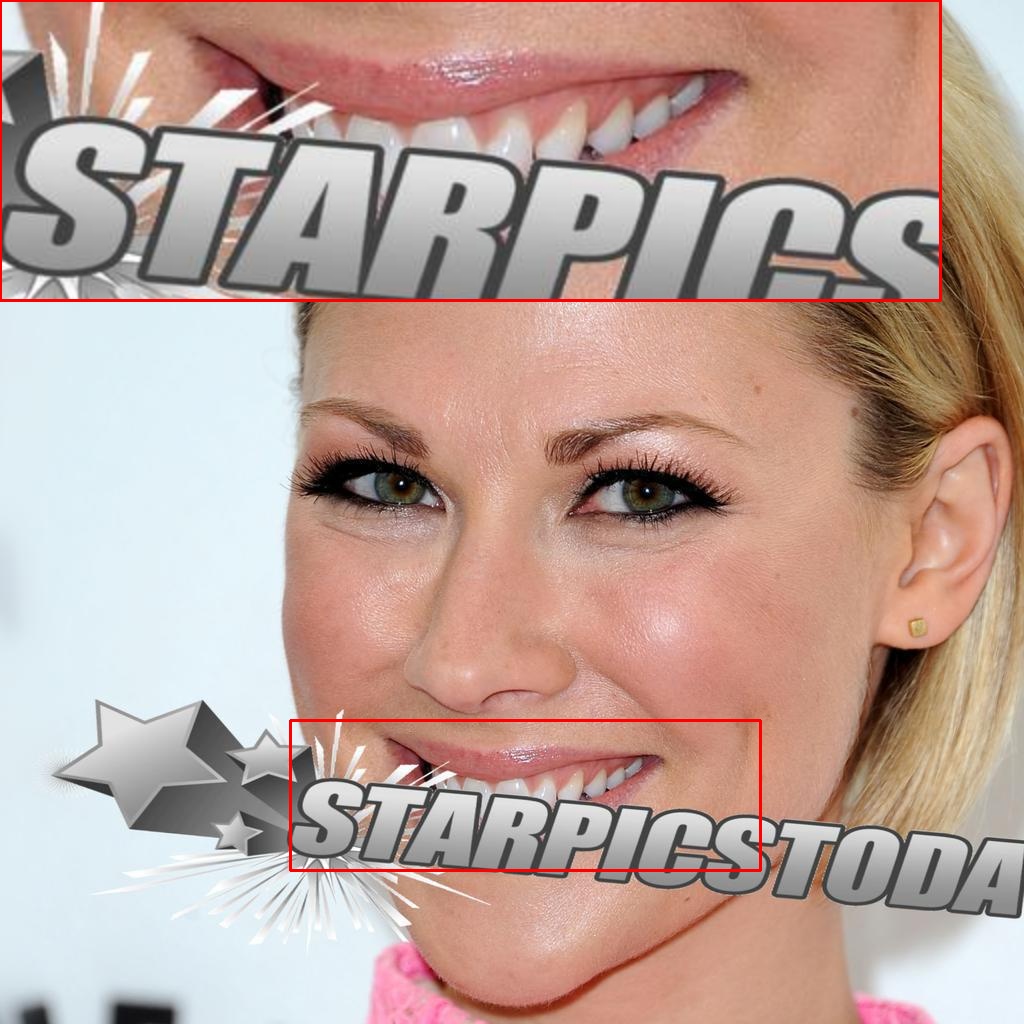}
			& \MyImbig{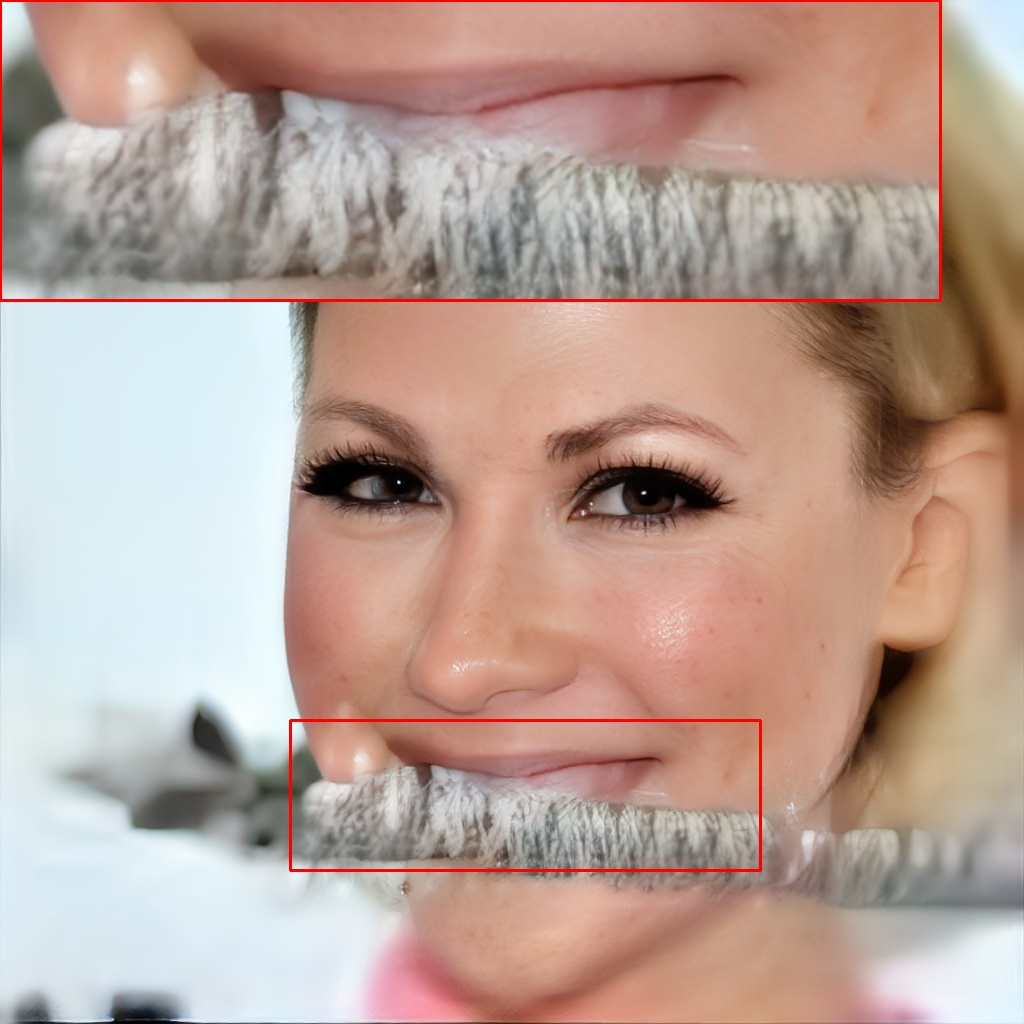} & \MyImbig{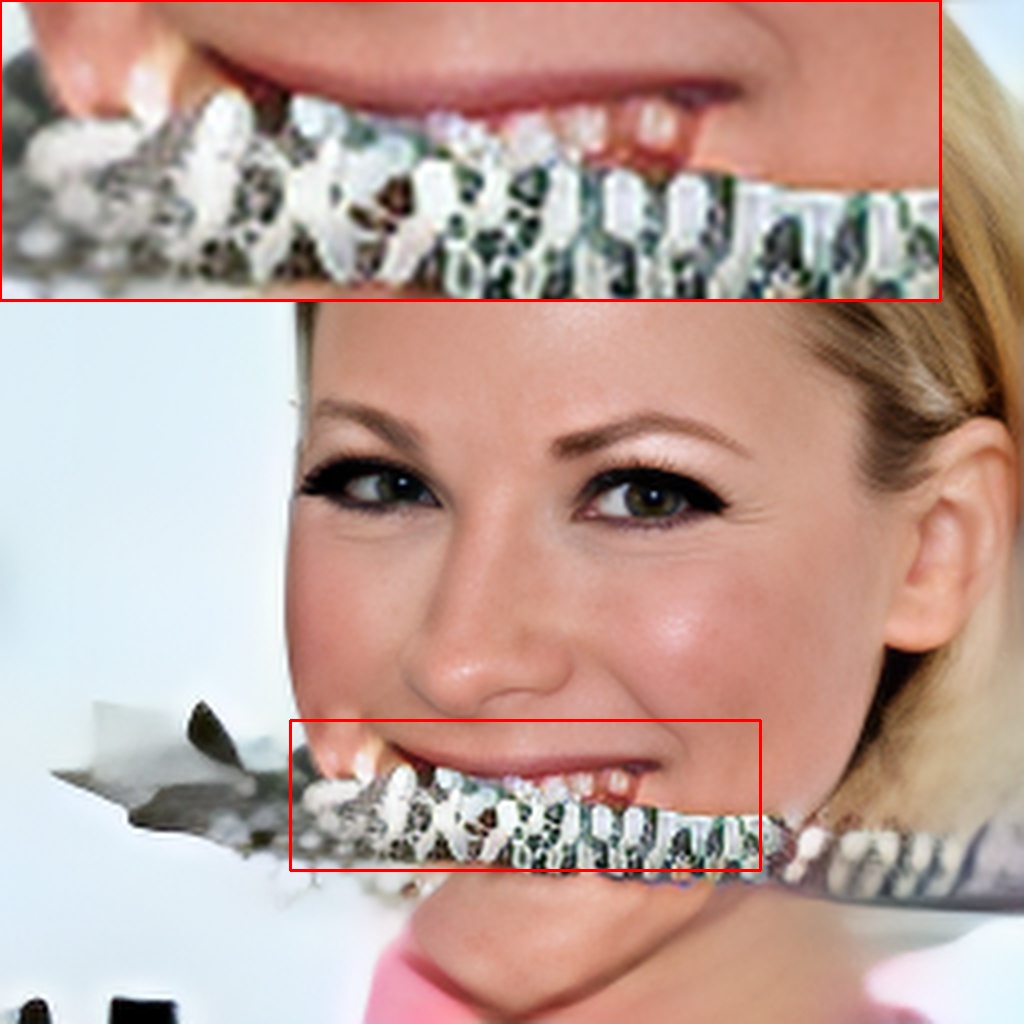}
			& \MyImbig{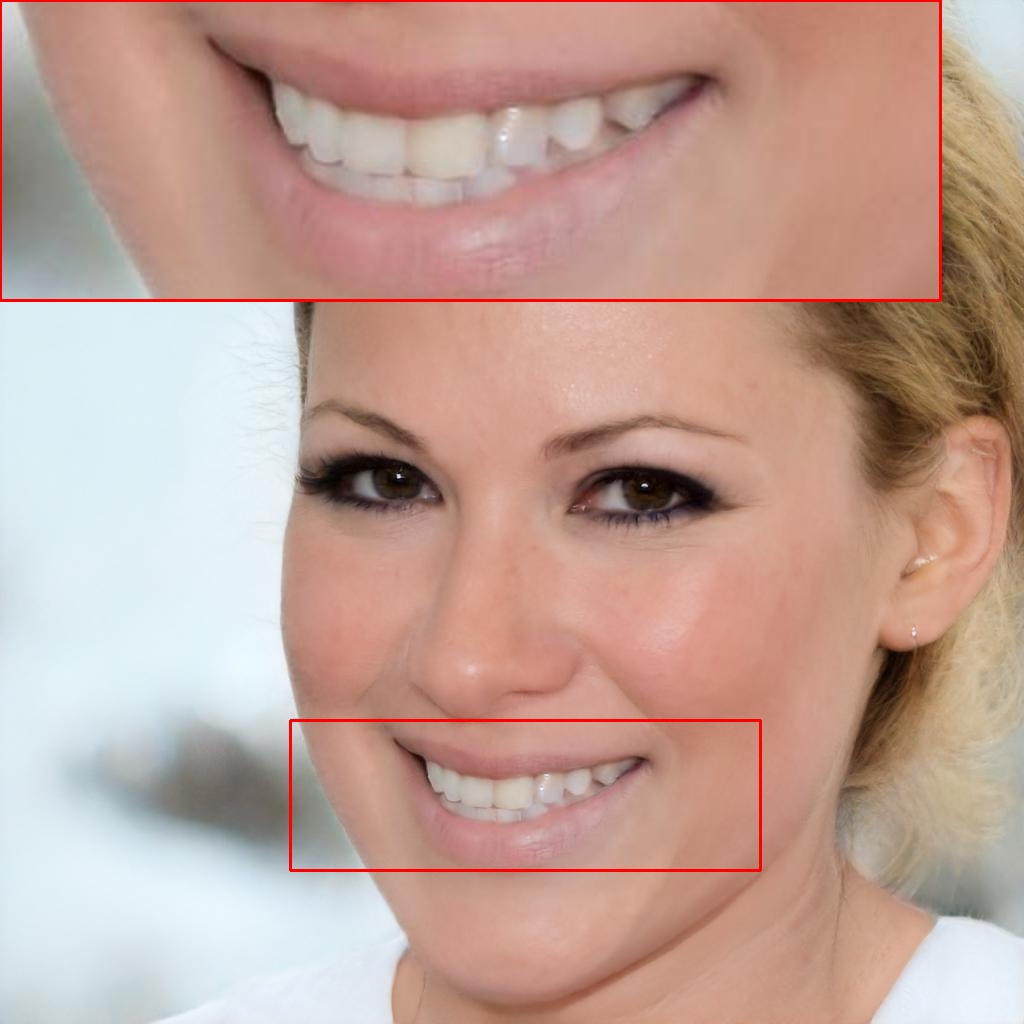}\\
			
			
			
			\MyImbig{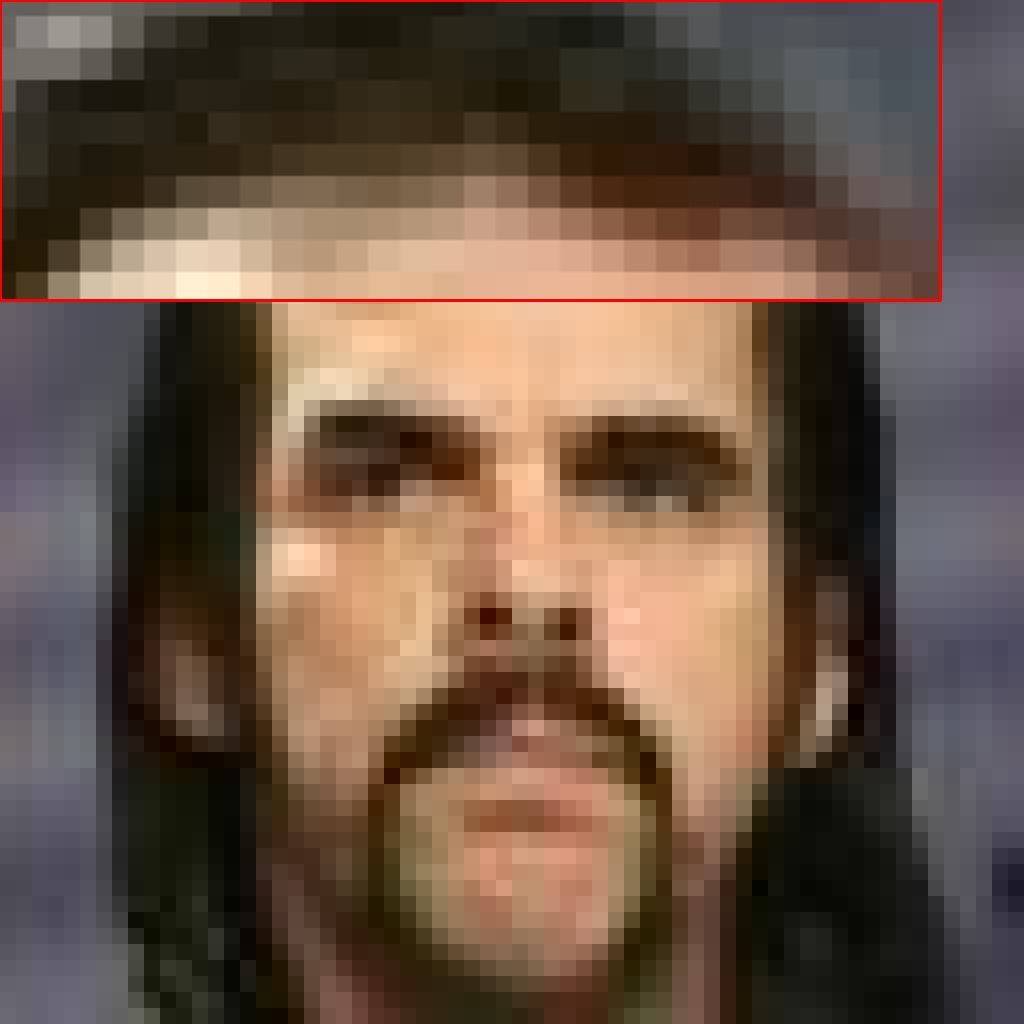} & \MyImbig{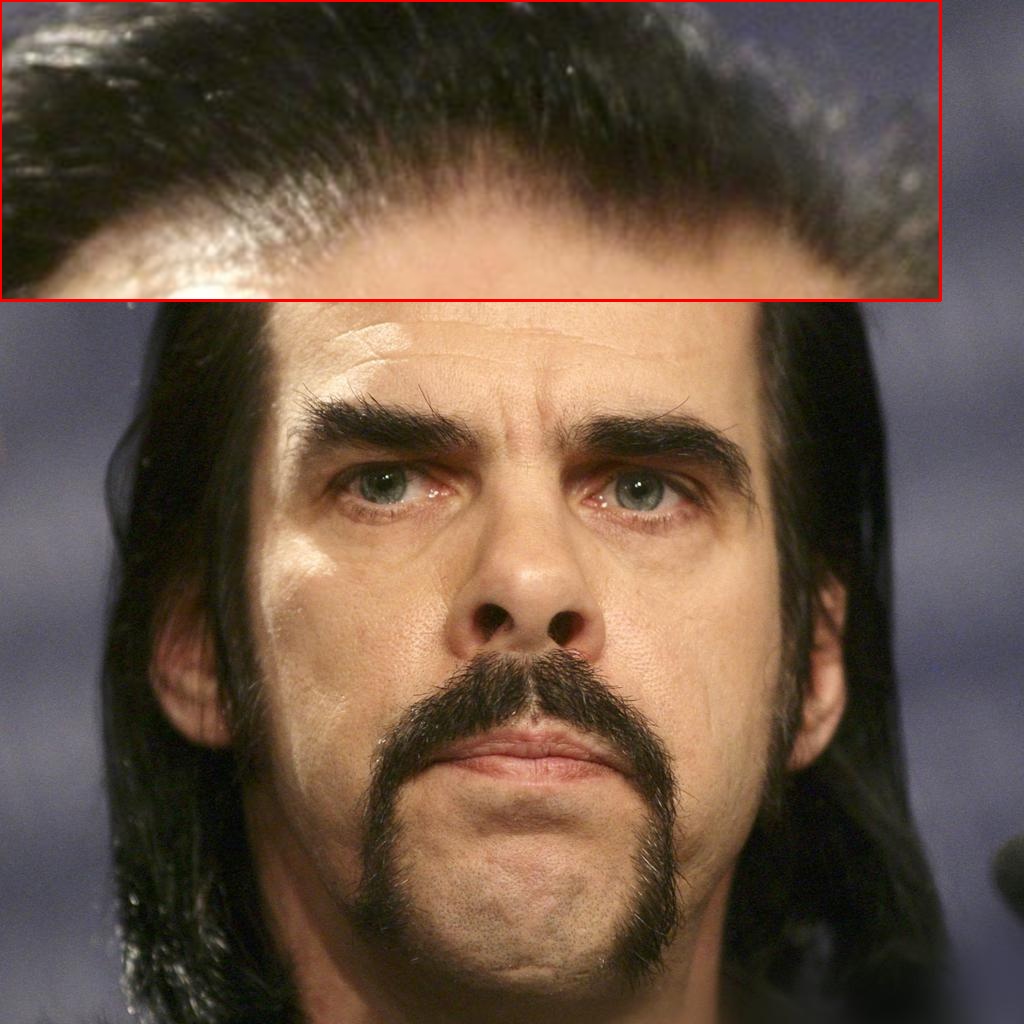}
			& \MyImbig{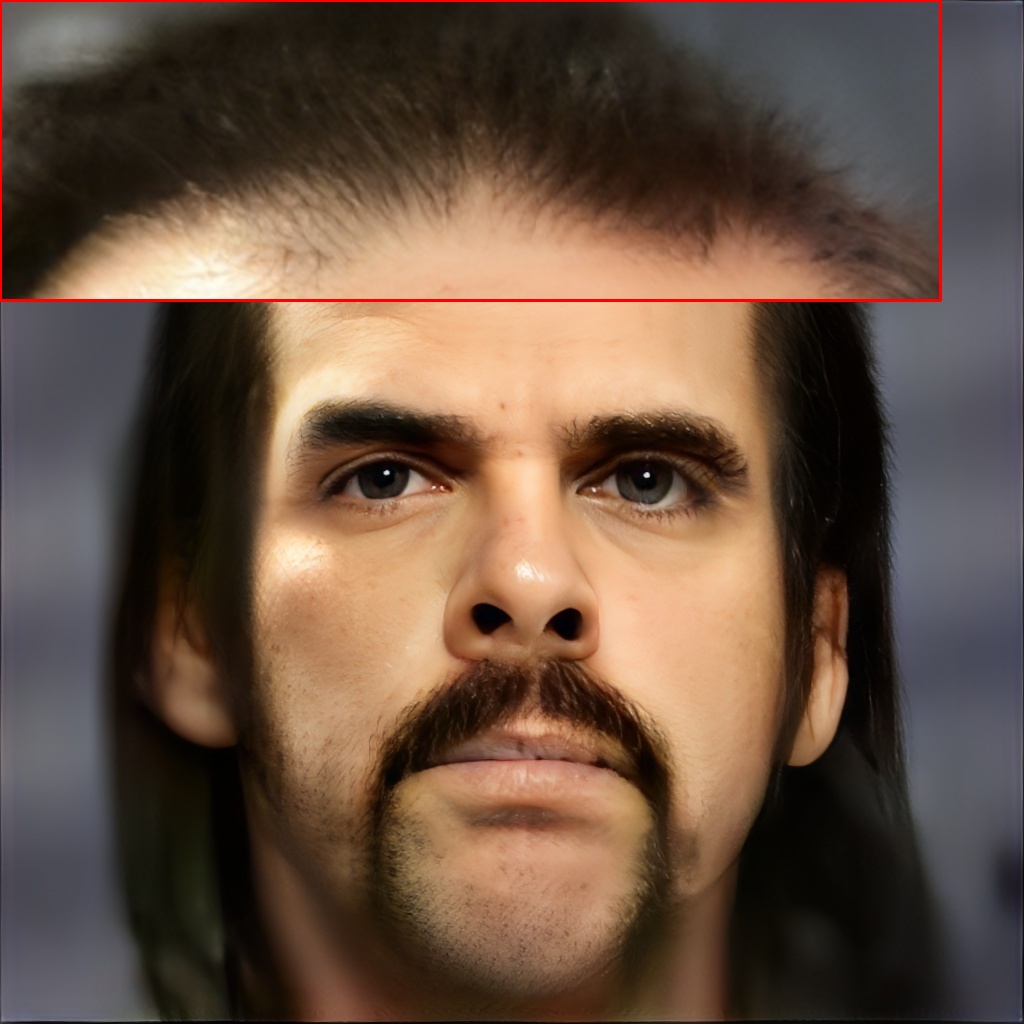} & \MyImbig{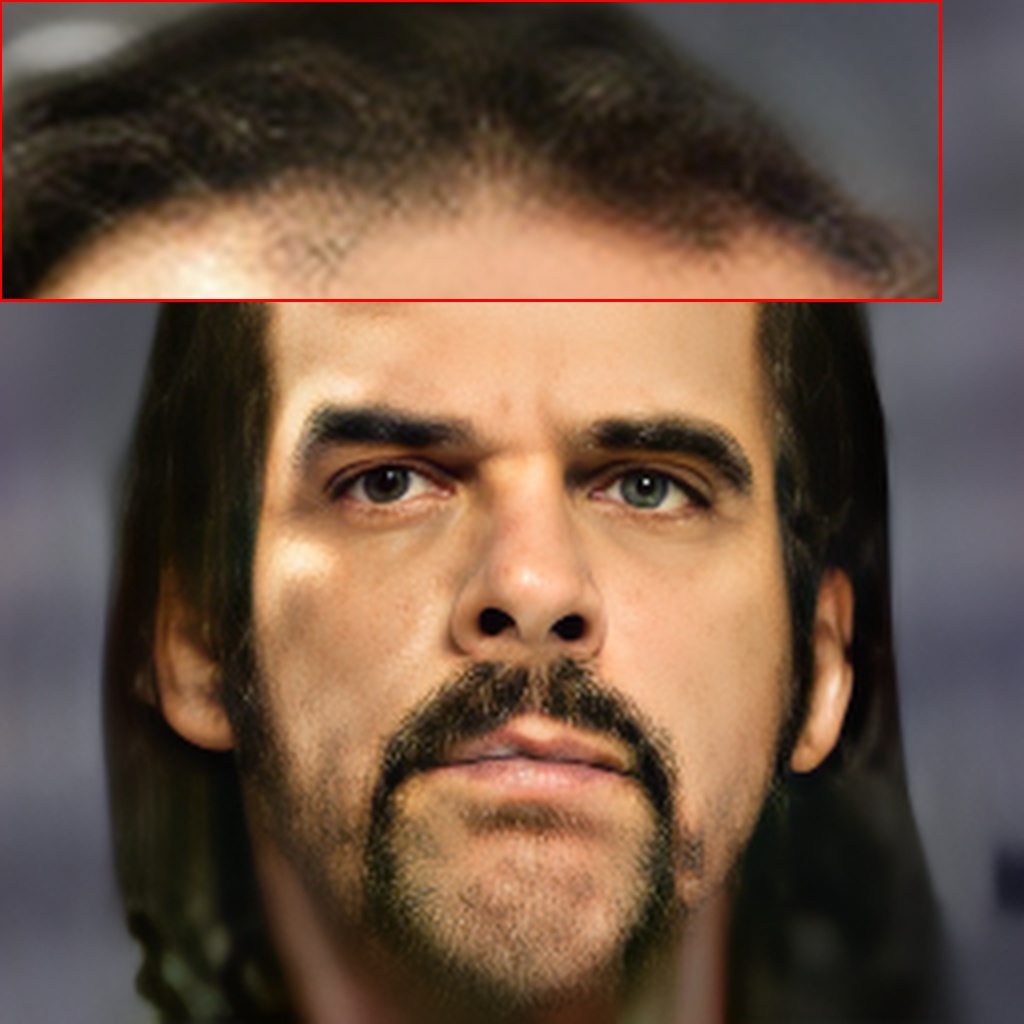}
			& \MyImbig{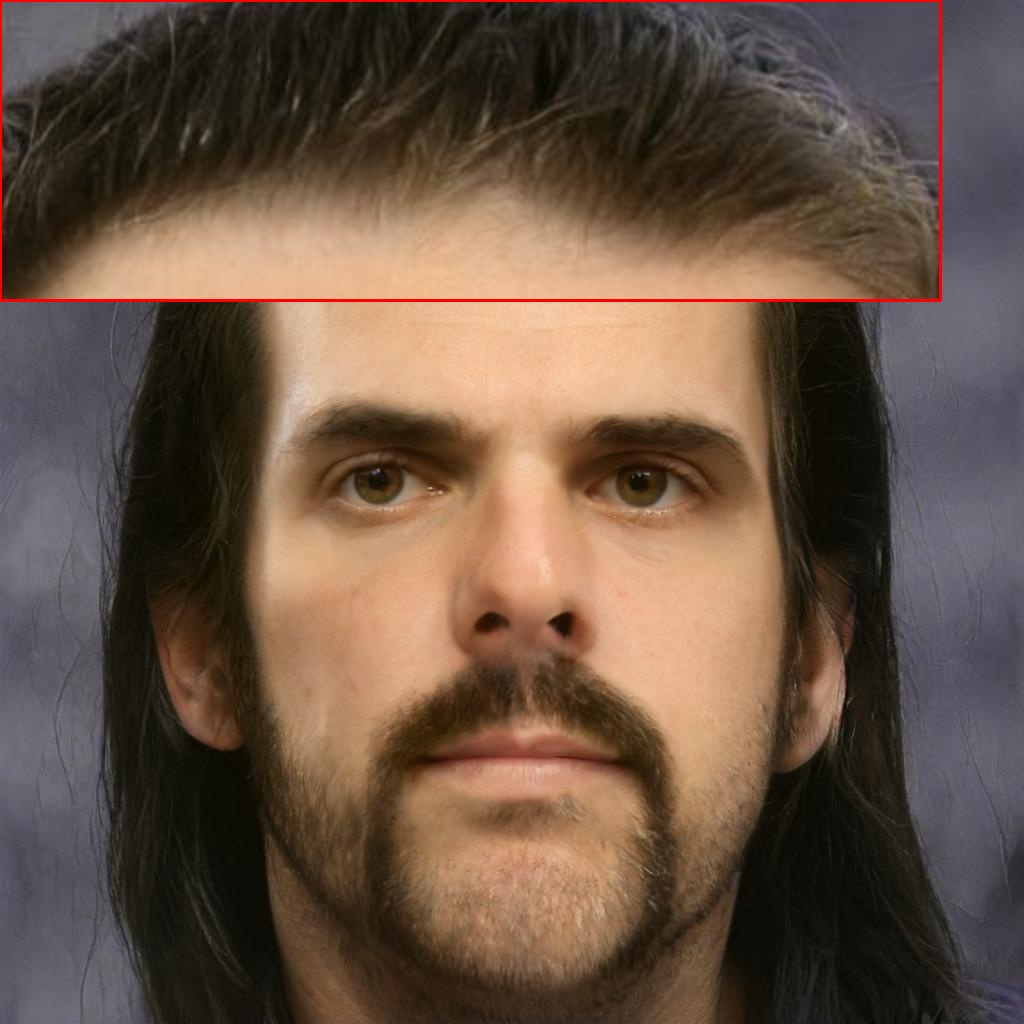}\\
			
			
		\end{tabular}
	\end{footnotesize}
	\caption{Comparison of reconstructions of a high resolution face from CelebA (16x). (Zoom-in for best view)}
	\label{fig:qualitative_app}
\end{figure*}

Additionally, we provide further examples that compare the output of RLS$^{+}$ with the baseline. These examples include out-of-domain images with challenging features such as extreme poses, heavy makeup, or occluded faces. \cref{fig:qualitative_app} shows that our approach can faithfully reconstruct images with challenging features, whereas the baselines struggle with such images. It produces realistic facial details and accurately preserves the individual's facial features.

\setlength{\tabcolsep}{4.5pt}
\begin{table}[!h]
	\centering	
	\begin{tabular}{@{}l|ll|ll@{}}
		\toprule
		Method & FID$\downarrow$ & KID$^{({\times10^3})}$$\downarrow$ 
		& LPIPS$\downarrow$ &PSNR$\uparrow$\\
		\hline
		GPEN~\cite{yang2021gan} & 36.0024 & 29.5146 
		& 0.3945 & 25.0845\\
		
		GFPGAN~\cite{wang2021towards} & 29.7925 & 18.3028 
		& 0.4221 & 24.2659\\
		
		RLS$^{+}$ & 26.3786 & 13.7664 
		& 0.3972 & 23.5242\\
		
		\bottomrule
	\end{tabular}
	\caption{Quantitative comparison on CelebA for 16x super-resolution.}
	\label{tab:quantitative appendix}
\end{table}

\setlength{\tabcolsep}{5pt}
\begin{table}[!h]
	\centering	
	\begin{tabular}{@{}l||lllll@{}}
		\toprule
		Method & NIQE$\downarrow$ & ID$\uparrow$ & LPIPS$\downarrow$ & MSSIM$\uparrow$ \\
		\hline
		RLS$^+$ & 4.2975 & 0.7976 & 0.4004 & 0.7223\\
		Gaussian Noise & 4.0545 & 0.7056 & 0.4390 & 0.6675\\
		Salt and Pepper & 4.2427 & 0.8045 & 0.4036 & 0.7215\\
		Gaussian Blur & 4.4677 & 0.7901 & 0.4414 & 0.6819\\
		Motion Blur & 4.3160 & 0.8077 & 0.4237 & 0.6821\\
		\bottomrule
	\end{tabular}
	\caption{Quantitative evaluation of robustness on 1000 images of CelebA (16x)}
	\label{tab:robust_16}
\end{table}

\paragraph{Ablation}
In this section, we first investigate the impact of parameters in the refinement step. To do so we compare four variants of RLS$^+$ and show the results both quantitatively (\cref{tab:quantitative_ablation}) and qualitatively (\cref{fig:ablation_RLS+}).

Firstly, ``w/o anchor'' is an optimization performed over $\w$, $\thetaa$, and $\noise$ that uses a mean value of random samples as initialization rather than the anchor point. The results produce faces with artifacts which suggests that it is crucial to first find the anchor point in the first step and then optimize from it.

Secondly, ``w/o noise'' and ``w/o g'' are variants where noise inputs and respectively generator's weights are removed from the optimization. Quantitative results show that training without noise inputs or generator weights achieves comparable performance. However, without optimizing these parameters, both reconstruction quality, and realism drop, suggesting they are necessary to synthesize facial details. This is further illustrated by qualitative results.

Thirdly, in the variant ``w/o w'', the latent code is fixed on the anchor point. One can see that w/o optimizing $\w$ the refinement can generate realistic face images by improving reconstruction loss; however, the identity of the face looks rather different from the ground-truth. 

Finally, we evaluate the $\ell_1$-norm ball constraint during the fine-tuning. One can see that this variant harm realism, since without this constraint, the refinement process leads to over-fitting to the LR image query, generating unrealistic facial details. The latter also aligns with quantitative results where  w/o $\ell_1$-norm ball outperforms RLS$^+$ on all metrics except NIQE and LPIPS.
Overall, RLS$^+$ achieves better quantitative measures than its variants, showing that our choice for refinement better balances realism and fidelity.\n

\begin{figure}[!h]
	\centering
	\setlength{\tabcolsep}{0pt}
	\begin{tiny}
		\begin{tabular}{EEEEE}
			LR & w/o Regu. & 
			w/o $\prior_{cross}$ & w/o $\wspace^+$ & RLS\\
			
			\MyImm{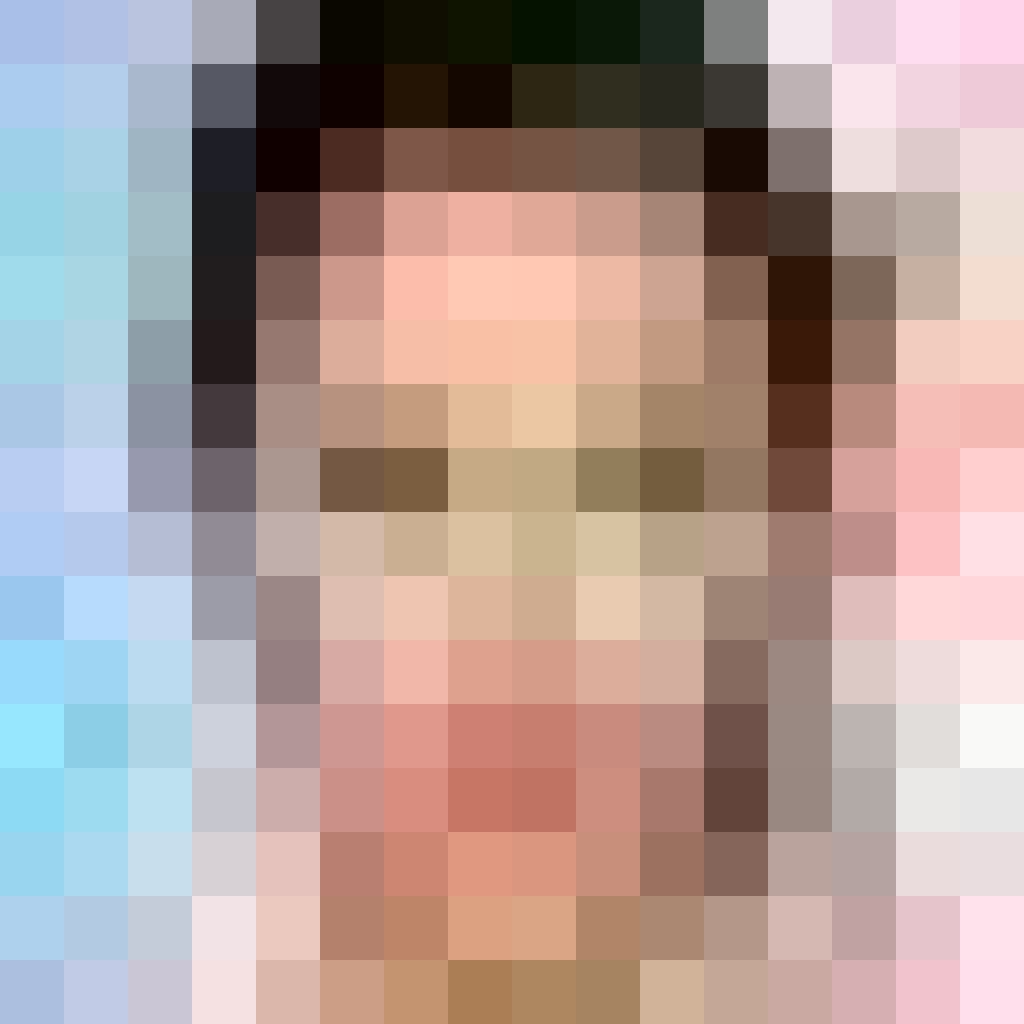} & 
			\MyImm{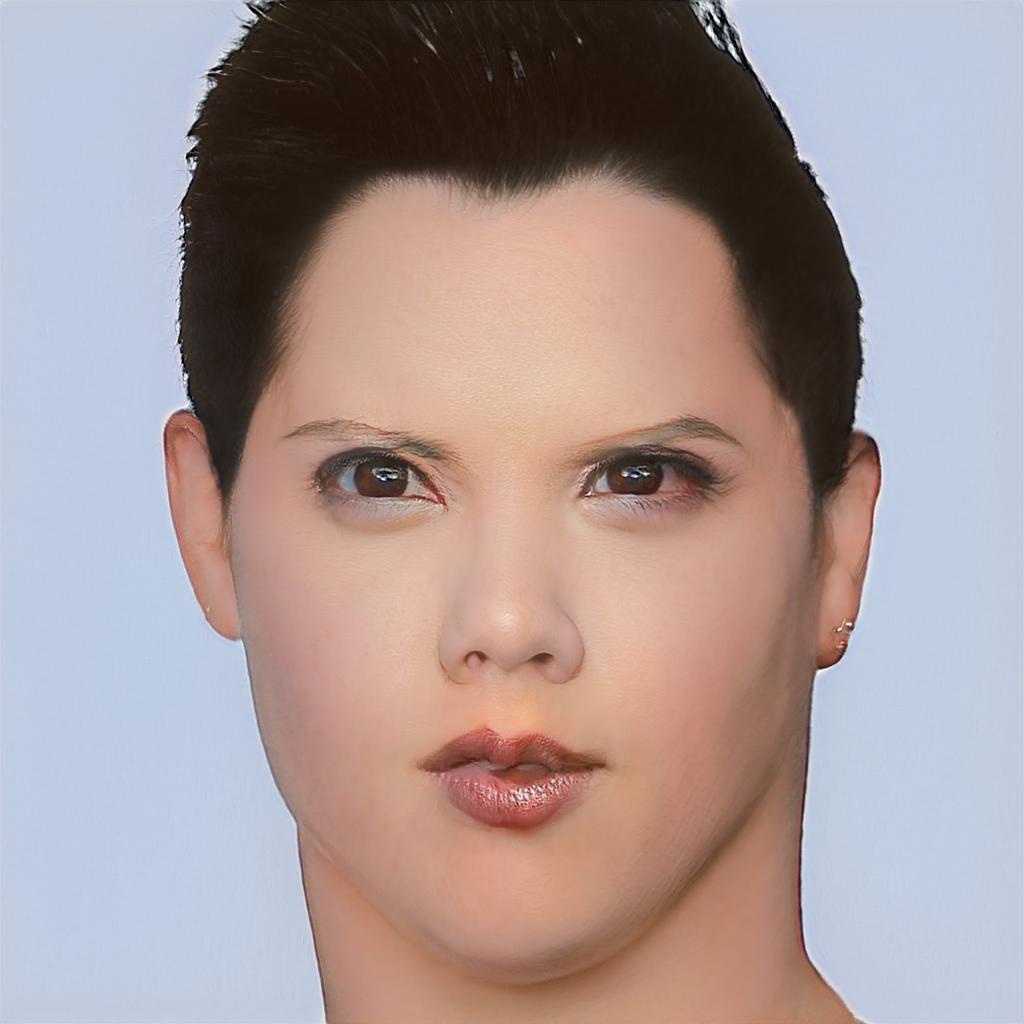} &
			\MyImm{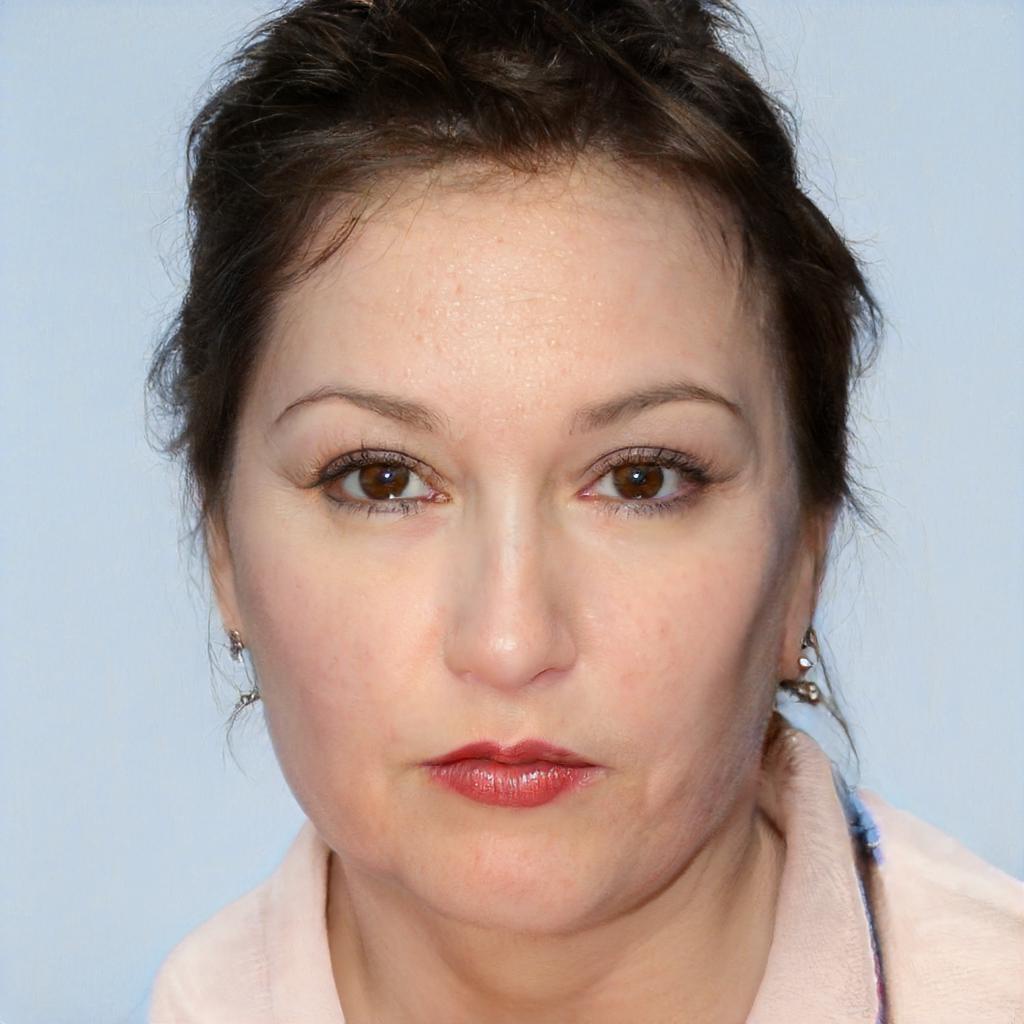} &
			\MyImm{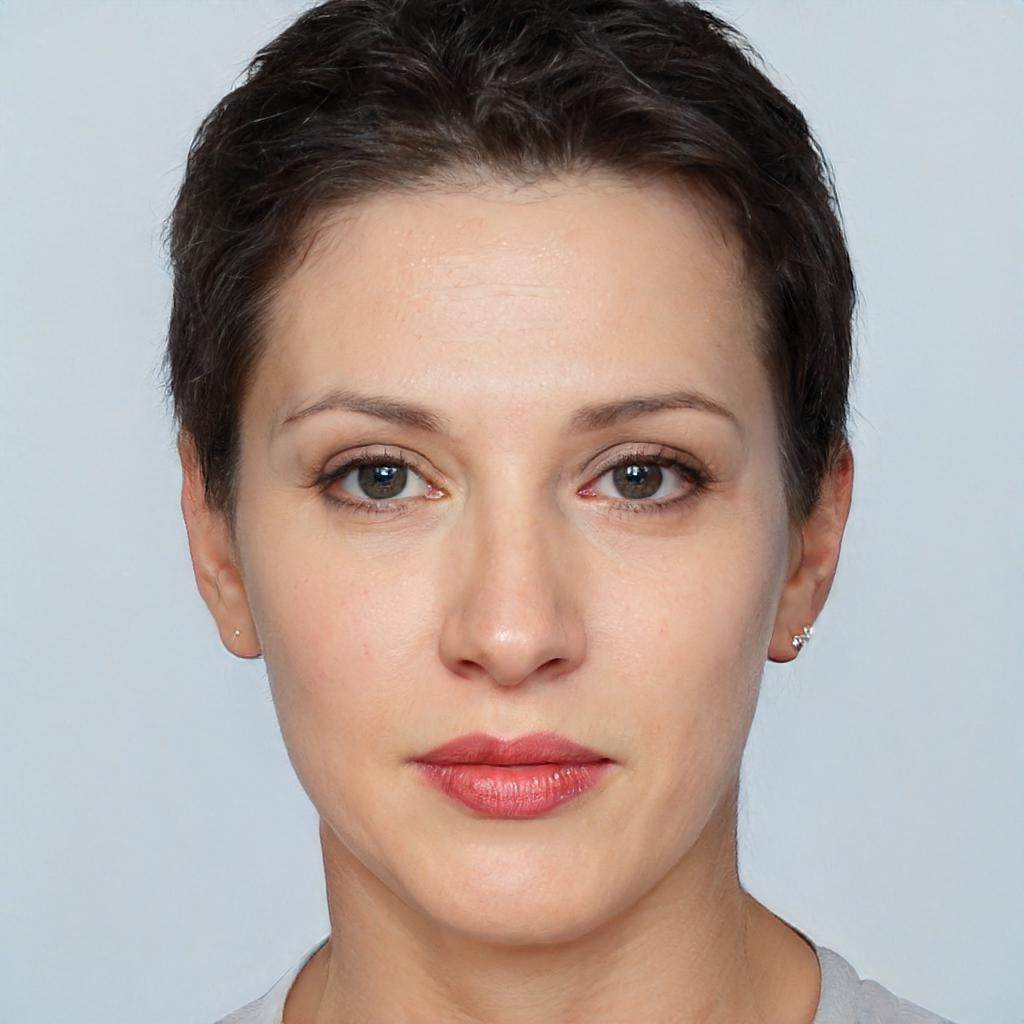} &
			\MyImm{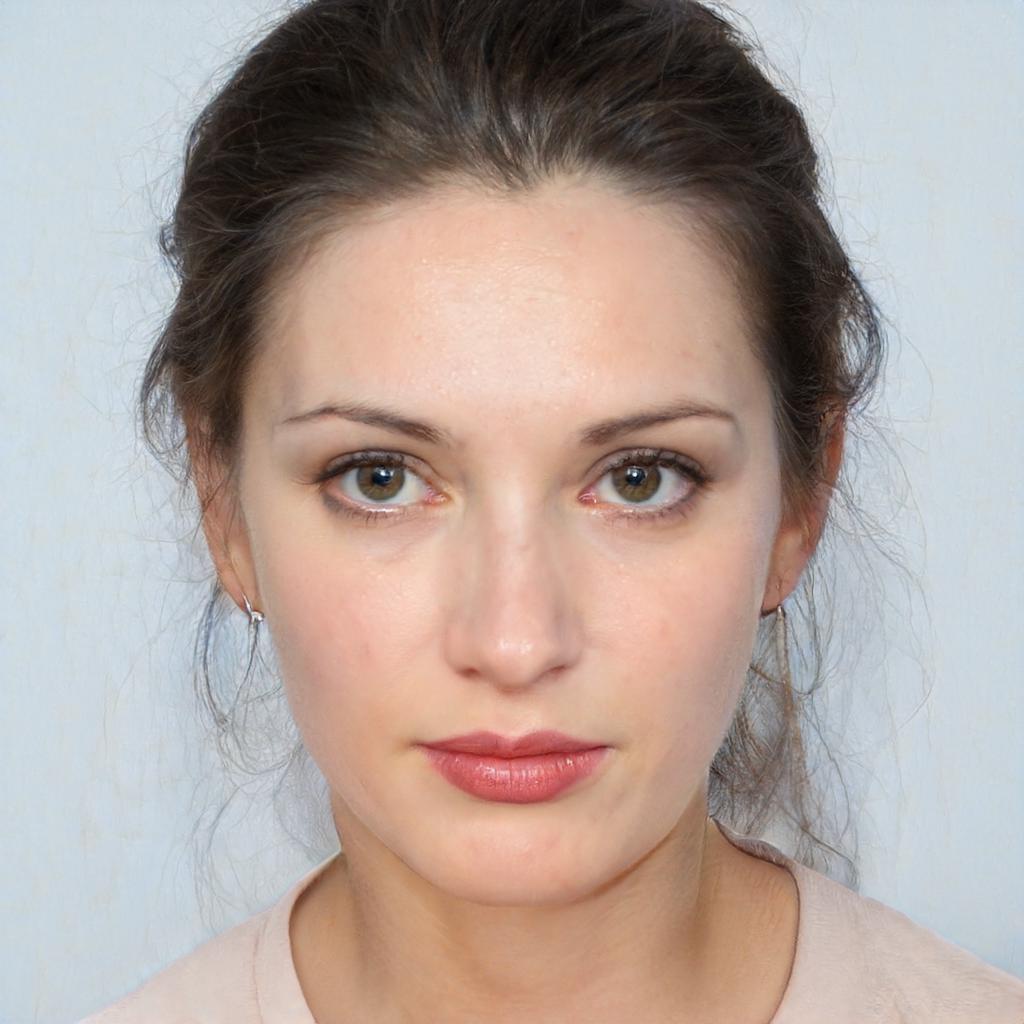} \\

			\MyImm{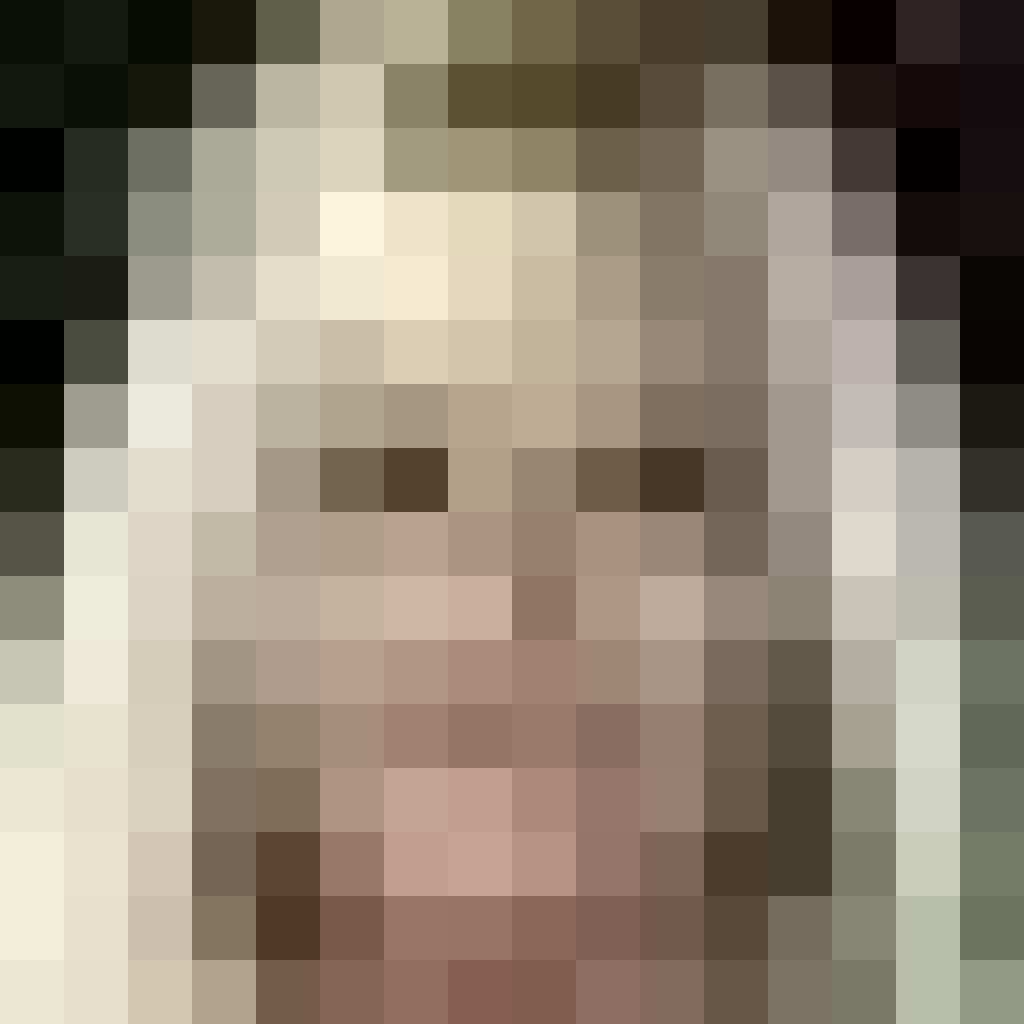} & 
			\MyImm{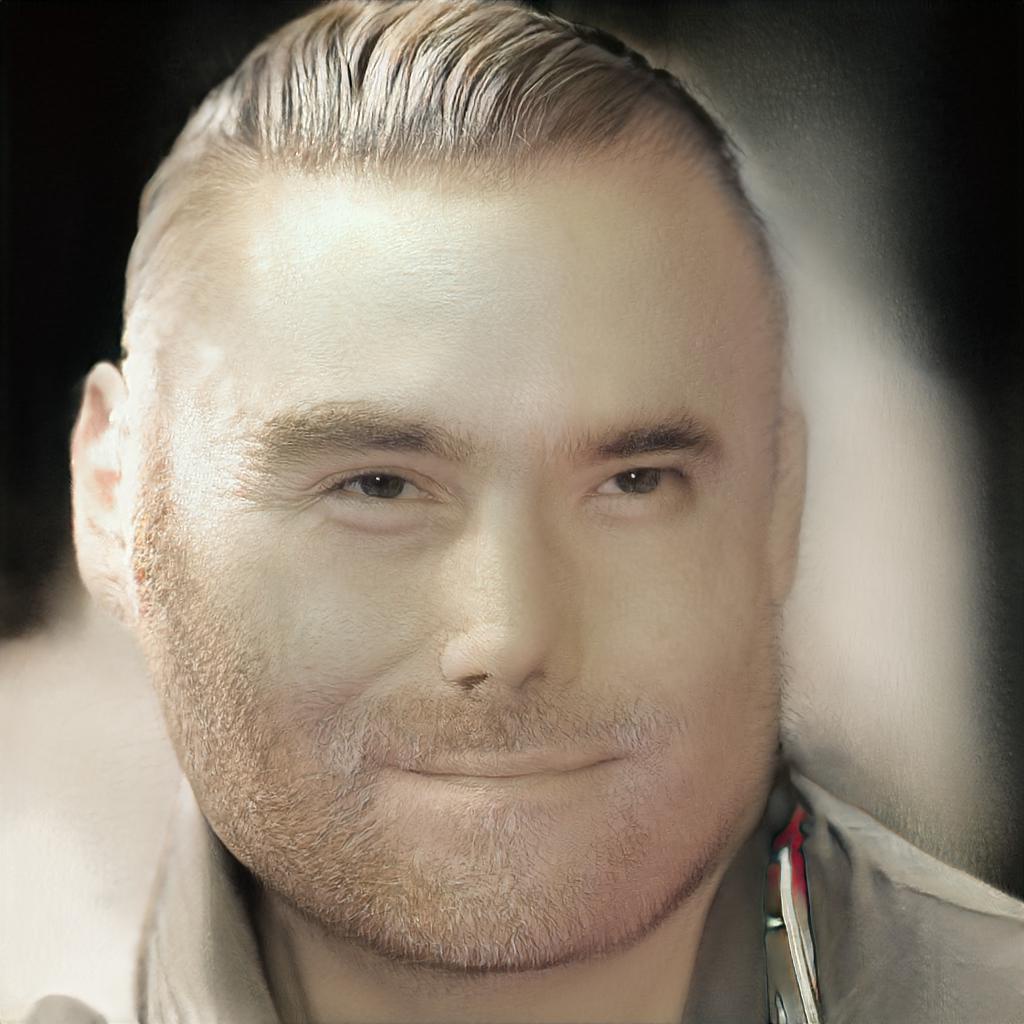} &
			\MyImm{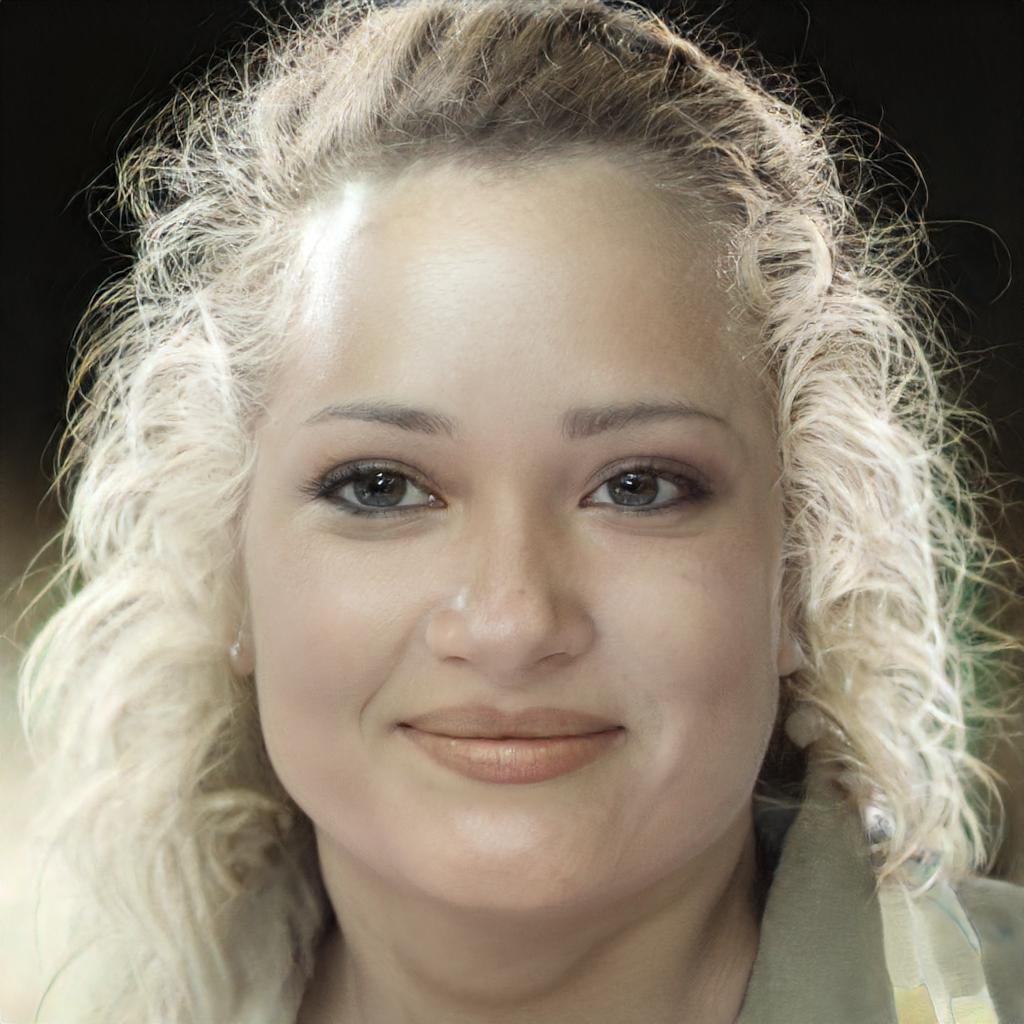} &
			\MyImm{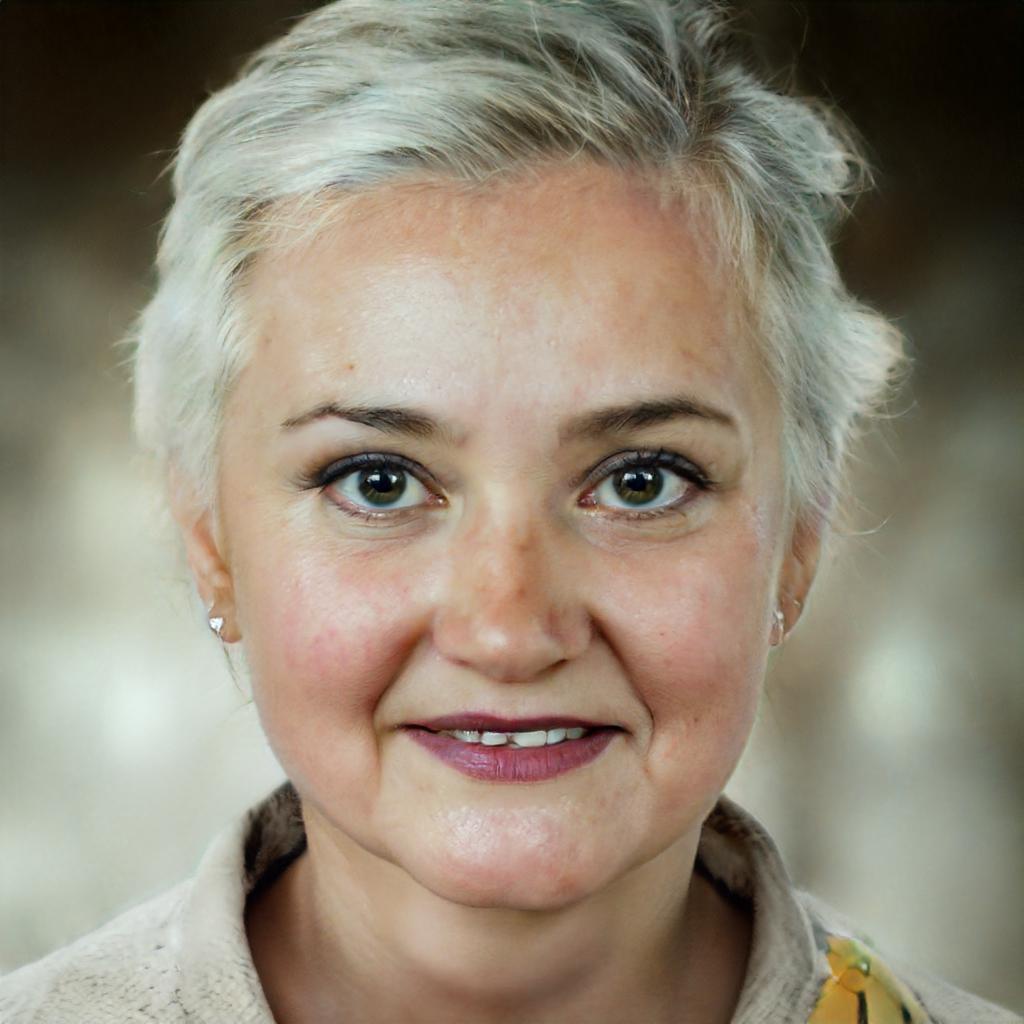} &
			\MyImm{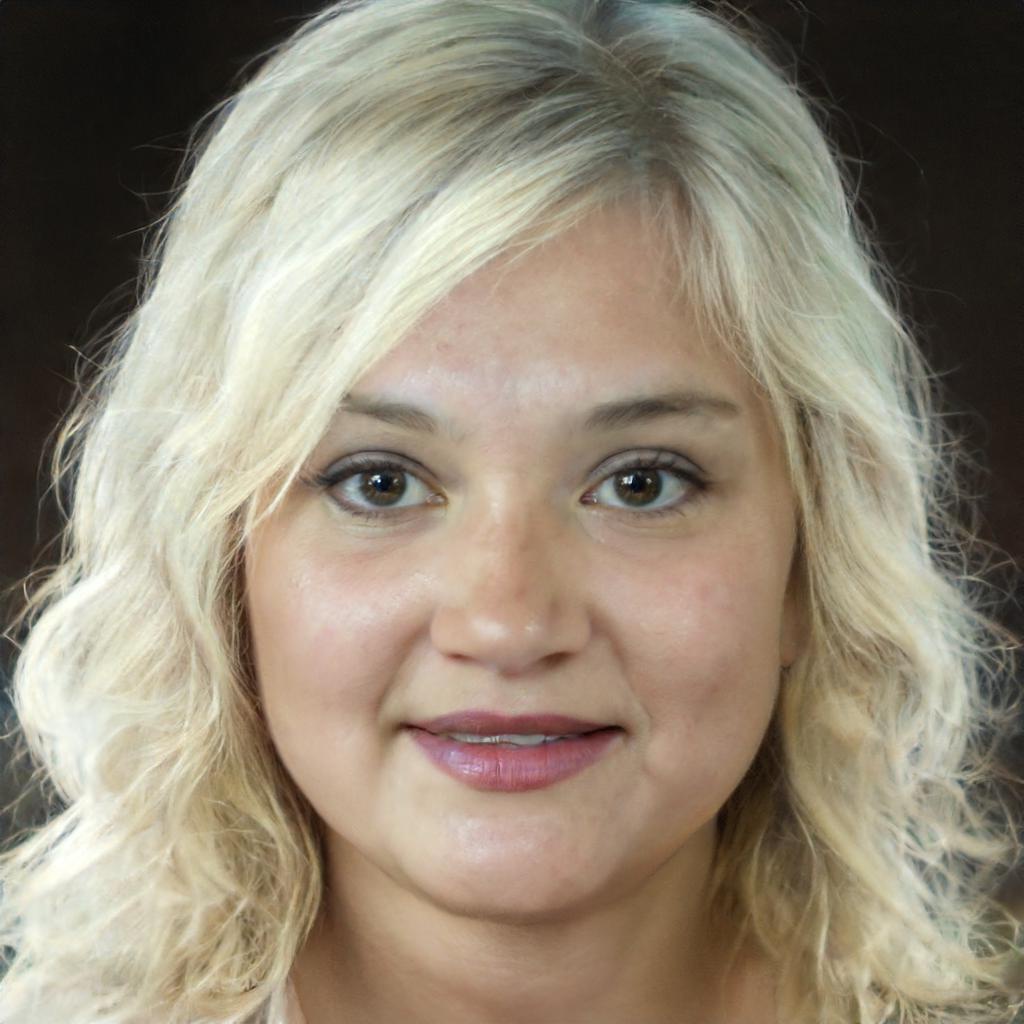} \\
			
			\MyImm{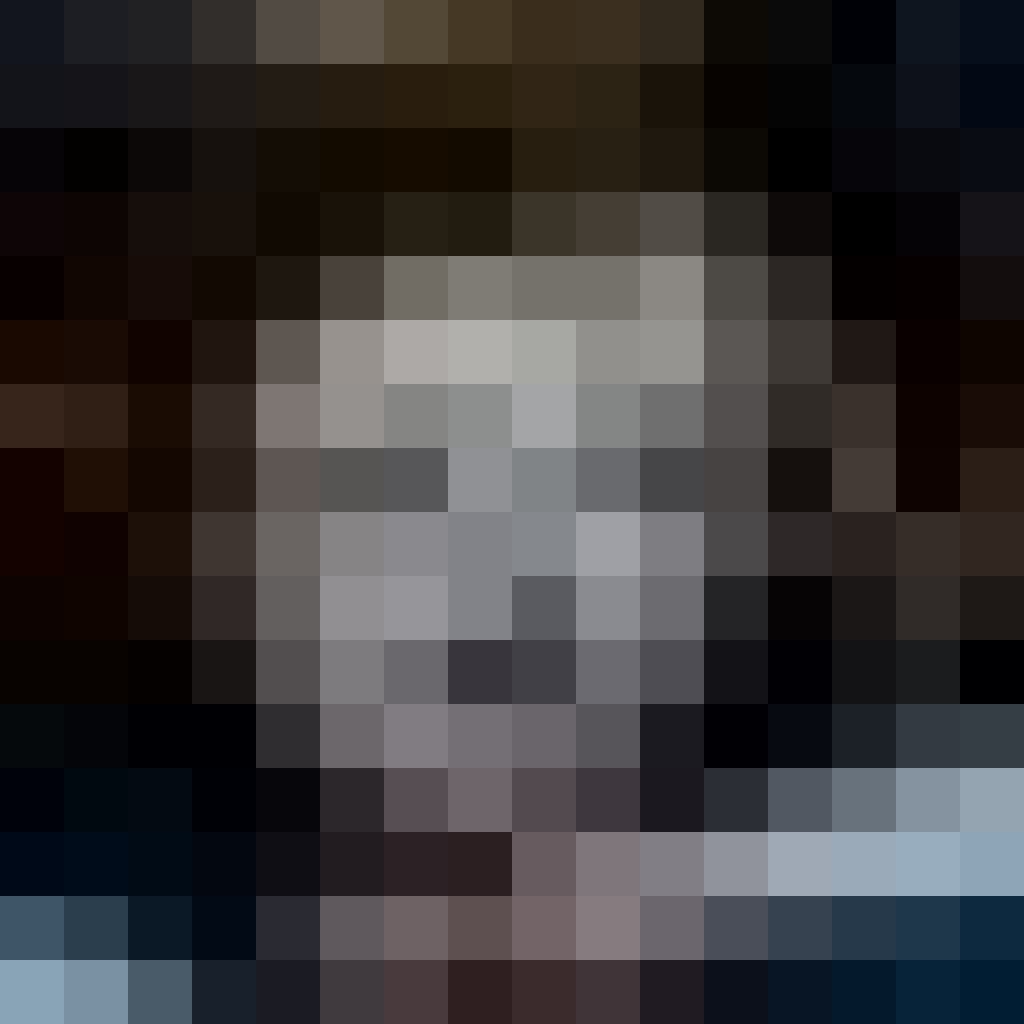} & 
			\MyImm{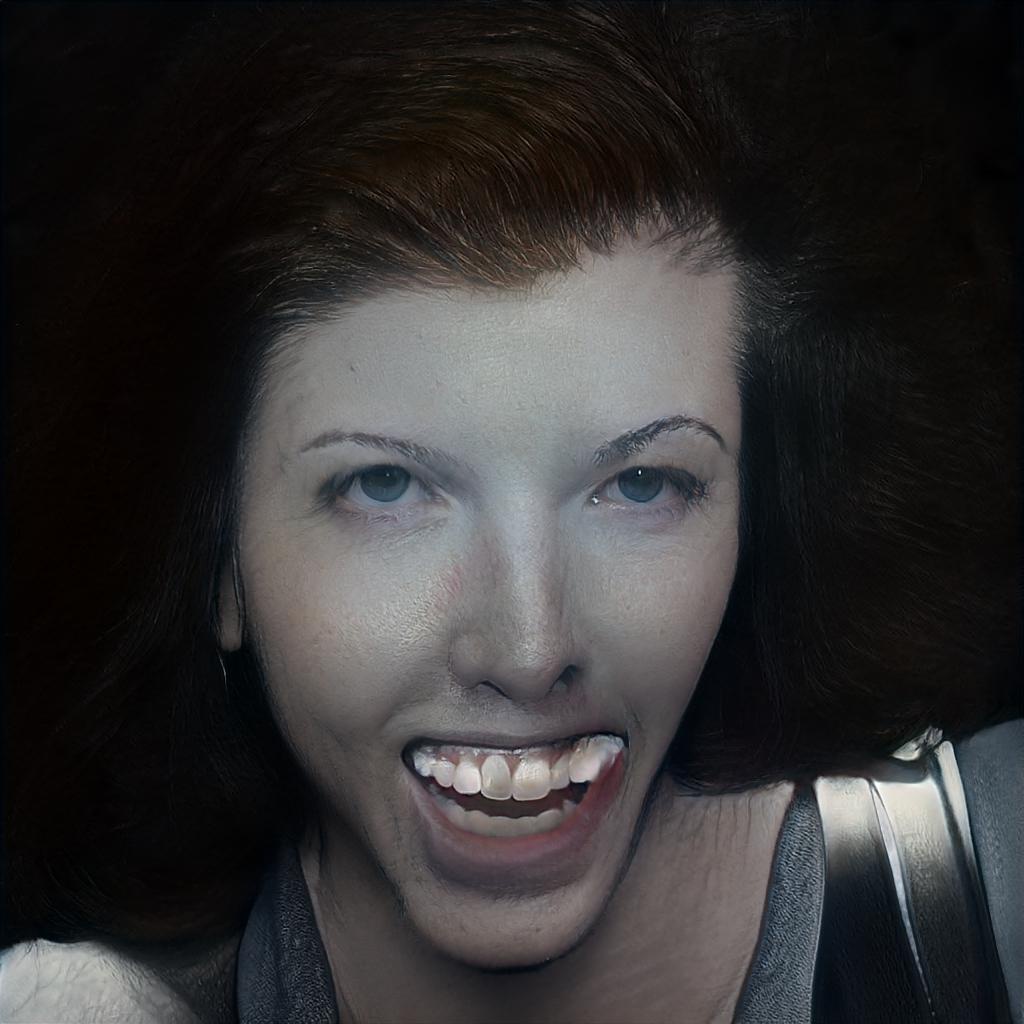} &
			\MyImm{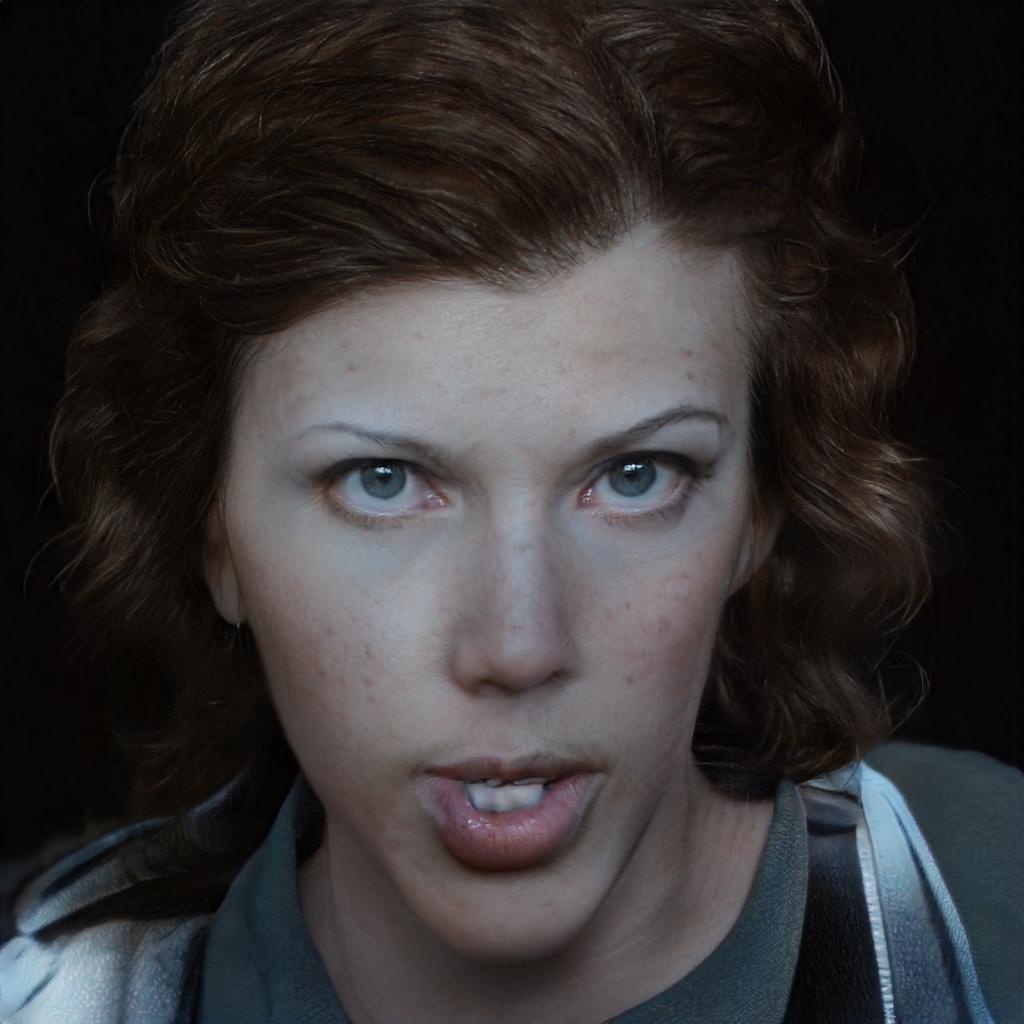} &
			\MyImm{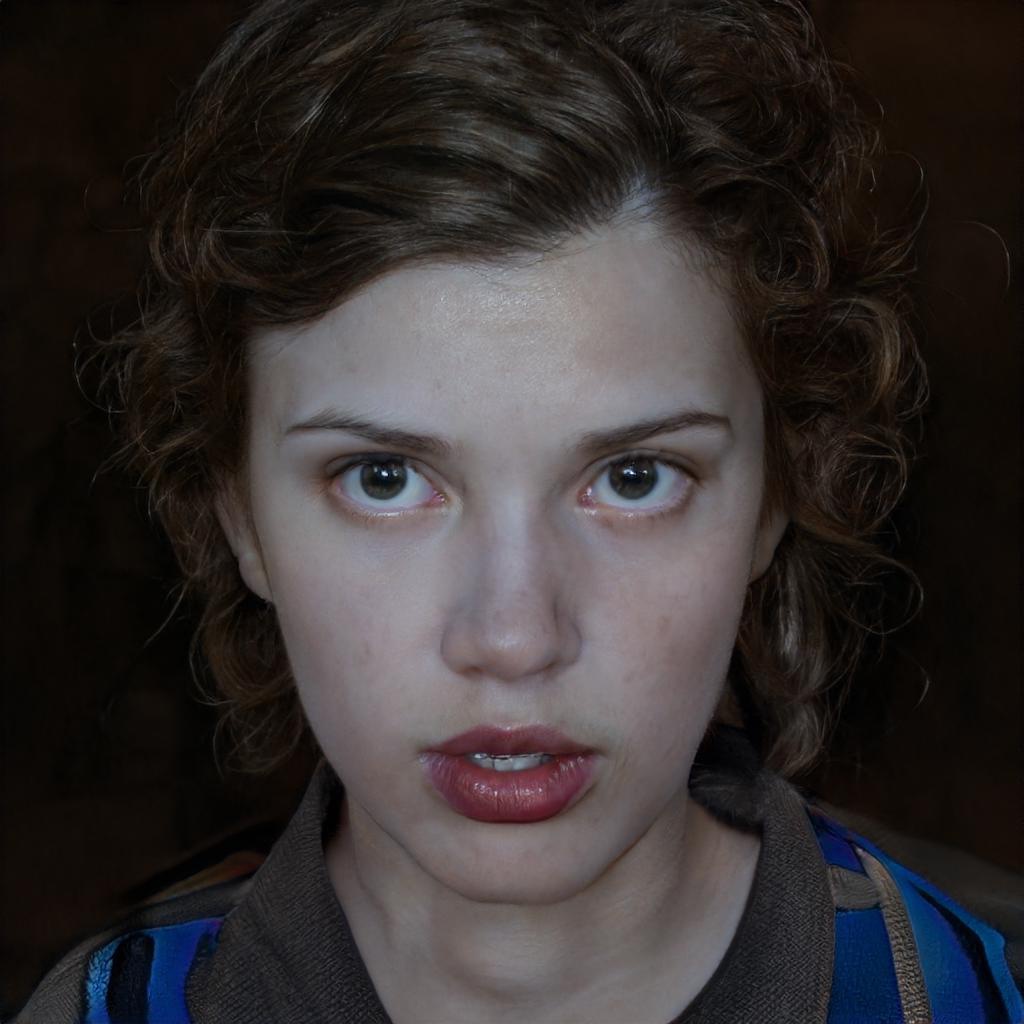} &
			\MyImm{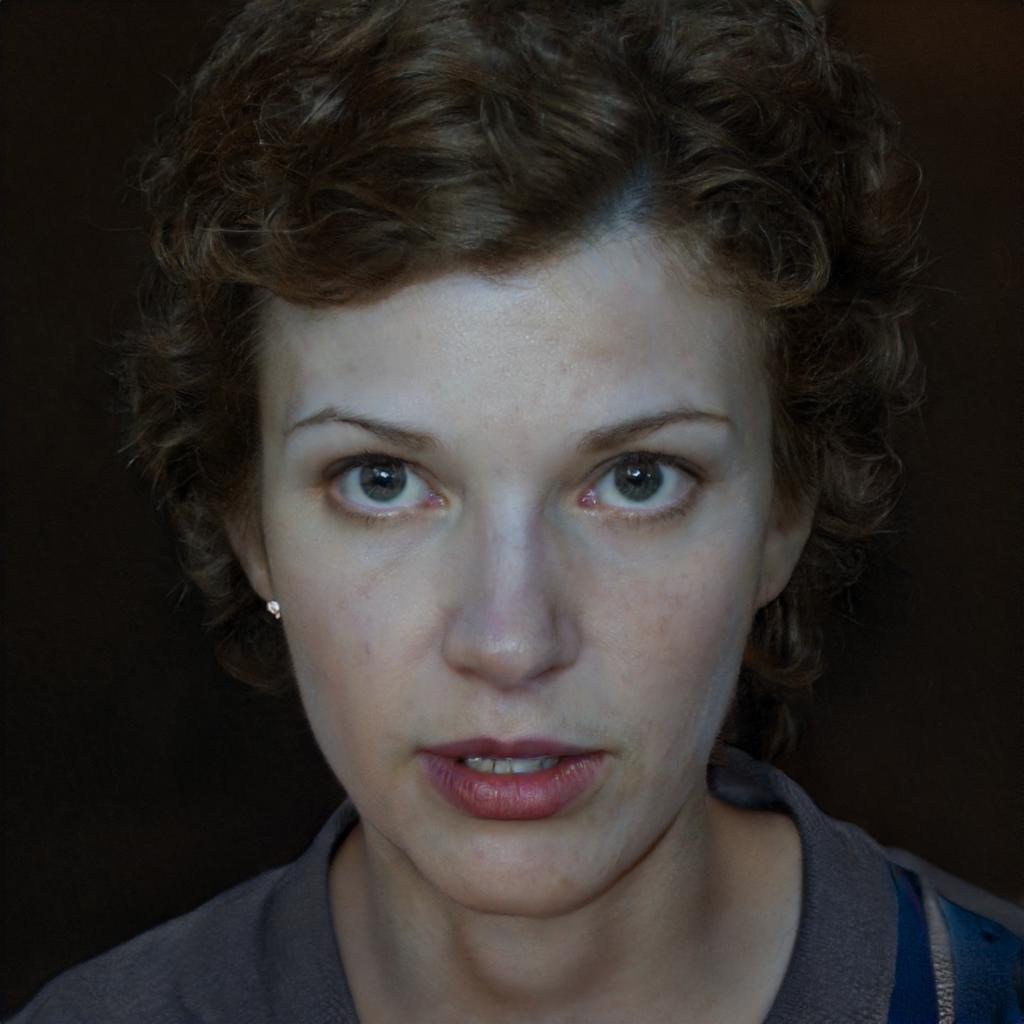} \\
			
			\MyImm{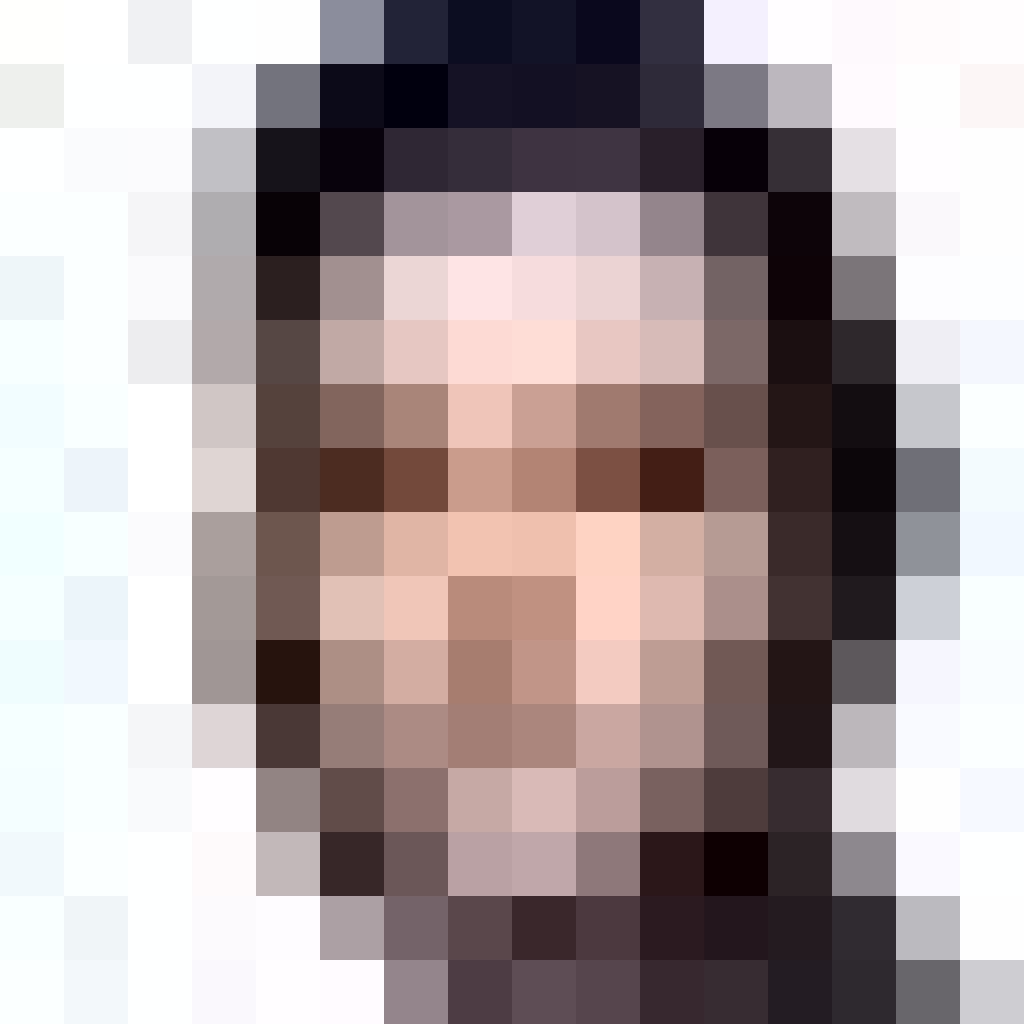} & 
			\MyImm{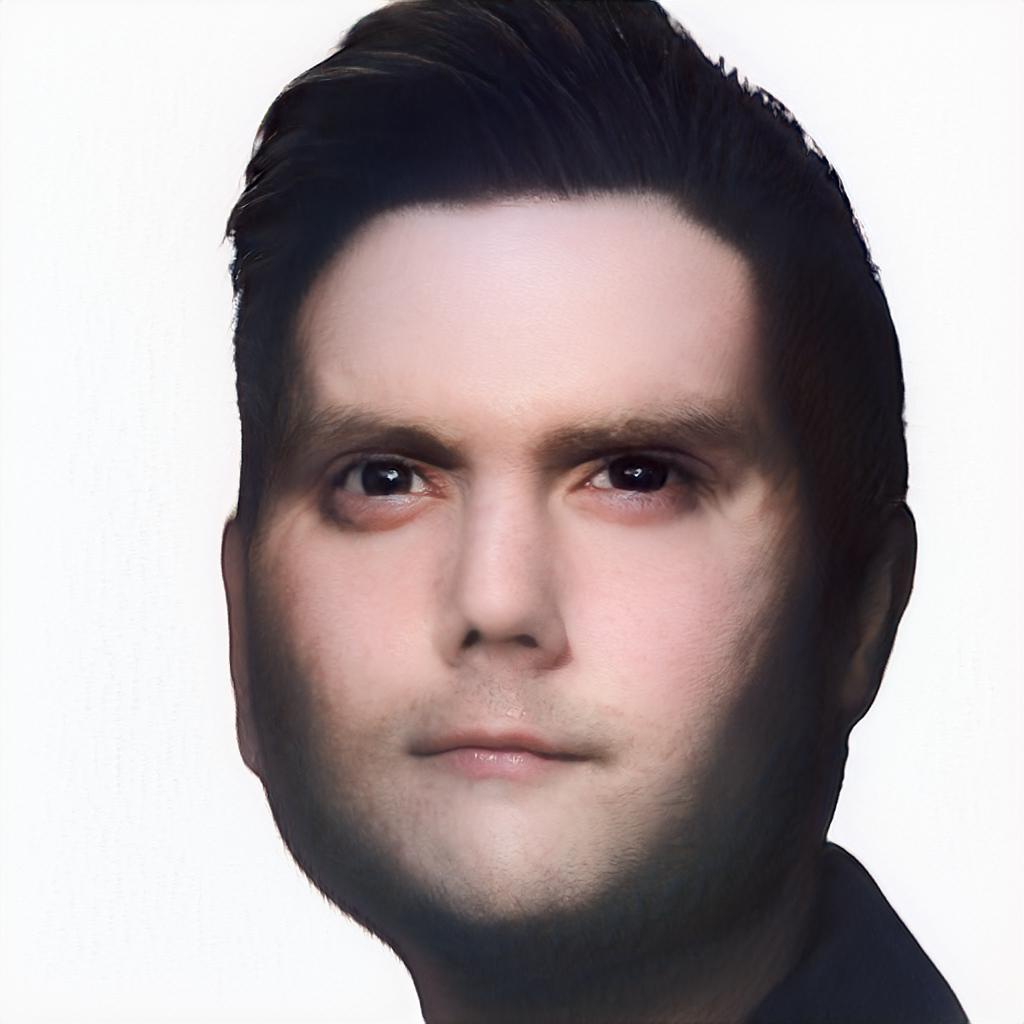} &
			\MyImm{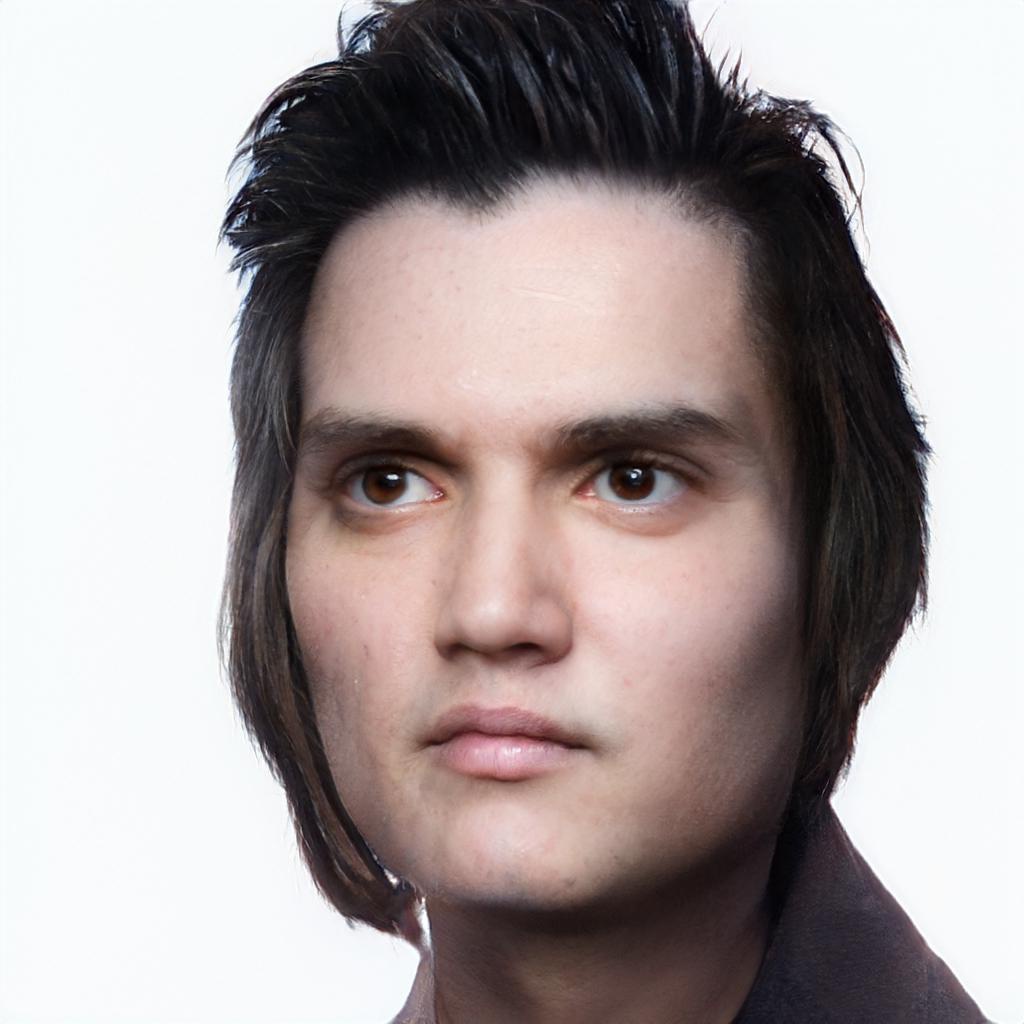} &
			\MyImm{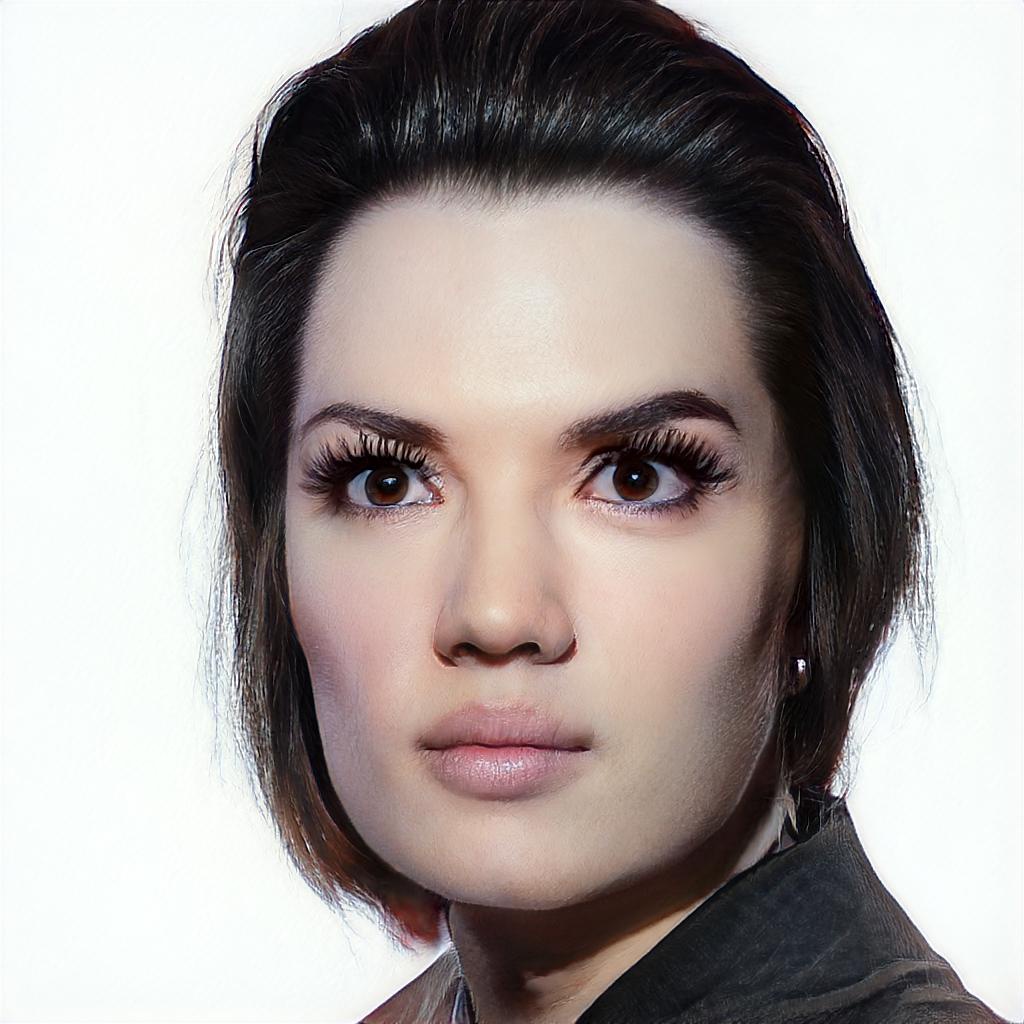} &
			\MyImm{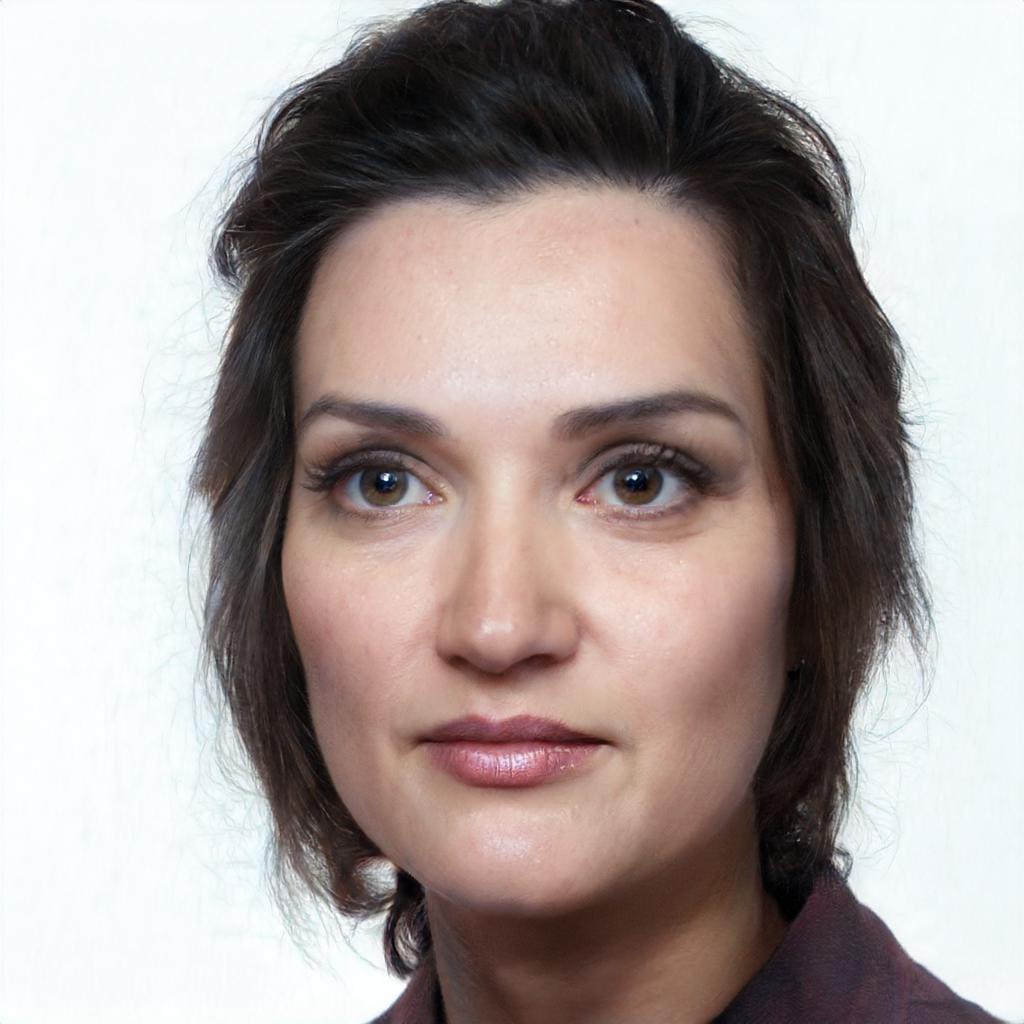} \\
			
			\MyImm{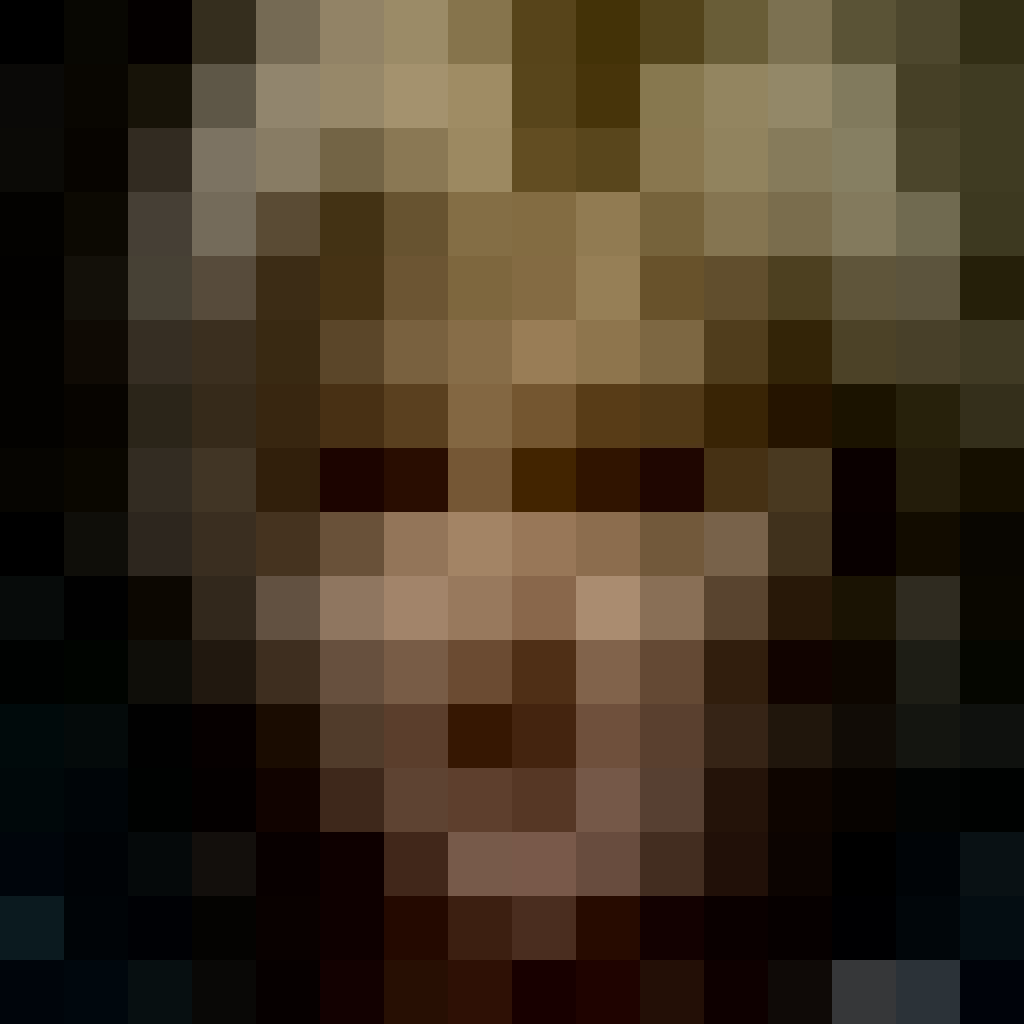} & 
			\MyImm{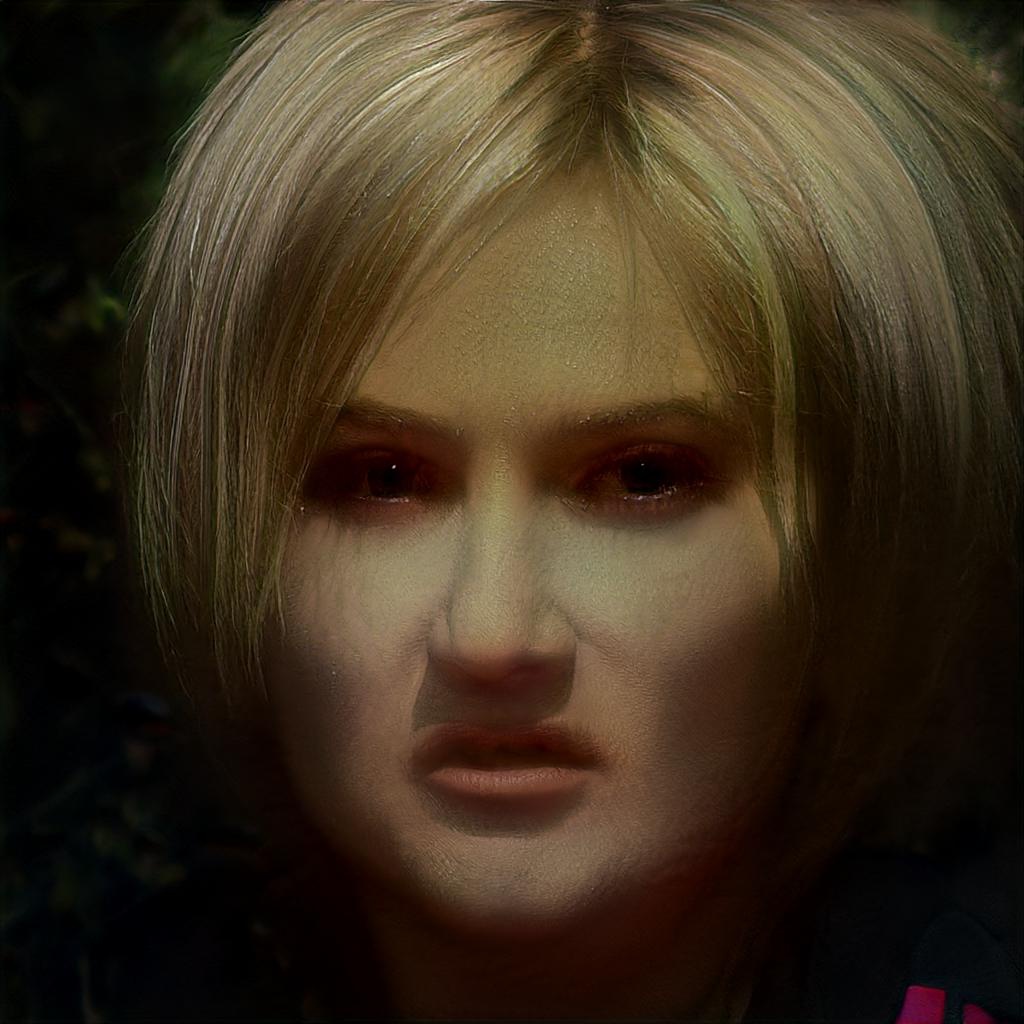} &
			\MyImm{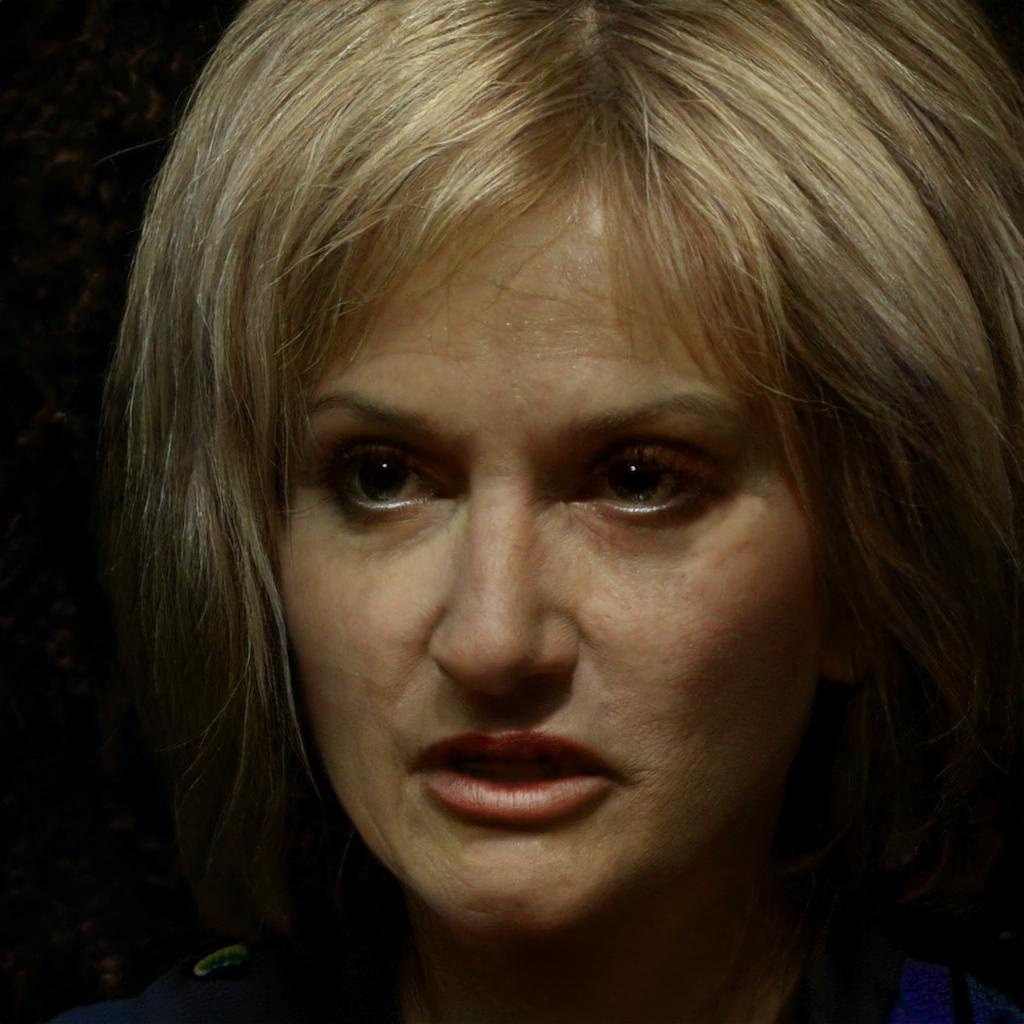} &
			\MyImm{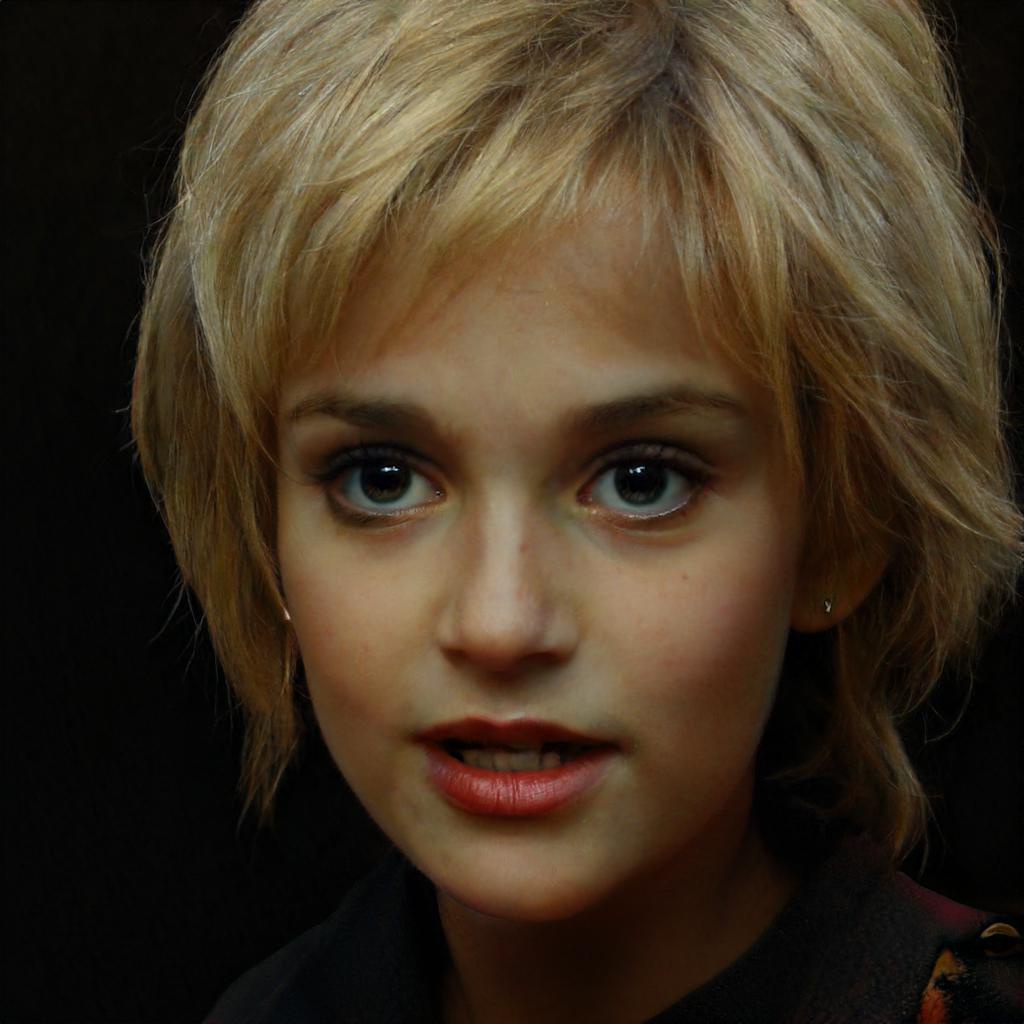} &
			\MyImm{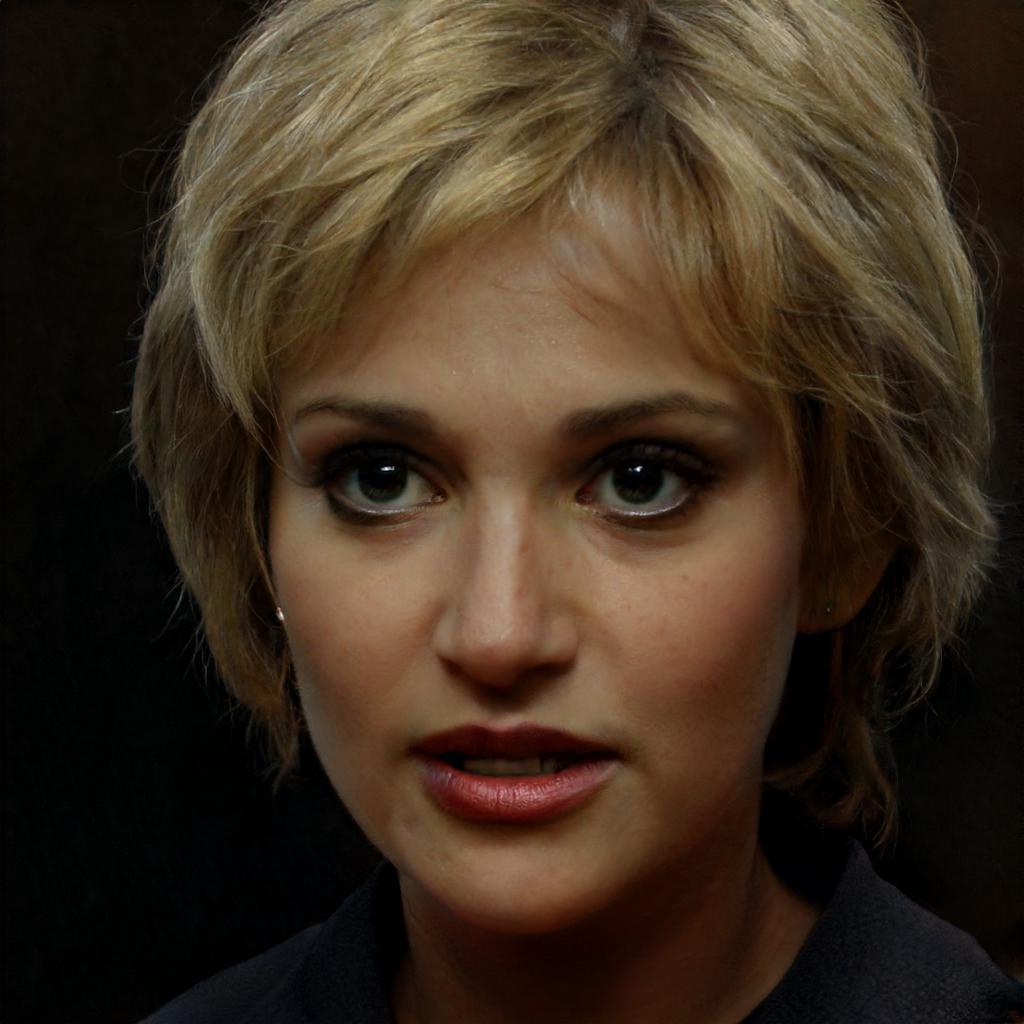} \\
		\end{tabular}
	\end{tiny}
	\caption{Qualitative comparison of different variants of RLS to evaluate the effectiveness of its image prior (64x). (Zoom-in for best view)}
	\label{fig:ablation1}
\end{figure}

\cref{fig:ablation1} demonstrates another ablation experiments highlighting the function of other components of RLS image prior. First, ``w/o Regu.'' searches the latent space without any regularization for the image that, once downscaled matches the LR image.
The second variant is denoted ``w/o $\prior_\text{cross}$'' refer to the suppression of  $\prior_{cross}$. The variant ``w/o ${\wspace^+}$'' refers to an  optimization in the $\wspace$ space rather than in the $\wspace^+$.

To evaluate the three variants, we use the same set of parameters used in \cref{sec:results}. One can see that searching the latent space without any regularization produces images that do not necessarily belong to the face domain and therefore do not appear realistic. It also produces faces with artifacts when $\prior_\text{cross}$ is discarded.
This implies that the cross prior plays an important role in generating realistic facial details. Moreover, we can see that w/o ${\wspace^+}$ generates realistic HR face images; however, as it is reported in \cref{tab:ablation_RLS}, RLS acheives higher values in terms of fidelity.
This is because the latent space $\wspace$ is less expressive than $\wspace^+$, reducing the range of images that can be reconstructed with high fidelity.

\setlength{\tabcolsep}{5pt}
\begin{table}
	\centering	
	\begin{tabular}{@{}l||llll@{}}
		\toprule
		Method & NIQE$\downarrow$ & ID$\uparrow$ & LPIPS$\downarrow$ & MSSIM$\uparrow$ \\
		\hline
		w/o Regu. & 4.2970 & 0.7003 & 0.5336 & \blue{0.5508}\\
		w/o $\prior_{cross}$ & 4.1946 & 0.7078 & \blue{0.4977} & \red{0.5488} \\
		w/o $\wspace^+$ & \blue{3.7184} & 0.7086 & 0.5106 & 0.5183 \\
		RLS & \red{4.1032} & \blue{0.7210} & \red{0.5037} & 0.5214 \\
		\bottomrule
	\end{tabular}
	\caption{Quantitative comparison of five variants of RLS (64x). (The \blue{best} and the \red{second-best} are emphasized by blue and red respectively.)}
	\label{tab:ablation_RLS}
\end{table}

\begin{figure}
	\centering
	\begin{tabular}{ccc}
		\includegraphics[scale=0.17]{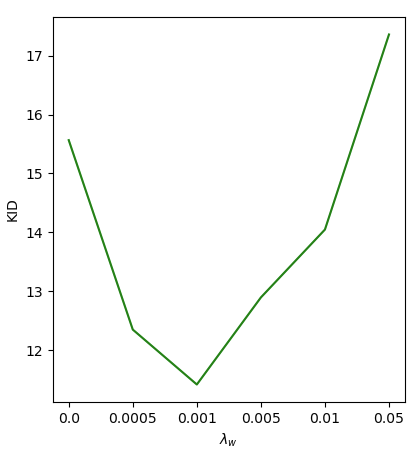}    &
		\includegraphics[scale=0.17]{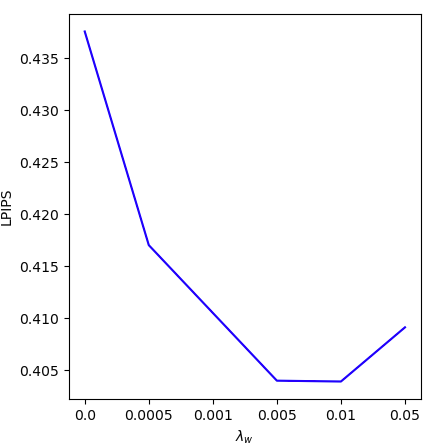} &
		\includegraphics[scale=0.17]{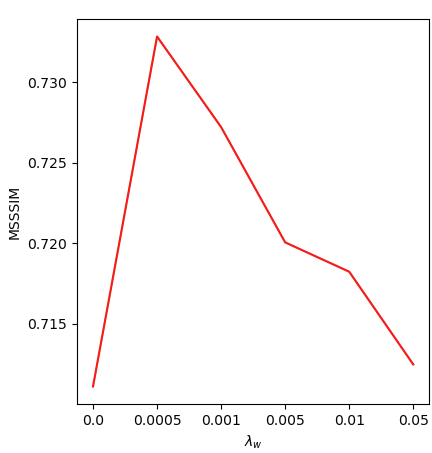}\\ 
	\end{tabular}
	\caption{Examining the impact of $P_w$ (16x SR)}
	\label{fig:abblation_pw}
\end{figure}

We also explored the impact of $P_w$ in \cref{equation imageprior}. \cref{fig:abblation_pw} shows the impact of $P_w$ on KID, LPIPS, and MSSIM metrics, supporting the claim that the incorporation of $P_w$ improves the overall realism-fidelity trade-off.

\paragraph{Robustness to artifacts.} 
In contrast to supervised methods, which are sensitive to the input image domain, this approach is not restricted to a particular degradation operator used during training.

To assess this aspect, we applied additional degradation operators such as Gaussian noise, Salt and Pepper, and Gaussian blur to a bicubic downscaled image. We also applied motion blur to the HR image followed by downscaling. \cref{fig:robustness} demonstrates the robustness evaluation when the low-resolution image is degraded with various types of noise. RLS$^+$ can still generate realistic images while preserving their identity even in the presence of noise. However, the quality of the output may decrease, particularly when subjected to Gaussian blur, as the generator adapts to produce "noisy" images when the noise becomes more intense. This evaluation justifies the use of the bicubic downscaling operator during training instead of more complicated specific degradations.

\begin{figure}[!h]
	\centering
	\setlength{\tabcolsep}{0pt}
	\begin{tiny}
		\begin{tabular}{EEEEEE}			
			GT & RLS$^{+}$ & \hspace{-10pt} \shortstack{Gaussian Noise \\ ($\sigma=0.1$)} & \shortstack{Salt$\&$Pepper \\ ($\sigma=0.05$)} & \shortstack{Gaussian Blur \\ ($\sigma=1$)} &\shortstack{Motion Blur \\ ($length=100$)}\\
			
			\MyImm{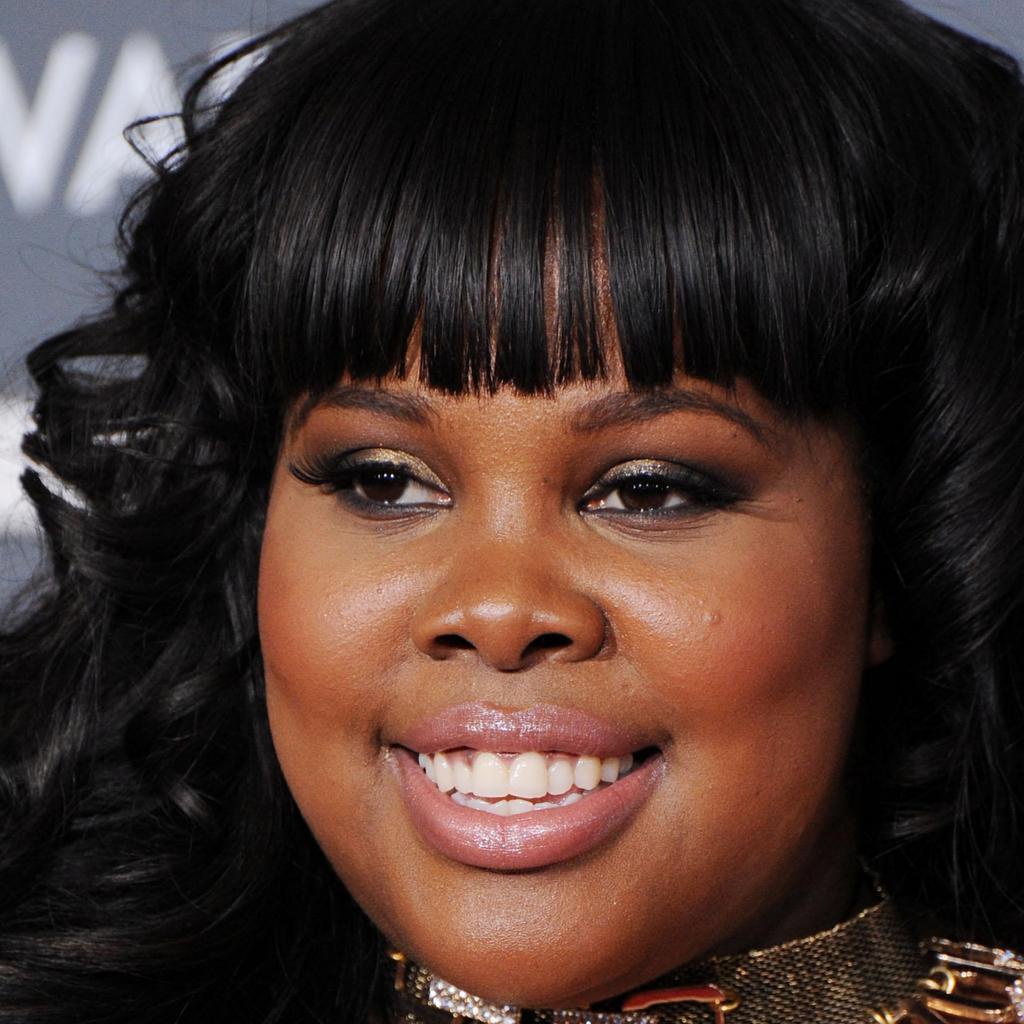} &
			\MyImm{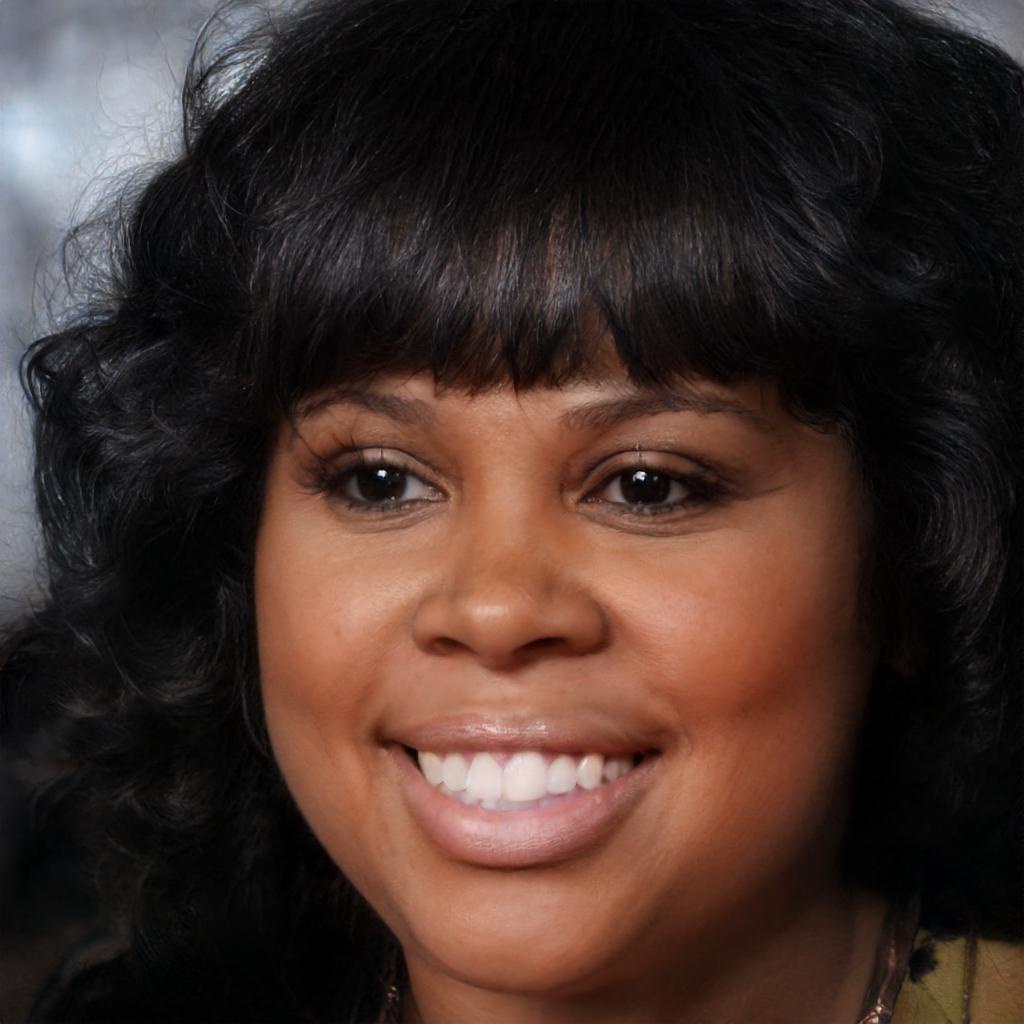} &
			\MyImm{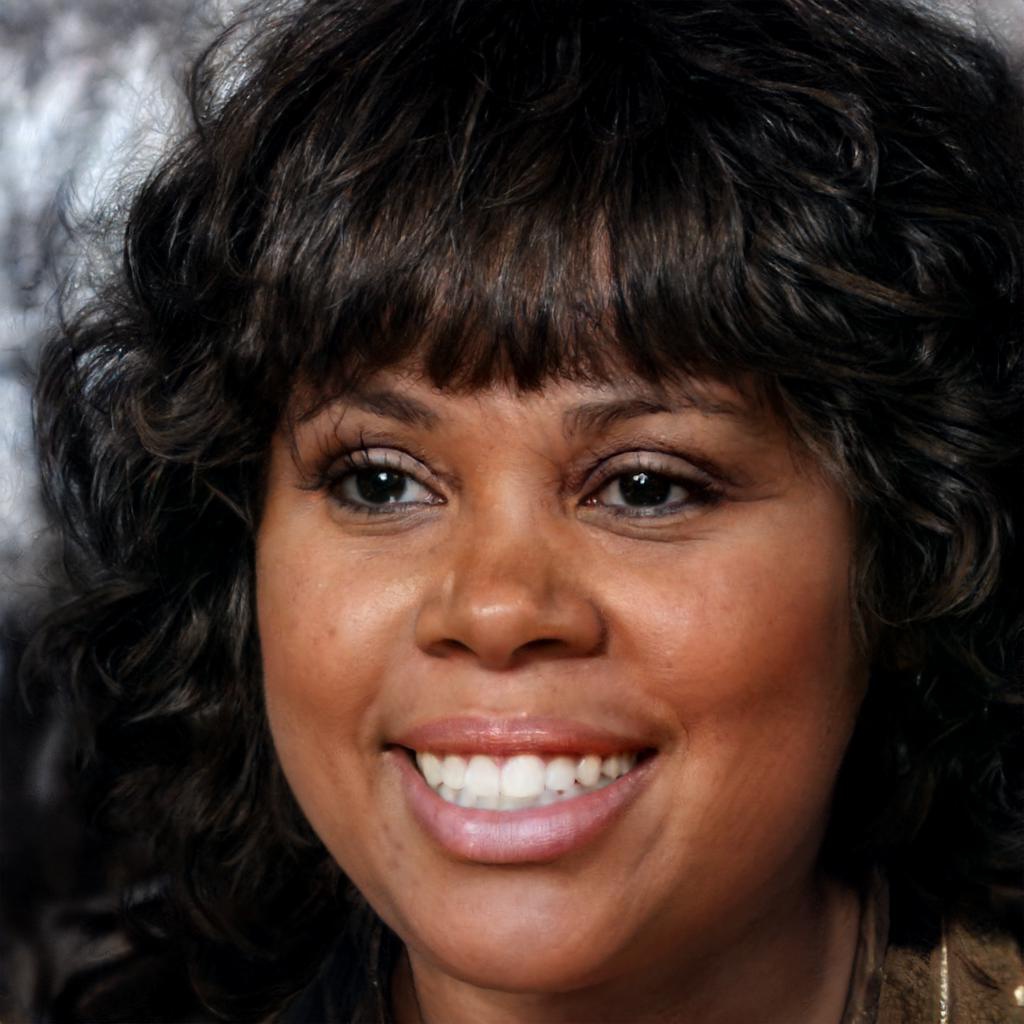} &
			\MyImm{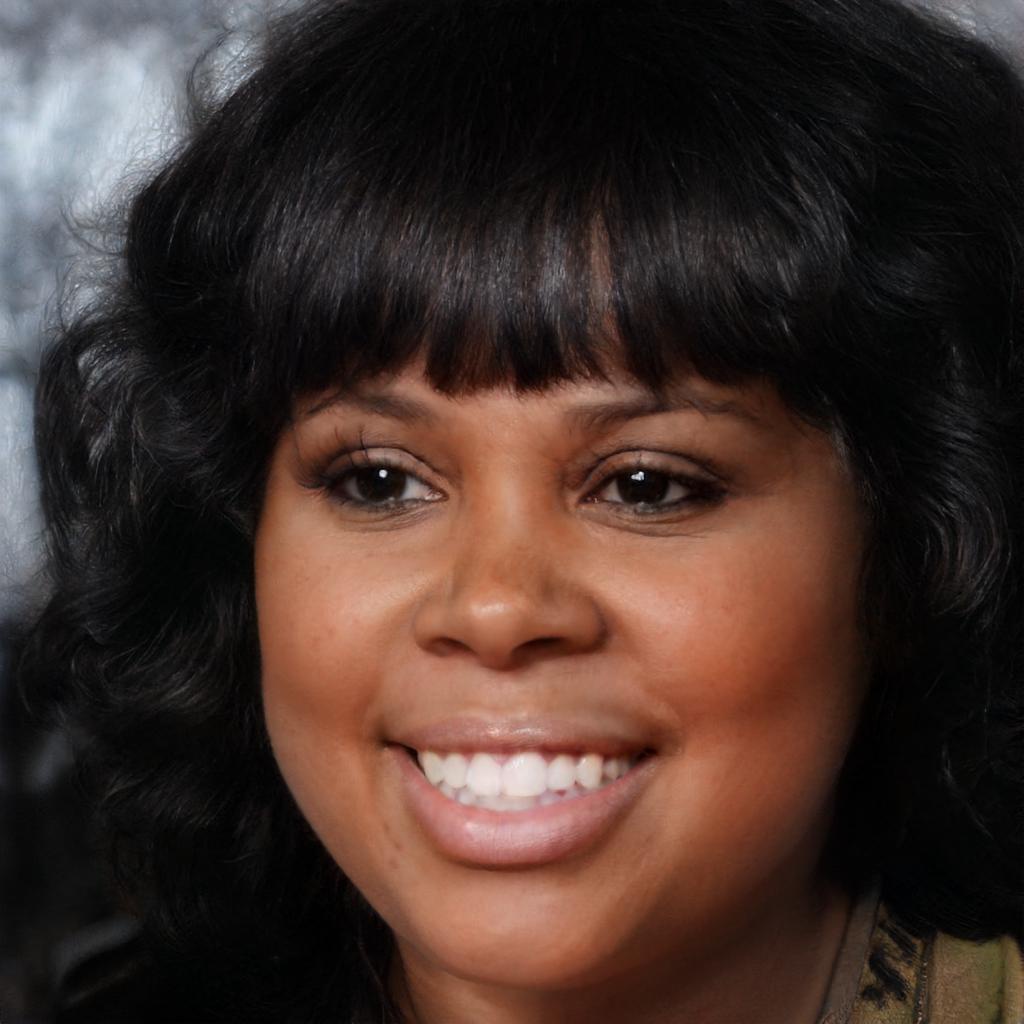} &
			\MyImm{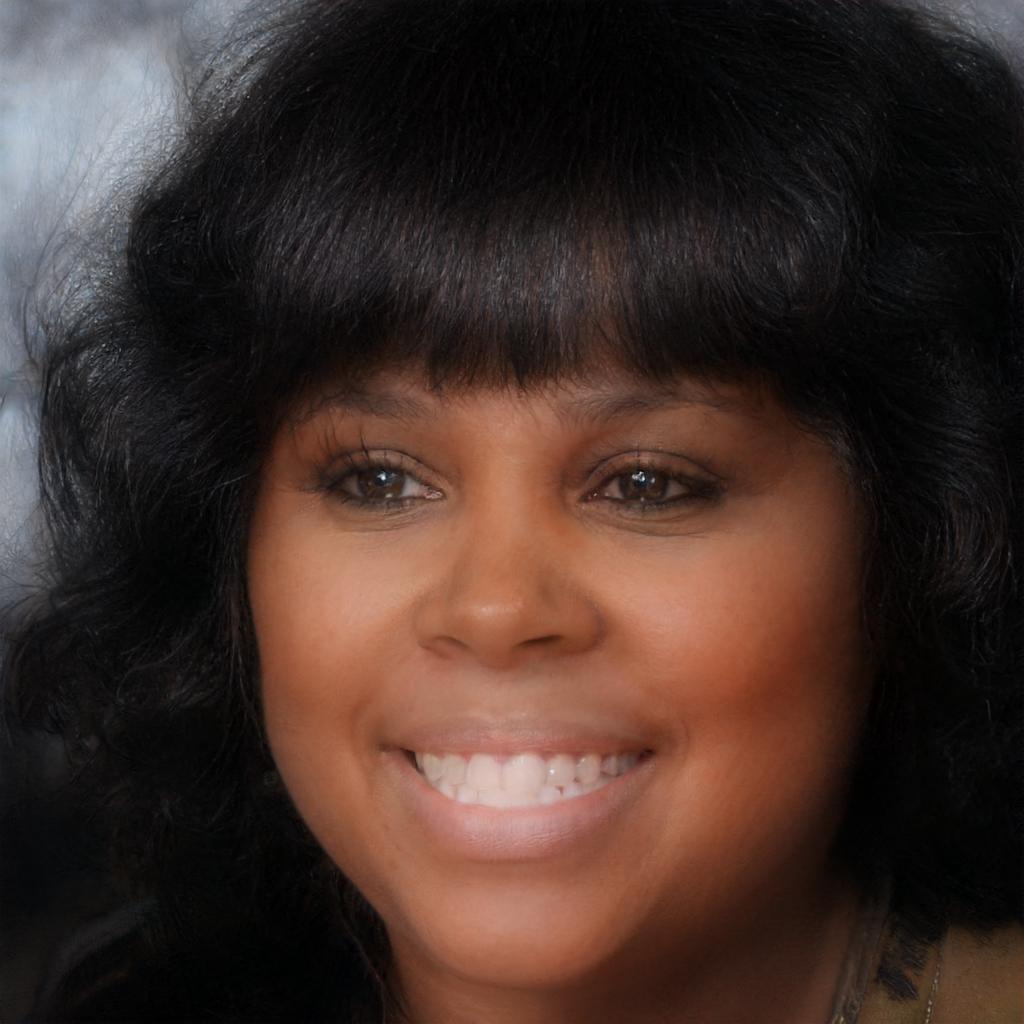} &
			\MyImm{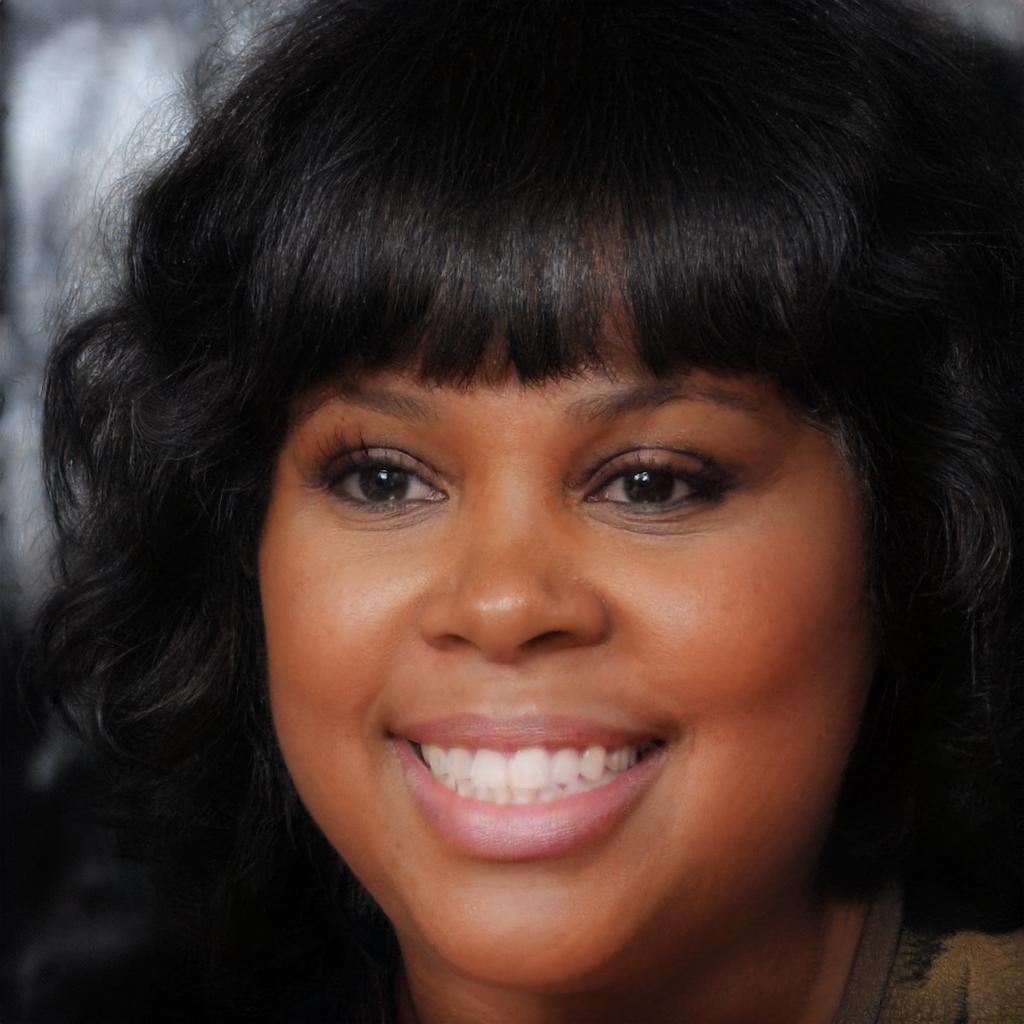} \\
			
			\MyImm{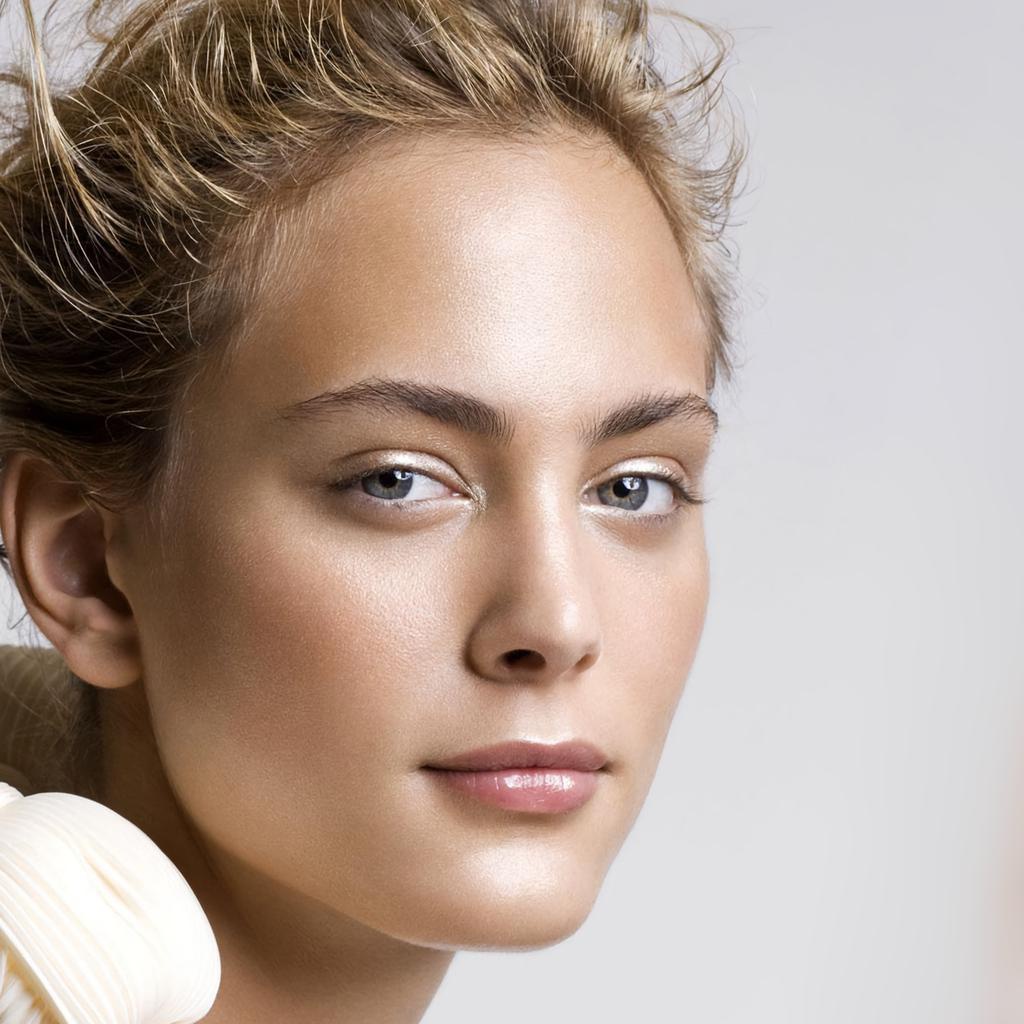} &
			\MyImm{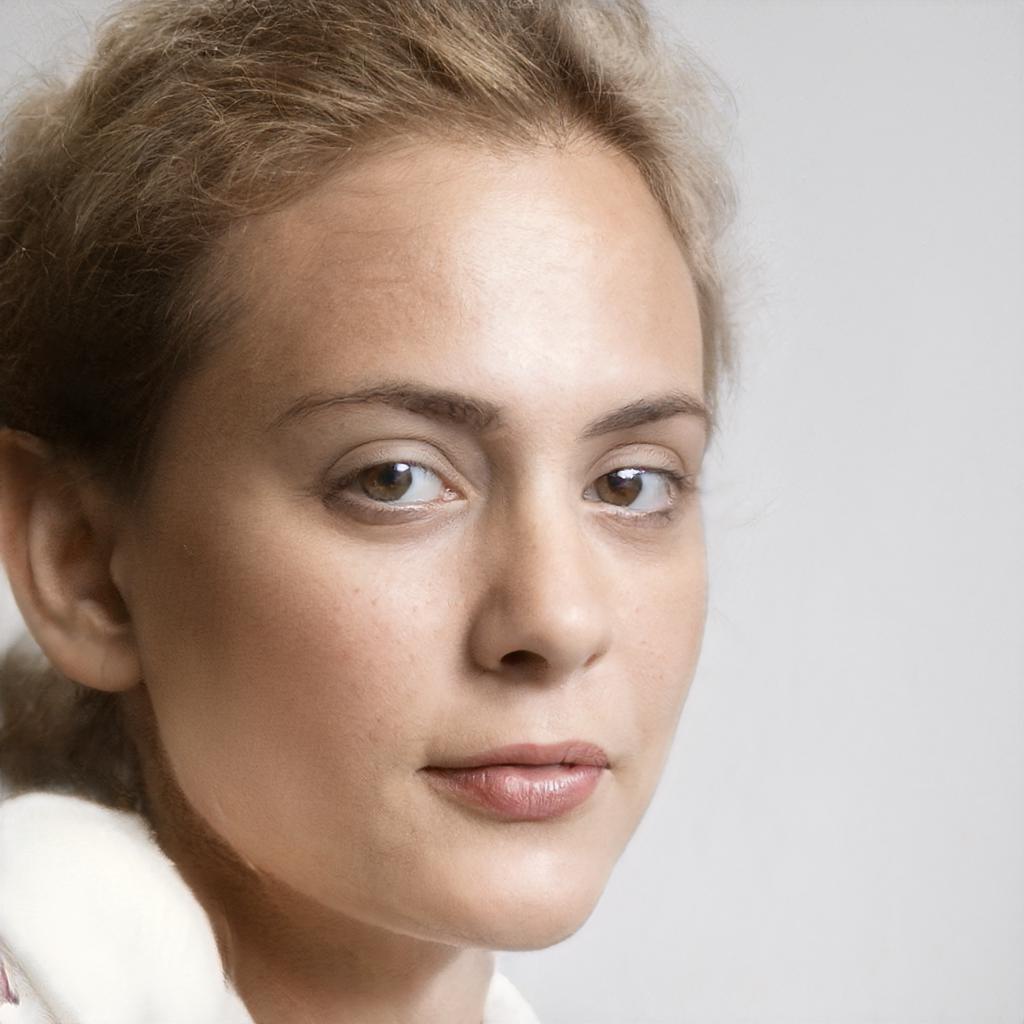} &
			\MyImm{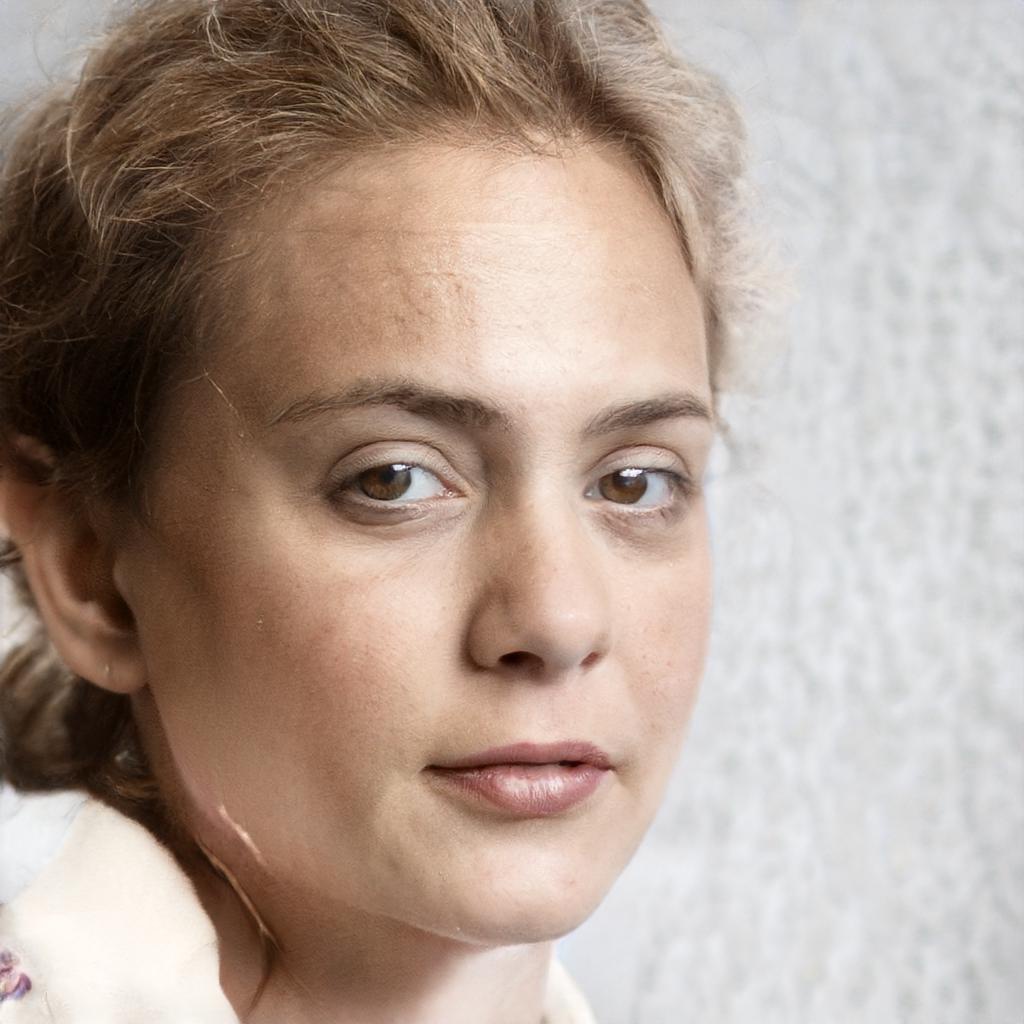} &
			\MyImm{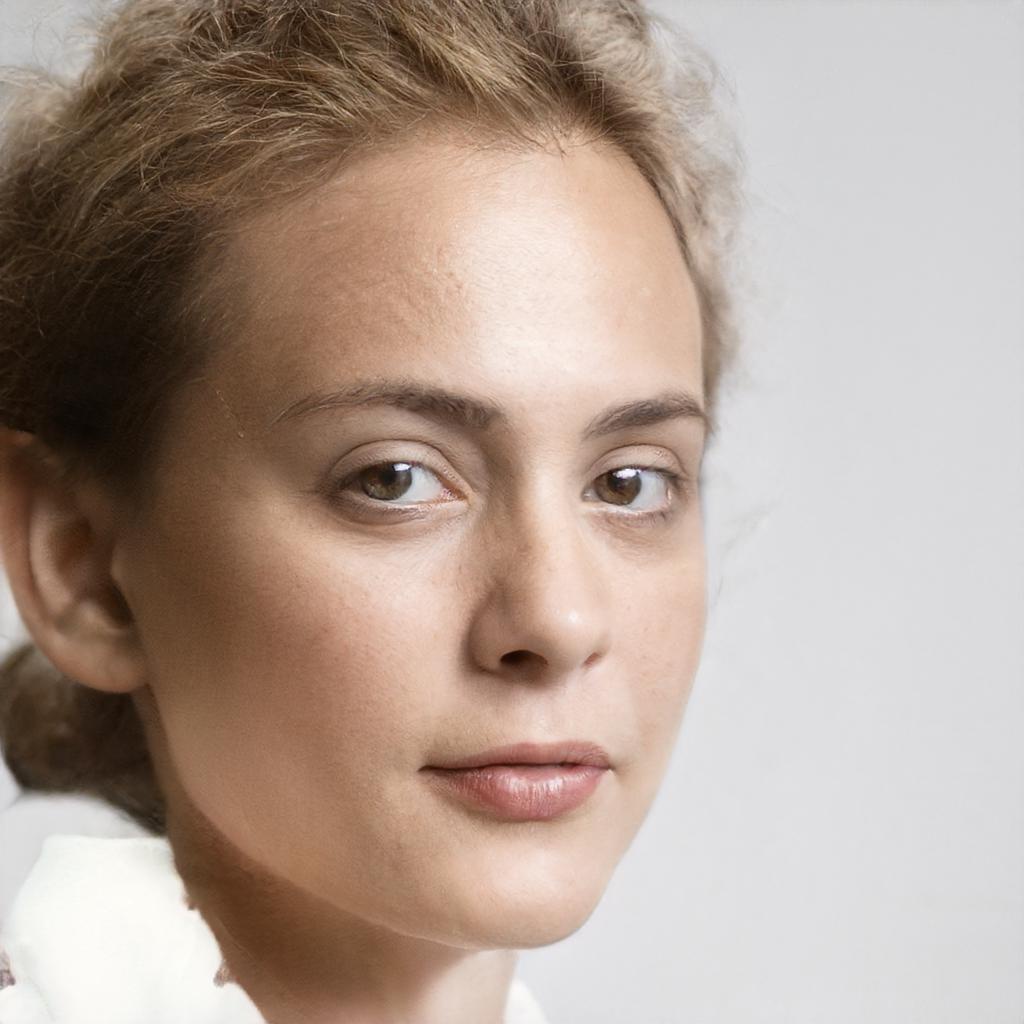} &
			\MyImm{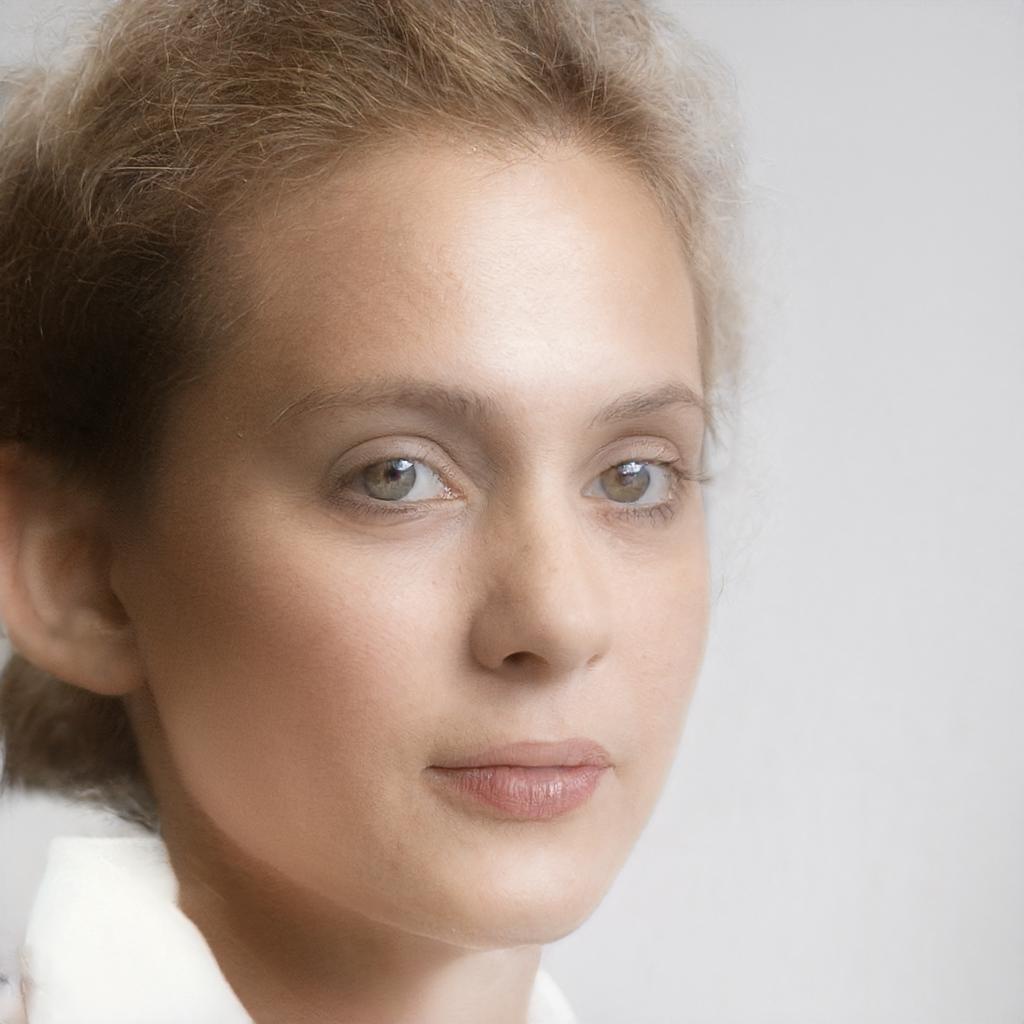} &
			\MyImm{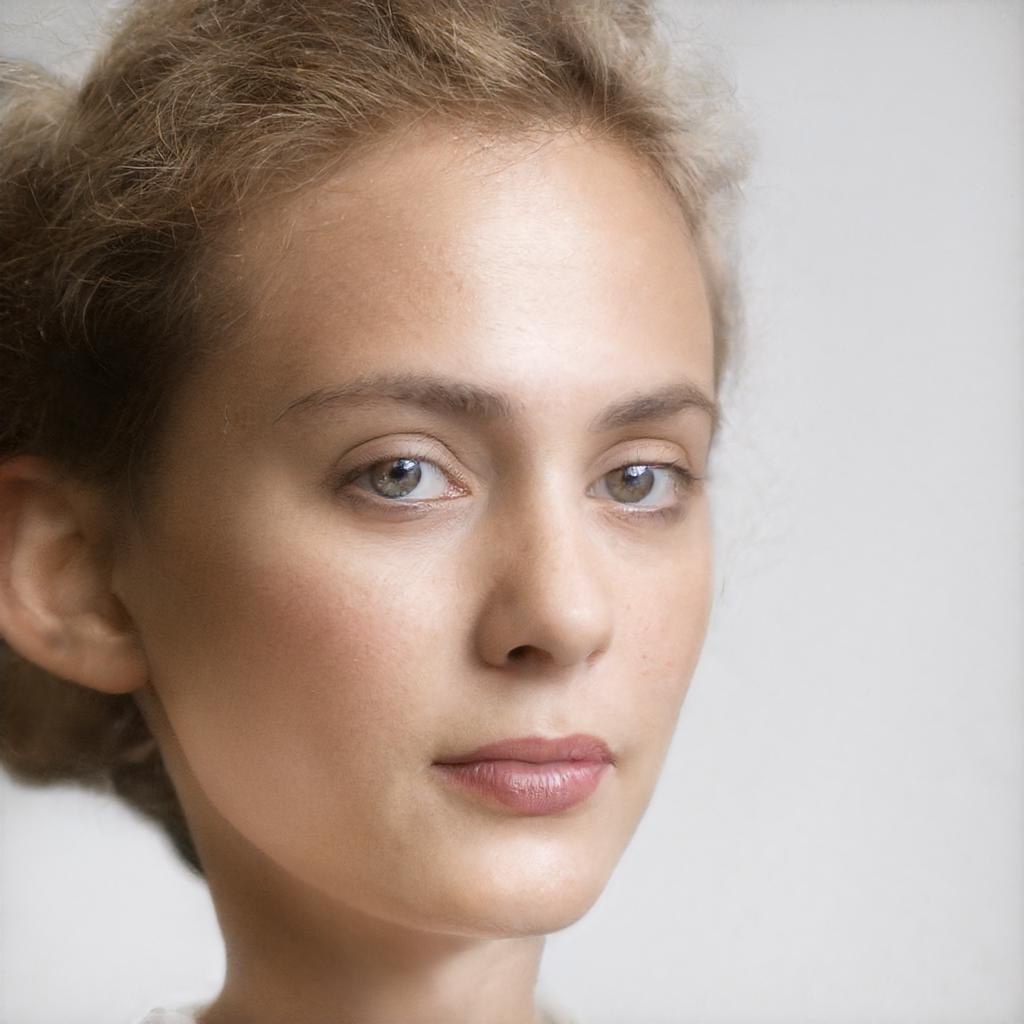} \\
			
			\MyImm{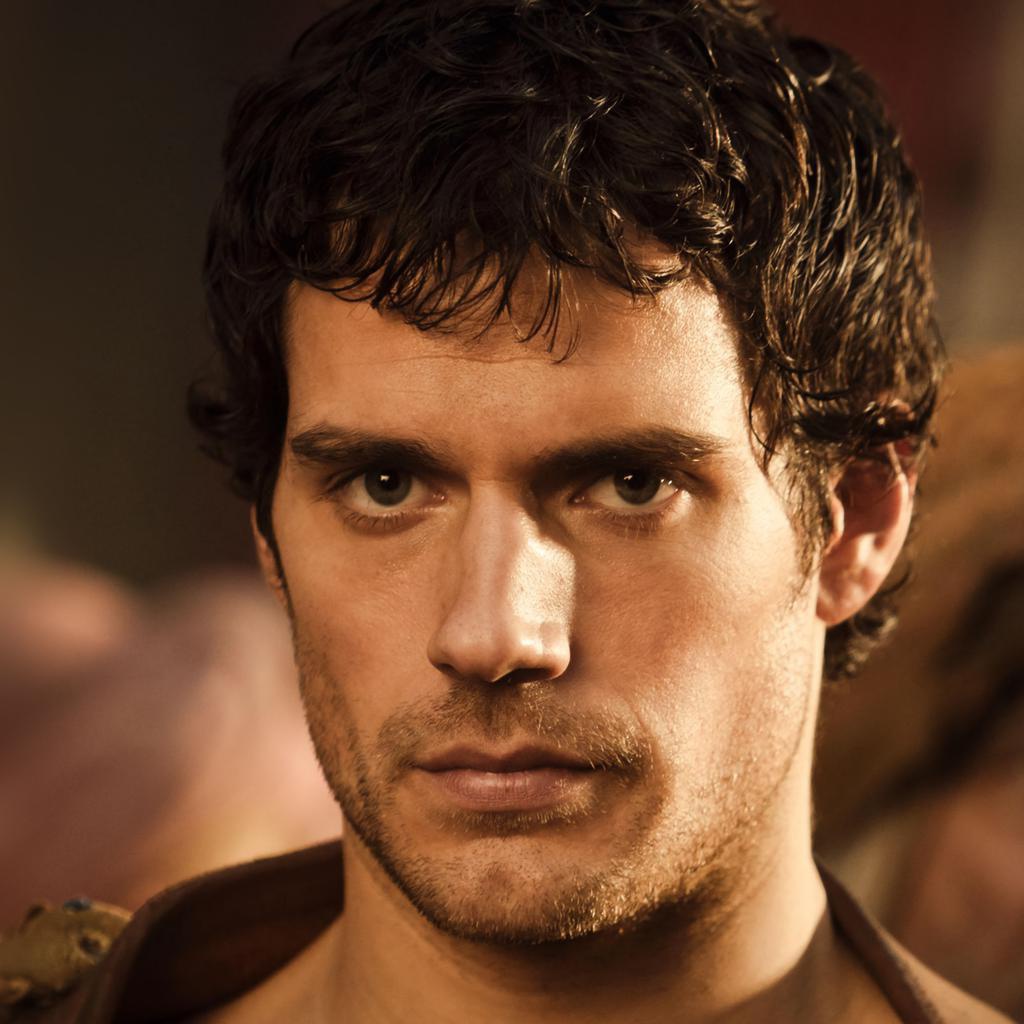} &
			\MyImm{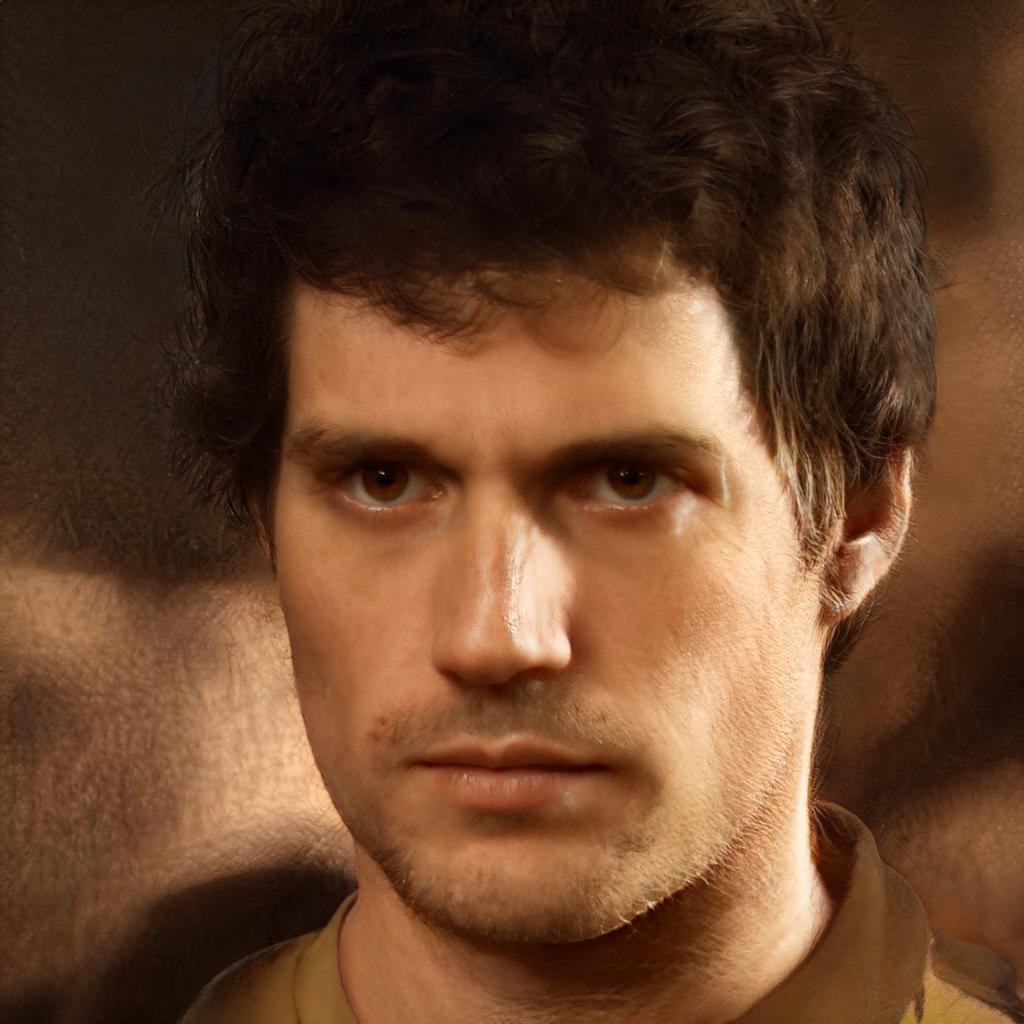} &
			\MyImm{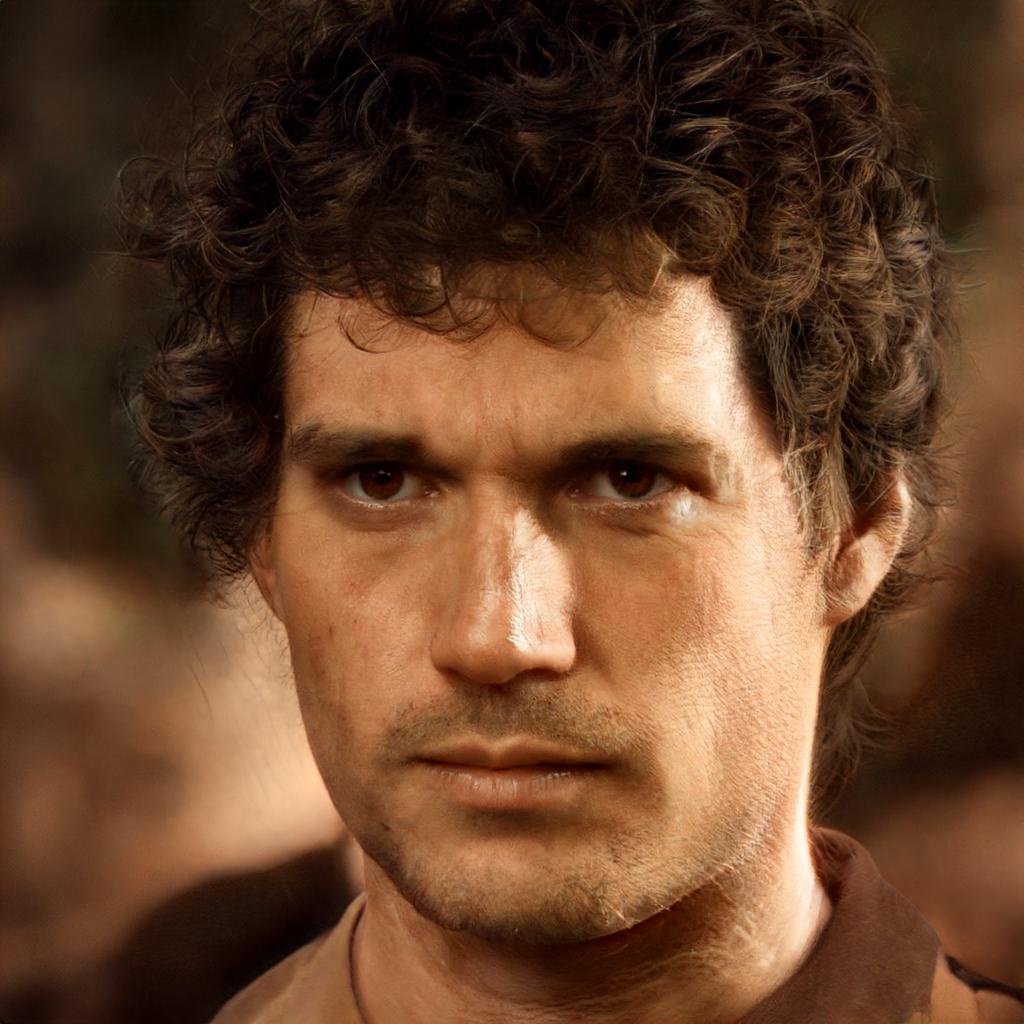} &
			\MyImm{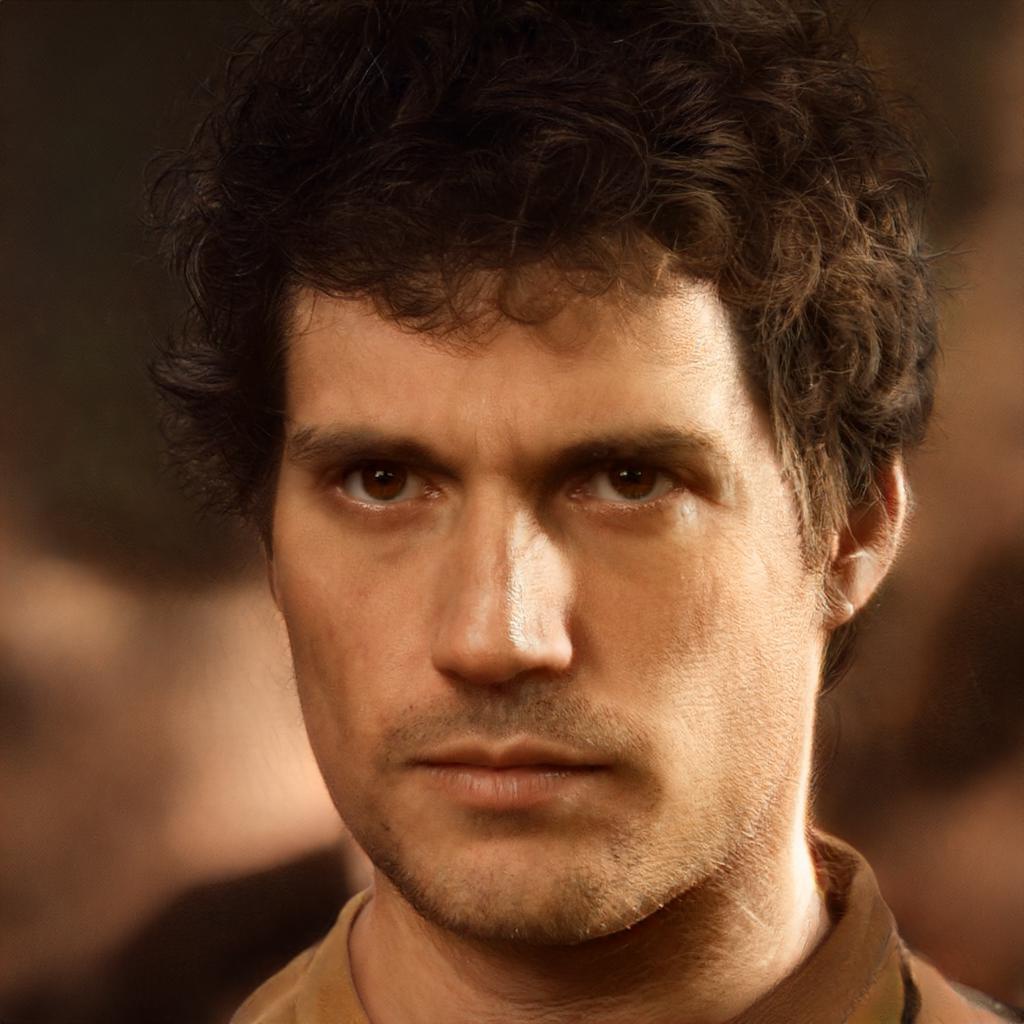} &
			\MyImm{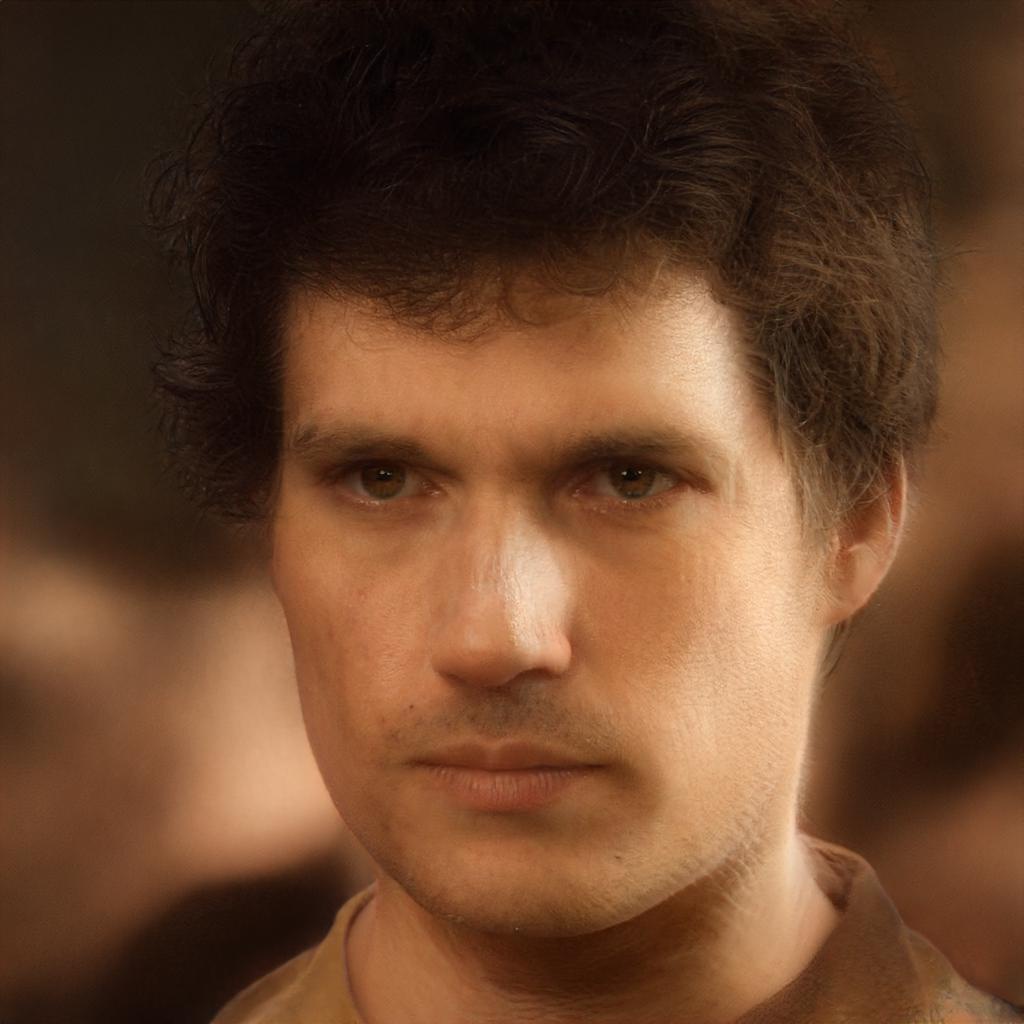} &
			\MyImm{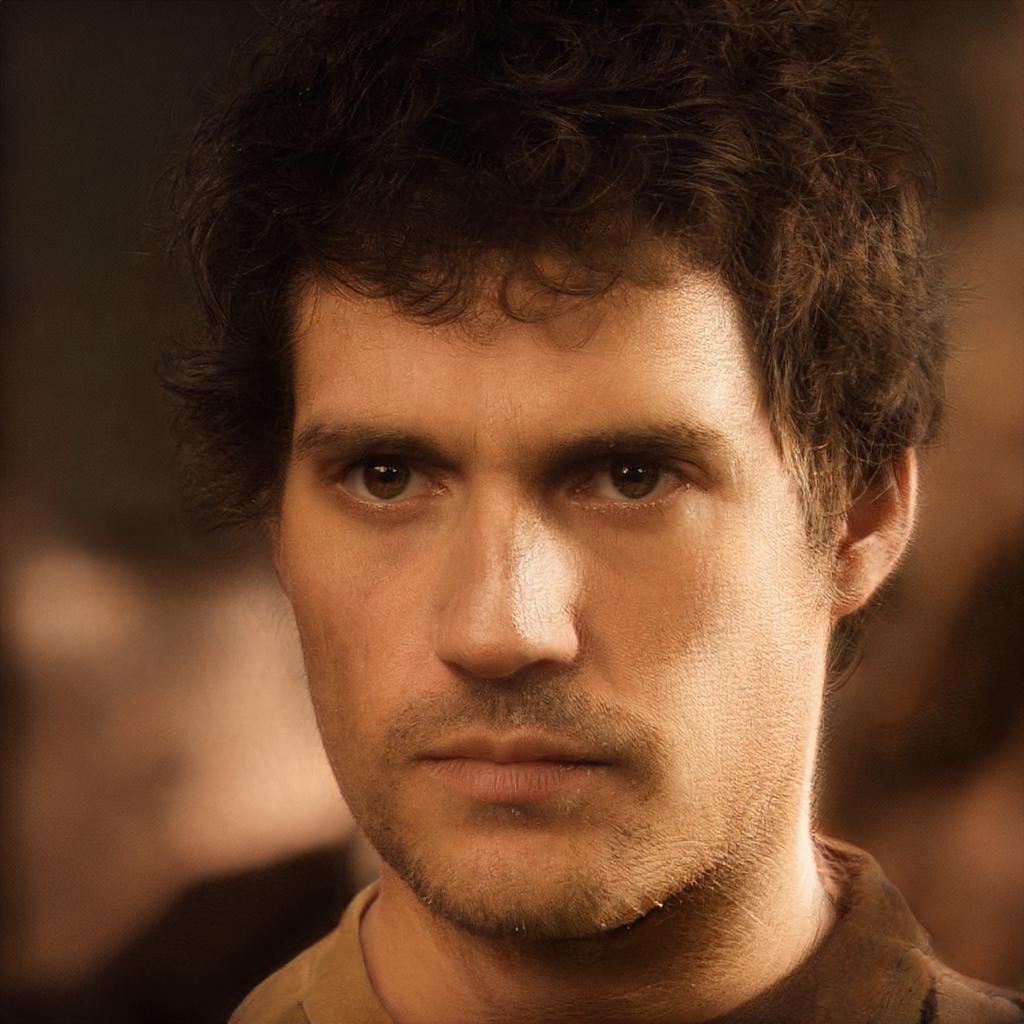} \\
			
			\MyImm{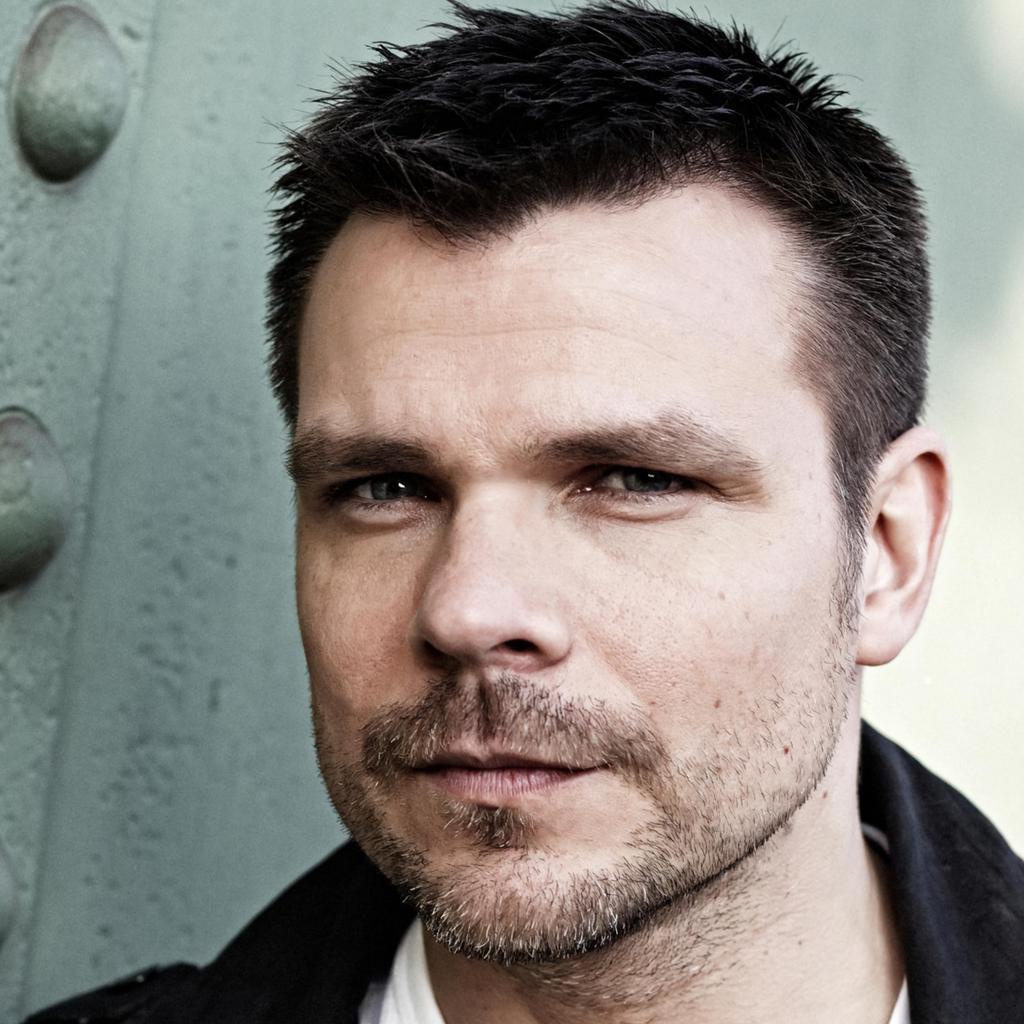} &
			\MyImm{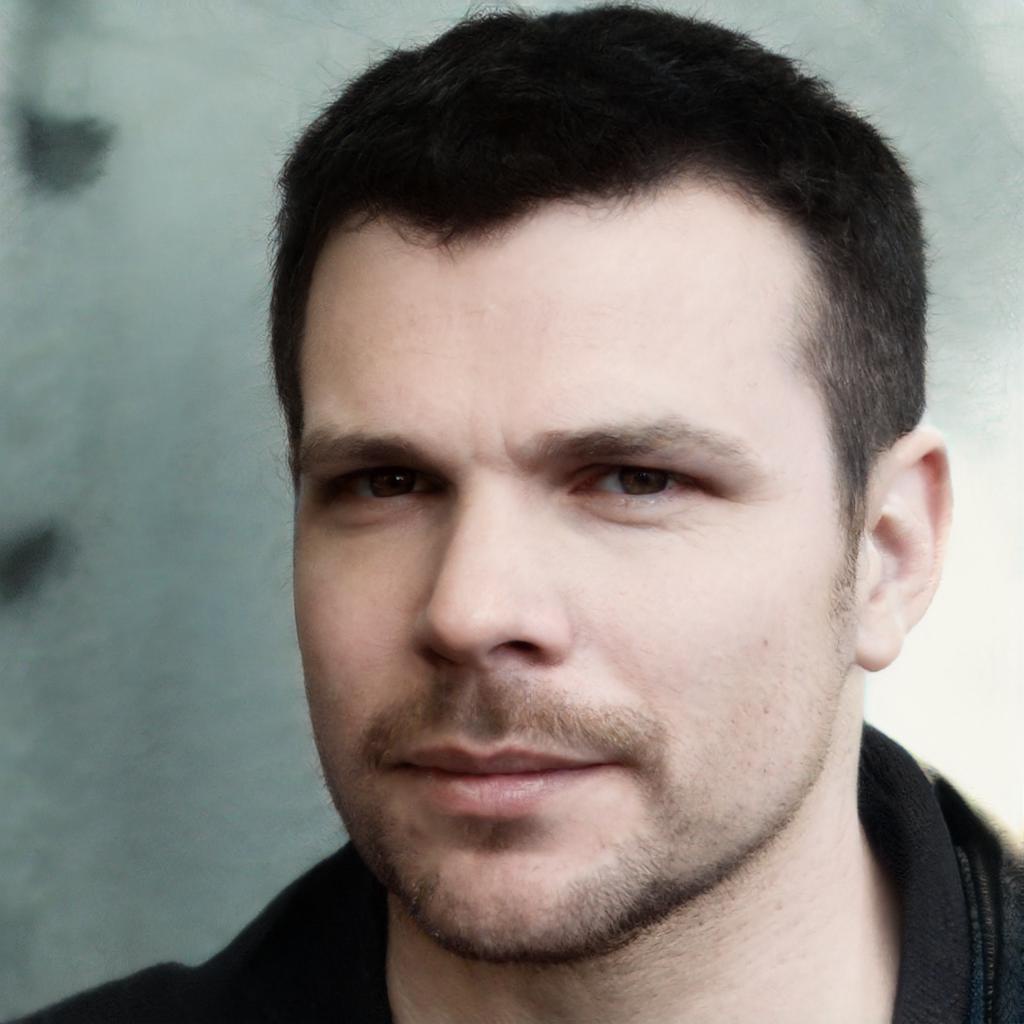} &
			\MyImm{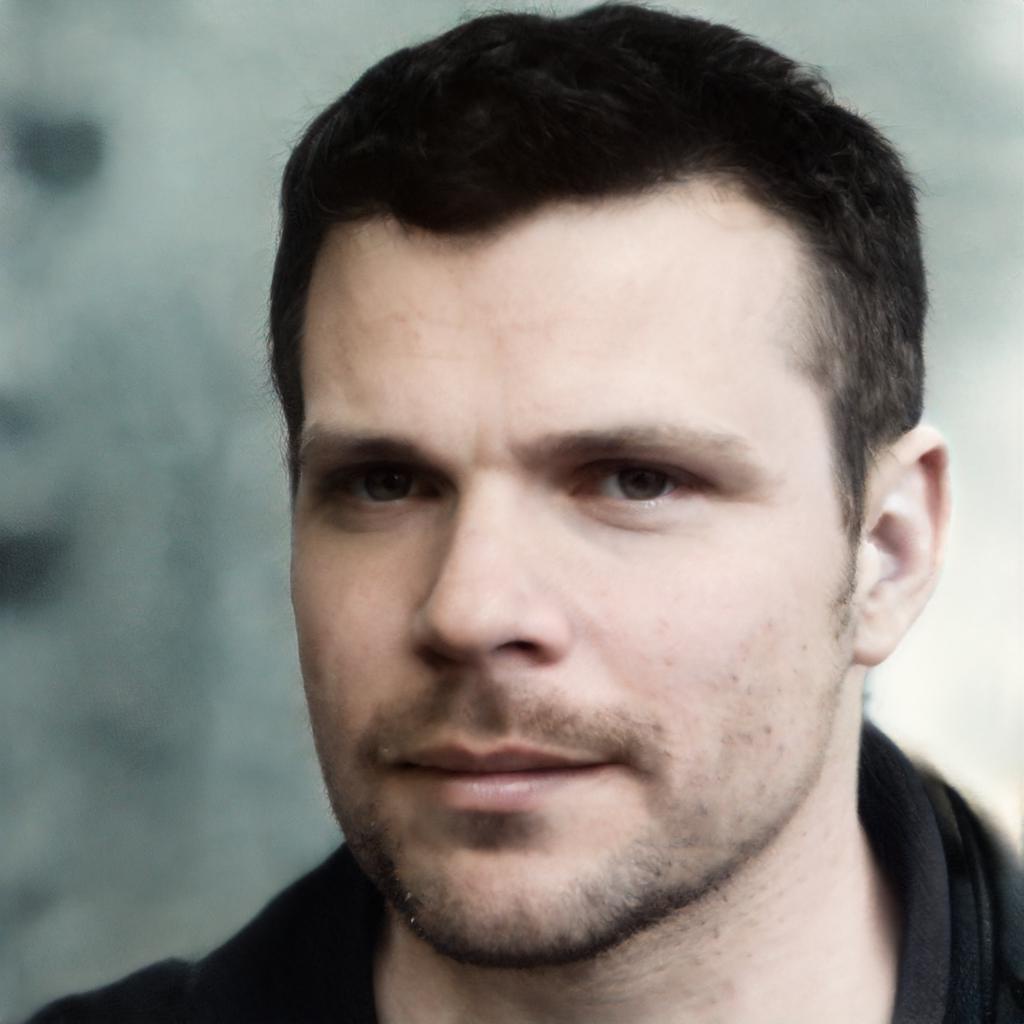} &
			\MyImm{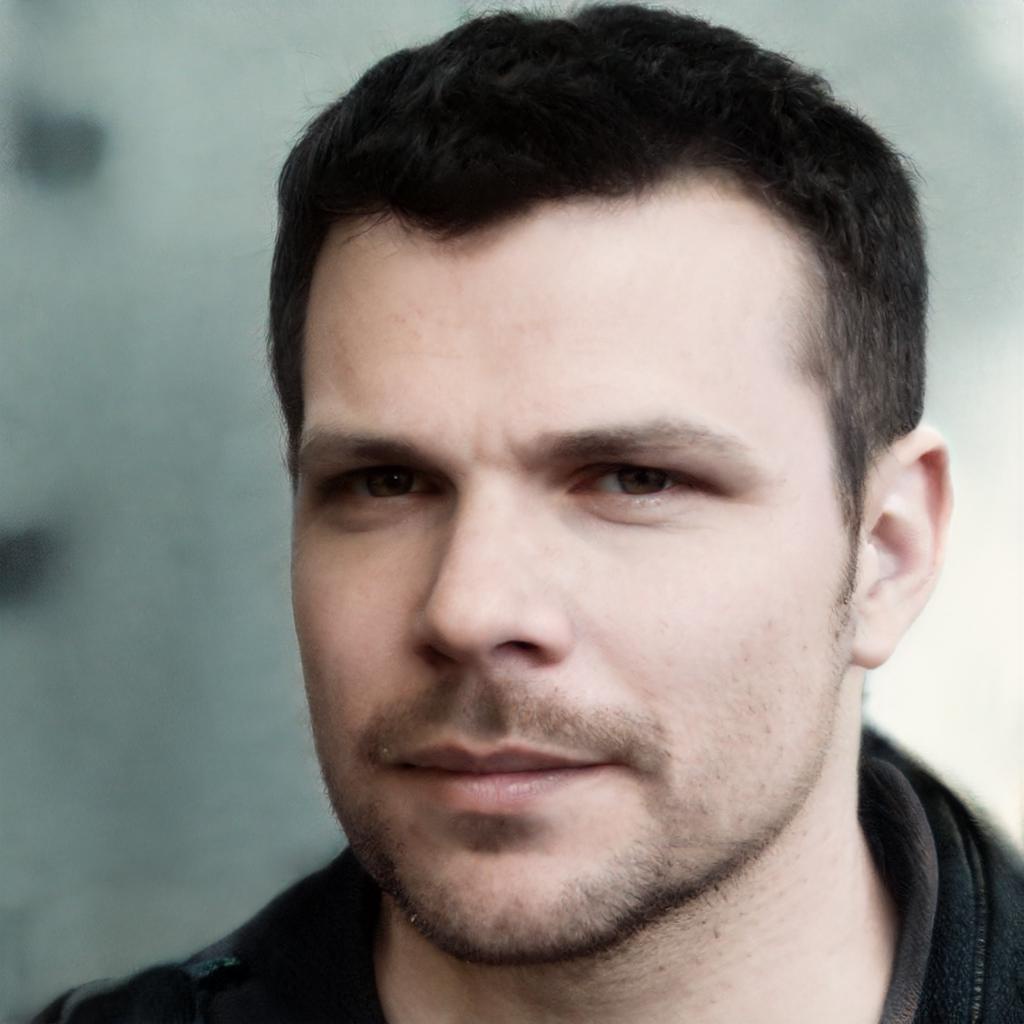} &
			\MyImm{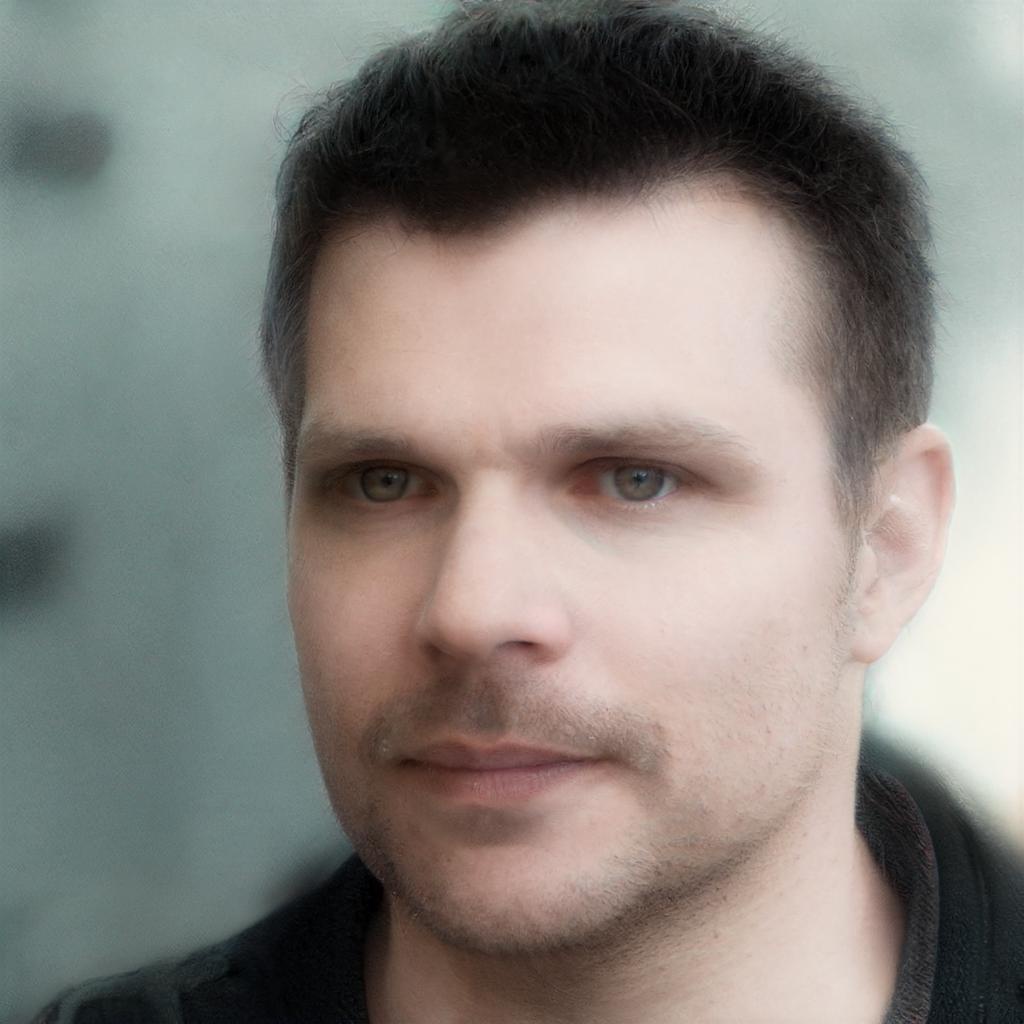} &
			\MyImm{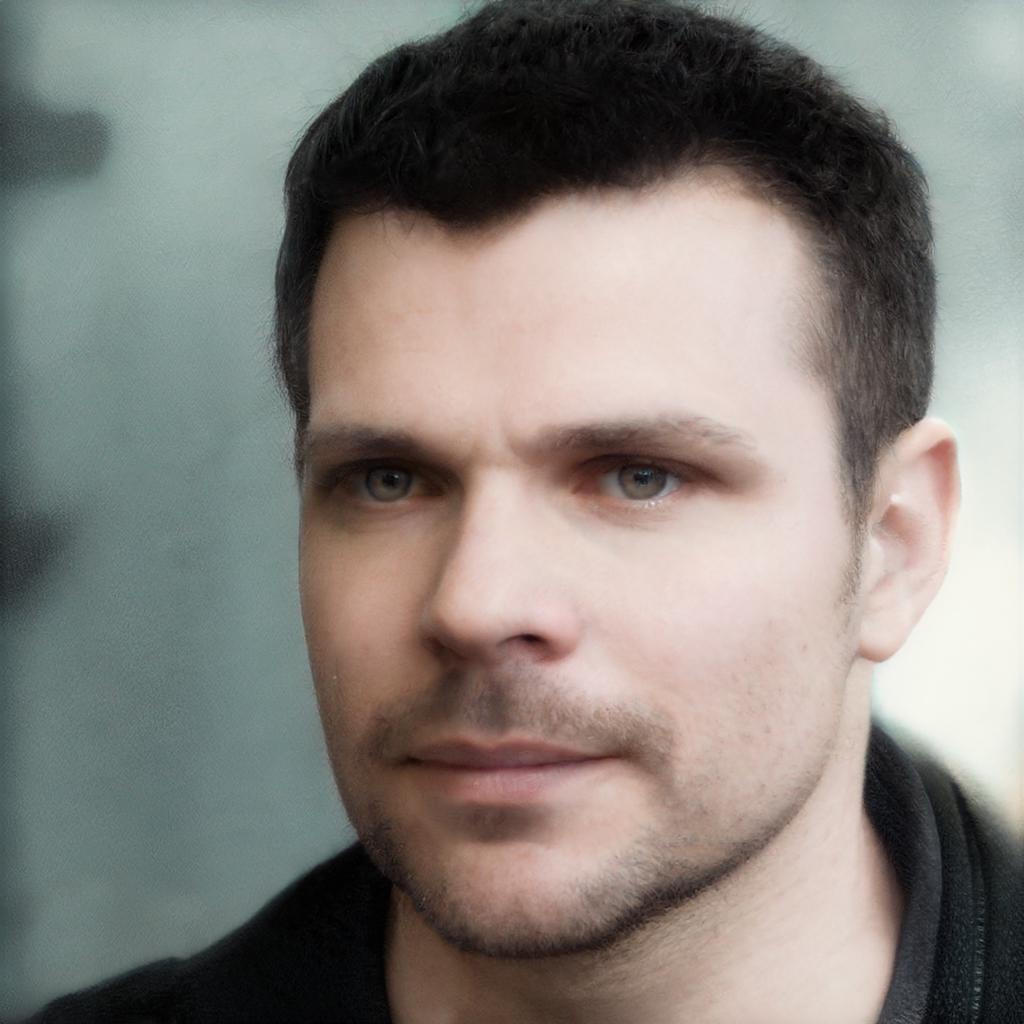} \\
			
			\MyImm{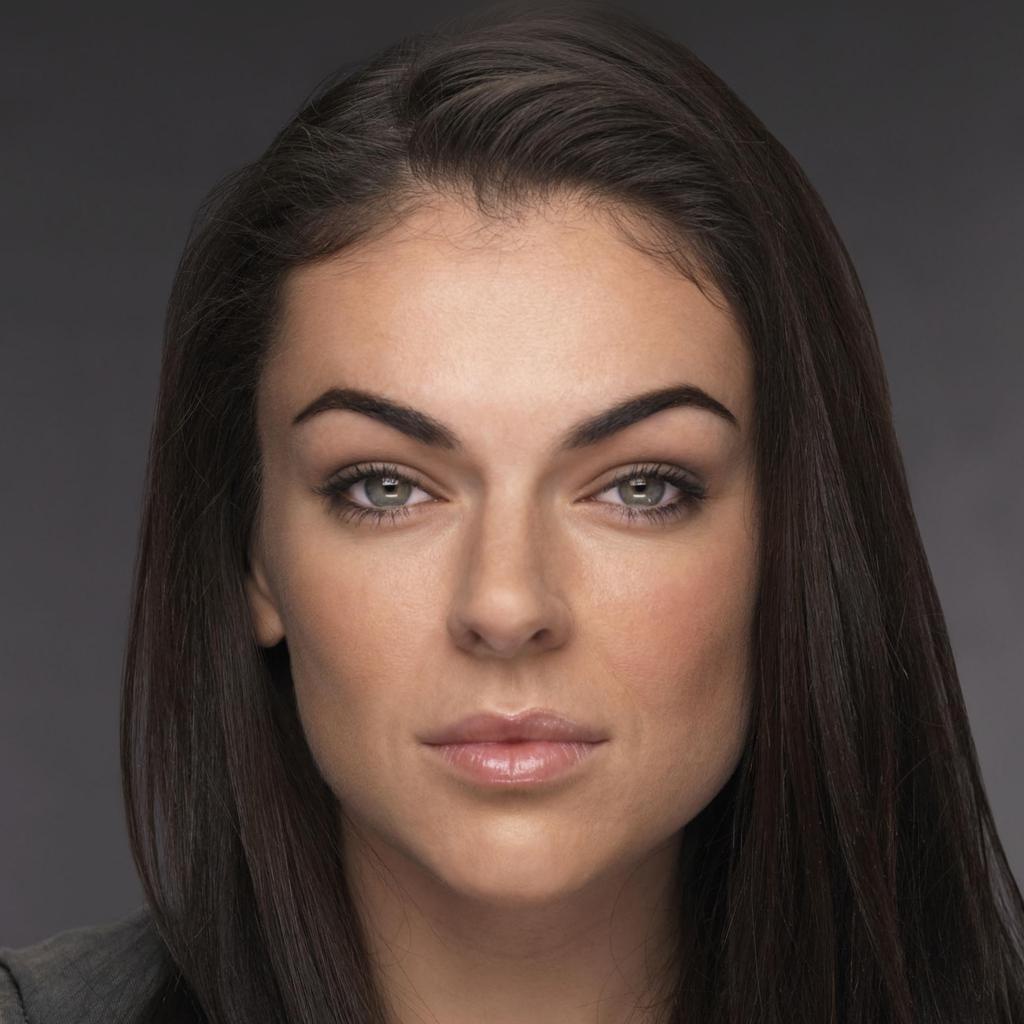} &
			\MyImm{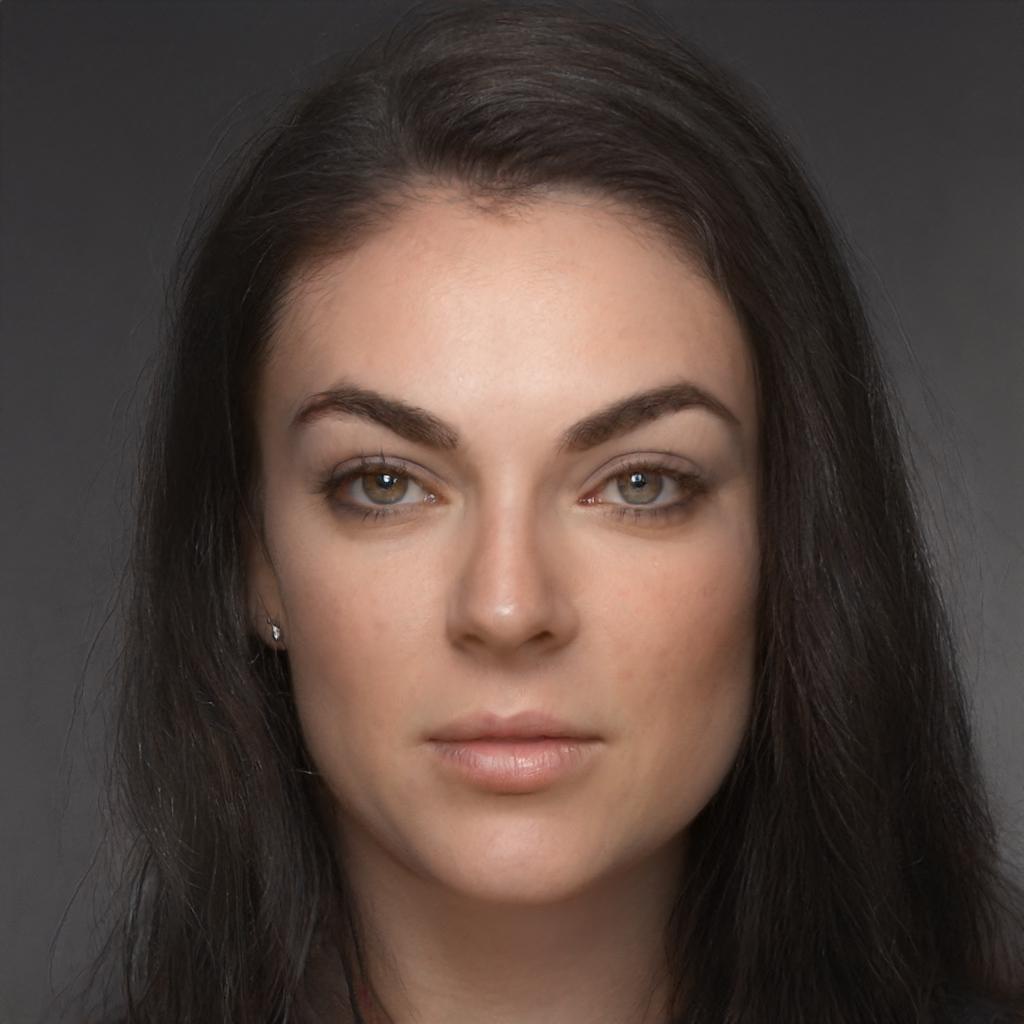} &
			\MyImm{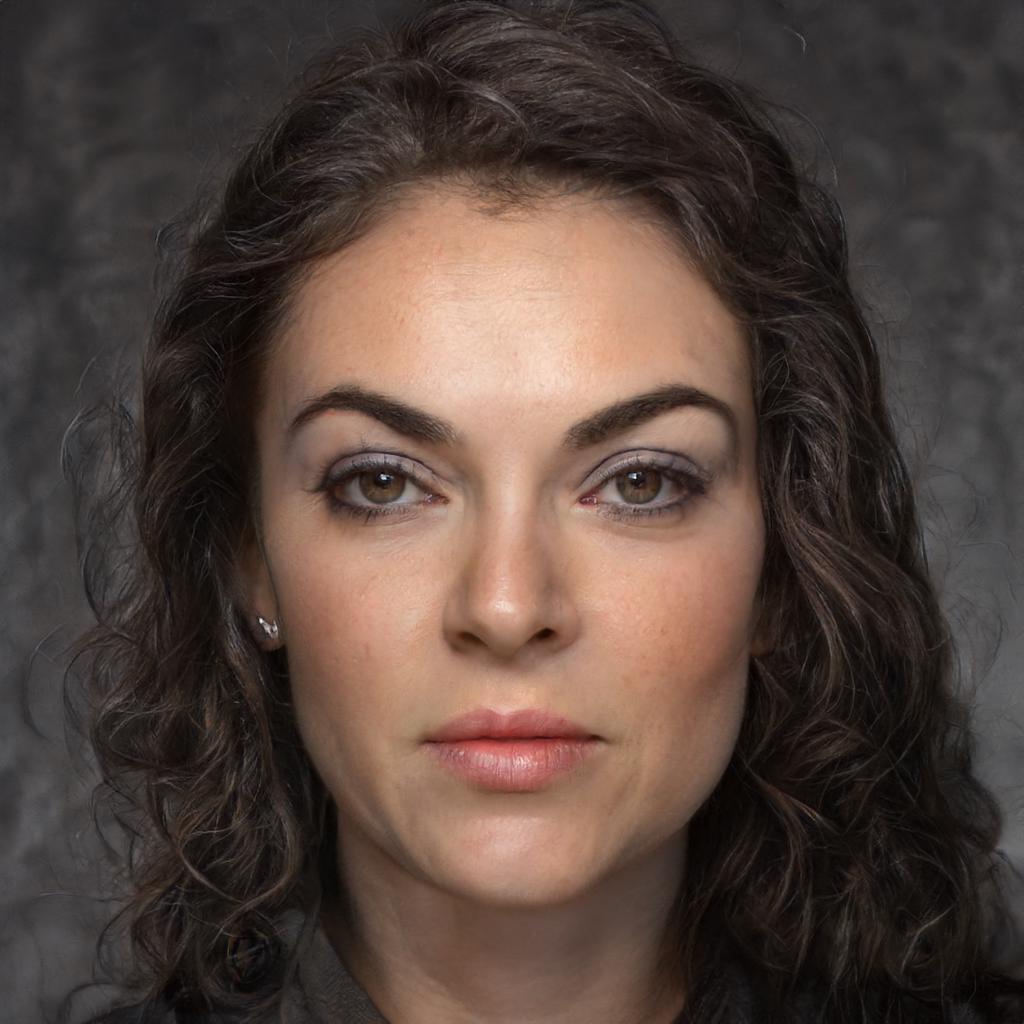} &
			\MyImm{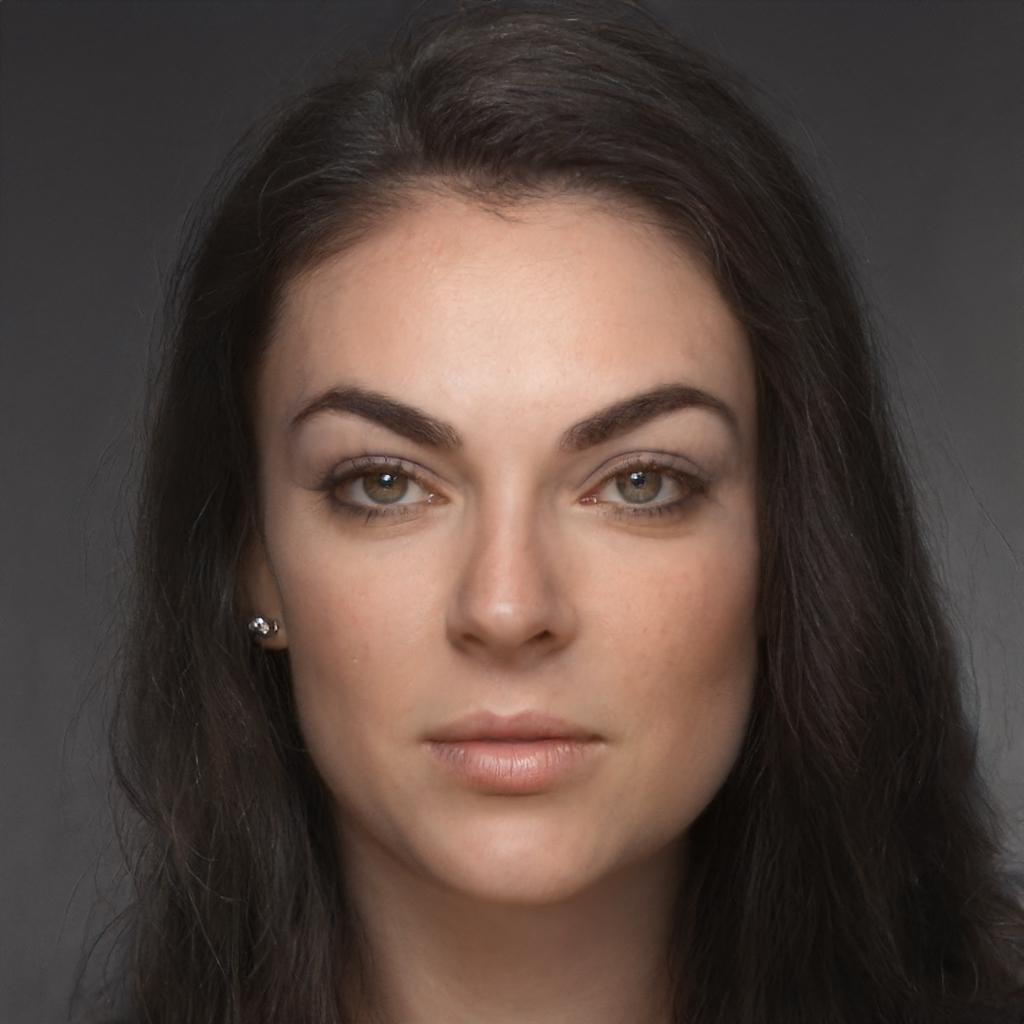} &
			\MyImm{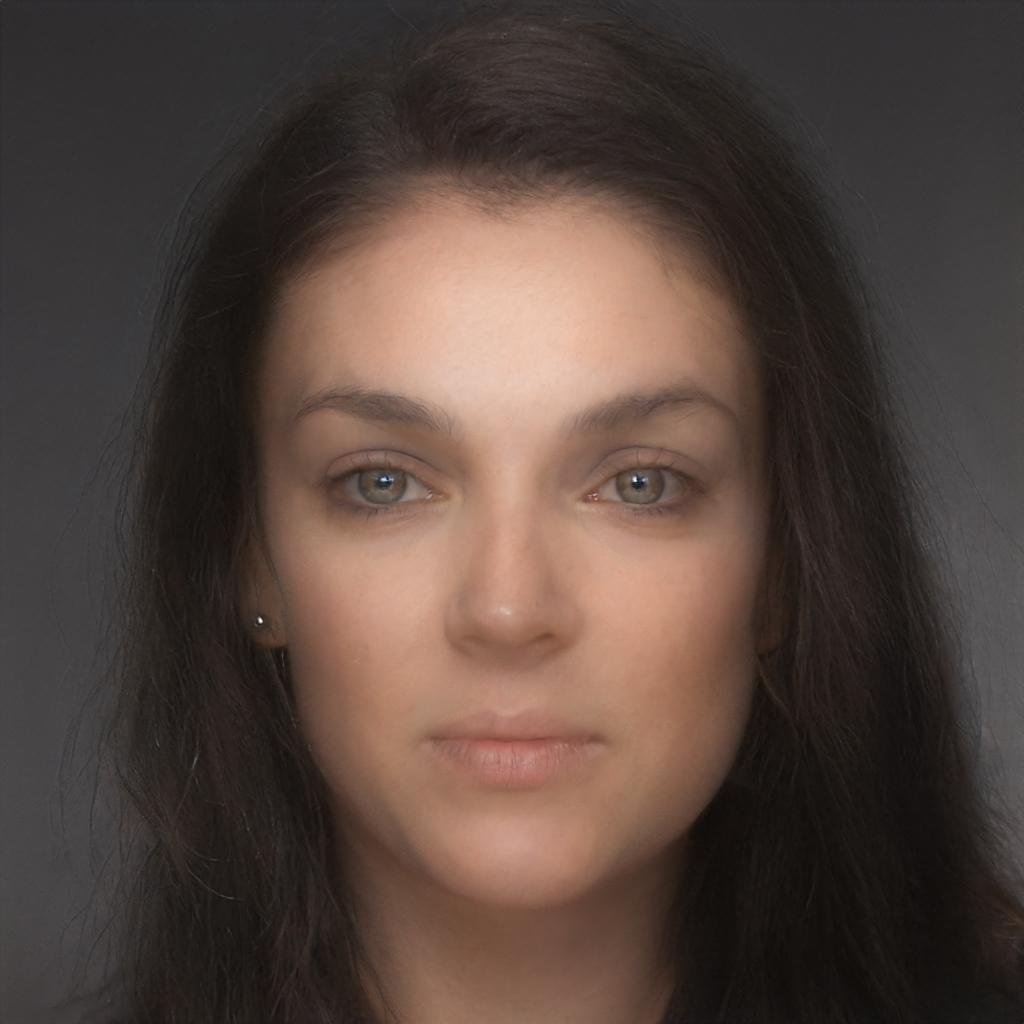} &
			\MyImm{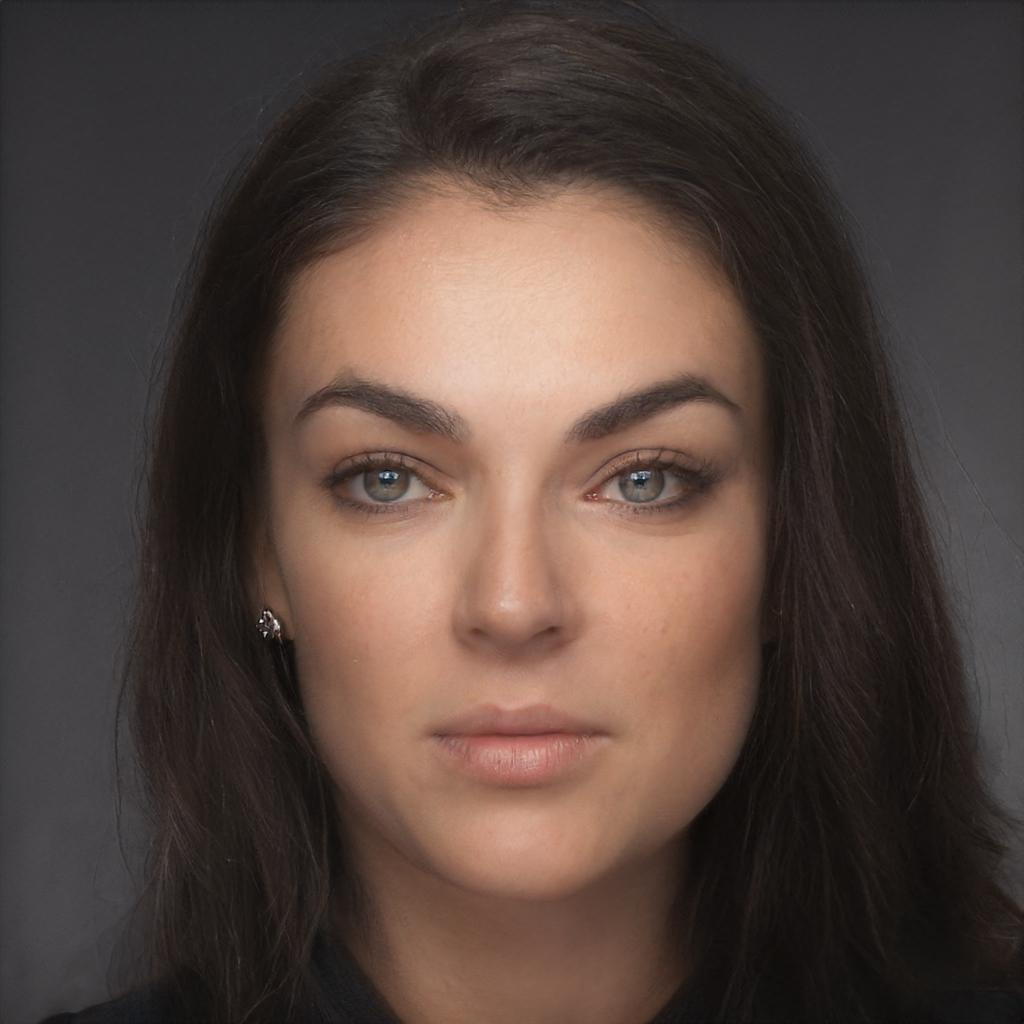} \\
			
		\end{tabular}
	\end{tiny}	
	\caption{Robustness evaluation on 16x super-resolution.}
	\label{fig:robust_app}
\end{figure}

It is worth noting that, although the quantitative results presented in \cref{tab:robust_16} indicate a reduction in identity-similarity when exposed to Gaussian noise, the qualitative outcomes suggest that the additional noise only affects some low-level attributes, such as skin tone and hairstyle, but does not impact the individual's identity. This is illustrated in \cref{fig:robust_app} through additional examples for further support.

\paragraph{Diversity of the output.}
At large magnification factors, many SR solutions could match the given LR input. Although these may not precisely match the HR ground-truth image, obtaining a range of diverse and plausible outputs can be advantageous. Different initialization of the latent code can be used to produce diverse SR outputs for the same LR input. In Figure \ref{fig:diversity}, the diversity of outputs generated by RLS and PULSE for a magnification of 64x is compared. It shows that RLS can better produce multiple consistent and realistic HR images from a single LR image. We hypothesize that this could be explained by the fact that PULSE limits the solutions to lie on a hypersphere which does not cover the whole distribution of images learned during StyleGAN training.

\section{Conclusion}
Super-resolved images reconstructed by existing self-supervised approaches, exploiting a pre-trained style-based generative model, seem to suffer from a lack of realism, especially when the original image is out-of-domain. With this work, we address this issue by first introducing a new regularization of the latent space exploration, which leverages a normalizing flow model, providing a more robust image prior to ensuring that the latent code remains in the original generative model manifold. Then, in order to reconstruct the image with higher fidelity, we slightly fine-tune the generative prior, within a small $\ell_1$-norm ball centered at the latent code obtained at the first step. Doing so enables us to mitigate the fidelity-realness trade-off. We performed extensive experiments demonstrating that our method can generate high-quality face images with clear facial details from severely degraded ones, outperforming prior works.

\bibliographystyle{ieee_fullname}
\bibliography{ms}
\end{document}